\documentclass[journal]{IEEEtran}
\usepackage{cite}
\usepackage{graphicx}
\usepackage{bm}
\usepackage{subfigure}
\usepackage{latexsym}
\usepackage{amsmath}
\usepackage{amssymb}
\usepackage{amsfonts}
\usepackage{color}
\usepackage{enumerate}
\usepackage{algorithm}
\usepackage{algpseudocode}
\usepackage{multirow}
\usepackage{tabularx}
\usepackage{booktabs}
\usepackage{lineno}
\usepackage{epstopdf}
\usepackage{makecell}
\usepackage[utf8]{inputenc}
\usepackage{hyperref}					
\usepackage{nomencl}
\makenomenclature
\hypersetup{colorlinks,
	linkcolor=black,%
	citecolor=black}

\graphicspath{{pic/}}

\makeatletter

\newcommand{\Rmnum}[1]{\expandafter\@slowromancap\romannumeral#1@}
\makeatother

\ifCLASSINFOpdf
\else

\fi

\begin{document}

\title{Hyperspectral Unmixing Based on Nonnegative Matrix Factorization: A Comprehensive  Review}

\author{Xin-Ru~Feng,
        Heng-Chao~Li,~\IEEEmembership{Senior Member,~IEEE,}
        Rui~Wang,
        Qian~Du,~\IEEEmembership{Fellow,~IEEE,}\\
        Xiuping Jia,~\IEEEmembership{Fellow,~IEEE,}
        and Antonio Plaza,~\IEEEmembership{Fellow,~IEEE}
\thanks{This work was supported in part by the National Natural Science Foundation of China under Grants 61871335 and 61801404, and in part by the Fundamental Research Funds for the Central Universities under Grant 2682020ZT35. \textit{(Corresponding author: Heng-Chao Li)}}
\thanks{X.-R. Feng and R. Wang are with the School of Information Science and Technology, Southwest Jiaotong University, Chengdu 611756, China.}
\thanks{H.-C. Li is with the School of Information Science and Technology, Southwest Jiaotong University, Chengdu 611756, China, and also with the National Engineering Laboratory of Integrated Transportation Big Data Application Technology, Southwest Jiaotong University, Chengdu 611756, China, (e-mail: lihengchao\_78@163.com)}
\thanks{Q. Du is with the Department of Electrical and Computer Engineering, Mississippi State University, Mississippi State, MS 39762 USA.}
\thanks{X. Jia is with the School of Engineering and Information Technology, University of New South Wales, Canberra, ACT 2612, Australia.}
\thanks{A. Plaza is with the Hyperspectral Computing Laboratory, Department of Technology of Computers and Communications, Escuela Polit$\acute{\textrm{e}}$cnica, University of Extremadura, 10071 C$\acute{\textrm{a}}$ceres, Spain.}
}

\markboth{IEEE Journal of Selected Topics in Applied Earth Observations and Remote Sensing}
{Shell \MakeLowercase{\textit{et al.}}: Bare Demo of IEEEtran.cls for IEEE Journals}

\maketitle

\begin{abstract}
Hyperspectral unmixing has been an important technique that estimates a set of endmembers and their corresponding abundances from a hyperspectral image (HSI). Nonnegative matrix factorization (NMF) plays an increasingly significant role in solving this problem. In this article, we present a comprehensive survey of the NMF-based methods proposed for hyperspectral unmixing. Taking the NMF model as a baseline, we show how to improve NMF by utilizing the main properties of HSIs (\emph{e.g.}, spectral, spatial, and structural information). We categorize three important development directions including constrained NMF, structured NMF, and generalized NMF. Furthermore, several experiments are conducted to illustrate the effectiveness of associated algorithms. Finally, we conclude the article with possible future directions with the purposes of providing guidelines and inspiration to promote the development of hyperspectral unmixing.
\end{abstract}

\printnomenclature

\nomenclature{HSIs}{Hyperspectral images}
\nomenclature{DL}{Deep learning}
\nomenclature{MV}{Minimum volume}
\nomenclature{PPI}{Pixel purity index}
\nomenclature{IEA}{Iterative error analysis}
\nomenclature{VCA}{Vertex component analysis}
\nomenclature{SGA}{Simplex growing algorithm}
\nomenclature{ICA}{Independent component analysis}
\nomenclature{NMF}{Nonnegative matrix factorization}
\nomenclature{LSTM}{Long short-term memory}
\nomenclature{LMM}{Linear mixture model}
\nomenclature{NLMM}{Nonlinear mixture model}
\nomenclature{ANC}{Abundance nonnegative constraint}
\nomenclature{ASC}{Abundance sum-to-one constraint}
\nomenclature{FCLS}{Fully constrained least squares}
\nomenclature{TV}{Total variation}
\nomenclature{GPU}{Graphics processing units}
\nomenclature{PSO}{Particle swarm optimization}
\nomenclature{BMM}{Bilinear mixture model}
\nomenclature{MRF}{Markov random field}
\nomenclature{SLIC}{Simple linear iterative clustering}
\nomenclature{NTF}{Nonnegative tensor factorization}
\nomenclature{CPD}{Canonical polyadic decomposition}
\nomenclature{BTD}{Block term decomposition}

\section{Introduction}
\IEEEPARstart{H}{yperspectral} images (HSIs) acquired by imaging spectrometers, record hundreds or thousands of spectral bands of the observed scene in a single acquisition \cite{DLandgrebe2002, GSwayze1992, ROGreen1998, JMBioucasDias2012, AZare2007, JRPatel2019}. Owing to the wealthy spectral information, HSIs have been applied to many applications including agricultural and military defense \cite{JSenthilnath2013, MJKhan2018, RPu2021}, food quality control \cite{AGowen2007, YFeng2012}, mineralogical mapping of earth surface \cite{ARHuete2003}, and pharmaceutical manufacturing industry \cite{CGendrin2003}. Due to the complexity of objects and the relatively low spatial resolution, pixels in HSIs are normally composed of mixed spectral responses from multiple ground objects \cite{JMBioucasDias2012, RHeylen2014}. Mixed pixels affect the performance of hyperspectral analysis, such as object classification and identification. To address this problem, hyperspectral unmixing is developed to decompose each pixel of an HSI into a set of endmembers and their corresponding abundances.

In general, unmixing algorithms can be divided into four categories: geometrical, sparse regression-based, statistical, and deep learning (DL)-based methods. Geometrical unmixing algorithms work under the assumption that the endmembers of an HSI are the vertices of a simplex with the minimum volume enclosing the data set or of a simplex with the maximum volume contained in the convex hull of the data set. Pure pixel-based and minimum volume (MV)-based methods belong to this category. The pure pixel-based algorithms assume that there is one pure pixel at least per endmember.  The classical methods include the pixel purity index (PPI) \cite{JBoardman1993}, N-FINDR \cite{MEWinter1999}, the iterative error analysis (IEA) \cite{RANeville1999}, the vertex component analysis (VCA) \cite{JMPNascimento2005}, and the simplex growing algorithm (SGA) \cite{CChang2006, CChang2017}. The MV-based approaches seek a mixing matrix that minimizes the volume of the simplex defined by its columns, such as the minimum volume enclosing simplex (MVSA) \cite{JLi2015} and the simplex identification via variable splitting and augmented Lagrangian (SISAL) \cite{JMBioucasDias2009}. To reduce the influence induced by the intrinsic nonlinearity of the geometric manifold of the HSI and extract the endmembers accurately, a novel nonlinear endmember extraction algorithm \cite{BYang2020} was proposed by combining the hypergraph framework-based manifold representation and fuzzy assessment. In \cite{CHLin2017}, maximum volume inscribed ellipsoid (MVIE) method was presented to attract endmembers effectively. As one of the mainstream methods, the geometrical unmixing approaches have shown their powerful ability in extracting endmembers from HSIs. However, the geometrical algorithms may hardly extract the endmembers from the highly mixed data since pure spectral signatures are not available.

With the increasing availability of spectral libraries for materials measured on the ground, sparse regression-based methods are proposed by expressing each mixed pixel in a scene as a linear combination of a finite set of pure spectral signatures in a spectral library. Sparse regression-based algorithms avoid estimating the number of endmembers and identifying the endmember signatures in the original data set \cite{MDIordache2011, MDIordache2014}. Owing to these two advantages, more efforts have been dedicated to improving the sparse unmixing performance. For example, double weights were introduced in \cite{RWang2017Hyperspectral} to improve the sparsity of fractional abundances in both spectral and spatial domains, where one is used to enhance the sparsity of endmembers in the spectral library, and the other is to encourage the sparsity of fractional abundances. To make full use of the spatial-contextual information, a new spectral-spatial weighted sparse unmixing (S$^2$WSU) framework \cite{SZhang2018} was developed for hyperspectral unmixing. Besides, the spatial correlation was incorporated to promote the abundance estimation in \cite{HLi2021, LQi2020, TInce2020}. Specifically, a superpixel-based reweighted low-rank and total variation (SUSRLR-TV) \cite{HLi2021} method was proposed to enhance the performance of the traditional spatial-regularization-based sparse unmixing approaches. By using multiview collaborative sparse and spectral-spatial-weights, the new sparse unmixing model \cite{LQi2020} took the advantage of spectral information as well as spatial information. In \cite{TInce2020}, graph Laplacian regularization was utilized to promote the smoothness of abundance maps in the sparse regression framework. These methods have obtained promising unmixing results. However, the spectra in the library have high coherence and are undesirable due to the diverse imaging conditions, which limit the applicability of these approaches.

The statistical algorithms identify the endmembers and their corresponding abundances at the same time by utilizing the statistical properties of the HSI. Popular statistical algorithms include independent component analysis (ICA) \cite{NWang2015, SJia2007}, nonnegative matrix factorization (NMF) \cite{RHuang2018, YYuan2021}, and Bayesian approaches \cite{NDobigeon2009, KEThemelis2012, JSBhatt2014}. Among them, NMF provides a good fit for hyperspectral unmixing owing to its nonnegativity and interpretability. Therefore, numerous NMF-based methods have been developed to pursue better unmixing performance.

In the last few years, DL \cite{GEHinton2006, JSBhatt2020} has shown great power and potential in pattern recognition. Therefore, many researchers have focused on hyperspectral unmixing using autoencoder and its variants, achieving more competitive unmixing performance \cite{YSu2018, BPalsson2018, YQu2019, YSu2019, MWang2019, JRPatel2021, BPalsson2022}. In addition, Hong \emph{et al.} \cite{DHong2021} proposed an effective guidance for real endmembers with shared weights in the autoencoder-like architecture. In \cite{MZhao2021}, the HSI was processed as sequential data, and a long short-term memory (LSTM) network was included in autoencoder architecture to capture spectral correlation information. These approaches can learn the abundance fractions from the original data via a series of the hierarchical layers, which are more suitable for coping with a variety of situations. For the availability in practice, \cite{VijayashekharSS2022} proposed a two-stage fully connected self-supervised DL network for alleviating some practical issues, such as the noise and perturbation. Nevertheless, there are still several drawbacks \cite{CZhou2022}. For instance, current approaches often require a lot of training samples and  network parameters to achieve satisfactory unmixing performance.

Compared with the geometrical methods and sparse regression-based methods, the NMF-based methods are powerful to extract simultaneously the endmembers and their associated abundances. Combining the ability to extract hierarchical features as DL-based approaches, multilayer/deep NMF \cite{RRajabi2015, XRFeng2018} models have been developed to explore hidden information with interpretability power as in classical NMF.

From the perspective of spectral signatures in HSIs, there are two main challenges. One is spectral variability \cite{BSomers2009, BSomers2011, AZare2014, RABorsoi2021}, which is often brought by many factors such as changes in illumination, environmental, atmospheric, and temporal conditions. It may lead to large amounts of errors in abundance estimation. To address this problem, a hierarchical sparse NMF (HSNMF) \cite{TUezato2020} introduced hierarchical sparsity constraints for describing endmember variability. Besides, endmember variability was considered by building a $4$D endmember tensor along with a new low-rank regularization \cite{TImbiriba2020} and relying on structured additively-tuned linear mixing model \cite{BSalah2020}. Another issue is the multiple physical interactions in the resulting observed spectrum \cite{RHeylen2014}. This challenge will reduce the generalization ability of unmixing methods based on the linear mixture model (LMM) and increase computational complexity. As such, combined with nonlinear mixture model (NLMM), NMF can also be applied to form some novel nonlinear unmixing methods\cite{NYokoya2014, CFevotte2015}.

In this article, we aim to provide a survey on NMF-based hyperspectral unmixing. We take the NMF model as a baseline to show how to improve NMF by utilizing the main properties of HSIs (\emph{e.g.}, spectral information, spatial information, and structural information). We introduce three important development directions for the NMF model and discuss their pros and cons, including
\begin{itemize}
  \item constrained NMF by introducing additional constraints or penalty terms to the cost function, such as sparsity constraints, smooth constraints, and graph constraints.
  \item structured NMF by modifying the structure of the cost function, \emph{e.g.}, weighted NMF, convex NMF, robust NMF, etc.
  \item generalized NMF by extending the decomposition form, involving nonnegative tensor factorization (NTF), multilayer NMF, deep NMF, etc.
\end{itemize}
In addition, we conduct several experiments to demonstrate the effectiveness of some associated algorithms. The purpose is to give guidelines and inspiration for the future improvement of hyperspectral unmixing.

The layout of this article is as follows. In Section II, we give a general introduction to spectral mixture model and the unmixing problem. Sections III, IV, V review classical NMF unmixing categories according to different constraints, structures of the cost function, and decomposition forms, respectively. Extensive experiments are conducted and the results are discussed in Section VI. Section VII draws comprehensive conclusions and presents a brief outlook on future possible research directions.

\section{Classical NMF for Hyperspectral Unmixing}
Let $\mathcal{X}\in\mathbb{R}^{r\times c\times B}$ denote an HSI with $B$ bands, $r$ rows, and $c$ columns. Through the unfolding operation, the HSI can be represented by a matrix
\begin{equation}
\label{eqn:1}
  \mathbf{X} = \begin{bmatrix}
      x_{11} & \cdots & x_{1P} \\
      \vdots & \ddots & \vdots\\
      x_{B1} & \cdots & x_{BP} \\
\end{bmatrix}\in \mathbb{R}^{B \times P},
\end{equation}
with the number of pixels $P=r\times c$. The element at ($b,p$) denoted by $x_{bp}$ represents the reflection value from $b$-th band of $p$-th pixel. From the row perspective, $\mathbf{X}=[\mathbf{x}^1,\mathbf{x}^2,\cdots,\mathbf{x}^B]$ where the $b$-th vector $\mathbf{x}^b$ is the ground information in the $b$-th band. Generally, $\mathbf{X}=[\mathbf{x}_1,\mathbf{x}_2,\cdots,\mathbf{x}_P]$ is given from the column perspective, where the $p$-th vector $\mathbf{x}_p$ is the spectrum of $p$-th pixel. The LMM assumes that an observed pixel spectrum in an HSI can be produced by a linear combination of endmember signatures and their corresponding abundances.
The matrix formulation of the LMM can be described as
\begin{equation}
\label{eqn:2}
  \mathbf{X} = \mathbf{AS} + \mathbf{G},
\end{equation}
where $\mathbf{A}=[\mathbf{a}_1,\mathbf{a}_2,\cdots,\mathbf{a}_M] \in \mathbb{R}^{B \times M}$ denotes endmember matrix, the vector $\mathbf{a}_m$ represents the $m$-th endmember signature, and $M$ denotes the number of endmembers. $\mathbf{S} \in \mathbb{R} ^{M \times P}$ represents the abundance matrix for all endmembers, and $\mathbf{G} \in \mathbb{R}^{B \times P} $ represents the noise matrix. Typically, two constraints are imposed on $\mathbf{S}$, \emph{i.e.}, the abundance nonnegative constraint (ANC) and the abundance sum-to-one constraint (ASC), given by $\mathbf{S}\geq0$ and $\mathbf{1}_M^T\mathbf{S}=\mathbf{1}_P^T$, here $\mathbf{1}_M$ and $\mathbf{1}_P$ are all-one column vectors with size $M$ and size $P$, respectively, and $(\cdot)^T$ denotes the transpose operation.

Given a matrix $\mathbf{X}$, NMF \cite{DDLee2001} focuses on decomposing it into the product of two nonnegative matrices $\mathbf{A}$ and $\mathbf{S}$, \emph{i.e.}, $\mathbf{X}\approx\mathbf{AS}$. Obviously, this decomposition form is consistent with LMM. Thus, NMF is attractive for hyperspectral unmixing. In general, Frobenius norm is utilized to measure the approximation between $\mathbf{X}$ and $\mathbf{AS}$, and the cost function is expressed as
\begin{equation}
\label{eqn:3}
  \min_{\mathbf{A},\mathbf{S}}~\lVert \mathbf{X-AS} \rVert_F^2,~\textrm{s.t.}~\mathbf{A}\geq 0,\mathbf{S}\geq 0, \mathbf{1}_M^T\mathbf{S}=\mathbf{1}_P^T,
\end{equation}
where the operator $\lVert \mathbf{\cdot} \rVert_F $ denotes the Frobenius norm. The multiplicative update rules are deduced as
\begin{subequations}
\label{eqn:4}
\begin{align}
\label{eqn:4a}
  \mathbf{A}&\leftarrow\mathbf{A}\odot(\mathbf{XS}^T)\oslash(\mathbf{ASS}^T),\\
\label{eqn:4b}
  \mathbf{S}&\leftarrow\mathbf{S}\odot(\mathbf{A}^T\mathbf{X})\oslash(\mathbf{A}^T\mathbf{AS}),
\end{align}
\end{subequations}
in which $\odot$ and $\oslash$ stand for the element-wise multiplication and division, respectively. Meanwhile, when abundance matrix $\mathbf{S}$ is updated, the ASC requires to be satisfied by redefining the observation and spectral signature matrices as
\begin{equation}
\label{eqn:5}
  \mathbf{\bar{X}} =
\begin{bmatrix}
      \mathbf{X}\\
      \delta \mathbf{1}_P^T\\
\end{bmatrix},
\quad\mathbf{\bar{A}} =
\begin{bmatrix}
      \mathbf{A}\\
      \delta \mathbf{1}_M^T\\
\end{bmatrix},
\end{equation}
where $\delta$ is a parameter to control the impact of the ASC. For the implementation of NMF, a crucial issue is how to initialize the related variables. The endmember matrix $\mathbf{A}$ can be initialized by the various methods such as random values from 0 to 1, vertex component analysis (VCA) \cite{JMPNascimento2005}, and automatic target generation process (ATGP) \cite{JCao2018}. Abundance matrix $\mathbf{S}$ is initialized according to the fully constrained least squares (FCLS) algorithm \cite{DCHeinz2001}. To speedup the NMF processing, an adaptive projected NMF (APNMF) algorithm was parallelized by introducing its parallel version in \cite{SARobila2009}. In addition, by applying Nesterov's optimal gradient method, NeNMF \cite{NGuan2012} was proposed to accelerate the optimization.

\begin{figure*}[!t]
\centering
  {\includegraphics[width=16.5cm]{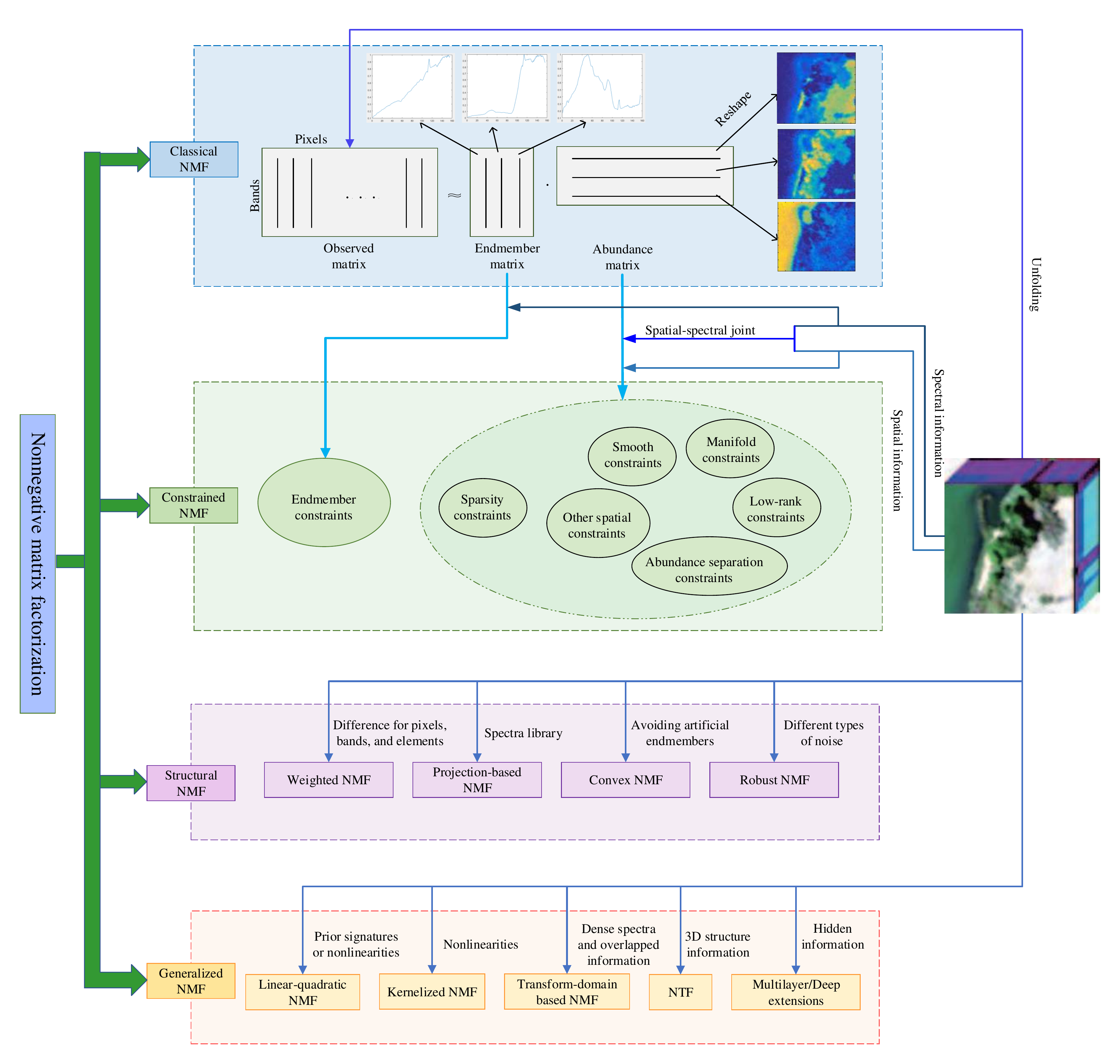}}
  \caption{Framework of the NMF-based methods for hyperspectral unmixing.}
\label{fig:1}
\end{figure*}

Nevertheless, the cost function of NMF is non-convex so that it easily falls into local optimal solutions. As such, to improve the unmixing performance, there are three important improvement directions for the NMF model. A considerable number of NMF-based methods address the spectral and spatial information by introducing additional constraints or penalty terms to the cost function. These methods are reviewed in Section III. As discussed in Section IV, many methods enable flexibility to account for more structures and details such as the difference of pixels, bands, and elements. Moreover, a lot of methods extend the decomposition form to acquire more essential characteristics, \emph{e.g.}, nonlinearities, $3$D structure information, hidden information. Such methods are considered in Section V. The framework of the NMF-based methods for hyperspectral unmixing is illustrated in Fig. \ref{fig:1}. The specifical unmixing methods are mainly summarized in Table \ref{tab:1}.

\begin{table*}[!t]
\renewcommand\arraystretch{1.5}
\caption{The Taxonomy Of the NMF-Based methods for Hyperspectral Unmixing.}
\label{tab:1}
\begin{tabular}{|c|c|c|}
  \hline
  \mbox{CATEGORY} & \mbox{SUBCATEGORY} & \mbox{TYPICAL EXAMPLES} \\ \hline
  \multirow{20}{*}{Constrained NMF}
   & \multirow{4}{*}{Endmember constraints}
                               & MVC-NMF \cite{LMiao2007}, \textbf{MPEC-NMF} \cite{KQu2020}, VRNMF \cite{AMSAng2019}, R-CoNMF \cite{JLi2016},
                                 \textbf{ICoNMF-TV} \cite{YYuan2020}, \\
   &                           & MDCNMF  \cite{YYu2007}, CSNMF-GPU \cite{ZWu2014}, APSO and MOPSO \cite{BYang2017}, BCNMF \cite{BYang2018}, \\
   &                           & IP-NMF \cite{CRevel2018}, PSnsNMF and PSNMFSC \cite{SJia2009}, MiniDisCo \cite{AHuck2010}, EDCNMF\cite{NWang2013}, \\
   &                           & Arctan-NMF \cite{YESalehani2017}, \textbf{MRS-PPK} \cite{LTong2016}, MCG-NMF \cite{WWang2020}. \\
   \cline{2-3}
   & \multirow{5}{*}{Sparsity constraints}
                & PSnsNMF and PSNMFSC \cite{SJia2009}, \textbf{$L_{1/2}$-NMF} \cite{YQian2011}, \textbf{HT$L_{1/2}$-NMF} \cite{WWang2015}, DgS-NMF\cite{FZhu2014}, \\
   &            & \textbf{DGC-NMF} \cite{RHuang2017}, NMF-SMC \cite{ZYang2011}, CSNMF-GPU \cite{ZWu2014}, Arctan-NMF \cite{YESalehani2017}, TV-RSNMF \cite{WHe2017}, \\
   &            & RNLCF \cite{TPeng2018}, DSPLR-NMF \cite{CGTsinos2017}, GMCA-HU \cite{XXu2020Generalized}, TS-NMF \cite{XXu2020Curvelet}, CSsRS-NMF \cite{XLi2021}, \\
   &                           & GMC-NMF \cite{FXiong2020}, R-CoNMF \cite{JLi2016}, Sparsity constrained distributed unmixing  \cite{SKhoshsokhan2019}, \\
   &                           & SGSNMF \cite{XWang2017}, NLNMF \cite{LYang2020}. \\
   \cline{2-3}
   & \multirow{3}{*}{Smooth constraints}
                               & PSnsNMF and PSNMFSC \cite{SJia2009}, ASSNMF \cite{XLiu2011}, WNMF \cite{JLiu2012}, DAC$^2$NMF \cite{RLiu2016}, \\
   &                           & Sparsity constrained distributed unmixing  \cite{SKhoshsokhan2019}, TV-RSNMF \cite{WHe2017}, \\
   &                         & \textbf{ICoNMF-TV} \cite{YYuan2020}, \textbf{MPEC-NMF} \cite{KQu2020}, NLHTV-LSRNMF \cite{JYao2019}. \\
   \cline{2-3}
   & \multirow{3}{*}{Manifold constraints}
                               & \textbf{GLNMF} \cite{XLu2013}, \textbf{MRS-PPK} \cite{LTong2016}, \textbf{SC-GLR$_{l_1}$NMF} \cite{BRathnayake2020}, HGNMF \cite{TPeng2020}, GNMF \cite{SYang2015}, \\
   &                           & \textbf{SS-NMF} \cite{FZhu2014ISPRS}, \textbf{BF-$L_2$SNMF} \cite{ZZhang2018}, GLrNMF \cite{MWang2018},
                                 \textbf{HG$L_{1/2}$-NMF} \cite{WWang2016}, \textbf{R-NMF}\cite{LTong2017}, \\
   &                           & \textbf{SRASU} \cite{JZhang2020}, SS-NMF \cite{GZhang2021}, \textbf{SC-NMF} \cite{XLu2020}. \\
   \cline{2-3}
   & Low-rank constraints      & DSPLR-NMF \cite{CGTsinos2017}, GLrNMF \cite{MWang2018}. \\
   \cline{2-3}
   & Abundance separation constraints
                               & ASSNMF \cite{XLiu2011}, DAC$^2$NMF \cite{RLiu2016}, Semi-supervised NMF \cite{YJia2020}. \\
   \cline{2-3}
   & \multirow{2}{*}{Other spatial constraints} & \textbf{CSNMF} \cite{XLu2014}, \textbf{KLSNMF}\cite{BDu2016},
                                 \textbf{Subspace structure regularized sparse NMF} \cite{LZhou2020}, \\
   &                           & \textbf{SDSNMF} \cite{YYuan2015TGRS}, S$^2$-NMF \cite{LDong2021}, Semi-supervised NMF \cite{YJia2020}. \\
  \hline
  \multirow{6}{*}{Structured NMF}
   & Weighted NMF              & CW-NMF \cite{XLv2021}, SpNMF \cite{JPeng2021}. \\
   \cline{2-3}
   & Projection-based NMF      & \textbf{PNMF} \cite{YYuan2015JSTARS}. \\
   \cline{2-3}
   & Convex NMF                & RCMF \cite{NAkhtar2017}, SC-NMF \cite{XLu2020}, HSNMF \cite{TUezato2020}. \\
   \cline{2-3}
   & \multirow{3}{*}{Robust NMF}
   & ISB-NMF$_{L_0}$ \cite{CLi2015}, rNMF \cite{CFevotte2015}, \textbf{$L_{1/2}$-RNMF} \cite{WHe2016Sep}, SSRNMF \cite{RHuang2019}, SS-NMF \cite{GZhang2021}, \\
   &                        & CENMF \cite{YWang2015}, CSsRS-NMF \cite{XLi2021}, RNLCF \cite{TPeng2018},  CANMF-TV \cite{XRFeng2020}, MCSNMF \cite{HWang2019}, \\
   &                        & GLNMF \cite{JPeng2020}.  \\
   \hline
  \multirow{10}{*}{Generalized NMF}
   &  Linear-quadratic NMF     & NMFupk \cite{WTang2012}, Method proposed in \cite{IMeganem2014}, Fan-NMF \cite{OEches2014}, Semi-NMF \cite{NYokoya2014},
                                 RASU \cite{XZhang2018}. \\
   \cline{2-3}
   & Kernelized NMF            & kernel NMF \cite{XLi2014}, biobjective NMF \cite{FZhu2016}, OKNMF \cite{FZhu2017}, IKNMF \cite{RHuang2018}. \\
   \cline{2-3}
   & Transform-domain based NMF & Method proposed in \cite{VijayashekharSS2021}, TS-NMF \cite{XXu2020Curvelet}. \\
   \cline{2-3}
   & \multirow{3}{*}{NTF}
                              & Method proposed in \cite{MAVeganzones2016}, MV-NTF \cite{YQian2017}, S-MV-NTF \cite{FXiong2018}, MV-NTF-TV \cite{FXiong2019}, \\
   &                           & Method proposed in \cite{BFeng2019}, DWSNTF \cite{HCLi2021}, EIC-NTF \cite{JJWang2021}, SPLRTF \cite{PZheng2021}, \textbf{SCNMTF} \cite{HCLi2020}, \\
   &                           & ULTRA \cite{TImbiriba2018}, ULTRA-V \cite{TImbiriba2020}, BUTTDL1 \cite{LSun2022}, Method proposed in \cite{KWang2020}. \\
   \cline{2-3}
   & \multirow{4}{*}{Multilayer/deep extensions}
                               & \textbf{MLNMF} \cite{RRajabi2015}, DCMLNMF \cite{HFang2019}, \textbf{SSTV-MLNMF} \cite{LTong2019}, \textbf{AGMLNMF} \cite{LTong2020}, \\
   &                           & \textbf{HR-MLNMF} \cite{LTong2020Homogeneous}, HP-MLNMF \cite{YYuan2021}, \textbf{CMLNMF} \cite{LChen2017}, 2LFKAA \cite{GZhao2016}, \\
   &                           & AC-MLKNMF \cite{JLiu2021}, \textbf{SDNMF-TV} \cite{XRFeng2018}, $L_1$-DNMF \cite{HFang2018}, SSRDMF \cite{HCLi2022}, \\
   &                           & MNN-BU \cite{YQian2020}, SNMF-Net \cite{FXiong2022}. \\
  \hline
\end{tabular}
\footnotesize{~~~}\\
\footnotesize{Boldfaced letters represent that $L_{1/2}$ regularizer is employed for the corresponding methods.}\\
\end{table*}

\section{Constrained NMF for Hyperspectral Unmixing}

By exploiting the spectral and spatial information in HSIs, additional constraints have been imposed on the endmembers and abundances to obtain better unmixing performance. The constrained NMF model can be integrated as
\begin{equation}
\begin{aligned}
\label{eqn:6}
  \min_{\mathbf{A},\mathbf{S}}~&\lVert \mathbf{X-AS} \rVert_F^2 + \alpha J(\mathbf{A}) + \beta J(\mathbf{S}),\\
  &\textrm{s.t.}~\mathbf{A}\geq 0,\mathbf{S}\geq 0, \mathbf{1}_M^T\mathbf{S}=\mathbf{1}_P^T,
\end{aligned}
\end{equation}
where $J(\mathbf{A})$ and $J(\mathbf{S})$ are regularization terms for endmembers and abundances, respectively, and $\alpha$ and $\beta$ are nonnegative parameters to balance the effect of the corresponding constraint terms. Next, we mainly describe the algorithms that incorporate the constraints on endmember matrix in Subsection III-A. By contrast, numerous works are reported in terms of imposing constraints for abundances, given in Subsection III-B to III-G. More details are presented as follows.

\subsection{Endmember constraints}

The constraints for endmembers are integrated into NMF by minimizing simplex volume \cite{LMiao2007, KQu2020, AMSAng2019, JLi2016, YYuan2020}, compacting endmember distance  \cite{YYu2007, ZWu2014, BYang2017, BYang2018, CRevel2018}, keeping signature smoothness \cite{SJia2009, AHuck2010, NWang2013, YESalehani2017}, introducing prior spectral information \cite{LTong2016}, and exploring high-level semantic information \cite{WWang2020}.

Motivated by geometric insights, the minimum volume constraint (MVC) is robust without the pure pixel assumption, which can be presented as
\begin{equation}
\label{eqn:7}
J(\mathbf{A}) = \emph{Vol}(\mathbf{A}), \\
\end{equation}
where $\emph{Vol}(\mathbf{A})$ is the volume of the simplex whose vertices correspond to the endmembers $\mathbf{A}$. As a typical one, the first minimum-volume NMF was proposed in \cite{LMiao2007} by incorporating the MVC into the NMF to effectively extract endmembers from highly mixed data. It should be noted that $J(\mathbf{A})$ is calculated with dimensionality reduction since the matrix determinant is valid only for square one. On this basis, Qu \emph{et al.} \cite{KQu2020} further introduced total variation (TV) and reweighted sparse regularizers to form a multiple-priors ensemble constrained NMF (MPEC-NMF) method. Although the above methods do not require the pure-pixel assumption, it is inconvenient to adopt appropriate dimensionality reduction. Therefore, more algorithms were proposed to define the volume of the simplex. In \cite{AMSAng2019}, three different volume regularizers were presented to form volume-regularized NMF (VRNMF), including Gram determinant, logarithm Gram determinant, and nuclear norm. Besides, volume minimization can be promoted by pushing endmembers towards a solution which is quadratically regularized by a given simplex \cite{JLi2016, YYuan2020}.

Different from using simplex volume as an endmember constraint, Yu \emph{et al.} \cite{YYu2007} used endmember distance (EMD) to keep the simplex as compact as possible, and proposed a minimum distance constrained NMF (MDC-NMF) method. The EMD is defined as the sum of distances from each endmember to their centroid, expressed as
\begin{equation}
\label{eqn:8}
  J(\mathbf{A}) = \sum_{m=1}^M\|\mathbf{a}_m-\frac{1}{M}\sum_{m=1}^M\mathbf{a}_m\|_2^2.
\end{equation}
This constraint is not only simpler but also convex for endmember matrix $\mathbf{A}$. Under the framework of MDC-NMF, \cite{ZWu2014} reported an unmixing method along with sparsity constraint and graphics processing units (GPU). Yang \emph{et al.} \cite{BYang2017} introduced particle swarm optimization (PSO) to solve the optimization problem, which possessed good global search ability and convergence. Afterwards, the bilinear mixture model (BMM)-based constrained NMF algorithm (BCNMF) was presented in \cite{BYang2018} with the EMD constraint for unsupervised nonlinear spectral unmixing, in which pixels were projected into their approximate linear mixture components based on the characteristics of BMM to reduce the collinearity greatly. Similarly, in \cite{CRevel2018}, an inertia constraint was presented so as to promote the homogeneity of estimated spectra from the same class using the trace of the covariance matrix. It can deal with intra-class variability by extracting a separate set of pure material spectra from each observed pixel spectrum.

To promote the smoothness, different metrics are applied on the matrix $\mathbf{A}$. Specifically, an adaptive potential function from discontinuity adaptive Markov random field (MRF) model was adopted in \cite{SJia2009}. Spectral dispersion function \cite{AHuck2010} encouraged the variance of each endmember spectrum to be as low as possible. In \cite{NWang2013}, an endmember dissimilarity function has been defined to make the estimated endmember signatures to be smooth. Besides, a quadratic weighted norm was used as a regularizer term in \cite{YESalehani2017} to exploit spectral smoothness.

By incorporating prior spectral information, Tong \emph{et al.} \cite{LTong2016} introduced a constraint to minimize the differences between the estimated endmember and the known one. It not only considers the discrepancies between the standard signature and the corresponding one in the image, but also shows flexibility in terms of its extensions.

Recently, in order to effectively learn high-level semantic information, a multiple clustering guided NMF (MCG-NMF) unmixing approach was proposed in \cite{WWang2020}. Specifically, clustering analysis via the $k$-means method is conducted on the data matrix $\mathbf{X}$. Then, selecting $K$ pixels in each cluster, according to the principle of $k$-nearest neighbors. Finally, the $\mathbf{R}^k,k=1,\cdots,K+1$ is constructed. Thus, the regularization term is expressed as
\begin{equation}
\begin{aligned}
\label{eqn:9}
  J(\mathbf{A}) = \frac{1}{K+1}\sum_{k=1}^{K+1}\|(\mathbf{R}^k-\mathbf{A})\mathbf{W}\|_F^2,
\end{aligned}
\end{equation}
where $\mathbf{W}$ is a diagonal matrix and depends on the similarity relationship between each pixel and their cluster mean.

\subsection{Sparsity constraints}

\begin{figure}[!t]
\centering
  \mbox{
  \subfigure[]{\includegraphics[width=2.5cm]{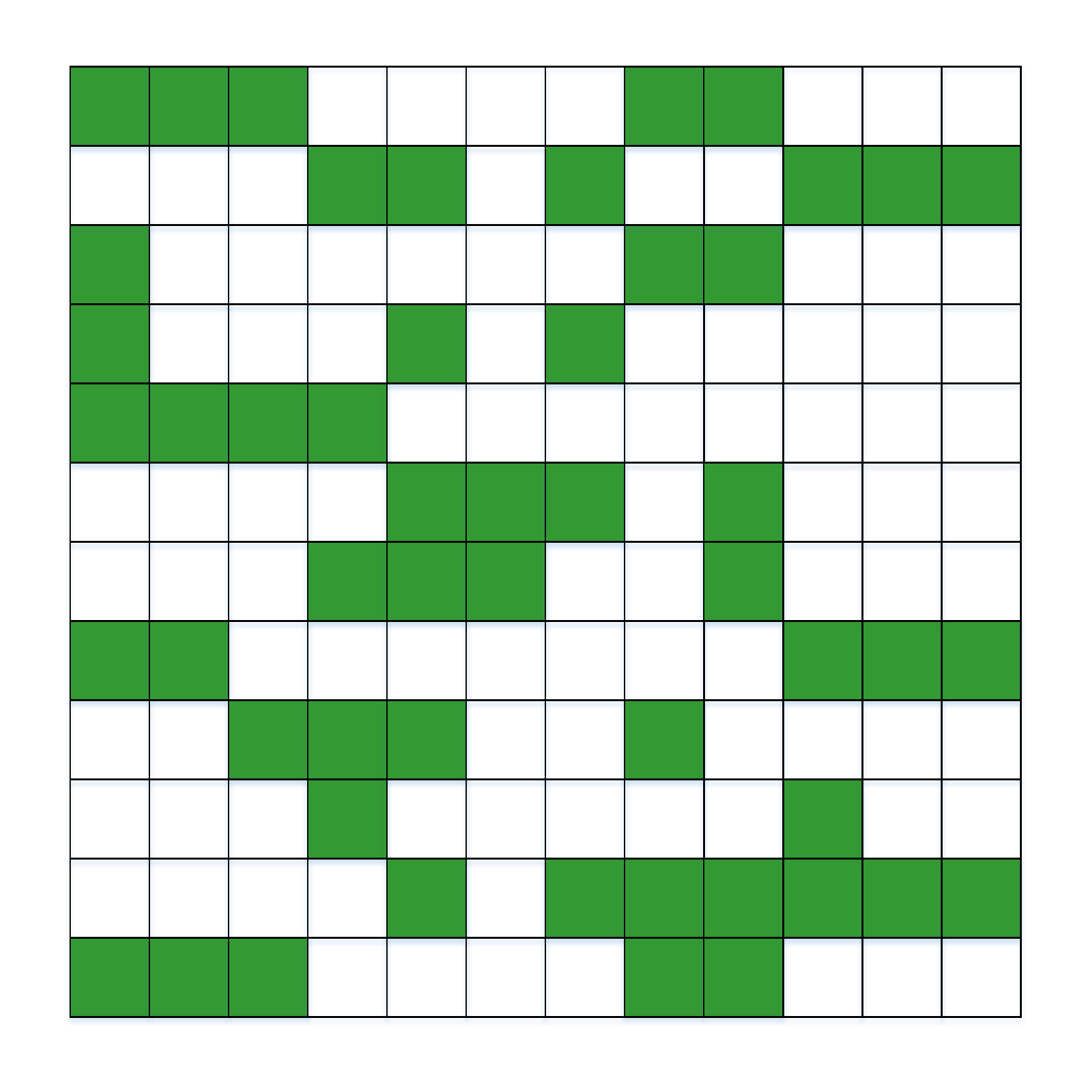}}
  \subfigure[]{\includegraphics[width=2.5cm]{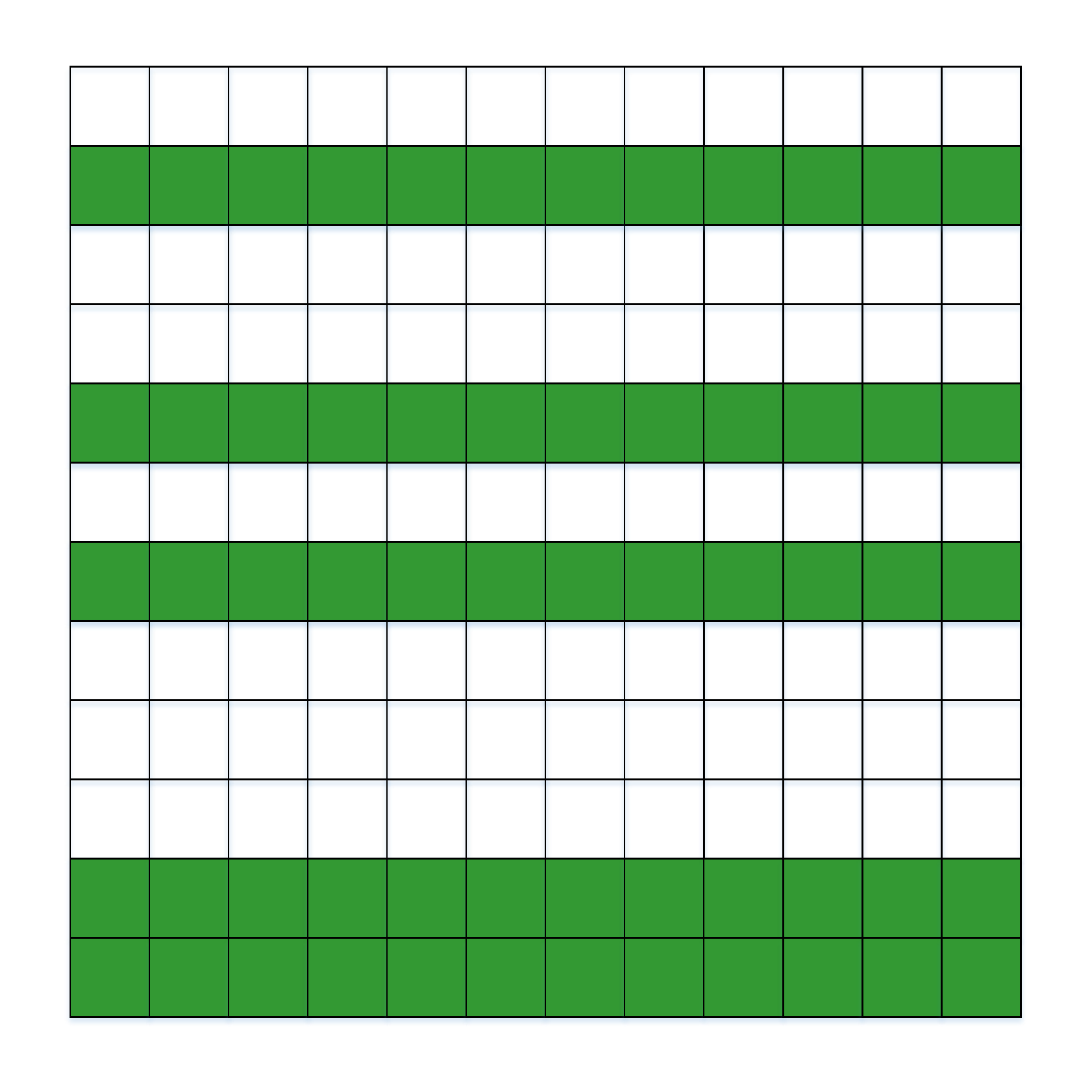}}
  \subfigure[]{\includegraphics[width=2.62cm]{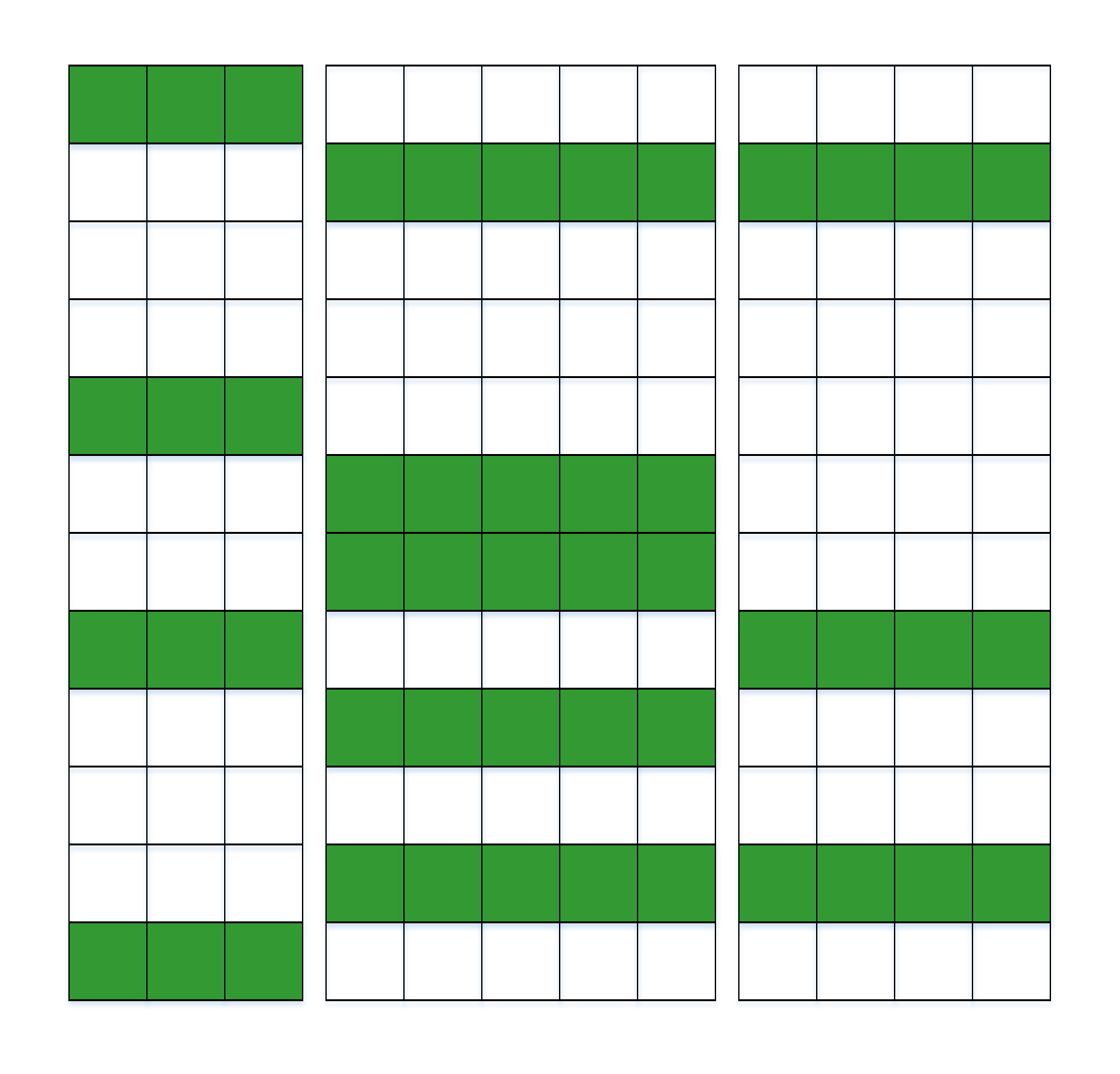}}
  }\\
  \caption{Illustration of (a) sparsity, (b) collaborative sparsity, and (c) group sparsity.}
\label{fig:2}
\end{figure}

Sparsity regularizer is extensively exploited during the hyperspectral unmixing procedure since the distribution of each endmember is sparse in general. Inspired by \cite{XWang2017}, we discuss the works on imposing sparsity regularizer from sparsity, collaborative sparsity, and group sparsity perspectives. As shown in Fig. \ref{fig:2}, diverse effects can be promoted for the abundances, where zero coefficients are denoted by white squares.

\emph{1) Sparsity}

Jia \emph{et al.} \cite{SJia2009} proposed two algorithms (called PSnsNMF and PSNMFSC) to explicitly represent sparsity. In particular, PSnsNMF controls the sparsity using the parameter of smoothing matrix in nonsmooth NMF (nsNMF) \cite{APascualMontano2006}. Thus, it is difficult to balance the smoothness and sparsity. PSNMFSC enforces sparsity by setting both $L_1$-norm and $L_2$-norm. However, the algorithm needs to assign an exact sparsity level which cannot be known a priori, and sparsity levels of different endmembers also vary from each other. In order to encourage the sparsity of abundance matrix, $L$-norm is widely used in the field of unmixing. Along this line, $L_0$ regularizer counts the number of non-zero elements in the abundances to yield the sparsest results, while its optimization belongs to an NP-hard problem. Therefore, $L_q$ regularizer is attractive in real applications, expressed as
\begin{equation}
\label{eqn:10}
  J(\mathbf{S}) = \|\mathbf{S}\|_q=\sum_{m,p=1}^{M,P}(s_{mp})^q,
\end{equation}
where $s_{mp}$ represents the element of the matrix $\mathbf{S}$ in the $m$-th row and $p$-th column.

\begin{itemize}
  \item By setting $q=1$, $L_1$ regularizer is a popular alternative for achieving a sparse abundance matrix   \cite{MDIordache2011, YLiu2019}. Nevertheless, it is not compatible with the ASC during the unmixing procedure.
  \item $L_q (0<q<1)$ regularizer achieves sparser results compared with the $L_1$ counterpart. Especially, Qian \emph{et al.} \cite{YQian2011} showed that $q = 1/2$ was an optimal choice and proposed an $L_{1/2}$-NMF algorithm for unmixing the hyperspectral data. Due to the simplicity and effectiveness, the $L_{1/2}$ regularizer is widely applied to develop novel unmixing approaches (see the boldfaced methods in Table \ref{tab:1}). To solve the nonconvex optimization caused by the $L_{1/2}$ regularizer, a fast and efficient adaptive half-thresholding algorithm was proposed in \cite{WWang2015}. Considering that the mixed level of each pixel may be different from each other, a data-guided sparsity was provided to adaptively employ $L_q (0<q<1)$ constraint \cite{FZhu2014}. In detail, a data-guided map (DgMap) was firstly learnt by measuring the uniformity of neighboring pixels, thus obtaining adaptive value for $q$.
  \item The problem \eqref{eqn:15} with $q=2$ is often considered to improve the unmixing performance as well. For instance, Huang \emph{et al.} \cite{RHuang2017} proposed a data-guided constrained NMF (DGC-NMF) model by imposing sparsity with either $L_{1/2}$-norm or $L_2$-norm. Specifically, the sparsity of each pixel is measured first. Then, the $L_{1/2}$ regularizer is utilized to constrain the abundances of the pixels with a high sparsity level, while the $L_2$ regularizer is adopted for generating smooth results with a low sparsity level.
\end{itemize}

Nevertheless, an $L_q (0<q<1)$ regularizer brings challenges as well since it is non-continuous and non-differentiable. Accordingly, more efforts are reported to achieve better abundances. Specifically, by adopting higher-order norm, NMF with S-measure constraint (NMF-SMC) was constructed \cite{ZYang2011}. Since NMF-SMC is a gradient based algorithm, the convergence speed may be slow for large scale data set. In \cite{ZWu2014}, Wu \emph{et al.} defined the sparsity constraint for the abundances using the difference between $L_1$-norm and $L_2$-norm. Furthermore, $\arctan$ function \cite{YESalehani2017} was introduced to explore the sparse property of the abundance matrix.

In order to pursue sparser representation, He \emph{et al.} \cite{WHe2017} proposed to utilize a weighted sparse regularizer for abundances, expressed as
\begin{equation}
\label{eqn:11}
  J(\mathbf{S}) = \|\mathbf{W}\odot\mathbf{S}\|_1,
\end{equation}
where $\mathbf{W}$ is a weight matrix whose element at $(m,p)$ is calculated by $w_{mp}=1/(|s_{mp}|+\varepsilon)$, and $\varepsilon$ is a positive parameter. It encourages the sparsity of the abundance matrix from the column perspective (\emph{i.e.}, spectral domain). Similarly, a double reweighted $L_1$-norm regularizer is designed to further exploit the sparsity of abundances in spatial domains \cite{RWang2017Hyperspectral, SZhang2018, HCLi2021}. Moreover, to ensure the sparsity of the learned coefficients, a local coordinate constraint was imposed to develop a robust nonnegative local coordinate factorization (RNLCF) method along with a correntropy induced metric (CIM) \cite{TPeng2018}, where the weight was defined as $w_{mp}=\|\mathbf{a}_m-\mathbf{x}_p\|^2$. The generalization in terms of an $L_q$-norm was presented in \cite{SKhoshsokhan2019}. By further accounting for local spatial information, the $L_1$-norm was added to each non-overlapping subblock \cite{CGTsinos2017}.

Based on generalized morphological component analysis (GMCA), each $\mathbf{s}^m$ (the vector of $\mathbf{S}$ in $m$-th row, $m=1,\cdots,M$) can be modeled as a linear combination of multiple morphological components ($s_{mk}$) that can be sparsely represented by its associated dictionary basis, \emph{i.e.}, $\mathbf{s}^m\approx\sum_k s_{mk}=\sum_k \mathbf{D}_k\varphi_{mk}$ \cite{XXu2020Generalized}. A concatenated dictionary $\mathbf{D}$ acts as a discriminator between different morphological components, and $\varphi_{mk}$ is the sparse representation coefficient constrained by the $L_1$-norm. Furthermore, in order to simplify the solution, the sparse constraint $\|\varphi\|_1$ is transferred as
\begin{equation}
\label{eqn:12}
  J(\mathbf{S}) = \sum_{m=1}^M\|\mathbf{s}^m\|_1.
\end{equation}
Through GMCA, spatial information can be naturally considered in the unmixing process by exploiting the sparsity and morphological diversity of the abundance maps associated with each endmember.

Unlike imposing a sparsity regularizer on abundances directly, a transform domain based sparse regularizer was proposed in \cite{XXu2020Curvelet}, expressed as
\begin{equation}
\label{eqn:13}
  J(\mathbf{S}) = \|\mathbf{S}W^T\|_1,
\end{equation}
where $W$ is the curvelet transform basis. Very recently, correntropy-based adaptive sparsity constraint \cite{XLi2021} was imposed to abundances for each pixel. Besides, a generalized minimax concave (GMC) sparsity regularizer was embedded into NMF \cite{FXiong2020}, which is nonconvex and nonseparable, avoiding systematic underestimation of high components of sparse vector and producing more accurate sparse approximation.

\emph{2) Collaborative  sparsity}

As shown in Fig. \ref{fig:2}(b), collaborative sparsity enforces the row sparsity (joint-sparsity), whose formulation is
\begin{equation}
\label{eqn:14}
  J(\mathbf{S}) = \|\mathbf{S}\|_{2,1}=\sum_{m=1}^M\|\mathbf{s}^m\|_2.
\end{equation}
Since it is challenging to correctly identify the number of endmembers, an overestimation for it was studied in \cite{JLi2016} and a collaborative sparsity regularizer is imposed to remove the redundant endmembers. Through this regularizer, the sparsity among the endmembers is achieved simultaneously (collaboratively) for all pixels, \emph{i.e.}, collaborative sparsity of the abundance matrix. In addition, a weighted $L_{2,1}$-norm regularizer was applied to local similar abundances (\emph{i.e.}, blocks) to consider both sparsity and spatial information \cite{JHuang2019}.

\emph{3) Group sparsity}

As shown in Fig. \ref{fig:2}(c), by fully exploring the spatial group structure and sparsity of the HSIs, spatial group sparsity regularizer was proposed in \cite{XWang2017}, defined as
\begin{equation}
\label{eqn:15}
  J(\mathbf{S}) = \sum_{g=1}^G\sum_{\mathbf{s}_p\in\vartheta_g}c_p\|\mathbf{W}^g\mathbf{s}_p\|_2,
\end{equation}
where $\mathbf{s}_p (p=1,\cdots,P)$ denotes the vector of $\mathbf{S}$ in the $p$-th column, and $G$ is the number of the spatial groups (\emph{i.e.}, superpixels), which was generated by an improved simple linear iterative clustering (SLIC) algorithm. That is, the abundance matrix is divided into $G$ groups as $\mathbf{s}_r = (\bar{\mathbf{s}}^1,\cdots,\bar{\mathbf{s}}^G)\in\mathbb{R}^{M\times P}$ in which $\bar{\mathbf{s}}^g$ denotes spatial group $\vartheta_g$. $c_p=\frac{1} {D_p^g}$ ($D_p^g$ is spatial-spectral distance) is a pixel-wise confidence index for relaxing the group sparsity constraints of heterogeneous pixels, such as boundaries and small targets. $\mathbf{W}^g$ is updated iteratively to appropriately control the non-zero abundances of $\vartheta_g$. This regularizer is expected to promote the same sparse structure for pixels within a local spatial group. By further combining with nonlocal spatial information, Yang \emph{et al.} \cite{LYang2020} developed NLNMF to address the unmixing problem. Compared to the unmixing methods with smooth constraints, it is more reasonable and effective to utilize this local spatial groups with irregular shapes. Furthermore, group sparsity \cite{LDrumetz2019} was investigated in unmixing process that accounts for spectral variability through the use of group two-level mixed norm, \emph{i.e.}, $L_{G,p,q}=\|\mathbf{S}\|_{G,p,q}$. More concretely, the fractional LASSO with $L_{G,1,q}, 0<q<1$ aims to simultaneously enforce group and within-group sparsity.

As discussed above, the $L_q$ regularizers draw much attention to enforce the sparsity of abundances, especially for $L_1$-norm and $L_{1/2}$-norm. Compared with the $L_1$-norm, the $L_{1/2}$-norm \cite{YQian2011} is excellent to obtain a sparser solution, thus enhancing the unmixing performance. Nevertheless, it requires to be combined with other constraints (\emph{e.g.}, graph constraint \cite{XLu2013}) such that the intrinsic structures of HSIs can be considered. In addition, many approaches have been devoted to incorporate the spectral and spatial information into the $L_1$-norm. For example, the reweighted $L_1$-norm regularizer \cite{WHe2017} promotes the sparsity of the abundance matrix from spectral domain, and double reweighted $L_1$-norm regularizer \cite{HCLi2021} aims to further describe the sparsity in spatial domains. In this way, the improvement can be obtained greatly in the unmixing process. However, they are sensitive to noise corruption. Collaborative sparsity \cite{JLi2016} is helpful to induce the row sparsity, whereas it is incompetent to explore the spatial information. Thus, Huang \emph{et al.} \cite{JHuang2019} applied a weighted $L_{2,1}$-norm regularizer on blocks and imposed TV regularizer. Meanwhile, it may be more reasonable to encourage the collaborative sparsity when the endmember matrix is composed of spectral library or bundles. Moreover, the group sparsity regularizer considers sparsity at the group level by integrating the spatial group structure \cite{XWang2017, LYang2020}, but how to effectively group the abundances deserves more investigation.

\subsection{Smooth constraints}
Neighboring pixels are more likely to be constituted by the same materials. Accordingly, it is significant to investigate the spatial correlation among the neighboring pixels. Smooth constraints are often used to preserve the spatial-contextual information.

An adaptive potential function from discontinuity adaptive MRF model \cite{SJia2009} was adopted to promote the piecewise smoothness of abundances, given as
\begin{equation}
\begin{aligned}
\label{eqn:16}
  J(\mathbf{S}) = g(\mathbf{S}-\mathbf{S}_{\mathcal{N}}),
\end{aligned}
\end{equation}
where $\mathbf{S}_{\mathcal{N}}$ is the neighborhood matrix of $\mathbf{S}$. However, PSNMFSC cannot perform well for many real data sets and has high computational complexities \cite{XLiu2011}. Hence, Liu \emph{et al.} proposed abundance separation and smoothness constrained NMF (ASSNMF) in \cite{XLiu2011}. They defined the smoothness function as
\begin{equation}
\begin{aligned}
\label{eqn:17}
  J(\mathbf{S}) = \frac{1}{2}\times\frac{1}{8}\sum_{h=1}^8\|\tilde{\underline{\mathbf{S}}}^m_h-\underline{\mathbf{S}}^m\|,
\end{aligned}
\end{equation}
where $\tilde{\underline{\mathbf{S}}}^m_h$ is a weight matrix corresponding to $\underline{\mathbf{S}}^m\in\mathbb{R}^{r\times c}$, which is obtained via the reshape operation to the $m$-th row of $\mathbf{S}$. Compared with PSNMFSC where only adjacent elements are considered, ASSNMF is more effective by exploiting its surrounding elements, and is acquired by a linear transform with a low computational cost. Nevertheless, it is not always true that the smoothness levels of two-pixel pairs are the same even if the spatial distances between them are the same.

To describe spatial correlations, a weighted nonnegative matrix factorization (WNMF) algorithm was presented in \cite{JLiu2012}, and the designed weight is expressed as
\begin{equation}
\begin{aligned}
\label{eqn:18}
  J(\mathbf{S}) = \sum_{i=1}^N\sum_{j\in\mathcal{N}_{(i)}}w_{ij}\|\mathbf{s}_i-\mathbf{s}_j\|_2^2,
\end{aligned}
\end{equation}
where $w_{ij}$ is a weight to characterize how much the neighboring pixel $\mathbf{x}_j$ contributes to the considered pixel $\mathbf{x}_i$. This regularizer integrates both spectral and spatial information. Similarly, an adaptive local neighborhood weight constraint was designed to propose a double abundance characteristics constrained NMF (DAC$^2$NMF) \cite{RLiu2016} along with a separation constraint to prevent over-smoothness. In \cite{SKhoshsokhan2019}, the weight constraint with the $L_2$-norm was generalized to use the $L_q$-norm.

Recently, the abundance maps are assumed to be piecewise smooth. Therefore, the TV regularizer is accounted for capturing the piecewise smoothness structure of each abundance map to enhance the unmixing results. He \emph{et al.} \cite{WHe2017} first embedded the TV regularizer into the NMF framework, expressed as
\begin{equation}
\begin{aligned}
\label{eqn:19}
  J(\mathbf{S}) = \|\mathbf{S}\|_{\textrm{HTV}}= \sum_{m=1}^M \|\mathcal{F}\mathbf{s}^m\|_{\text{TV}},
\end{aligned}
\end{equation}
where $\mathcal{F}$ denotes the reshape operation from a vector with $P$ pixels to a matrix with the size of $c\times r$ and $\|\cdot\|_{\text{TV}}$ is the anisotropic TV norm. Together with collaborative sparsity and endmember constraint, \cite{YYuan2020} presented an improved collaborative NMF and TV algorithm (ICoNMF-TV) for the unmixing task. Similarly, multiple-priors ensemble constrained NMF (MPEC-NMF) \cite{KQu2020} integrated the MVC, reweighted $L_{1/2}$-norm, and TV regularizers. Besides, \cite{JYao2019} extended the piecewise smoothness to the non-local smoothness, developing a non-local TV and log-sum regularized NMF (NLTV-LSRNMF) method.

For the above methods, the neighborhood is determined by a set of pixels involved in a predefined regular shape, such as a cross or square window. Consequently, smooth constraints can be utilized to exploit spatial-contextual information adequately, while ignoring the edges and local spatial details. By comparison, it may be more reasonable to use irregular shapes adaptively.

\subsection{Manifold constraints}

The above methods exploit the Euclidean structure of the hyperspectral data space. Considering that the data are more likely to lie on a low-dimensional submanifold embedded in the high-dimensional ambient space \cite{DCai2011}, the intrinsic manifold structure receives increasing attention in the hyperspectral unmixing field. The graph regularizer can be expressed as
\begin{equation}
\label{eqn:20}
  J(\mathbf{S}) =\sum_{i,j=1}^P \|\mathbf{s}_i-\mathbf{s}_j\|^2\mathbf{W}_{ij}
                =\textrm{Tr}(\mathbf{SLS}^T),
\end{equation}
where $\textrm{Tr}(\cdot)$ denotes the trace of a matrix, $\mathbf{W}$ is an affinity matrix,  $\mathbf{L}=\mathbf{D}-\mathbf{W}$, and $\mathbf{D}$ is a diagonal matrix with its ($i,i$) element calculated by $\mathbf{D}_{ii}=\sum_j\mathbf{W}_{ij}$. Next, we introduce how to construct the affinity matrix.

\emph{1) Spectral similarity}

In \cite{XLu2013}, a manifold regularizer was incorporated into the sparsity constrained NMF, presenting graph-regularized $L_{1/2}$-NMF (GLNMF) for hyperspectral unmixing, where the affinity matrix of the nearest neighbor graph $\mathbf{W}\in \mathbb{R}^{N\times N}$ is built to model the local structural information, given as
\begin{equation}
\label{eqn:21}
  \mathbf{W}_{ij}=\exp(-\frac{\|\mathbf{x}_i-\mathbf{x}_j\|^2}{\sigma}),
\end{equation}
which is known as the heat kernel. Here, $\mathbf{x}_i$ is one of the $k$-nearest neighbors of $\mathbf{x}_j$. When $\mathbf{x}_i$ and $\mathbf{x}_j$ are similar, $\mathbf{W}_{ij}$ is relatively large, and $\sigma$ denotes the standard deviation. This regularizer was leveraged in \cite{LTong2016} along with partially known endmembers for unmixing. Inspired by the graph regularizer based on $L_2$-Laplacian, Rathnayake \emph{et al.} \cite{BRathnayake2020} developed an $L_1$-norm based graph, called GLR$_{l_1}$. Besides, a Hessian graph \cite{TPeng2020} was adopted for hyperspectral unmixing. Compared with Laplacian graph, it is more stable and further captures the relationship of the nonlinear mapping.

\emph{2) Spectral-spatial similarity}

By respectively defining spatial geometric distance and spectral geometric distance of these two pixels in local window, dual Laplacian manifold regularizer \cite{SYang2015} was established to exploit the geometric structure of the HSI. In order to encode the manifold structures embedded in the hyperspectral data space, a graph Laplacian is incorporated from spatial distance and feature distance perspectives simultaneously \cite{FZhu2014ISPRS}, where the weight matrix is obtained by using spectral angle distance (SAD) metric and two conditions are adopted to determine the neighbors: one is the nearest spatial distance via a local window, and the other is the nearest feature distance via the SAD similarities in the local window. In this way, the edges can be preserved by avoiding the graph across dissimilar pixels. Likewise, a bilateral filter regularizer based on graph theory was adopted to utilize the correlation information in both spatial and spectral spaces \cite{ZZhang2018}. Moreover, GLrNMF \cite{MWang2018} defined a spatial-spectral semantic weight based on the intersection (denoted by $QG$) of spatial neighbor (denoted by $Q$) and spectral neighbor (denoted by $G$).
Motivated by hypergraph learning, the spectral-spatial joint structure was modeled by a hypergraph to capture the high-order relation among multiple pixels \cite{WWang2016}.

\emph{3) Region-based similarity}

Considering that the spectra are similar in the same region while different between regions, Tong \emph{et al.} \cite{LTong2017} presented a region-based structure preserving NMF (R-NMF) to explore consistent data distribution in the same region, aiming at discriminating different data structures across regions. Furthermore, in view of the difference in spatial structure, an HSI was partitioned into homogeneous region and detail region based on a sketch map \cite{JZhang2020}, in which the manifold constraint and the $L_{1/2}$ regularizer were employed for the homogeneous region, while the $L_1$ regularizer for detail region.

\emph{4) Other weight learning}

Different from the above perspectives to establish a graph regularizer, \cite{GZhang2021} was based on the local linear embedding assumption, where the weight matrix $\mathbf{W}$ is learned by minimizing the following equation:
\begin{equation}
\label{eqn:22}
  \mathbf{W}_{ij}=\arg \min\|\mathbf{x}_i-\sum_{\mathbf{x}_j\in\mathcal{N}(\mathbf{x}_i)}\mathbf{W}_{ij}\mathbf{x}_j\|_2,
\end{equation}
to exploit the local geometric structure. In addition, the clustering algorithm is also beneficial to characterizing the structure information of the HSIs. In \cite{XLu2020}, subspace clustering was applied to capture the latent characteristic structure, for which the weight matrix $\mathbf{W}$ is constructed by
\begin{equation}
\label{eqn:23}
  \mathbf{W}=\frac{\mathbf{H}+\mathbf{H}^T}{2},
\end{equation}
to form a graph regularizer. Here, $\mathbf{H}$ is the subspace structure matrix that is learned by minimizing  $\|\mathbf{X}-\mathbf{X}\mathbf{H}\|^2_F,~\text{s.t.}~ \text{diag}(\mathbf{H}) = 0$. It is noteworthy that only the largest $k$ values are remained for each column of $\mathbf{H}$.

\subsection{Low-rank constraints}

The high spatial correlation of HSIs can be also translated into the low rank of the involved abundance matrices.

Let $G$ denote the number of non-overlapping subimages from the input HSI. By enforcing simultaneously the local low-rank and sparse structures to the abundance matrix $\mathbf{S}_G$, DSPLR-NMF \cite{CGTsinos2017} was reported for unmixing, in which, the local low-rank constraint is given as
\begin{equation}
\label{eqn:24}
  J(\mathbf{S}) = \sum_{g=1}^G\|\mathbf{S}_g\|_*,
\end{equation}
where $\|\cdot\|_*$ denotes the nuclear norm. In order to naturally incorporate spatial priors, superpixels were generated by employing SLIC algorithm to the HSI and constrained using the low-rank penalty \cite{MWang2018}. For the learning of subspace structure \cite{LZhou2020}, the low-rank constraint was utilized to construct a self-representation matrix.

\subsection{Abundance separation constraints}

Taking the spatial correlation into consideration, the unmixing results can be improved and robust to noise generally. However, the over-smoothing problem may occur since dissimilar pixels are often ignored in a local window. Therefore, an abundance separation constraint \cite{XLiu2011} was imposed based on the KL divergence to minimize the correlation between different endmembers. Likewise, Liu \emph{et al.} \cite{RLiu2016} introduced the separation constraint to preserve the inner diversity of the same type of materials. Moreover, in \cite{YJia2020}, a dissimilarity regularizer constructed by label information was incorporated into the NMF.

As discussed above, numerous works are devoted to obtaining better spatial structure from different insights, such as sparsity, spatial-contextual information, low-dimensional manifold structure, low-rank structure, and global spatial information. Among these constraints, it has considerable potential to make use of the spatial and spectral information simultaneously for HSIs applications.

\subsection{Other spatial constraints}
To maintain the structural information, clustering has contributed to the regularization term in \cite{XLu2014, BDu2016}. A subspace structure learned from the HSIs was introduced to form a subspace regularizer $\|\mathbf{S}-\mathbf{S}\mathbf{H}\|^2_F + \tau \|\mathbf{H}\|_*$, thereby capturing the global correlation of all pixels \cite{LZhou2020}. In view of the similarities between the substances, Yuan \emph{et al.} \cite{YYuan2015TGRS} introduced a substance dependence constraint. To be more specific, the substance dependence is considered in the whole space to describe the global spatial information, which is reflected by the abundance. In \cite{LDong2021}, spectral-spatial joint sparse NMF (S$^2$-NMF) was proposed by combining the global spatial information and local spectral information simultaneously. Based on the label information, Jia \emph{et al.} \cite{YJia2020} developed a similarity regularizer  to compensate the dissimilarity regularizer.

\section{Structured NMF}

\subsection{Weighted NMF}

NMF-based methods achieve the unmixing by utilizing the statistical properties of HSIs. Thus, it is closely related to the number of samples. However, there is a large difference for the number of pixels concerning different endmembers in many scenarios. Therefore, a cluster-wise weighted NMF (CW-NMF) \cite{XLv2021} method was provided to explore the information of imbalanced pixels. In particular, a weight matrix based on the result of clustering is integrated into the NMF, expressed as
\begin{equation}
\label{eqn:25}
  \min_{\mathbf{A},\mathbf{S}}~ \|(\mathbf{X}-\mathbf{AS})\mathbf{W}\|_F^2,
\end{equation}
where $\mathbf{W}$ is the diagonal weight matrix. Then, the effects of the pixels involving imbalanced endmembers can be enhanced by giving larger weight values, while the effects of the pixels that involving majority endmembers are reduced by assigning smaller weight values. Similarly, self-paced NMF (SpNMF) \cite{JPeng2021} was presented by adopting weighted least-squares losses, given as
\begin{equation}
\label{eqn:26}
  \min_{\mathbf{A},\mathbf{S}}~\sum_{b=1}^B\Big{\{}w_b\|\mathbf{x}^b-(\mathbf{AS})^b\|_2^2+h(\gamma, w_b)\Big{\}},
\end{equation}
where $h$ is a self-paced function associated with ``model age'' parameter $\gamma$ to learn the weights adaptively. There are many self-paced functions such as binary, linear, logarithmic, and mixture functions. Besides, the SpNMF can be also extended for pixel weighting (\emph{i.e.}, $\min_{\mathbf{A},\mathbf{S}}~\sum_{p=1}^P\big{\{}w_p\|\mathbf{x}_p-(\mathbf{AS})_p\|_2^2+h(\gamma, w_p)\big{\}}$) and element weighting (\emph{i.e.}, $\min_{\mathbf{A},\mathbf{S}}~\sum_{b,p=1}^{B,P}\big{\{}w_{bp}[x_{bp}-(\mathbf{AS})_{bp}]^2+h(\gamma, w_{bp})\big{\}}$). Meanwhile, the weighted NMF is also effective to improve the robustness of NMF.

\subsection{Projection-based NMF}

Inspired by the sparse regression-based unmixing methods, a projection-based NMF (PNMF) algorithm was proposed by introducing a spectral library into the NMF framework \cite{YYuan2015JSTARS}. In detail, related spectra in the library $\mathbf{U}$ are projected onto a subspace based on a transformation matrix $\mathbf{V}$ to obtain the projected endmembers, \emph{i.e.}, $\mathbf{A}=\mathbf{UV}$, the cost function is
\begin{equation}
\label{eqn:27}
  \min_{\mathbf{V},\mathbf{S}}~\|\mathbf{X}-\mathbf{UVS}\|_F^2.
\end{equation}

In this way, the endmembers are not only adaptively generated from the spectral library, but also related to the HSIs. Meanwhile, the number of endmembers, which is a difficulty in sparse regression, is controlled by the dimension of the subspace.

\subsection{Convex NMF}

In NMF, some of the computed endmembers in $\mathbf{A}$ may be artificial, which do not belong to any real material in the scene, but exist only in the solution space of the problem. To keep the association between extracted and real endmembers, each spectrum $\mathbf{a}_m$ was assumed to be nonnegative linear combinations of the observed pixels \cite{NAkhtar2017}, \emph{i.e.}, $\mathbf{A}=\mathbf{X\Xi}$. Thus, the cost function is
\begin{equation}
\label{eqn:28}
  \min_{\mathbf{\Xi},\mathbf{S}}~\|\mathbf{X}-\mathbf{X\Xi}\mathbf{S}\|_F^2.
\end{equation}
Along with subspace clustering, Lu \emph{et al.} \cite{XLu2020} also proposed to extract endmembers by linearly combining of all pixels in a spectral subspace, avoiding the generation of artificial endmembers. Under this assumption, the hierarchical sparse NMF (HSNMF) \cite{TUezato2020} introduced hierarchical sparsity constraints to accommodate endmember variability.

\subsection{Robust NMF}

NMF is applied to the unmixing under the assumption of Gaussian noise. However, HSIs are inevitably contaminated by different types of noise, \emph{e.g.}, Gaussian noise, impulse noise, stripes, and deadlines. Hence, the classical NMF model defined by the least-squares loss is sensitive to noise, leading to dramatically degrading the unmixing performance. To improve the robustness of NMF, many models have been reported based on certain metrics, including but not limited to bounded Itakura-Saito (IS) divergence \cite{CLi2015}, $L_{2,1}$-norm regularizer \cite{CFevotte2015, WHe2016Sep, RHuang2019,GZhang2021}, CIM \cite{YWang2015, XLi2021, TPeng2018, XRFeng2020}, Cauchy function \cite{HWang2019}, and general robust loss function \cite{JPeng2020}. The bounded IS divergence was employed to address the additive, multiplicative, and mixed noises in HSIs \cite{CLi2015}.

The $L_{2,1}$-norm regularizer proposed by Ding \emph{et al.} \cite{CDing2006} can effectively handle noise and outliers. Based on this regularizer, different robust models are designed to reduce the impact of noise. In \cite{CFevotte2015}, robust NMF (rNMF) was proposed by introducing a residual term $\mathbf{E}\in \mathbb{R}^{B \times P}$ accounting for outliers (\emph{i.e}., nonlinear effects), whose cost function is written as
\begin{equation}
\label{eqn:29}
  \min_{\mathbf{A},\mathbf{S},\mathbf{E}}~ D\left(\mathbf{X}|(\mathbf{AS}+\mathbf{E})\right)+\lambda\|\mathbf{E}\|_{2,1},
\end{equation}
where $D\left(\mathbf{X}|(\mathbf{AS}+\mathbf{E})\right)$ measures dissimilarity between $\mathbf{X}$ and $(\mathbf{AS}+\mathbf{E})$, $\lVert\mathbf{E}\rVert_{2,1} =\sum_{p=1}^P\lVert\mathbf{e}_p\rVert_2$, and $\mathbf{e}_p$ denotes the $p$-th vector of $\mathbf{E}$. As such, the problem \eqref{eqn:29} addresses the nonlinearities and improves the robustness against the noisy pixels. Based on the $L_{1,2}$-norm regularizer, He \emph{et al.} \cite{WHe2016Sep} used specific bands of the HSIs and modeled the sparse noise explicitly for significantly improving the robustness of NMF. The $L_{2,1}$-norm was employed to replace Euclidean distance or KL divergence directly in \cite{GZhang2021}, and the spectral-spatial constrained NMF (SS-NMF) model was developed to cope with non-Gaussian noises or outliers, whose reconstruction error is calculated by
\begin{equation}
\label{eqn:30}
  \min_{\mathbf{A},\mathbf{S}}~\|\mathbf{X}-\mathbf{A}\mathbf{S}\|_{2,1}.
\end{equation}
In this way, the model is robust to noisy pixels by columns. To further achieve robustness to band noise by rows, the $L_{1,2}$-norm was also incorporated to form a spectral-spatial robust NMF model for hyperspectral unmixing \cite{RHuang2019}.

Correntropy  is a nonlinear similarity measure, which is based on  Gaussian kernel
\begin{equation}
\label{eqn:31}
  \kappa(x) = \frac{1}{2\pi\sigma}\exp\left(\frac{-x^2}{2\sigma^2}\right).
\end{equation}
Recently, CIM is employed to replace the least-squares loss to develop some robust models. For example, a correntropy based NMF (CENMF) \cite{YWang2015} was proposed to suppress the influence of noisy bands efficiently. Considering the diversity of the noise levels of pixels, correntropy-based spatial-spectral robust sparsity-regularized NMF (CSsRS-NMF) was proposed in \cite{XLi2021} by adaptive assigning weights to noisy pixels. Furthermore, robustness can be achieved from an element-wise noise perspective \cite{TPeng2018, XRFeng2020}.

By cutting off the large error via the truncation operation, the truncated Cauchy loss \cite{NGuan2019} exhibits robustness to outliers. Accordingly, the reconstruction error was measured via the truncated Cauchy function \cite{HWang2019}, which is expressed as
\begin{equation}
\label{eqn:32}
  h(x) = \left\{
                    \begin{array}{ll}
                      \ln\left(1+\frac{x^2}{\gamma^2}\right), & |x|\leq\epsilon, \\
                      \ln\left(1+\frac{\epsilon^2}{\gamma^2}\right), & |x|\geq\epsilon,
                    \end{array}
                  \right.
\end{equation}
where $\gamma$ and $\epsilon$ are the scale parameter and truncated coefficient, respectively.

Furthermore, a general loss-based NMF (GLNMF) model \cite{JPeng2020} was developed by introducing a general robust loss function defined in \cite{JTBarron2017}, given as
\begin{equation}
\label{eqn:33}
  f(x) = \frac{|2-\upsilon|}{\upsilon}\left(\left(\frac{(x/c)^2}{|2-\upsilon|}+1\right)^{\upsilon/2}-1\right),
\end{equation}
to downplay the large noise. It is a generalization of the $L_2$ loss ($\upsilon\rightarrow 2$), Cauchy loss ($\upsilon\rightarrow 0$), and Welsch loss ($\upsilon\rightarrow -\infty$), etc.

Through the $L_{2,1}$-norm, the model can effectively address the pixels with noise or outliers. By contrast, CIM can be applied flexibly to relieve the impact of noise from bands, pixels, and elements insights. Meanwhile, the algorithms with CIM are effective to process non-Gaussian and impulsive noise. The truncated Cauchy loss can suppress the outliers effectively.

\section{Generalized NMF}
\subsection{Linear-quadratic NMF}

Combined with the linear-quadratic model, NMF can be extended as $\mathbf{X}=\mathbf{AS}=\mathbf{A}_a\mathbf{S}_a+\mathbf{A}_b\mathbf{S}_b$. Subsequently, the following cost function requires to be minimized:
\begin{equation}
\label{eqn:34}
  \min_{\mathbf{A}_a,\mathbf{A}_b,\mathbf{S}_a,\mathbf{S}_b}~ \|\mathbf{X}-\mathbf{A}_a\mathbf{S}_a-\mathbf{A}_b\mathbf{S}_b\|_F^2.
\end{equation}

Under this framework, the unmixing task can be addressed from the following two perspectives.
\begin{itemize}
  \item First, Tang \emph{et al.} \cite{WTang2012} proposed an unmixing method by using the prior knowledge of some signatures in the scene. To be specific, $\mathbf{A}_a$ and $\mathbf{A}_b$ represent the matrices of known and unknown endmembers with related abundance fractions $\mathbf{S}_a$ and $\mathbf{S}_b$, respectively.
  \item Then, the second-order scattering of light can be considered to improve the performance \cite{IMeganem2014, OEches2014, NYokoya2014, XZhang2018}. In this case, $\mathbf{A}_a$ and $\mathbf{A}_b$ are the endmember matrix and bilinear endmember matrix, while $\mathbf{S}_a$ and $\mathbf{S}_b$ are abundance matrix and interaction abundance matrix, respectively. Among them, semi-nonnegative matrix factorization (semi-NMF) was used for the optimization to process an entire image in the matrix form \cite{NYokoya2014, XZhang2018}.
\end{itemize}

\subsection{Kernelized NMF}
Kernel methods can be introduced for nonlinear hyperspectral unmixing without estimating the nonlinear mixture model. In \cite{XLi2014}, a constrained kernel NMF (CKNMF) was proposed for dealing with nonlinearities. Through kernel mappings (denoted by $\phi$), the observed matrix $\mathbf{X}$ and endmember matrix $\mathbf{A}$ are transformed to high-dimensional feature space, obtaining $\phi(\mathbf{X})=[\phi(\mathbf{x}_1),\cdots,\phi(\mathbf{x}_N)]$ and $\phi(\mathbf{A})=[\phi(\mathbf{a}_1),\cdots,\phi(\mathbf{a}_M)]$. As a result, the data are linearly separable in the feature space. The cost function is given as
\begin{equation}
\label{eqn:35}
  \min_{\mathbf{A},\mathbf{S}}~\|\phi(\mathbf{X})-\phi(\mathbf{A})\mathbf{S}\|_F^2,
\end{equation}
where $\phi$ is a nonlinear function. Generally, the Gaussian kernel $ k(\mathbf{x}_i,\mathbf{x}_j)=\langle\phi(\mathbf{x}_i)\cdot\phi(\mathbf{x}_j)\rangle=\exp(-\frac{\|\mathbf{x}_i-\mathbf{x}_j\|^2}{2\sigma^2})$ is utilized to achieve dot product operator in a high-dimensional kernel feature space. Taking into account both the input and feature spaces, a biobjective NMF \cite{FZhu2016} was formulated by combining the linear and kernel-based models. However, the kernel-based methods suffer from computation burden. For dealing with large-scale and streaming dynamic data, Zhu \emph{et al.} \cite{FZhu2017} proposed an online KNMF (OKNMF) framework to control the computational complexity via adopting the stochastic gradient descent (SGD), mini-batch, and averaged SGD (ASGD) strategies. In addition, the KNMF was extended to incremental KNMF (IKNMF) and improved IKNMF (IIKNMF) for desired unmixing accuracy and efficiency \cite{RHuang2018}.

\subsection{Transform-domain based NMF}
Due to the dense spectra (typically $10$ nm) and overlapped information, HSIs are compressible in a suitable transformed domain. Very recently, wavelet transform has been exploited to express the hyperspectral data compactly \cite{VijayashekharSS2021}. Specifically, biorthogonal discrete wavelet transform was employed to represent the hyperspectral data in the wavelet domain, denoted as $\mathbf{x}_{p\textrm{w}}$. Accordingly, the LMM in the wavelet domain can be written as $\mathbf{X}_\textrm{w}=\mathbf{A}_\textrm{w}\mathbf{S}_\textrm{w}+\mathbf{N}_\textrm{w}$. Hence, the cost function is
\begin{equation}
\label{eqn:36}
  \min_{\mathbf{A}_\textrm{w},\mathbf{S}_\textrm{w}}~\|\mathbf{X}_\textrm{w}-\mathbf{A}_\textrm{w}\mathbf{S}_\textrm{w}\|_F^2.
\end{equation}
On this basis, three prior terms (\emph{i.e.}, volume regularizer, spatial smoothness prior, and sparseness constraint) in the wavelet domain are integrated to better handle the ill-posedness.
Similarly, a wavelet-based approach was proposed for estimating abundances in \cite{VijayashekharSS2019}.
Furthermore, considering that the curvelet is capable of characterizing anisotropic singularity such as curves or edges in the image, Xu \emph{et al.} \cite{XXu2020Curvelet} adopted fast discrete curvelet transform to impose a sparse regularizer on the transformed domain of abundances, thereby enhancing both sparsity and diversity.

\subsection{Nonnegative tensor factorization (NTF)}
A third-order tensor, which is the high-dimensional extension of a matrix, can be used to represent the HSI for preserving the intrinsic structure information. Accordingly, nonnegative tensor factorization (NTF) has been successfully applied to HSIs processing such as denoising \cite{XGuo2013} and classification \cite{ZZhong2015}. In \cite{MAVeganzones2016}, the NTF method was used to the unmixing task by using canonical polyadic decomposition (CPD). However, a limitation is the lack of an explicit link with LMM. In \cite{YQian2017}, a matrix-vector NTF (MV-NTF) unmixing method was proposed based on block term decomposition (BTD). The MV-NTF factorizes the HSI tensor into a sum of several component tensors as
\begin{equation}
\label{eqn:37}
  \mathcal{X}=\sum_{m=1}^M\mathbf{S}_m\circ\mathbf{a}_m+\mathcal{G}=\sum_{m=1}^M(\mathbf{C}_m\mathbf{B}_m^T)\circ\mathbf{a}_m+\mathcal{G},
\end{equation}
where $\mathcal{X}\in\mathbb{R}^{r\times c\times B}$ is a third-order HSI tensor, $\mathbf{a}_m$ is regarded as an endmember, $\mathbf{S}_m$ is the corresponding abundances represented by the product of two low-rank matrices $\mathbf{C}_m\in\mathbb{R}^{r\times R}$ and $\mathbf{B}_m^T\in\mathbb{R}^{R\times c}$, and $\mathcal{G}$ denotes the noise term. $R$ is related to the rank of abundance matrix and $\circ$ denotes the outer product. Apparently, this model constructs a straightforward link between LMM and tensor factorization. The cost function is expressed as
\begin{equation}
\label{eqn:38}
  \min_{\mathbf{A},\mathbf{B},\mathbf{C}}~\|\mathcal{X}-\sum_{m=1}^M(\mathbf{C}_m\mathbf{B}_m^T)\circ\mathbf{a}_m\|_F^2.
\end{equation}

Although intrinsic structure information is preserved, the local spatial information is not fully exploited due to the strict rank constraint. Subsequently, under this framework, Xiong \emph{et al.} further incorporated the similarity graph regularizer \cite{FXiong2018} and the TV regularizer \cite{FXiong2019} so as to describe the local spatial information. Likewise, three constraints were embedded into MV-NTF, including sparsity, minimum volume, and nonlinearity in \cite{BFeng2019}. In addition, some unmixing methods were also presented by combining additional constraints \cite{HCLi2021, JJWang2021, PZheng2021}. Considering that NMF characterizes more local spatial details through dealing with HSI at the pixel level, MV-NTF and NMF were coupled to make full use of their individual advantages \cite{HCLi2020}. In \cite{TImbiriba2018}, a low-rank tensor regularization was introduced during the learning process, allowing flexibility to the rank of the estimated abundance tensor. Endmember variability was considered based on the $4$D endmember tensor that was constrained by a new low-rank regularization\cite{TImbiriba2020, LSun2022}. Besides, a nonlocal Tucker decomposition method \cite{KWang2020} was provided to exploit the spectral-spatial correlations and the nonlocal self-similarity.

\subsection{Multilayer/Deep extensions}
\begin{figure*}[!t]
\centering
  \mbox{
  \subfigure[]{\includegraphics[width=4.2cm]{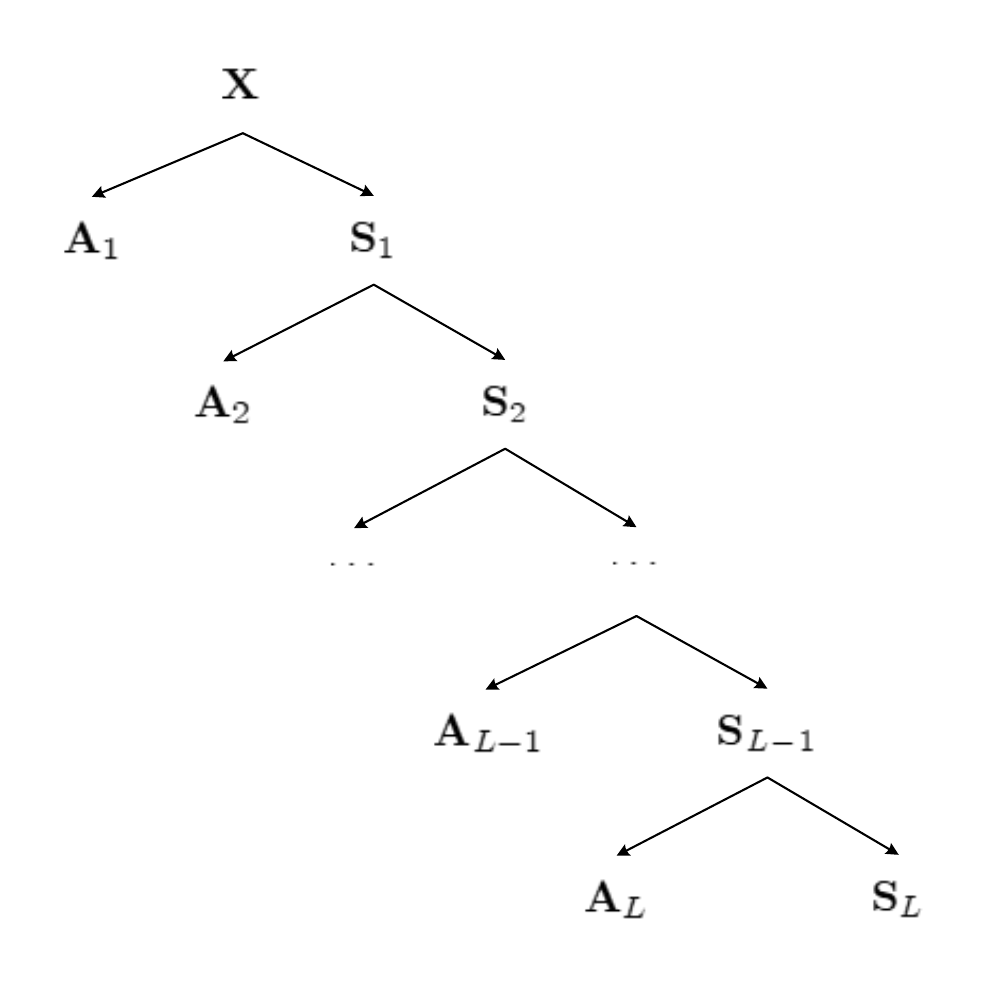}}
  \subfigure[]{\includegraphics[width=4.6cm]{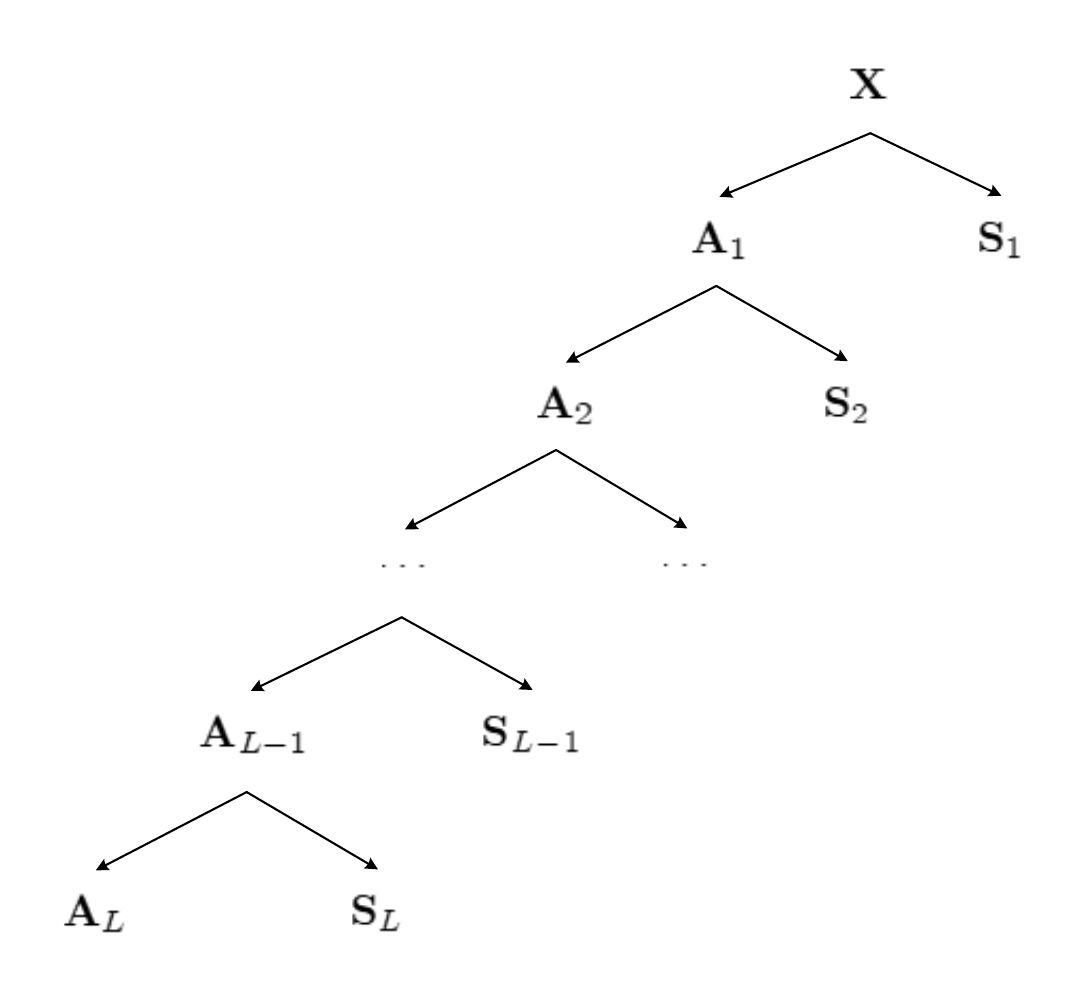}}
  \subfigure[]{\includegraphics[width=4.1cm]{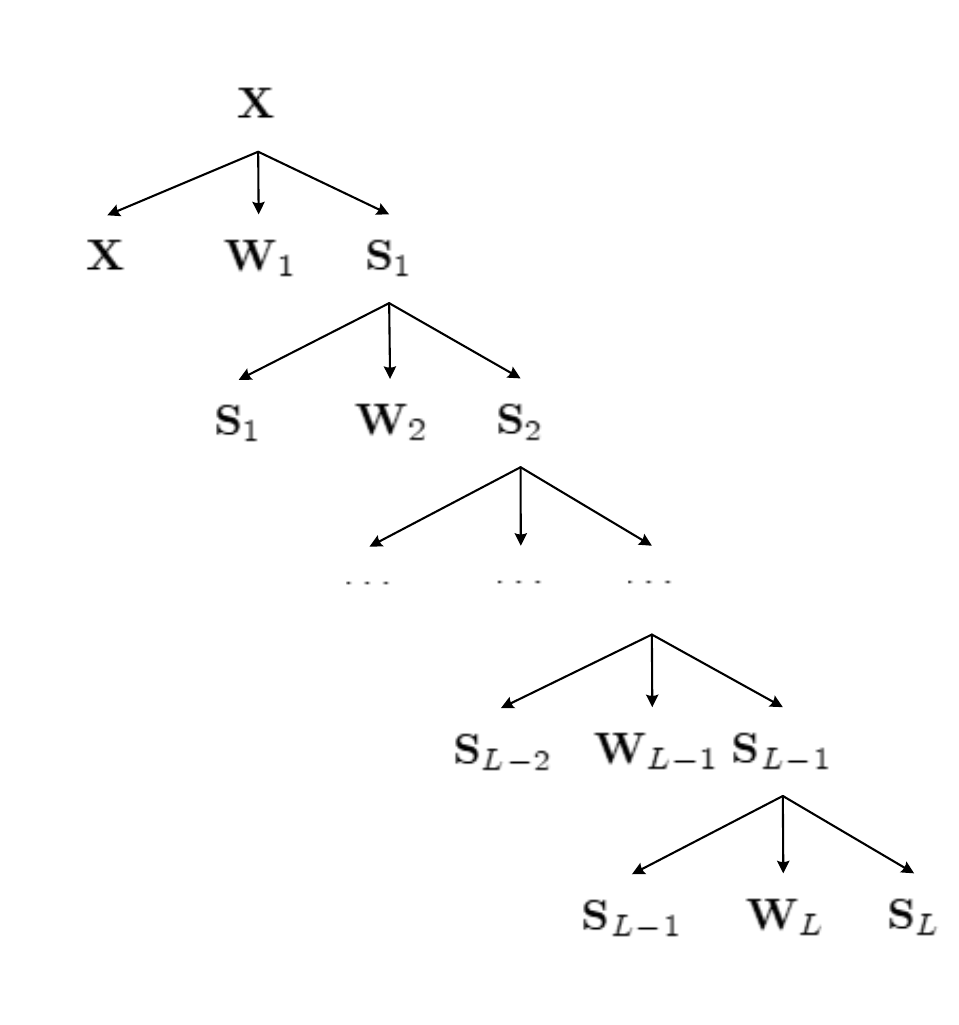}}
  \subfigure[]{\includegraphics[width=4.3cm]{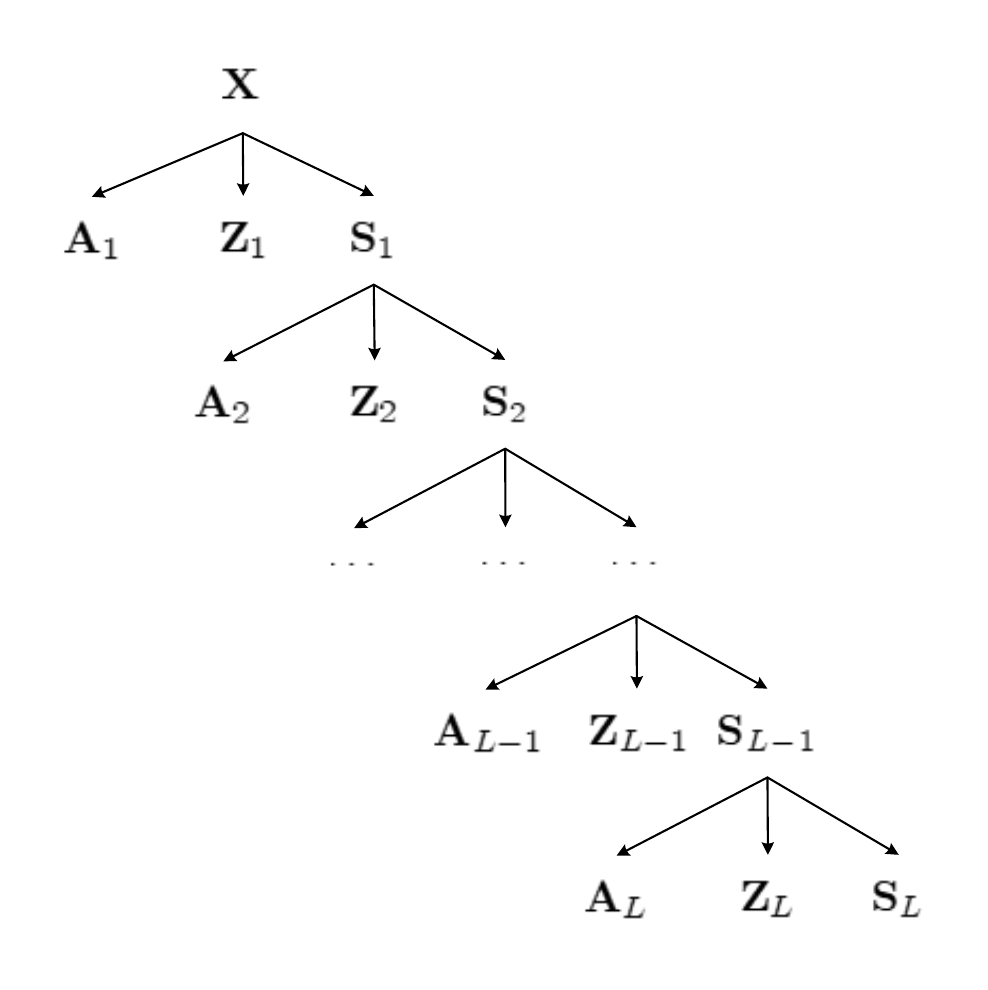}}
  }\\
  \caption{Architecture of decomposition for (a) Multilayer NMF by factorizing the coefficient matrix , (b) Multilayer NMF by factorizing the basis matrix, (c) Multilayer convex NMF, and (d) Multilayer nsNMF.}
\label{fig:3}
\end{figure*}

The aforementioned approaches explore the information in a single-layer manner, which do not allow for hierarchical refinement of the obtained endmembers and abundances. In order to extract hierarchical features with hidden information, DL \cite{GEHinton2006} has achieved commendable success in pattern recognition \cite{LZhang2016, BQian2016, GTrigeorgis2017}. Hence, Rajabi and Ghassemian \cite{RRajabi2015} unfolded NMF into multilayer architecture (\emph{i.e.}, multiple basis matrices and one abundance matrix), and proposed the multilayer NMF (MLNMF) model for hyperspectral unmixing. As shown in Fig. \ref{fig:3} (a), in the first layer, the matrix $\mathbf{X}$ can be factorized into $\mathbf{A}_1$ and $\mathbf{S}_1$. Then, in the next layer, $\mathbf{S}_1$ is further factorized into $\mathbf{A}_2$ and $\mathbf{S}_2$. A similar process is continued until the factorization of the $L$-th layer is completed. Here $L$ represents the maximum number of layers. Accordingly, the observation matrix is decomposed into $L+1$ nonnegative factors, \emph{i.e.}, $\mathbf{X}=\mathbf{A}_1\mathbf{A}_2\cdots\mathbf{A}_L\mathbf{S}_L$. The latent factors can be obtained by minimizing the following cost function:
\begin{equation}
\label{eqn:39}
  \min_{\mathbf{A}_l\mathbf{S}_l}~\|\mathbf{S}_{l-1}-\mathbf{A}_l\mathbf{S}_l\|_F^2,
\end{equation}
with $l=1,\cdots,L$, and $\mathbf{S}_0=\mathbf{X}$ when $l=1$. Then, the endmember and abundance matrices are $\mathbf{A}=\mathbf{A}_1\mathbf{A}_2\cdots\mathbf{A}_L$ and $\mathbf{S}=\mathbf{S}_L$, respectively. Based on this multilayer architecture, a double-constrained multilayer NMF (DCMLNMF) \cite{HFang2019} method was proposed to jointly explore the sparsity and the geometric structure. Besides, Tong \emph{et al}. \cite{LTong2019} developed novel unmixing approaches by further imposing constraints, such as a spectral and spatial total variation regularizer, an adaptive graph regularizer \cite{LTong2020}, and a homogeneous region regularizer \cite{LTong2020Homogeneous}. Moreover, the classical MLNMF was restructured and improved by integrating the Hoyer's projector in \cite{YYuan2021}. Different from factorizing $\mathbf{S}$ mentioned above, $\mathbf{A}$ is decomposed in \cite{LChen2017} to form constrained multilayer NMF (CMLNMF) along with MVC and sparsity constraints, shown in Fig. \ref{fig:3} (b). Thus, $\mathbf{X}=\mathbf{A}_L\mathbf{S}_L\cdots\mathbf{S}_2\mathbf{S}_1$. Here, the endmember and abundance matrices are $\mathbf{A}=\mathbf{A}_L$ and $\mathbf{S}=\mathbf{S}_L\cdots\mathbf{S}_2\mathbf{S}_1$, respectively. Furthermore, multilayer factorization was investigated with fast kernel archetypal analysis (KAA) \cite{GZhao2016} and kernel NMF \cite{JLiu2021} for unmixing.

However, these models are optimized by only minimizing the cost function of each layer, which fail to reduce the total reconstruction error. Trigeorgis \emph{et al}. \cite{GTrigeorgis2017} formulated deep Semi-NMF, achieving a significant breakthrough. Motivated by this, a deep NMF structure \cite{XRFeng2018} was proposed to address the unmixing task, whose cost function is
\begin{equation}
\label{eqn:40}
  \min_{\mathbf{A}_1,\mathbf{A}_2,\cdots,\mathbf{A}_L,\mathbf{S}_L}~\|\mathbf{X}-\mathbf{A}_1\mathbf{A}_2\cdots\mathbf{A}_L\mathbf{S}_L\|_F^2,
\end{equation}
where $\mathbf{A}=\mathbf{A}_1\mathbf{A}_2\cdots\mathbf{A}_L$ and $\mathbf{S}=\mathbf{S}_L$ denote the endmembers and abundances, respectively. The proposed model consists of \emph{pretraining stage} and \emph{fine-tuning stage}, where the former pretrains all factors layer by layer and the latter is used to reduce the total reconstruction error. Likewise, a sparsity-constrained deep NMF ($L_1$-DNMF) was proposed for hyperspectral unmixing \cite{HFang2018}. Multiview concept learning was incorporated to explicitly model the consistent and complementary information \cite{WZhao2021}.

In addition, an asymmetric \emph{encoder}-\emph{decoder} framework was presented in \cite{HCLi2022} for hyperspectral unmixing, where a multilayer nonlinear network was designed to powerfully encode the original data, and the resulting abundances were then decoded by the decoder part with one layer. The cost function is given by
\begin{equation}
\label{eqn:41}
  \min_{\mathbf{A}, \mathbf{E}, \mathbf{W}_1,\cdots,\mathbf{W}_L}
  \|\mathbf{X}-\mathbf{E}-\mathbf{A}\sigma(\mathbf{W}_L\cdots\sigma(\mathbf{W}_2\sigma(\mathbf{W}_1\mathbf{X})))\|_F^2,
\end{equation}
where $\sigma(\cdot)$ is a nonlinear activation function, abundances $\mathbf{S}=\mathbf{S}_L=\sigma(\mathbf{W}_L\cdots\sigma(\mathbf{W}_2\sigma(\mathbf{W}_1\mathbf{X})))$, and $\mathbf{E}$ is introduced in the \emph{decoder} to characterize sparse noise.

Besides, various deep NMF models \cite{PDHandschutter2020} have been proposed in an increasing number of applications.

(1) Deep orthogonal NMF \cite{BLyu2017, YQiu2017, PDHandschutter2020} is a variant of deep NMF by imposing orthogonality constraint to $\mathbf{S}_l$, where the decomposition is the same as in \cite{LChen2017}.

(2) Deep convex NMF \cite{BQian2016} was developed by extending convex NMF \cite{CDing2010} that is also known as archetypal analysis (AA) or concept factorization (CF), where each basis vector (named concept) is modeled as a linear combination of data points, \emph{i.e.}, $\mathbf{a}_m=\sum_{n=1}^N\mathbf{x}_nw_{nm}$. Fig. \ref{fig:3} (c) shows the factorization process of multilayer concept factorization \cite{XLi2017}.
Accordingly, $\mathbf{X}=\mathbf{X}\mathbf{W}_1\mathbf{S}_1\cdots\mathbf{W}_L\mathbf{S}_L$. After layerwise factorization, the cost function of deep convex NMF is
\begin{equation}
\label{eqn:42}
  \min_{\mathbf{W}_1,\cdots,\mathbf{W}_L,\mathbf{S}_1,\cdots,\mathbf{S}_L} \|\mathbf{X}-\mathbf{X}\mathbf{W}_1\mathbf{S}_1\cdots\mathbf{W}_L\mathbf{S}_L\|_F^2.
\end{equation}

Different from directly using the output of the previous layer as the input of subsequent layer, Zhang \emph{et al}. proposed a novel deep self-representative concept factorization network (DSCF-Net) \cite{YZhang2020} and a deep semisupervised coupled factorization network (DS$^2$CF-Net) \cite{YZhang2020dual}.

(3) By introducing a smoothing matrix at each layer, deep nsNMF (dnsNMF) \cite{JYu2018} was reported to learn features hierarchically in the context of text mining. Let $\mathbf{Z}_l$ denote the ``smoothing'' matrix. The matrix $\mathbf{X}$ is factorized into $\mathbf{A}_1$, $\mathbf{Z}_1$, and $\mathbf{S}_1$ in the first layer. Then, $\mathbf{S}_1$ is factorized into $\mathbf{A}_2$, $\mathbf{Z}_2$, and $\mathbf{S}_2$ in the next layer. The same process will be continued until the $L$-th layer is reached, shown in Fig. \ref{fig:3} (d). Hence, the observation can be represented using $\mathbf{X}=\mathbf{A}_1\mathbf{Z}_1\cdots\mathbf{A}_L\mathbf{Z}_L\mathbf{S}_L$. The cost function of dnsNMF is expressed as
\begin{equation}
\label{eqn:43}
  \min_{\mathbf{A}_1,\mathbf{Z}_1,\cdots,\mathbf{A}_L,\mathbf{Z}_L,\mathbf{S}_L} \|\mathbf{X}-\mathbf{A}_1\mathbf{Z}_1\cdots\mathbf{A}_L\mathbf{Z}_L\mathbf{S}_L\|_F^2.
\end{equation}

\begin{table*}[!t]
\centering
\renewcommand\arraystretch{1.3}
\caption{SAD (Average of 10 Runs) Along With Standard Deviation on the AVIRIS Cuprite Data Set for Different Methods. Boldfaced Number Denotes the Best Result Under Each Condition.}
\label{tab:2}
\setlength{\tabcolsep}{1.0mm}{
\begin{tabular}{c|c|c|c|c|c|c|c}
\hline\hline
\rule[-1ex]{0pt}{3.5ex} \mbox{Methods}&\mbox{$L_{1/2}$-NMF}&
\mbox{SGSNMF}&\mbox{TV-RSNMF}&\mbox{$L_{1/2}$-RNMF}&\mbox{MV-NTF-TV}&\mbox{MLNMF}&\mbox{SSRDMF} \\
\hline
Alunite GDS82 Na82
&0.1340$\pm$6.29\%& 0.1219$\pm$\textbf{1.72\%}&0.1012$\pm$3.12\%&\textbf{0.0891}$\pm$2.04\%&0.1047$\pm$2.76\%&0.1132$\pm$3.42\%&0.1165$\pm$2.72\%\\
Andradite WS487
&\textbf{0.0731}$\pm$1.66\%& 0.0945$\pm$5.49\%&0.0756$\pm$1.42\%&0.0832$\pm$2.14\%&0.0894$\pm$2.44\%&0.0748$\pm$\textbf{1.40\%}&0.0936$\pm$1.84\%\\
Buddingtonite GDS85 D-206
&0.1032$\pm$3.20\%& \textbf{0.0903$\pm$1.60\%}&0.0976$\pm$2.37\%&0.1009$\pm$2.48\%&0.0941$\pm$2.27\%&0.0935$\pm$2.55\%&0.1281$\pm$2.19\%\\
Chalcedony CU91-6A
&0.1504$\pm$\textbf{1.53\%}& 0.1495$\pm$1.64\%&0.1420$\pm$2.34\%&0.1500$\pm$1.59\%&0.1505$\pm$1.67\%&0.1539$\pm$1.57\%&\textbf{0.1269}$\pm$1.86\%\\
Kaolin/Smect H89-FR-5 30K
&0.1004$\pm$3.83\%& \textbf{0.0631$\pm$1.20\%}&0.0689$\pm$1.53\%&0.0699$\pm$1.39\%&0.0845$\pm$1.83\%&0.0914$\pm$3.31\%&0.0789$\pm$1.96\%\\
Kaolin/Smect KLF508 85\%K
&0.1163$\pm$3.78\%& 0.0903$\pm$\textbf{1.47\%}&0.1099$\pm$4.25\%&0.1091$\pm$3.34\%&0.1057$\pm$2.98\%&0.1190$\pm$3.81\%&\textbf{0.0901}$\pm$2.72\%\\
Kaolinite KGa-2
&0.1520$\pm$5.67\%& 0.1348$\pm$4.16\%&0.1621$\pm$5.19\%&0.1300$\pm$3.62\%&\textbf{0.1283}$\pm$4.12\%&0.1547$\pm$5.91\%&0.1486$\pm$\textbf{2.90\%}\\
Montmorillonite+Illi CM37
&0.0583$\pm$1.19\%& 0.0569$\pm$2.41\%&0.0564$\pm$\textbf{0.95\%}&\textbf{0.0528}$\pm$1.07\%&0.0662$\pm$2.60\%&0.0563$\pm$1.18\%&0.0617$\pm$1.06\%\\
Muscovite IL107
&0.1045$\pm$2.59\%& 0.1047$\pm$1.44\%&0.0975$\pm$3.27\%&0.1212$\pm$4.11\%&\textbf{0.0870}$\pm$1.82\%&0.0924$\pm$\textbf{0.93\%}&0.1025$\pm$2.10\%\\
Nontronite NG-1.a
&0.1261$\pm$1.34\%& 0.1186$\pm$1.18\%&0.1263$\pm$1.73\%&0.1292$\pm$\textbf{1.10\%}&0.1403$\pm$4.48\%&0.1273$\pm$1.37\%&\textbf{0.1033}$\pm$1.47\%\\
Pyrope WS474
&0.1134$\pm$2.55\%& 0.1220$\pm$\textbf{0.99\%}&0.1222$\pm$3.35\%&0.0993$\pm$4.47\%&0.0876$\pm$2.97\%&0.0810$\pm$2.34\%&\textbf{0.0721}$\pm$1.67\%\\
Sphene HS189.3B
&0.0680$\pm$1.69\%& 0.0742$\pm$\textbf{0.65\%}&0.0681$\pm$1.03\%&0.0829$\pm$2.00\%&\textbf{0.0606}$\pm$0.83\%&0.0716$\pm$1.80\%&0.0668$\pm$0.84\%\\
\hline
Mean
&0.1083$\pm$0.81\%& 0.1017$\pm$0.60\%&0.1023$\pm$0.90\%&0.1015$\pm$0.61\%&0.0999$\pm$\textbf{0.41\%}&0.1024$\pm$0.57\%&\textbf{0.0991}$\pm$0.53\%\\
\hline
\hline
\end{tabular}}
\end{table*}

\begin{table*}[!t]
\centering
\renewcommand\arraystretch{1.3}
\caption{SAD Scores (Average of 10 Runs) Along With Their Standard Deviation on the Samson Data Set for Different Methods. Boldfaced Number Denotes the Best Result Under Each Condition.}
\label{tab:3}
\setlength{\tabcolsep}{1.8mm}{
\begin{tabular}{c|c|c|c|c|c|c|c}
\hline\hline
\rule[-1ex]{0pt}{3.5ex} \mbox{Methods}&\mbox{$L_{1/2}$-NMF}&
\mbox{SGSNMF}&\mbox{TV-RSNMF}&\mbox{$L_{1/2}$-RNMF}&\mbox{MV-NTF-TV}&\mbox{MLNMF}&\mbox{SSRDMF} \\
\hline
\ ~~~Soil~~~
&0.0620$\pm$11.28\%&\textbf{0.0085$\pm$0.00\%}&0.0275$\pm$0.47\%&0.0281$\pm$0.39\%&0.0228$\pm$0.58\%&0.0754$\pm$16.39\%&0.0320$\pm$0.94\%\\
\ ~~~Tree~~~
&0.0647$\pm$5.03 \%&0.0402$\pm$\textbf{0.01\%}&0.0481$\pm$0.41\%&0.0514$\pm$0.30\%&0.0474$\pm$0.55\%&0.0590$\pm$3.87 \%&\textbf{0.0376}$\pm$0.32\%\\
\ ~~~Water~~~
&0.1167$\pm$2.24 \%&0.1364$\pm$\textbf{0.02\%}&0.1086$\pm$0.14\%&0.0998$\pm$0.67\%&0.1114$\pm$0.18\%&0.1001$\pm$0.66 \%&\textbf{0.0903}$\pm$1.52\%\\
\hline
~~~Mean~~~
&0.0811$\pm$4.72 \%&0.0617$\pm$\textbf{0.01\%}&0.0614$\pm$0.34\%&0.0598$\pm$0.45\%&0.0605$\pm$0.22\%&0.0781$\pm$6.91 \%&\textbf{0.0533}$\pm$0.46\%\\
\hline
\hline
\end{tabular}}
\end{table*}

\begin{table*}[!t]
\centering
\renewcommand\arraystretch{1.3}
\caption{SAD Scores (Average of 10 Runs) Along With Their Standard Deviation on the Jasper Ridge Data Set for Different Methods. Boldfaced Number Denotes the Best Result Under Each Condition.}
\label{tab:4}
\setlength{\tabcolsep}{1.8mm}{
\begin{tabular}{c|c|c|c|c|c|c|c}
\hline\hline
\rule[-1ex]{0pt}{3.5ex} \mbox{Methods}&\mbox{$L_{1/2}$-NMF}&\mbox{SGSNMF}&\mbox{TV-RSNMF}&\mbox{$L_{1/2}$-RNMF}&\mbox{MV-NTF-TV}&\mbox{MLNMF}&\mbox{SSRDMF} \\
\hline
\ ~~~Tree~~~
&0.1959$\pm$9.56 \%&\textbf{0.0818$\pm$0.63 \%}&0.1854$\pm$26.21\%&0.1349$\pm$7.84 \%&0.2091$\pm$3.98 \%&0.1456$\pm$8.52 \%&0.1518$\pm$4.44 \%\\
\ ~~~Water~~~
&0.1548$\pm$6.56 \%&0.1783$\pm$6.31 \%&\textbf{0.0991}$\pm$2.78 \%&0.1069$\pm$3.01 \%&0.3443$\pm$\textbf{1.90} \%&0.1421$\pm$5.19 \%&0.2061$\pm$7.75 \%\\
\ ~~~Soil~~~
&0.2661$\pm$25.31\%&0.3738$\pm$19.67\%&0.1780$\pm$16.56\%&0.2090$\pm$29.59\%&0.2497$\pm$21.54\%&0.1576$\pm$21.42\%&\textbf{0.1223$\pm$3.18 \%}\\
\ ~~~Road~~~
&0.5453$\pm$8.19 \%&0.3112$\pm$31.86\%&0.4957$\pm$\textbf{3.36 \%}&0.4870$\pm$8.70 \%&\textbf{0.2474}$\pm$9.52\%&0.6736$\pm$11.68\%&0.4355$\pm$12.45\%\\
\hline
~~~Mean~~~
&0.2905$\pm$7.36 \%&0.2363$\pm$\textbf{3.54 \%}&0.2395$\pm$8.09 \%&0.2345$\pm$9.30 \%&0.2626$\pm$4.91 \%&0.2797$\pm$4.09 \%&\textbf{0.2289}$\pm$4.03 \%\\
\hline
\hline
\end{tabular}}
\end{table*}

(4) Deep autoencoder-like NMF (DANMF) \cite{FYe2018} was proposed for community detection, whose cost function is expressed as
\begin{equation}
\begin{aligned}
\label{eqn:44}
  \min_{\mathbf{A}_1,\mathbf{A}_2,\cdots,\mathbf{A}_L,\mathbf{S}_L}\|\mathbf{X}&-\mathbf{A}_1\mathbf{A}_2\cdots\mathbf{A}_L\mathbf{S}_L\|_F^2 \\
                  +&\|\mathbf{S}_L-\mathbf{A}_L^T\cdots\mathbf{A}_2^T\mathbf{A}_1^T\mathbf{X}\|_F^2.
\end{aligned}
\end{equation}
Similar to deep autoencoder, DANMF consists of an \emph{encoder} component and a \emph{decoder} component. This architecture empowers DANMF to learn the hierarchical mappings between the original network and the final community assignment with implicit low-to-high level hidden attributes of the original network learned in the intermediate layers.

Meanwhile, the hierarchical factorization has applied to tensor, developing some multilayer frameworks of tensor decomposition \cite{XBi2018}, and deep tensor decompositions \cite{CJia2016, SOymak2018, JCasebeer2019}. In addition, the deep unfolding technique was used for unrolling the iteration inference algorithm into a layerwise structure to obtain novel neural network-like architectures that enjoy the advantages of well-defined interpretability, strong learning power, and little computational cost \cite{CZhou2022, YQian2020, FXiong2022}.

Multilayer/deep extensions of NMF combine both interpretability and the extraction of multiple hierarchical features. Nevertheless, it is also an important and challenging research issue such as how to determine the parameters (\emph{e.g.}, the inner ranks and the number of layers) and loss function, and how to choose efficient optimization algorithms and initial conditions.

\begin{figure}[!t]
\centering
\subfigure[]{\includegraphics[width=2.6cm]{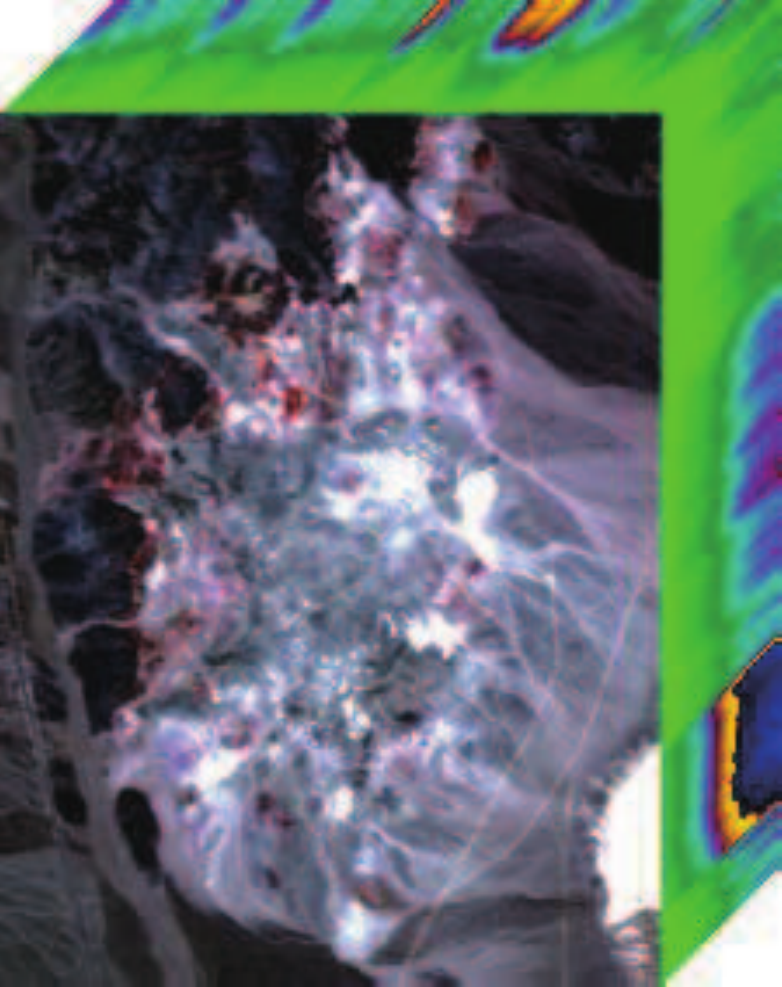}}~~
\subfigure[]{\includegraphics[width=2.19cm]{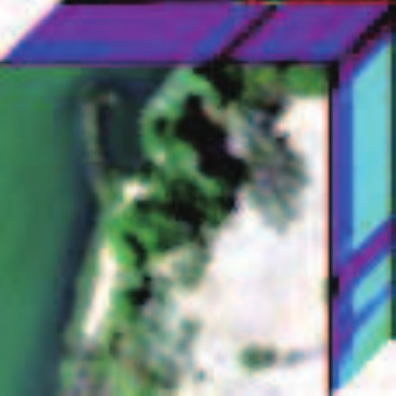}}~~
\subfigure[]{\includegraphics[width=2.56cm]{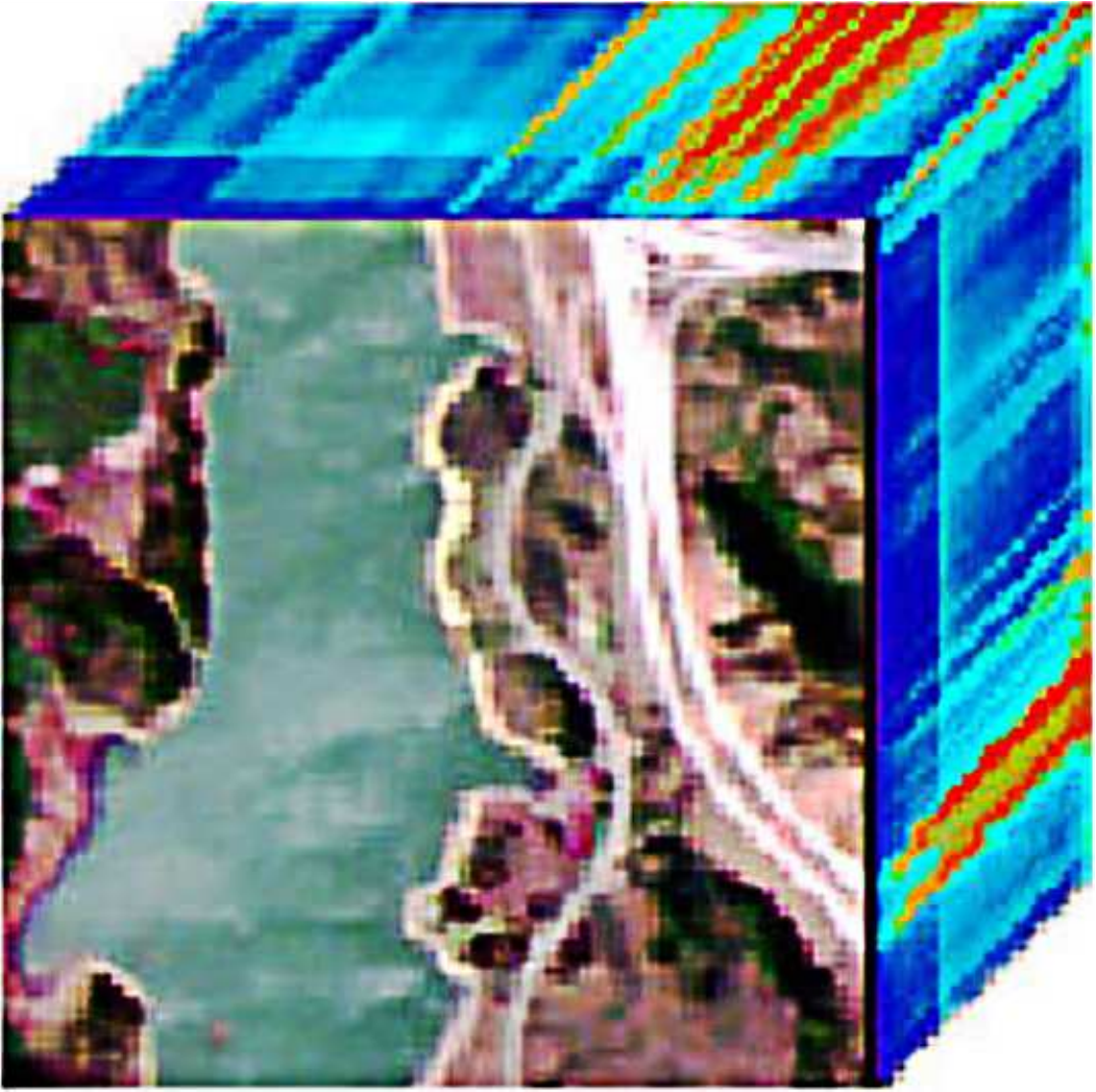}}
\caption{Visualization of (a) Cuprite image, (b) Samson image, and (c) Jasper Ridge image.}
\label{fig:4}
\end{figure}

\section{Experimental Results and Discussion}
Spectral angle distance (SAD) is utilized to assess unmixing performance quantitatively, given as:
\begin{equation}
\label{eqn:45}
  \textrm{SAD}_m= \textrm{arccos}\Bigl(\frac{\mathbf{A}^T_
  m\hat{\mathbf{A}}_m}{\lVert\mathbf{A}^{T}_m\rVert\lVert\hat
  {\mathbf{A}}_m\rVert}\Bigr),
\end{equation}
where $\mathbf{A}_m$ and $\hat{\mathbf{A}}_m$ are the $m$-th original and estimated endmember spectral signatures, respectively.

\begin{figure*}[!t]
\centering
\mbox{
{\includegraphics[width=2.45cm]{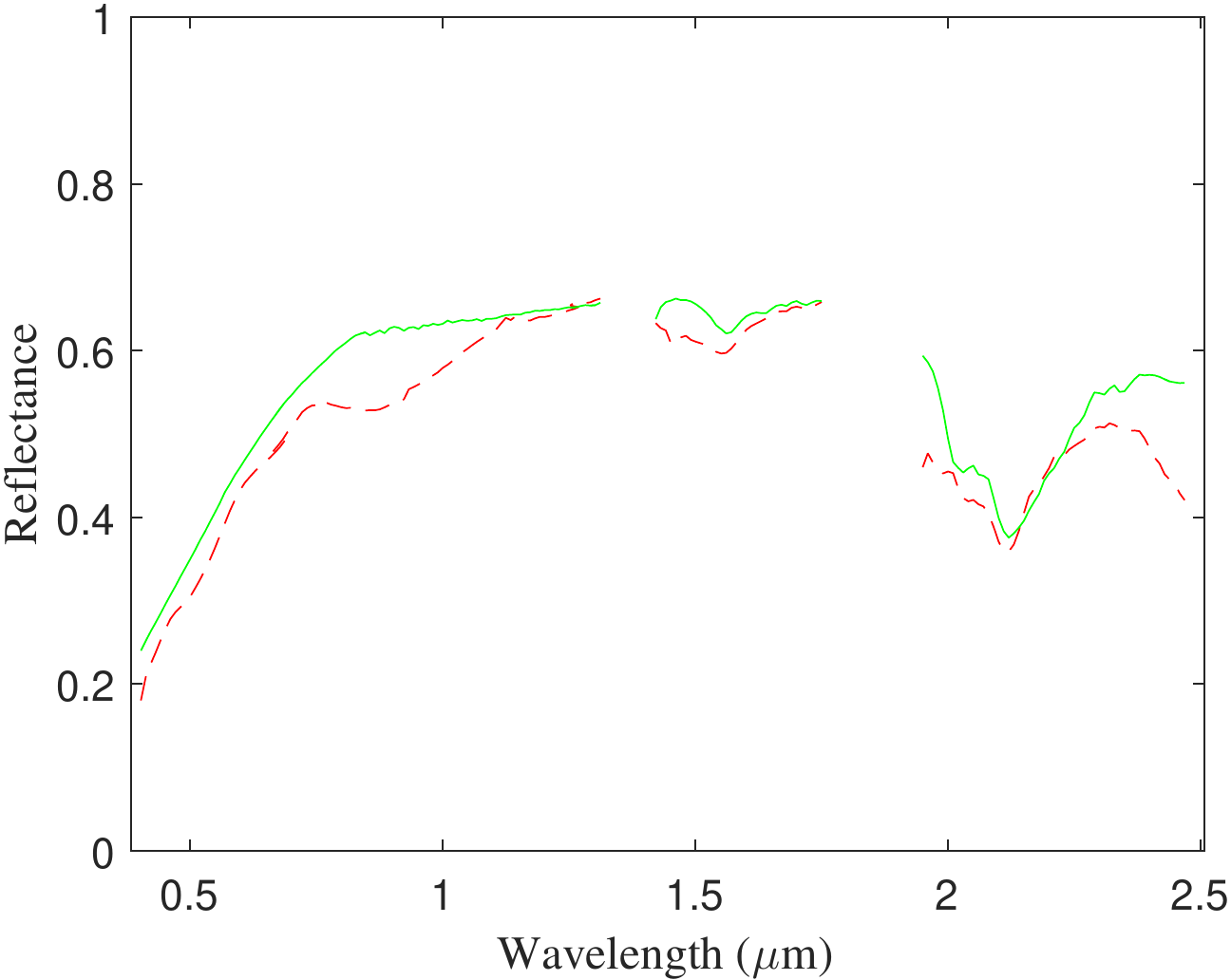}}
{\includegraphics[width=2.45cm]{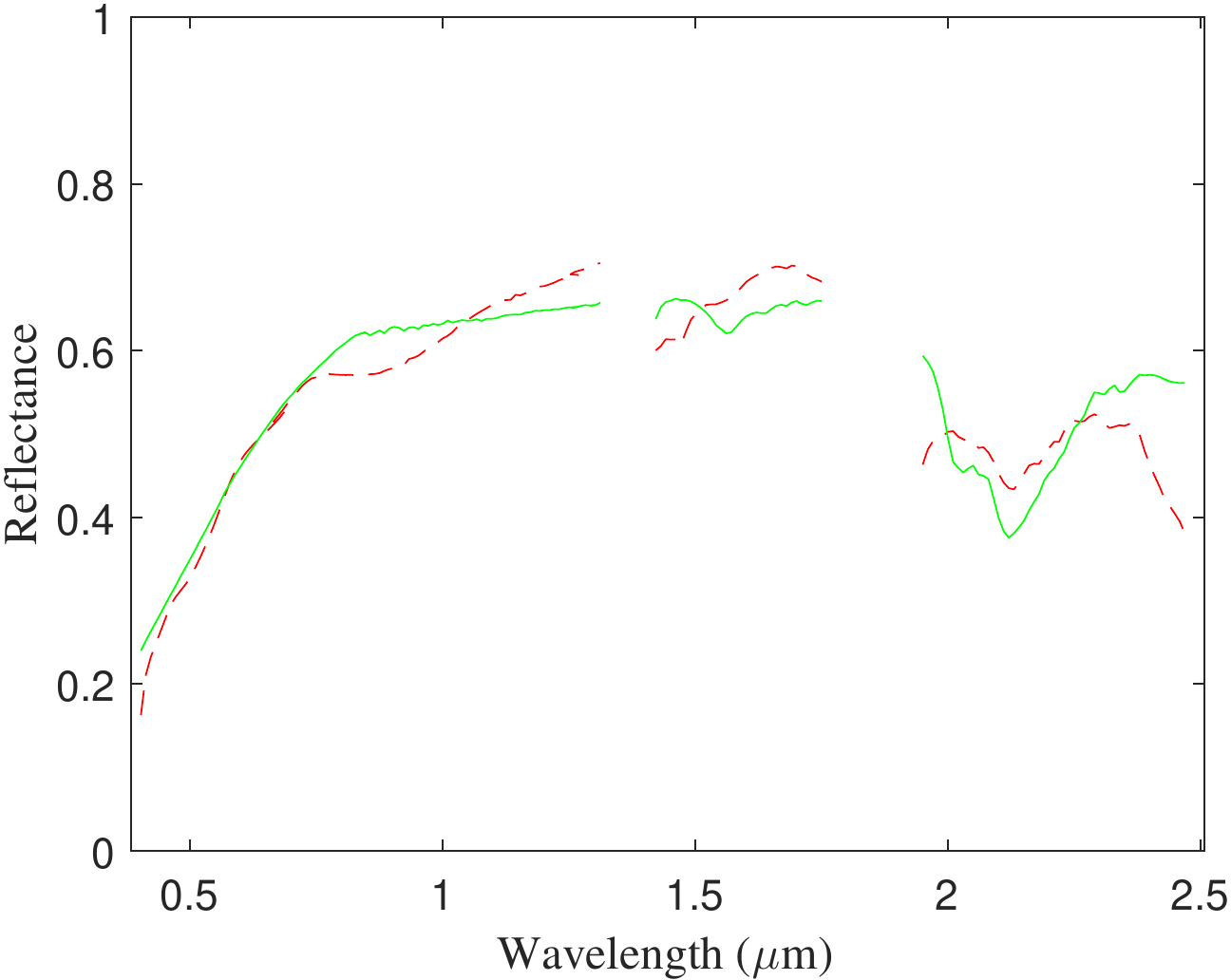}}
{\includegraphics[width=2.45cm]{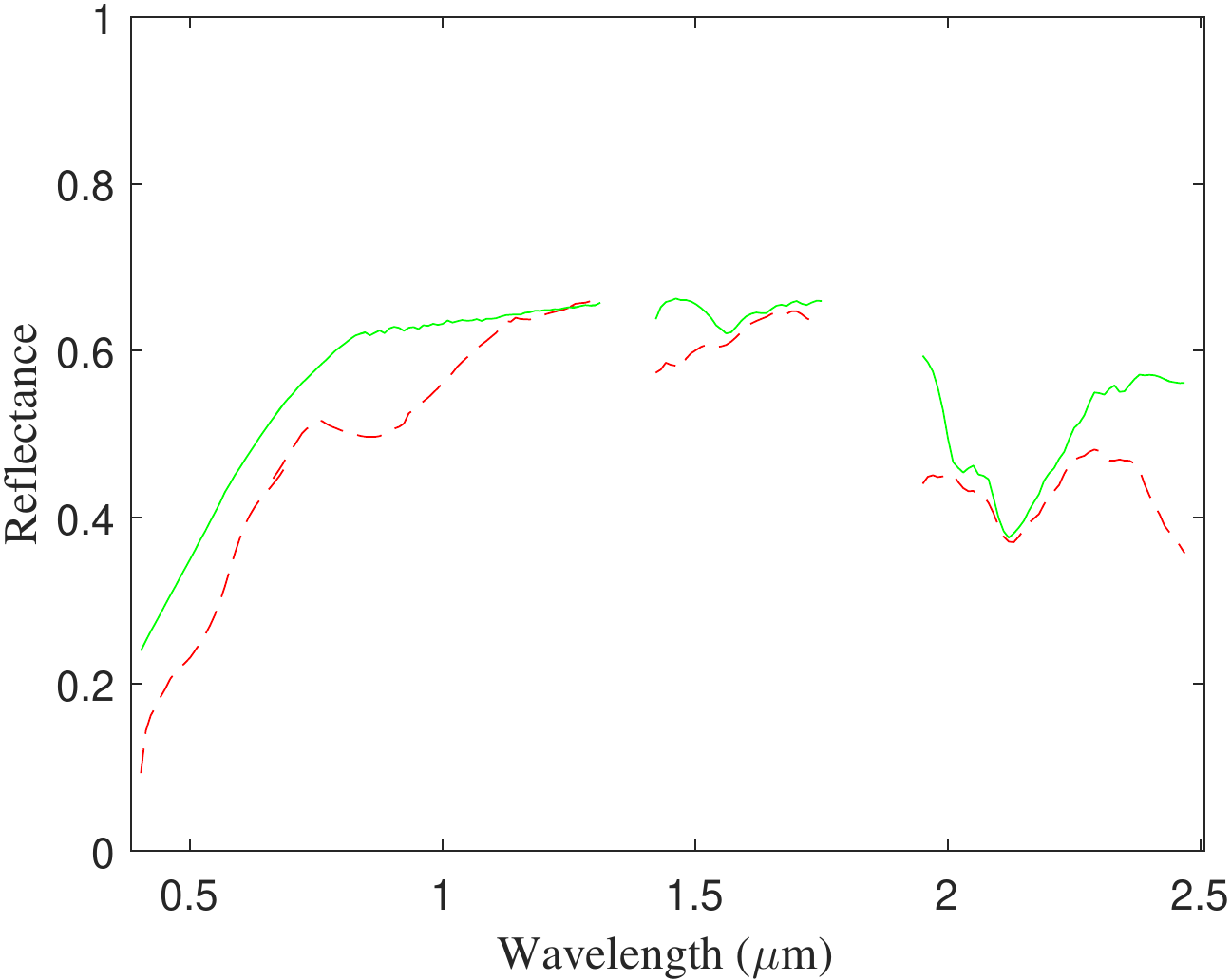}}
{\includegraphics[width=2.45cm]{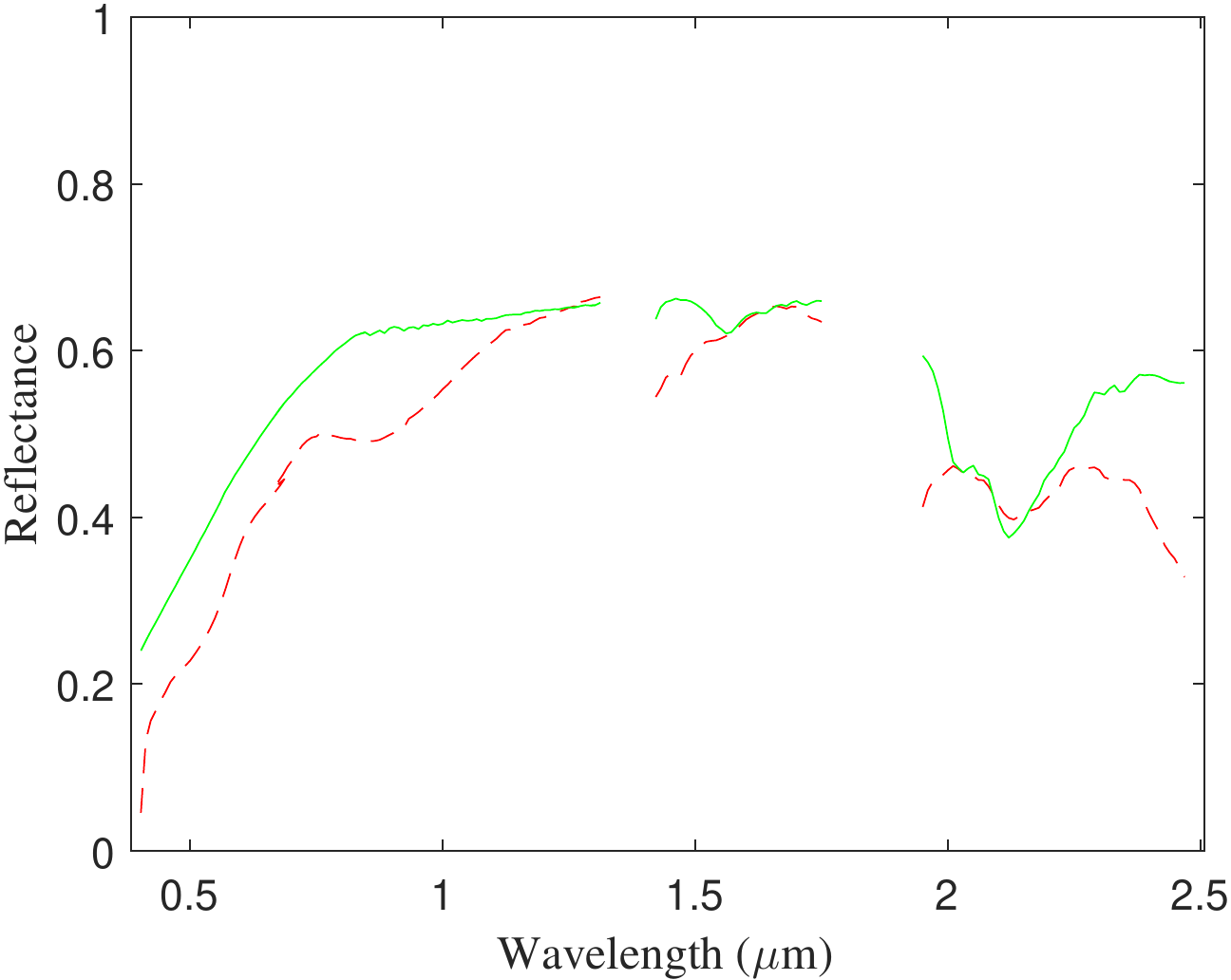}}
{\includegraphics[width=2.45cm]{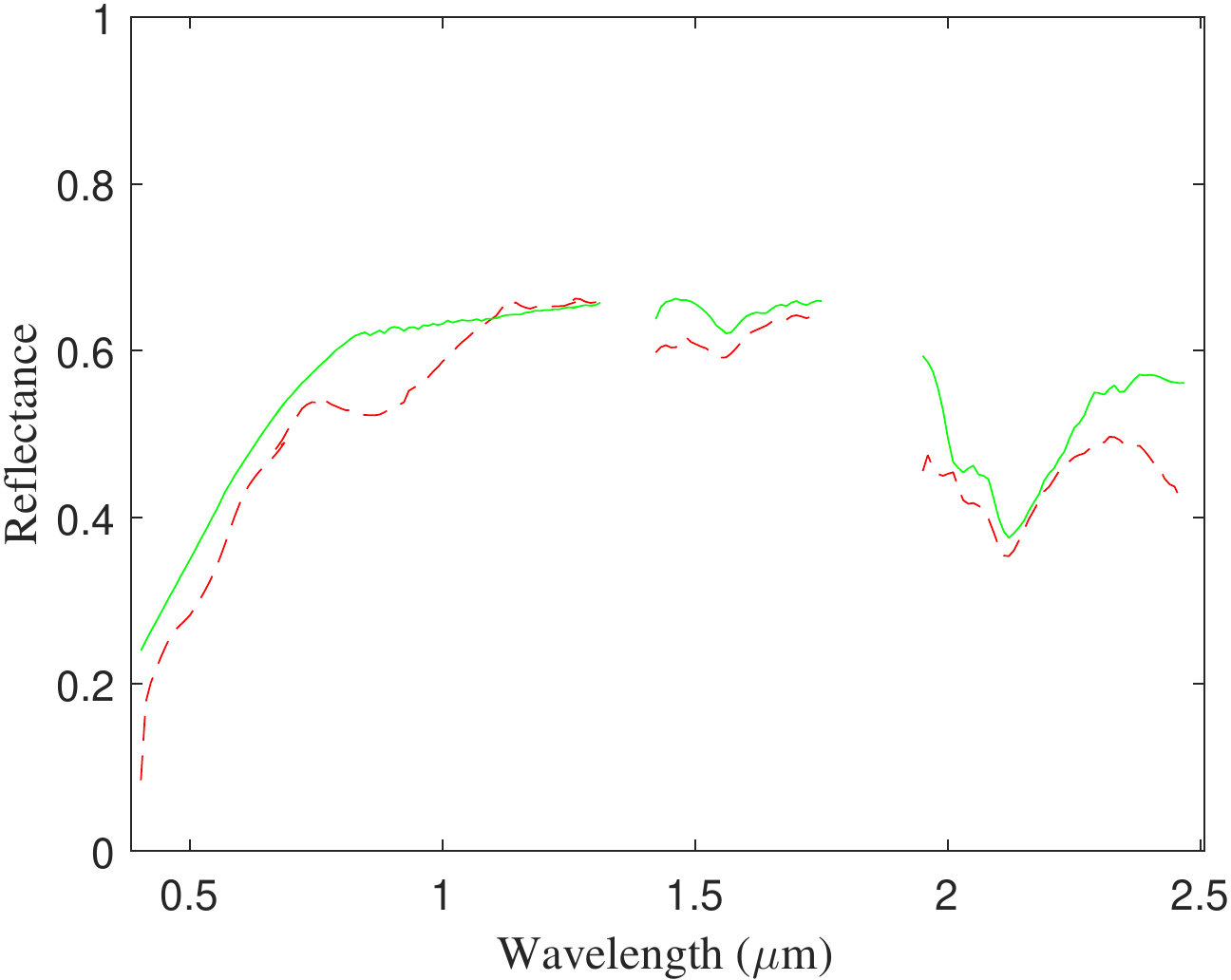}}
{\includegraphics[width=2.45cm]{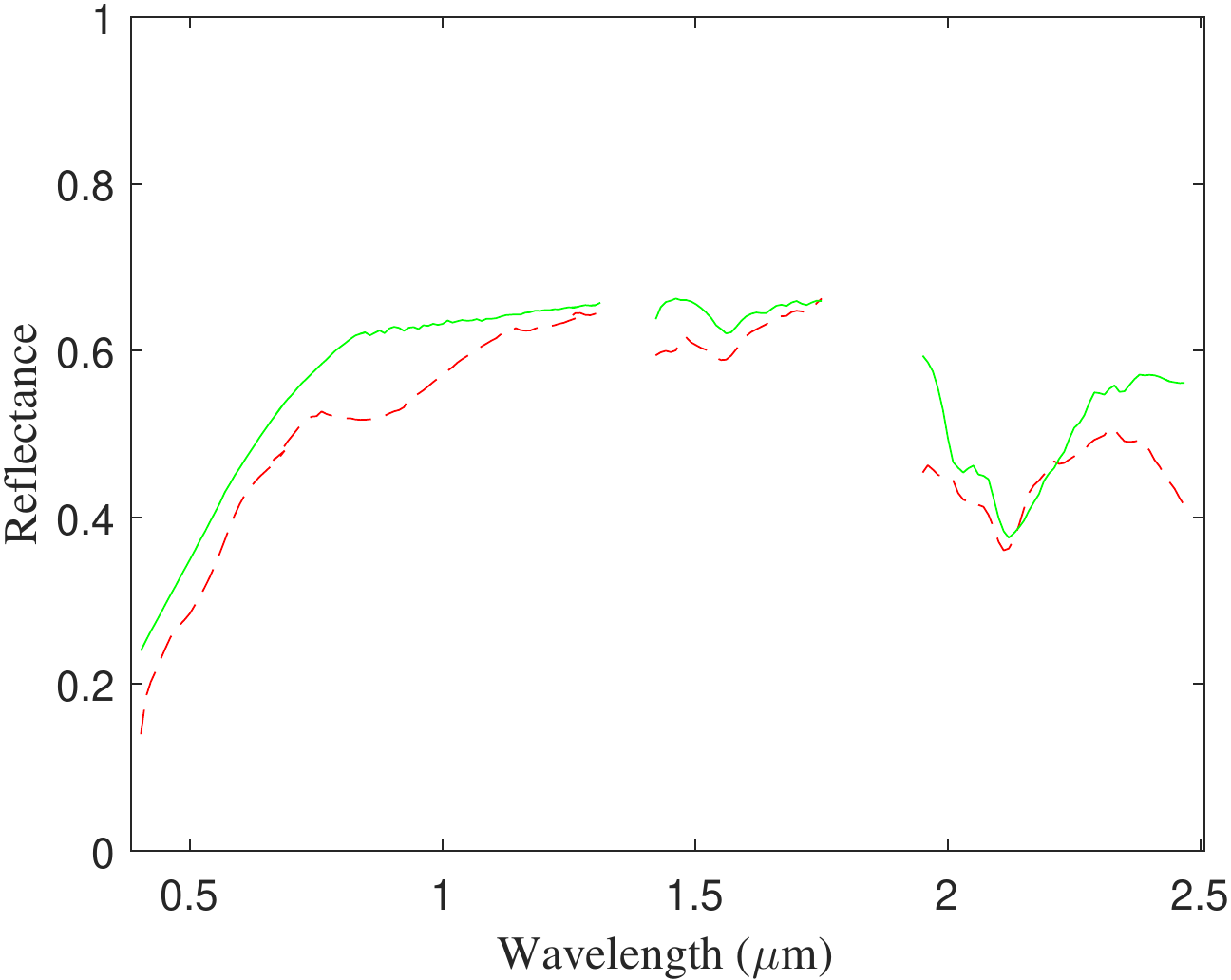}}
{\includegraphics[width=2.45cm]{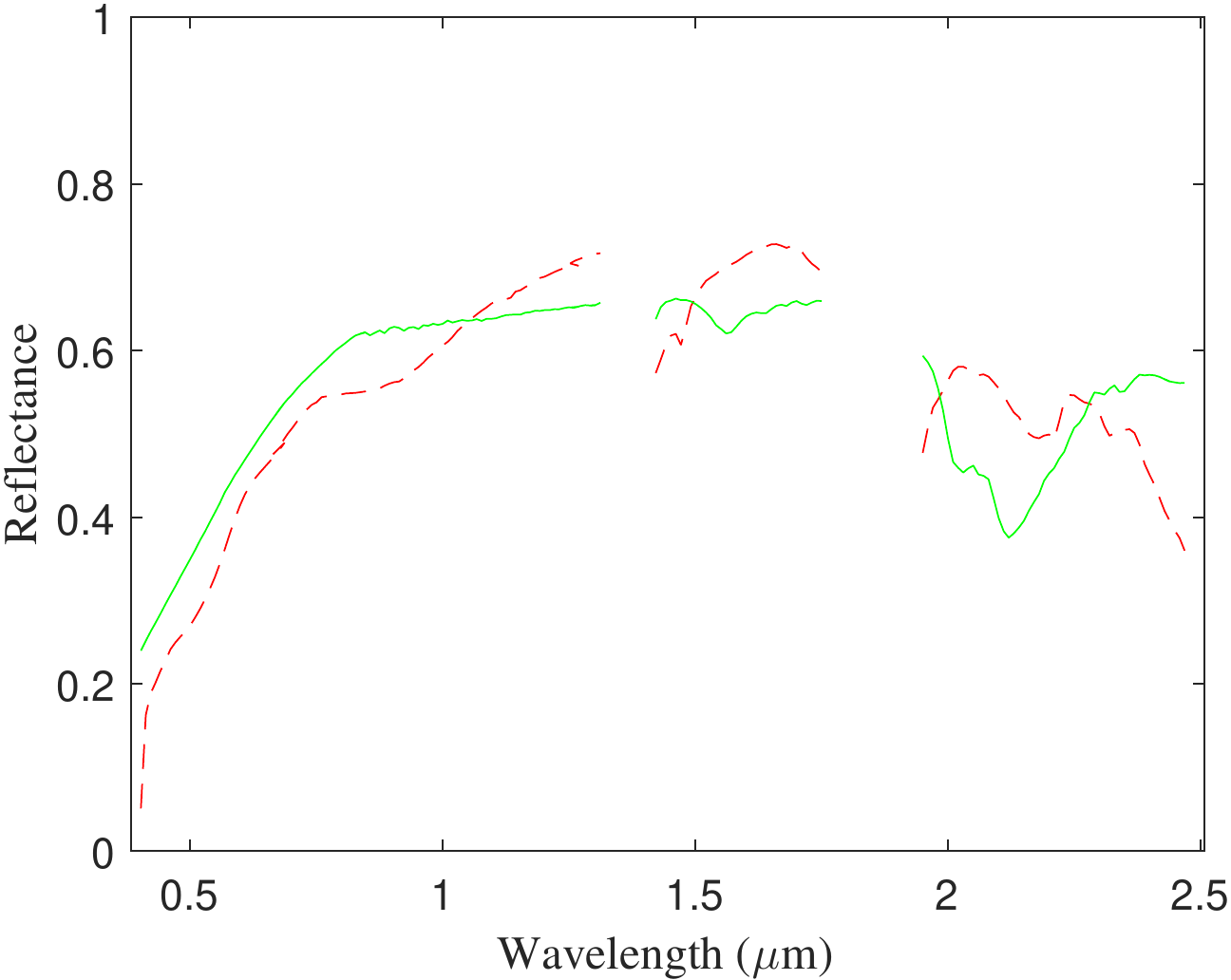}}}
\\
\mbox{
{\includegraphics[width=2.45cm]{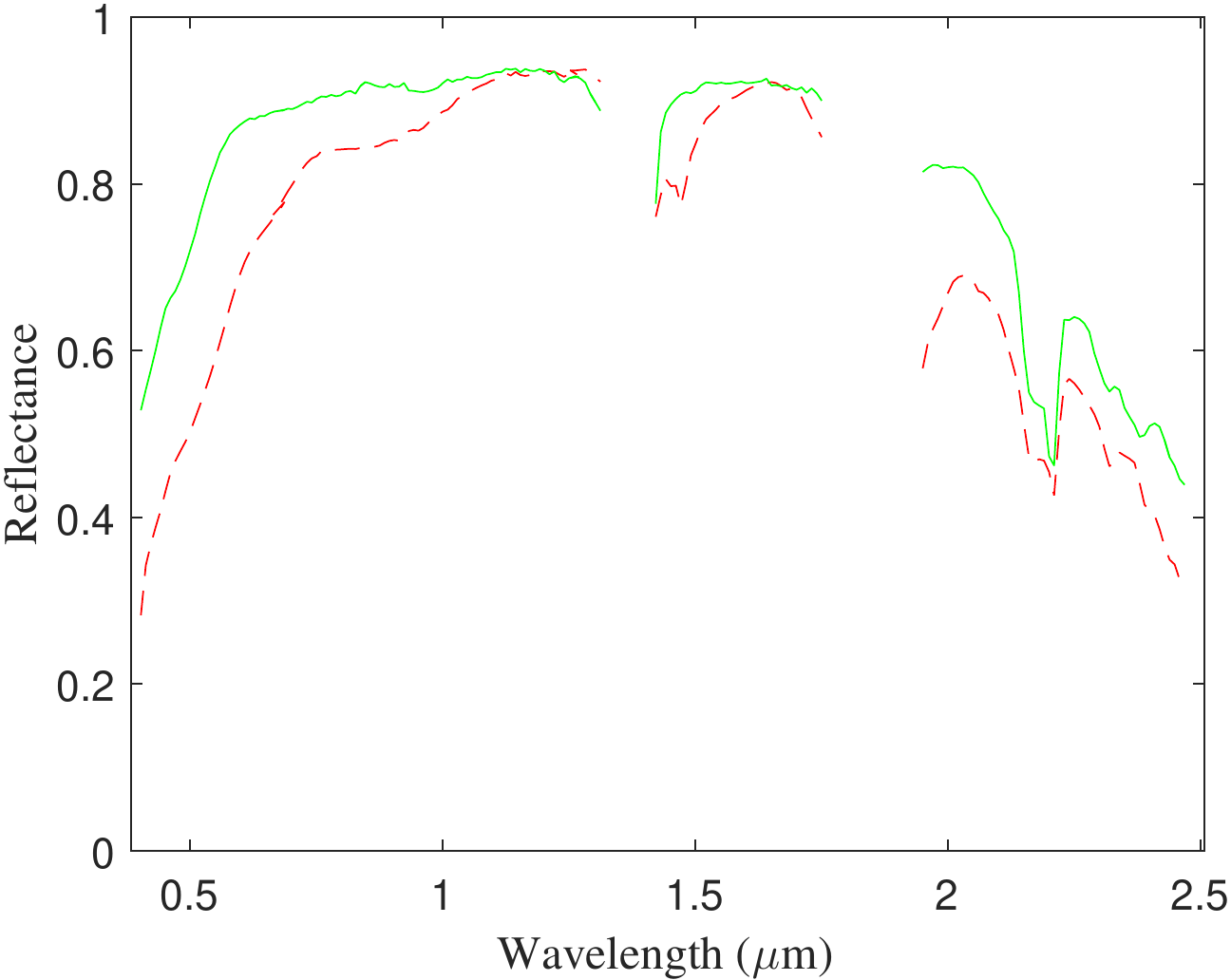}}
{\includegraphics[width=2.45cm]{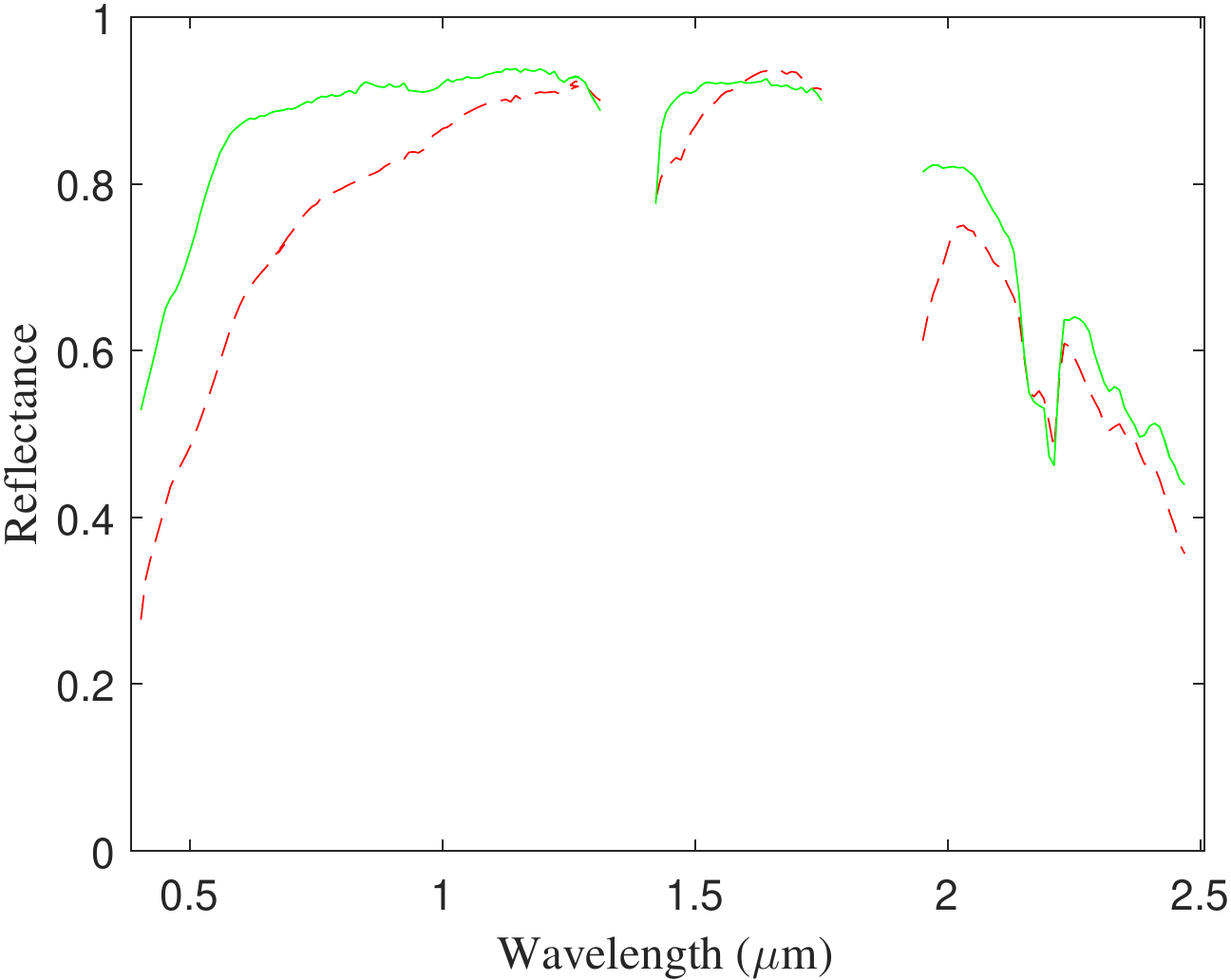}}
{\includegraphics[width=2.45cm]{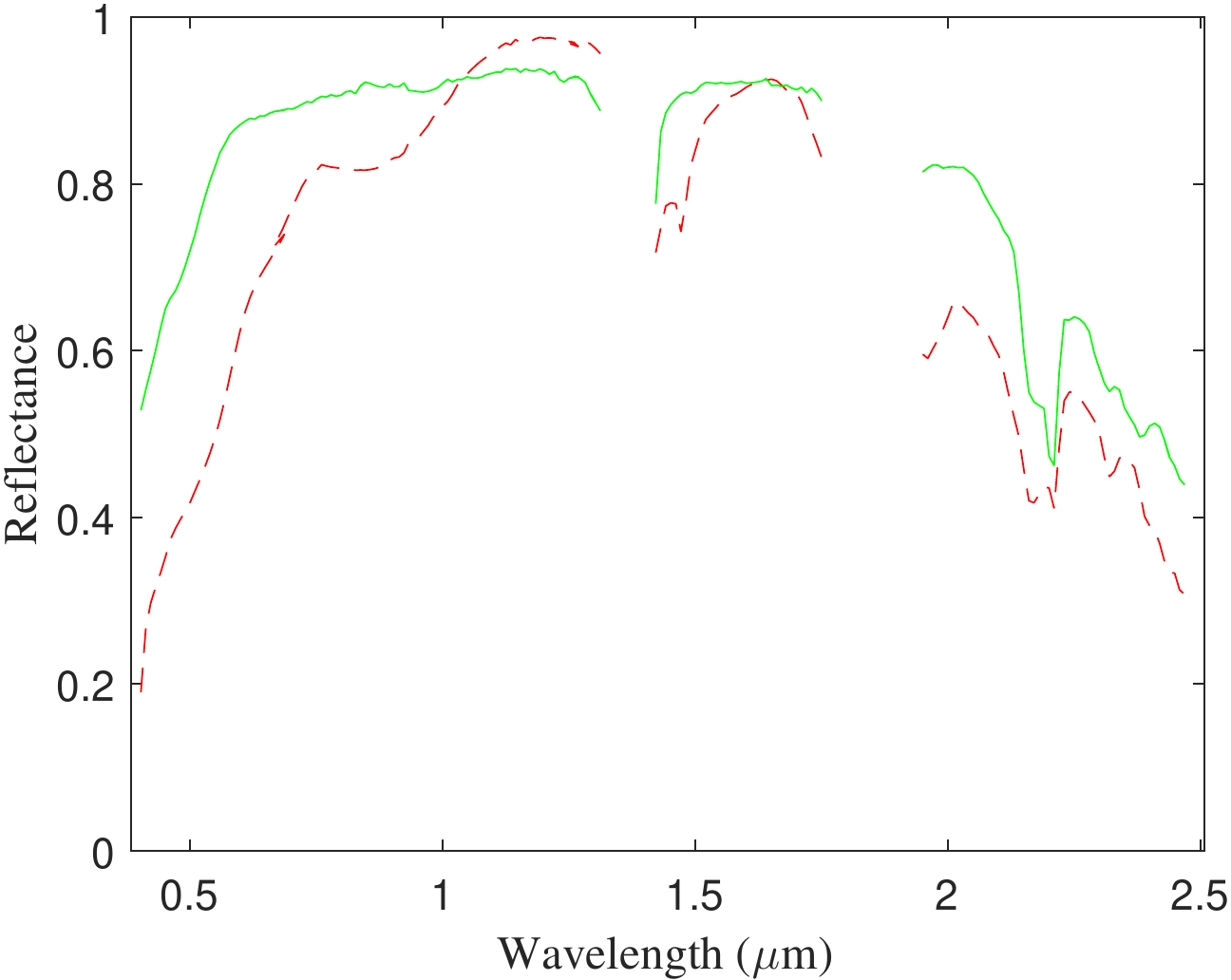}}
{\includegraphics[width=2.45cm]{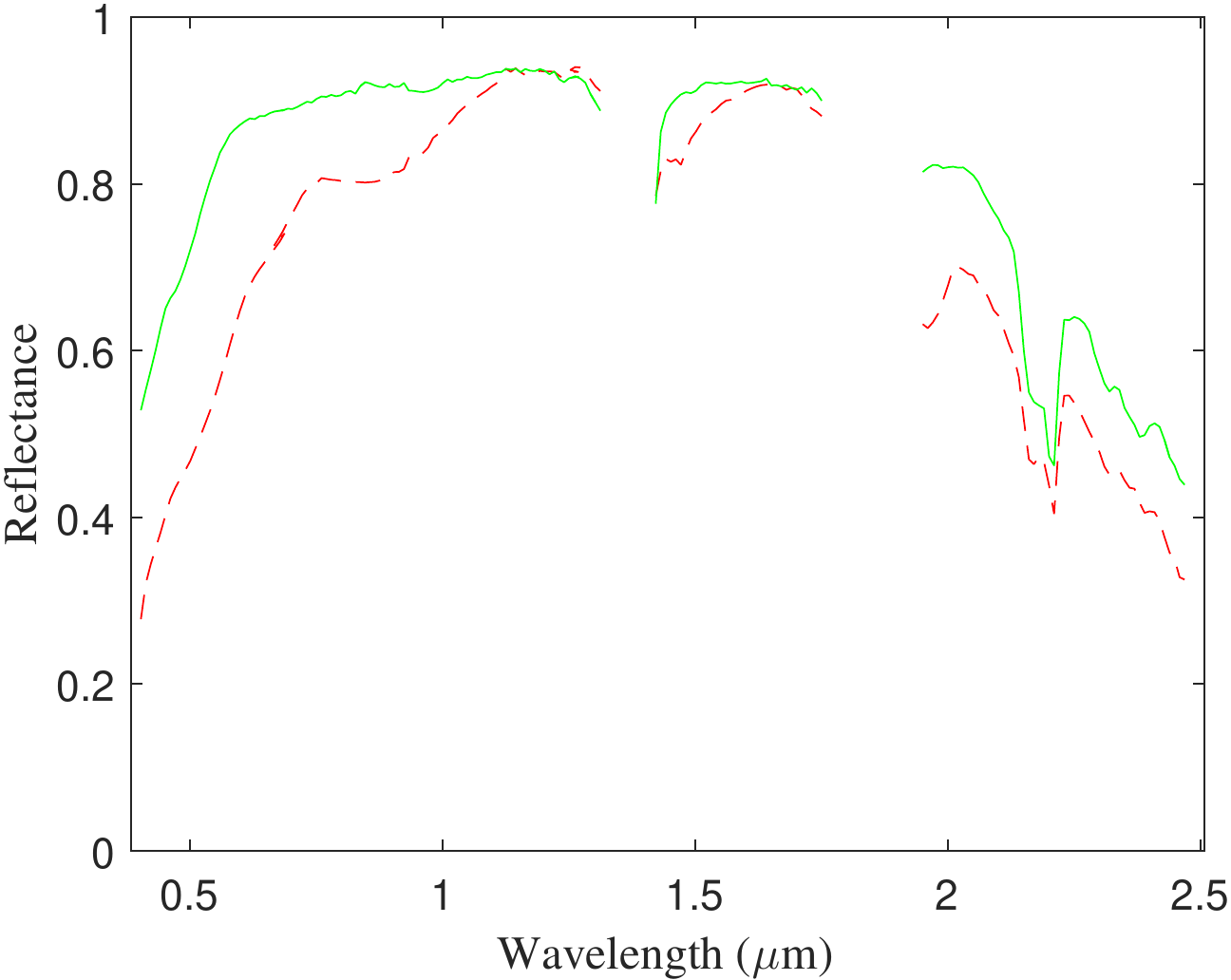}}
{\includegraphics[width=2.45cm]{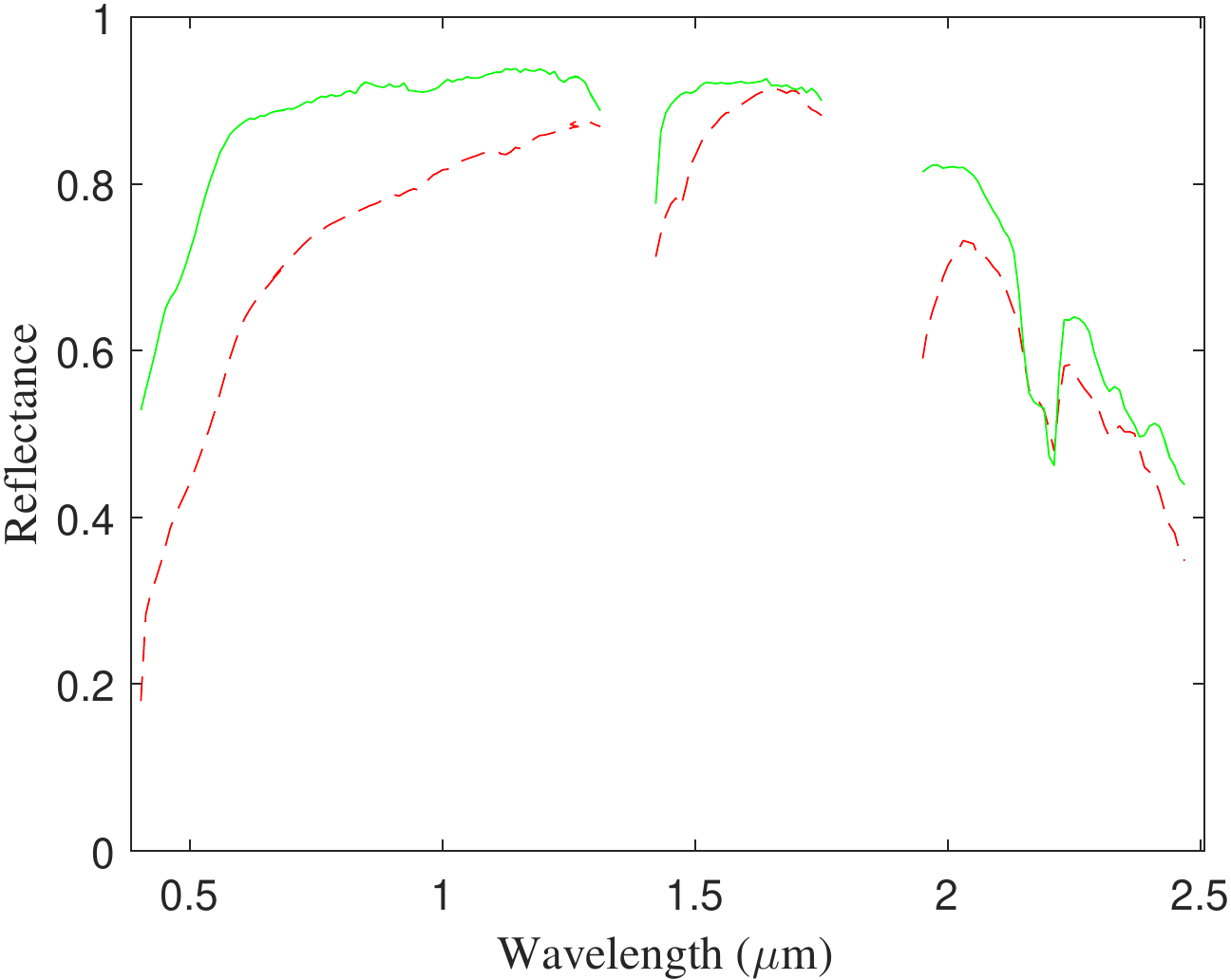}}
{\includegraphics[width=2.45cm]{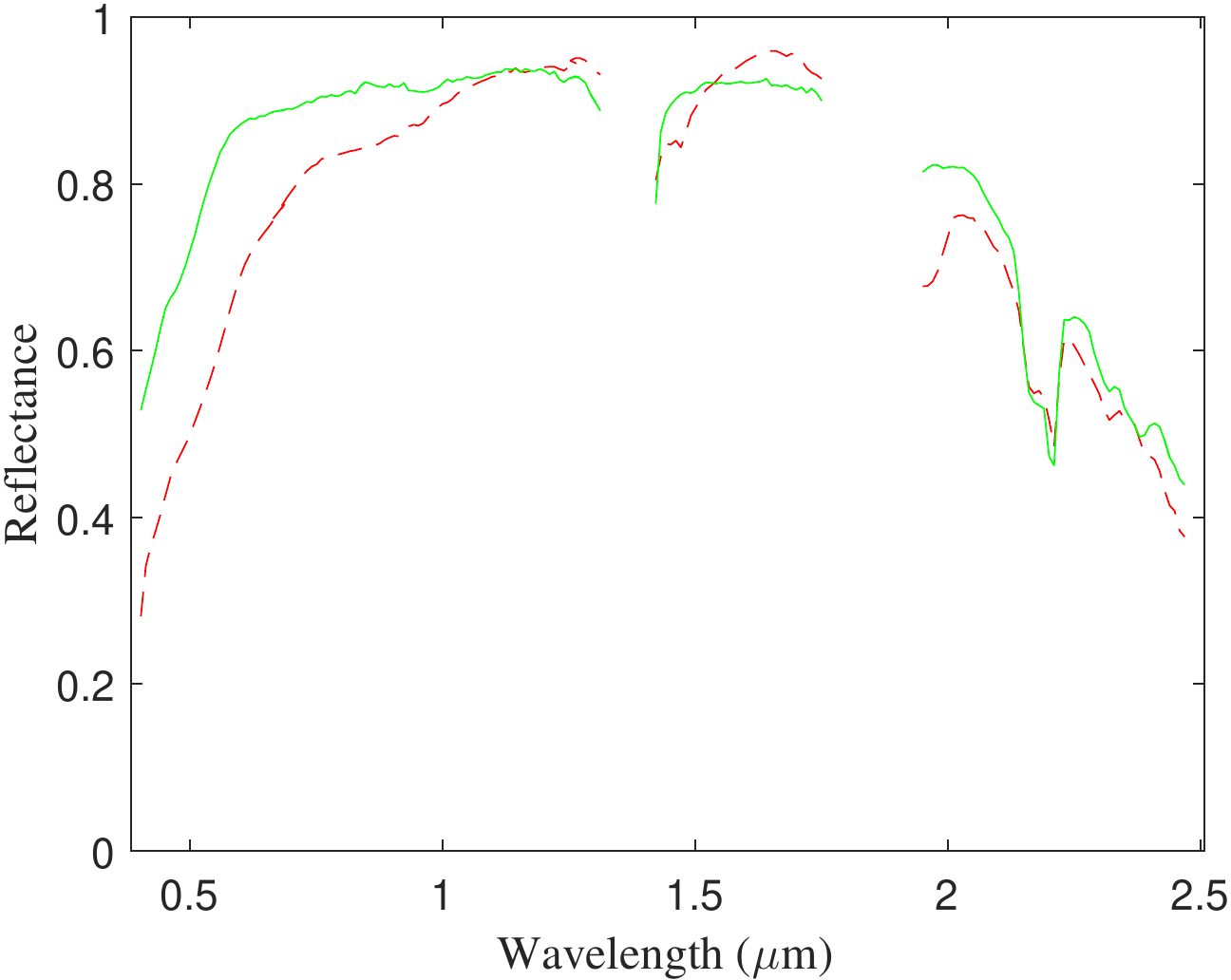}}
{\includegraphics[width=2.45cm]{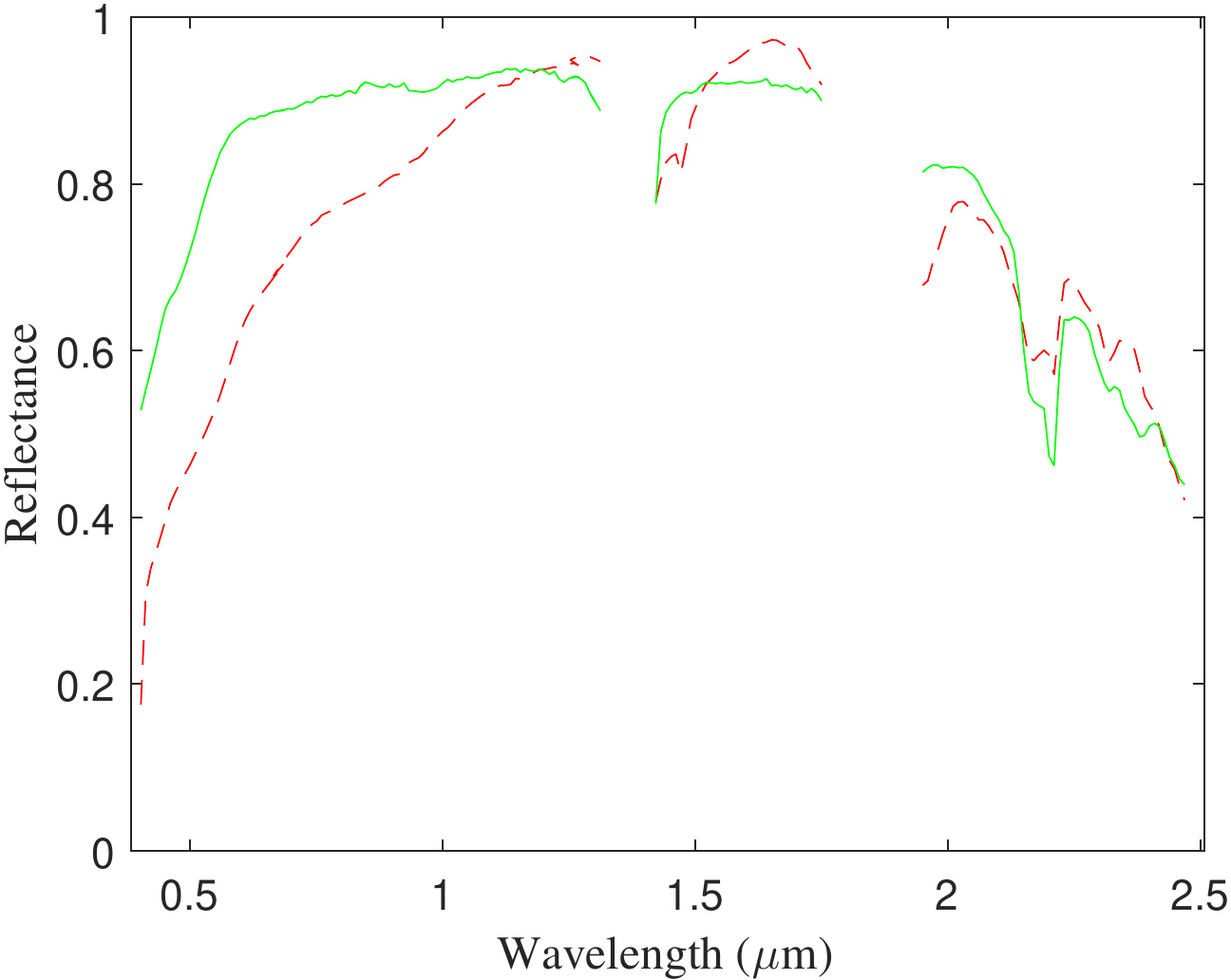}}}
\\
\mbox{
{\includegraphics[width=2.45cm]{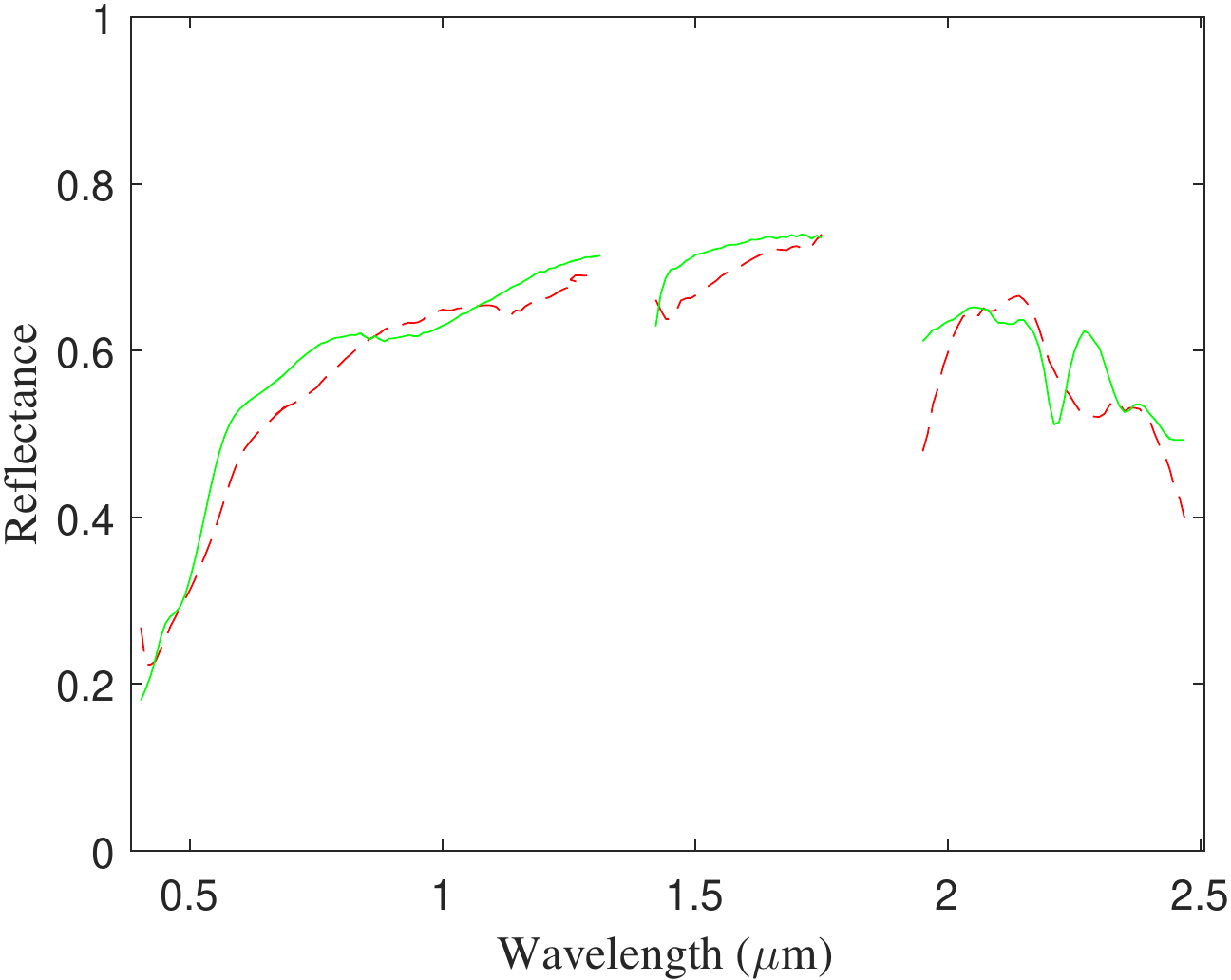}}
{\includegraphics[width=2.45cm]{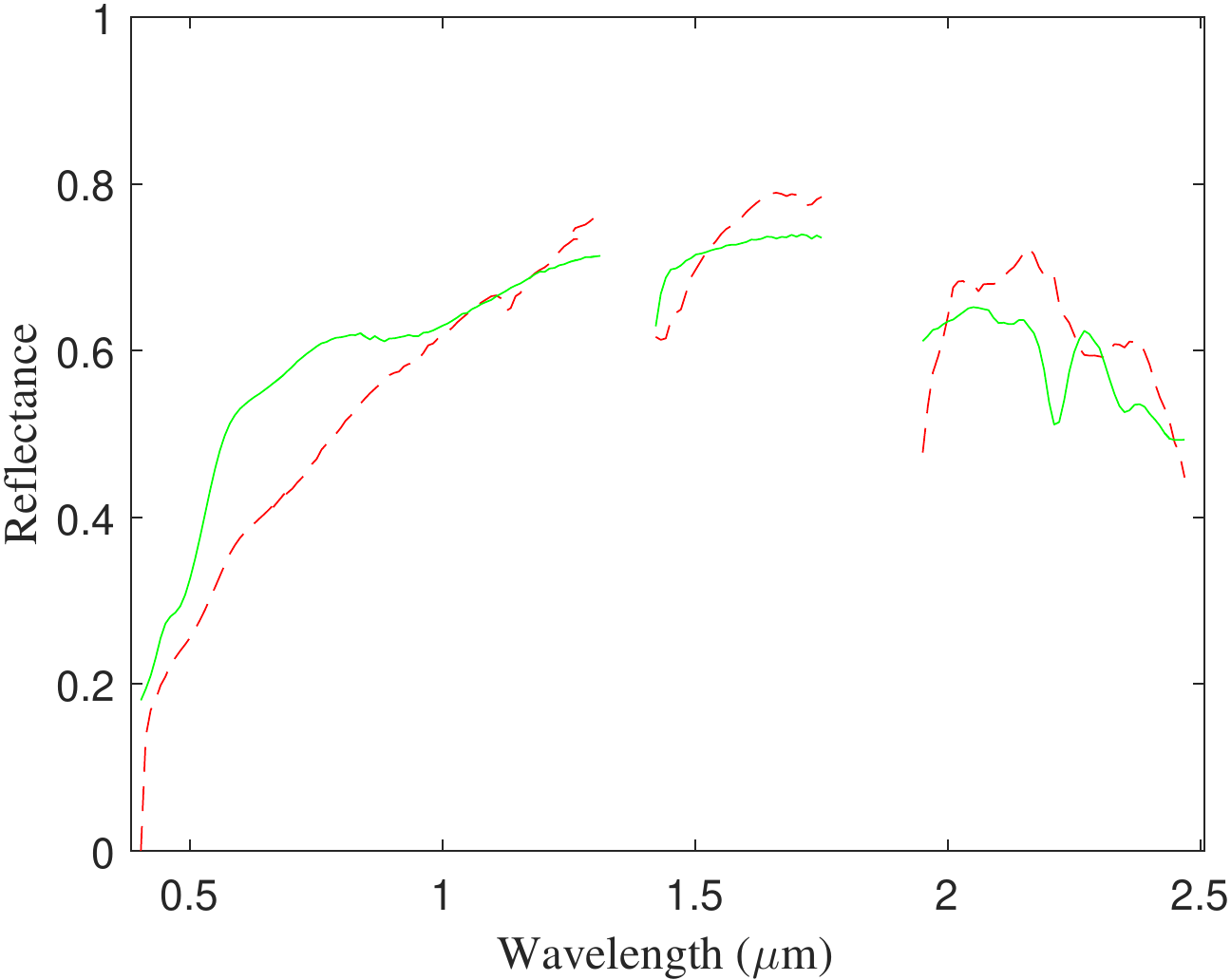}}
{\includegraphics[width=2.45cm]{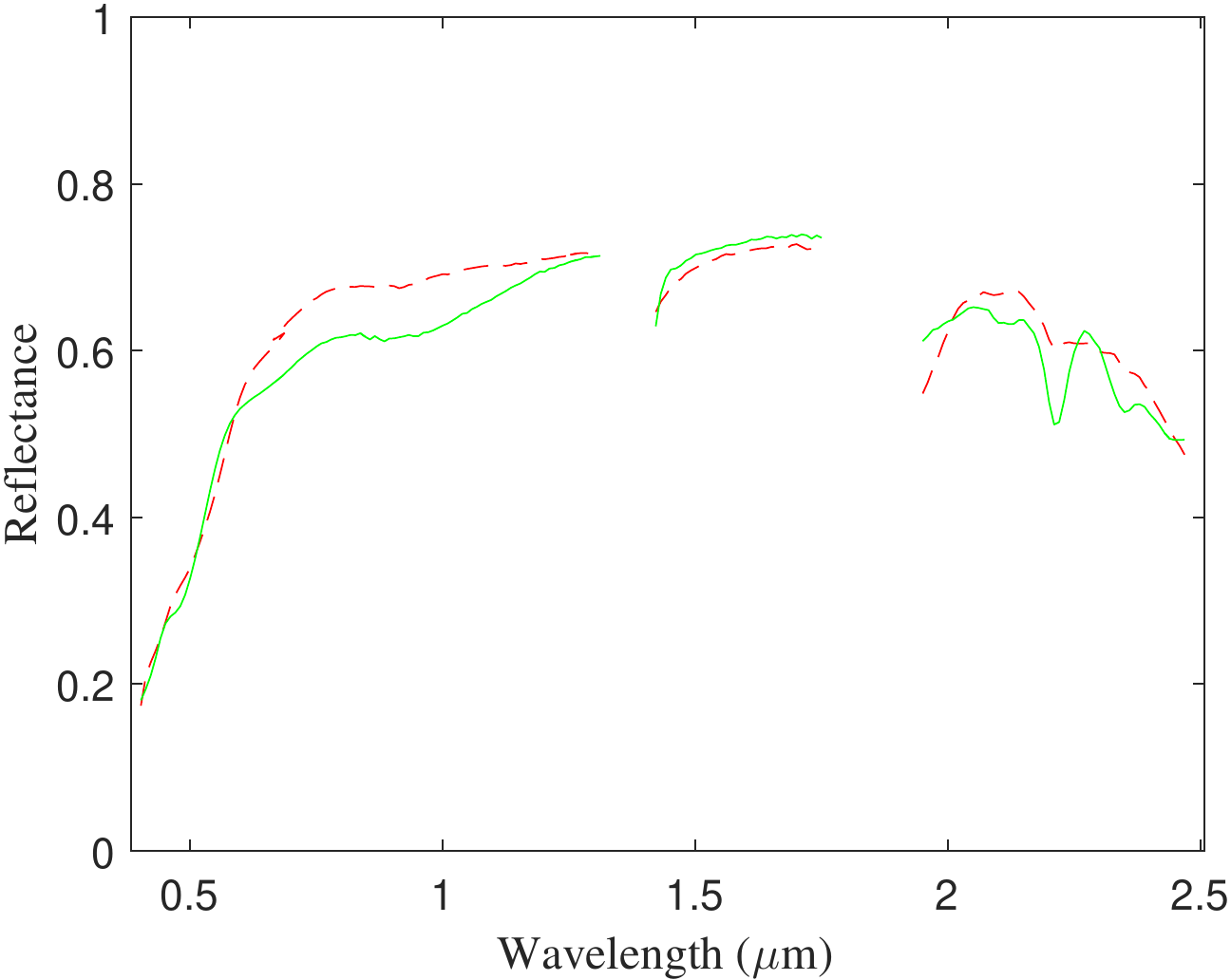}}
{\includegraphics[width=2.45cm]{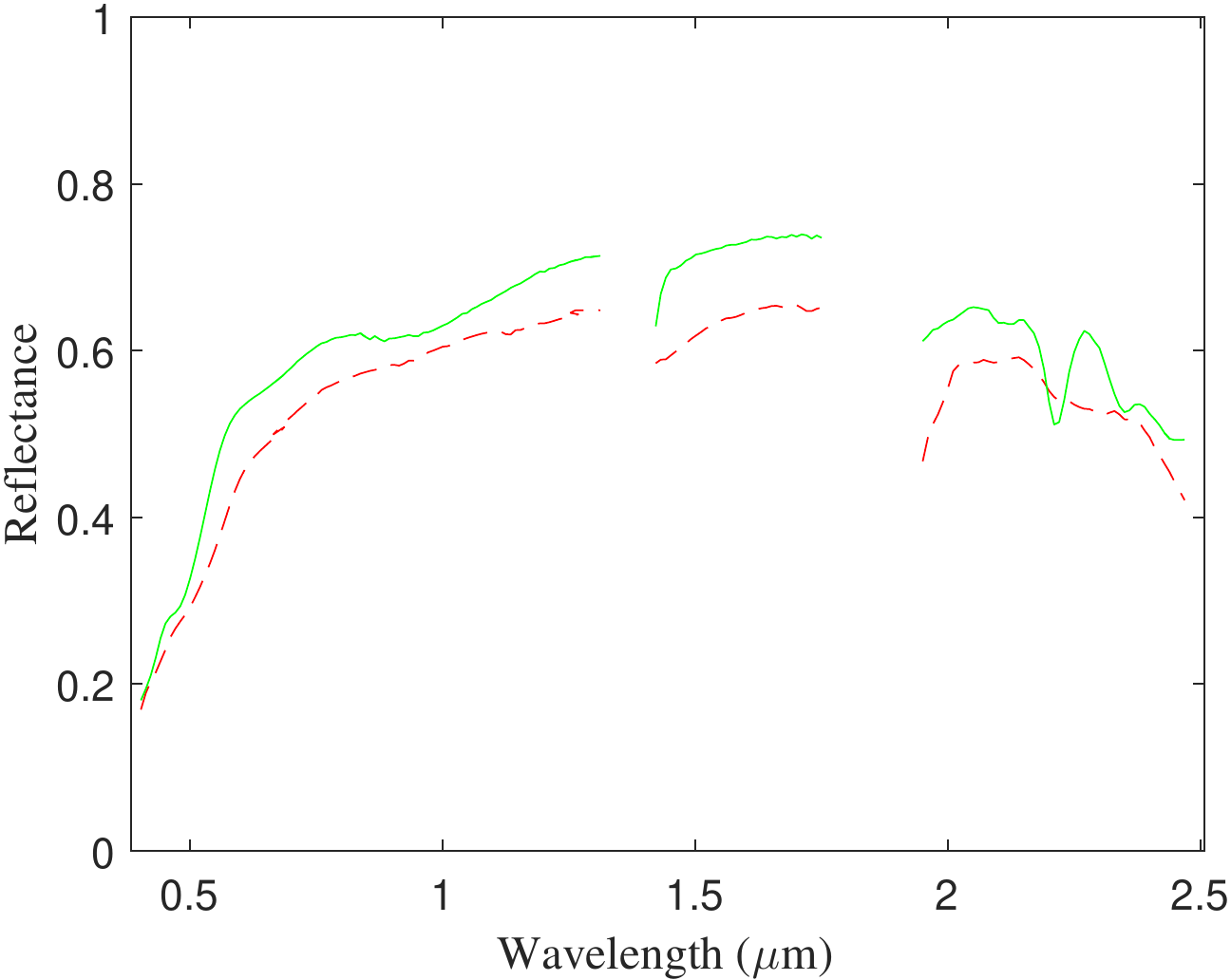}}
{\includegraphics[width=2.45cm]{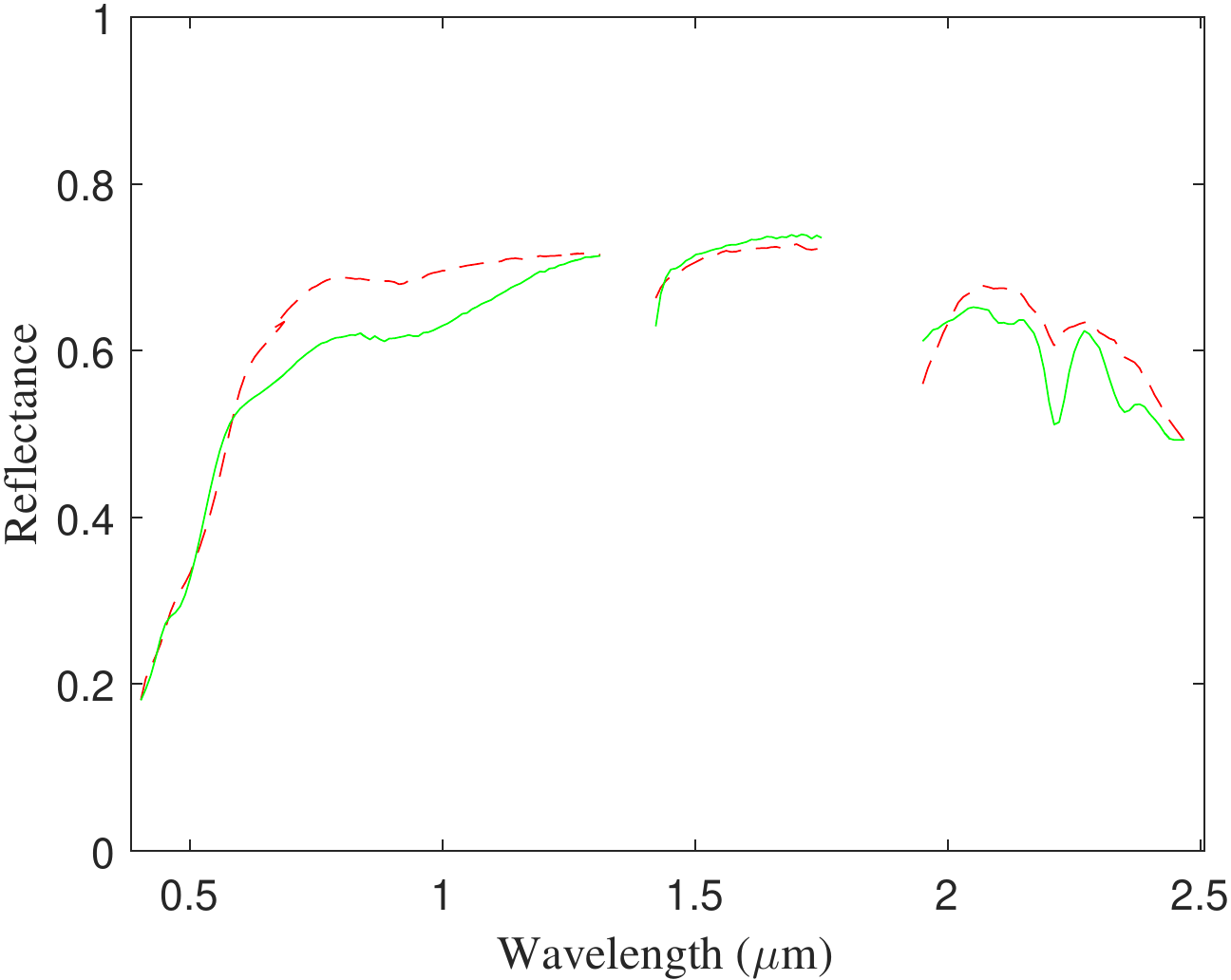}}
{\includegraphics[width=2.45cm]{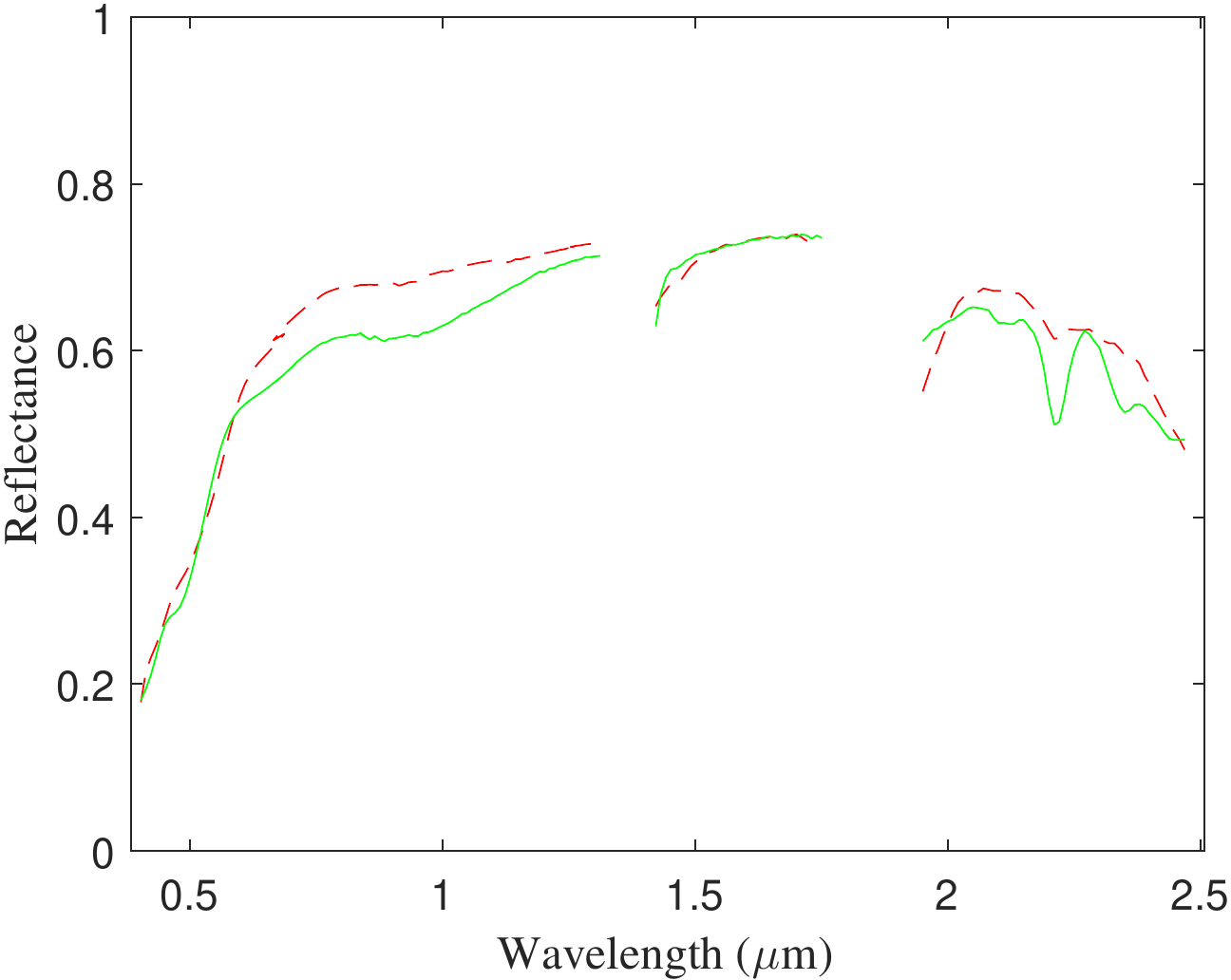}}
{\includegraphics[width=2.45cm]{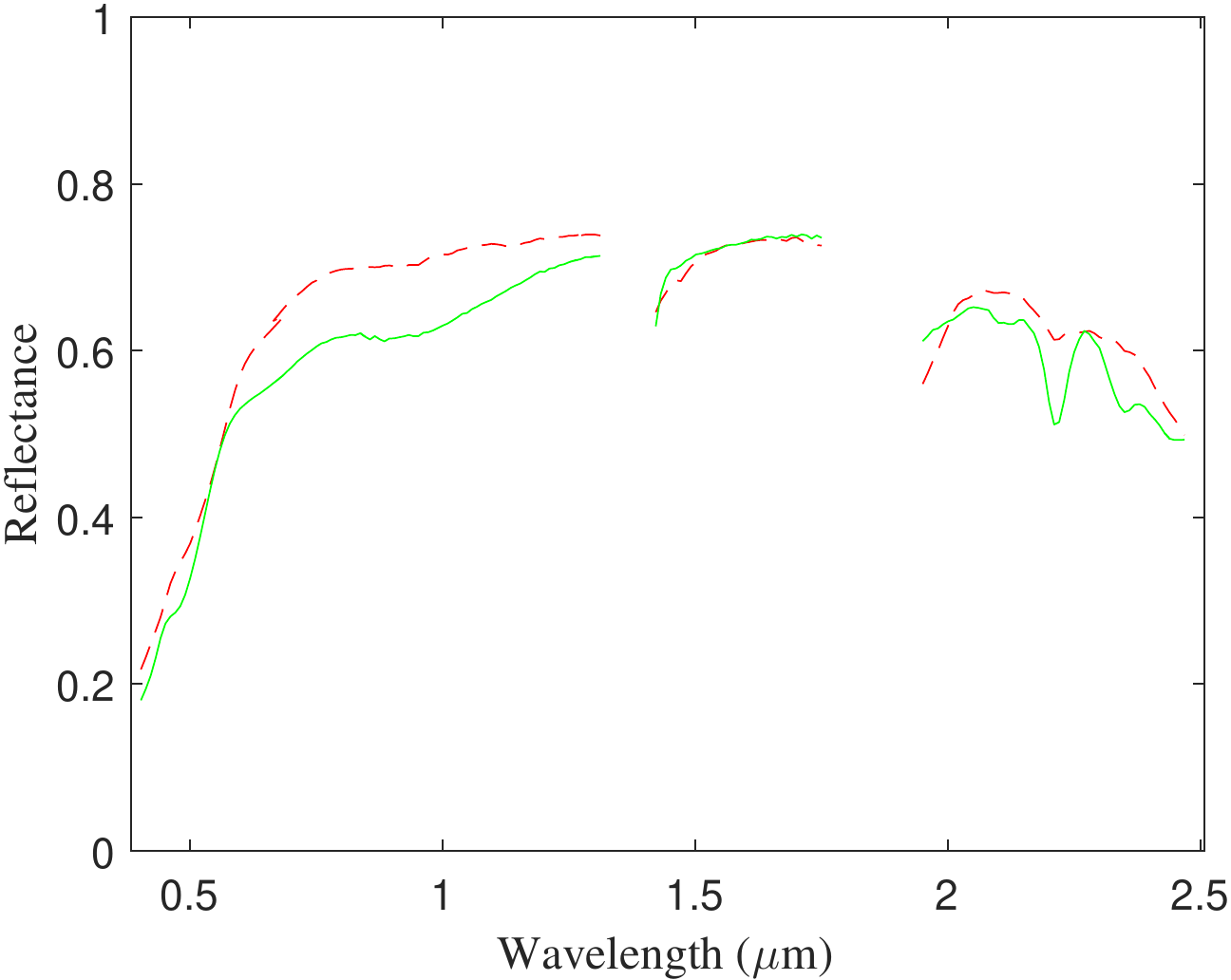}}}
\\
\mbox{
\subfigure[]{\includegraphics[width=2.45cm]{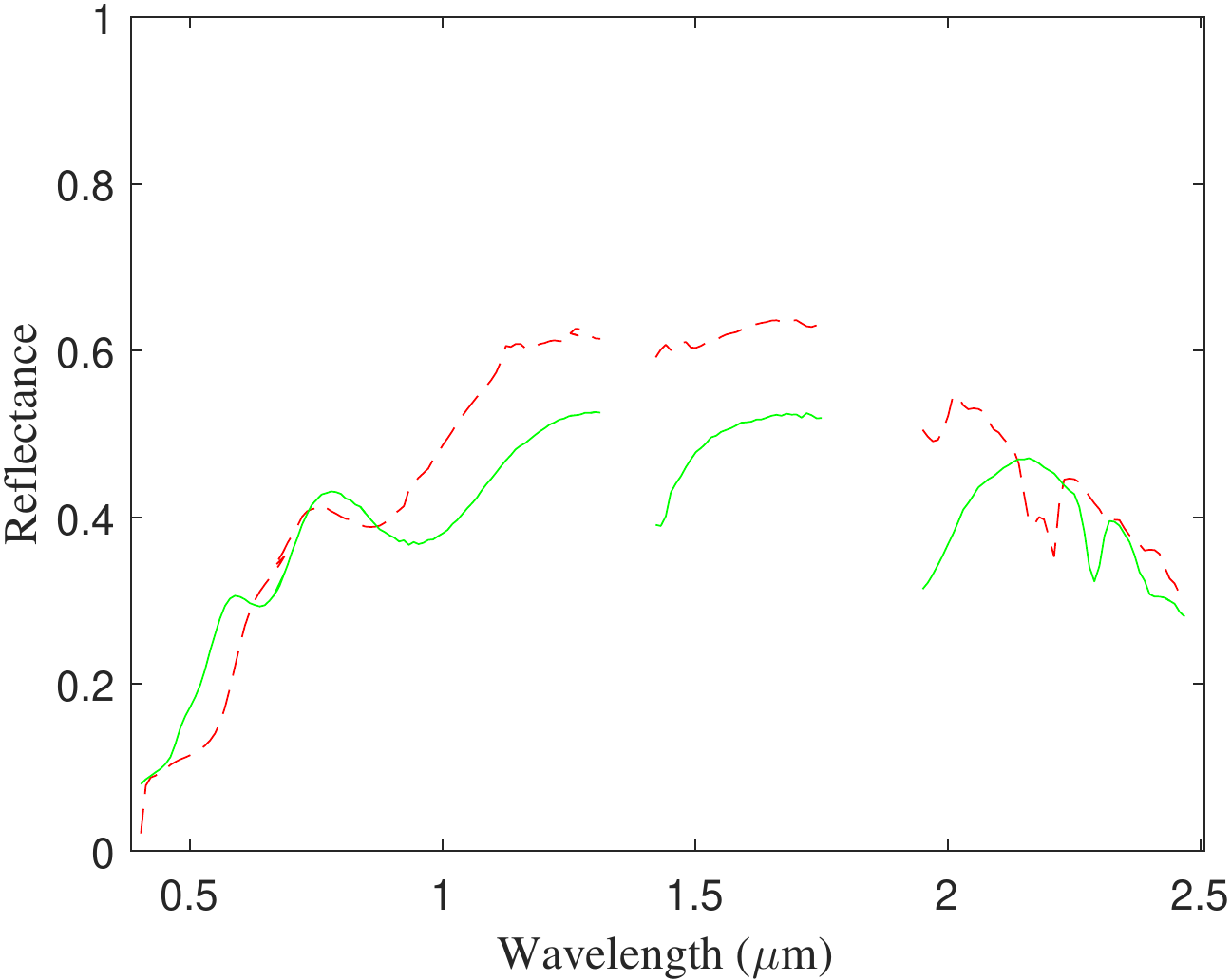}}
\subfigure[]{\includegraphics[width=2.45cm]{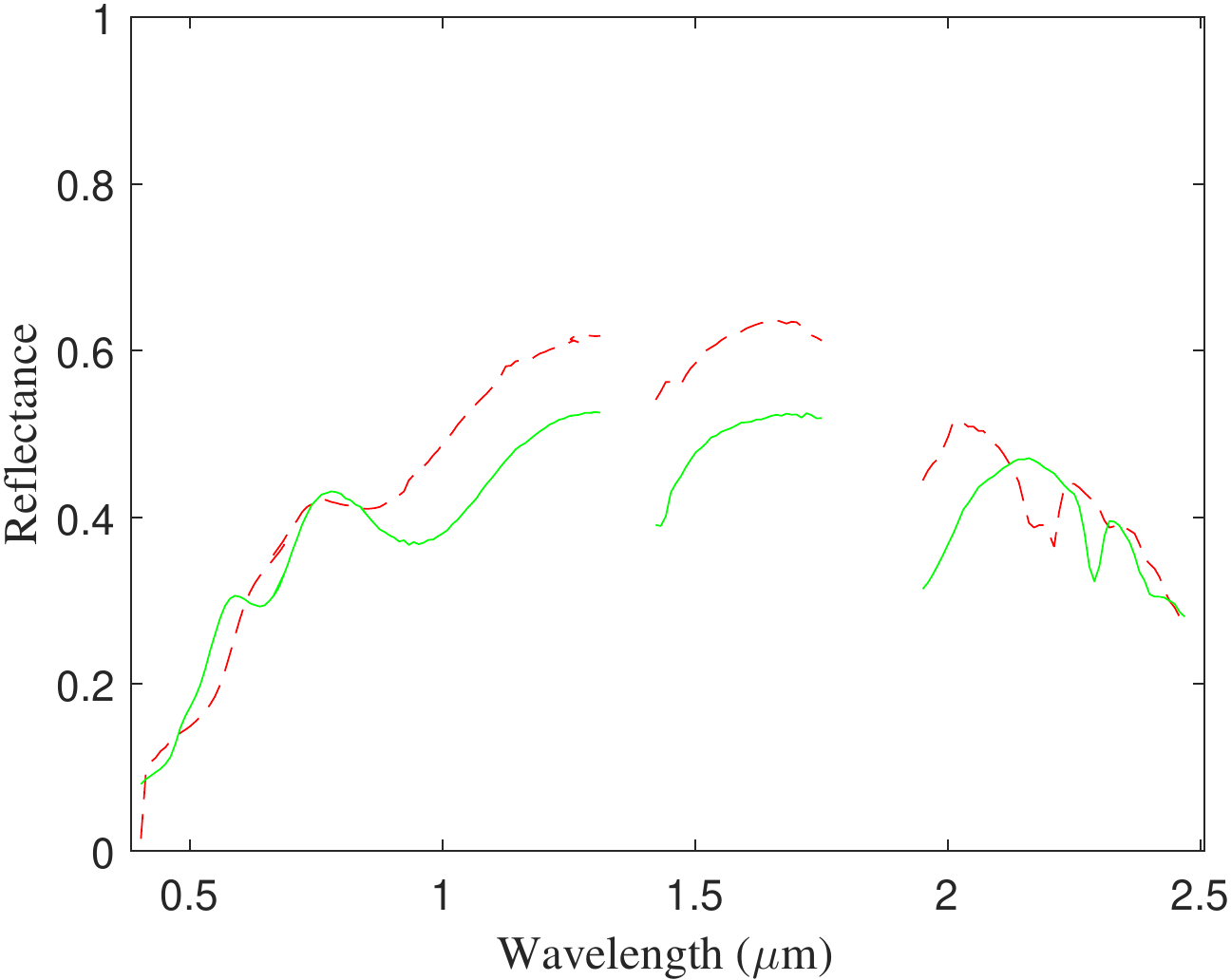}}
\subfigure[]{\includegraphics[width=2.45cm]{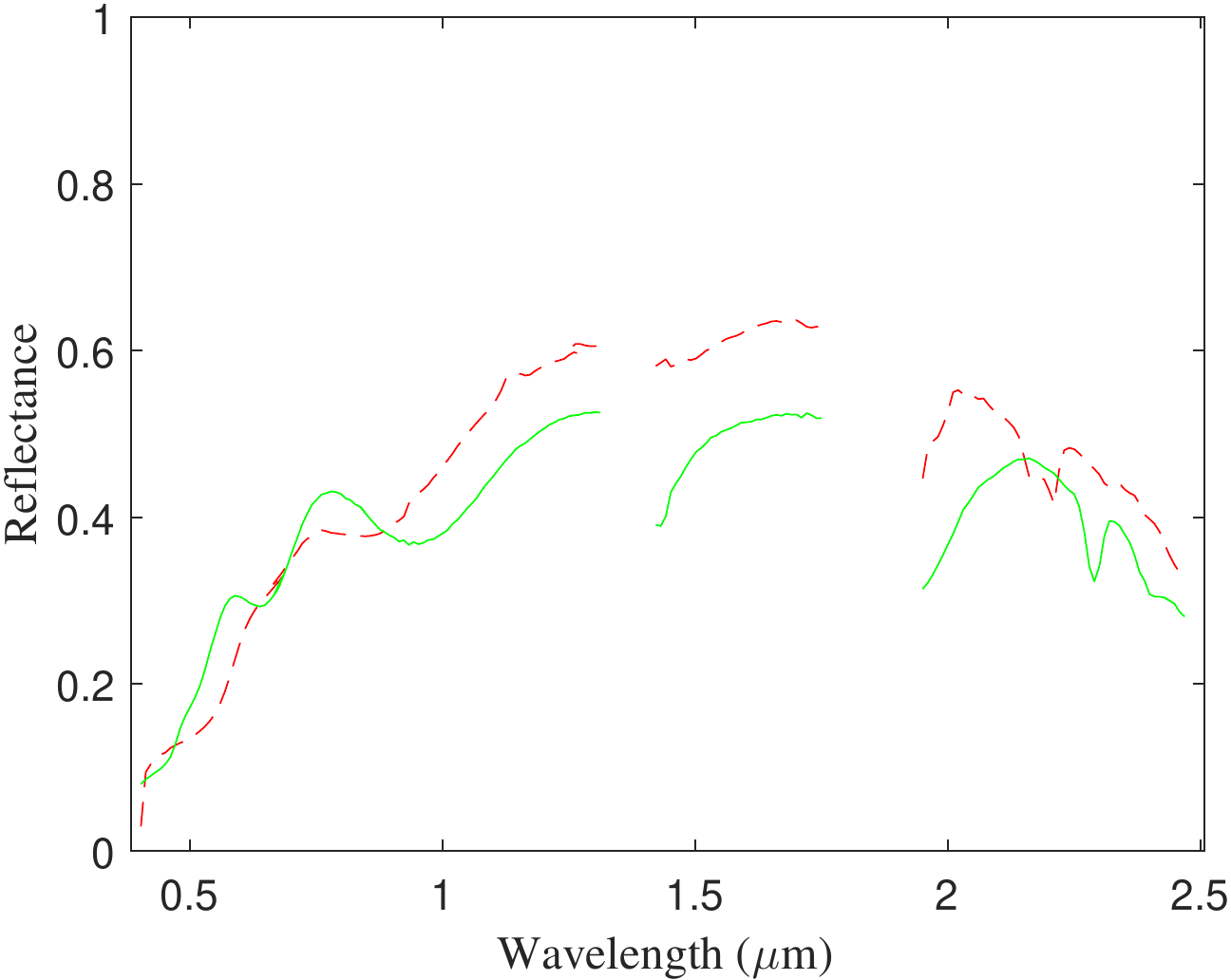}}
\subfigure[]{\includegraphics[width=2.45cm]{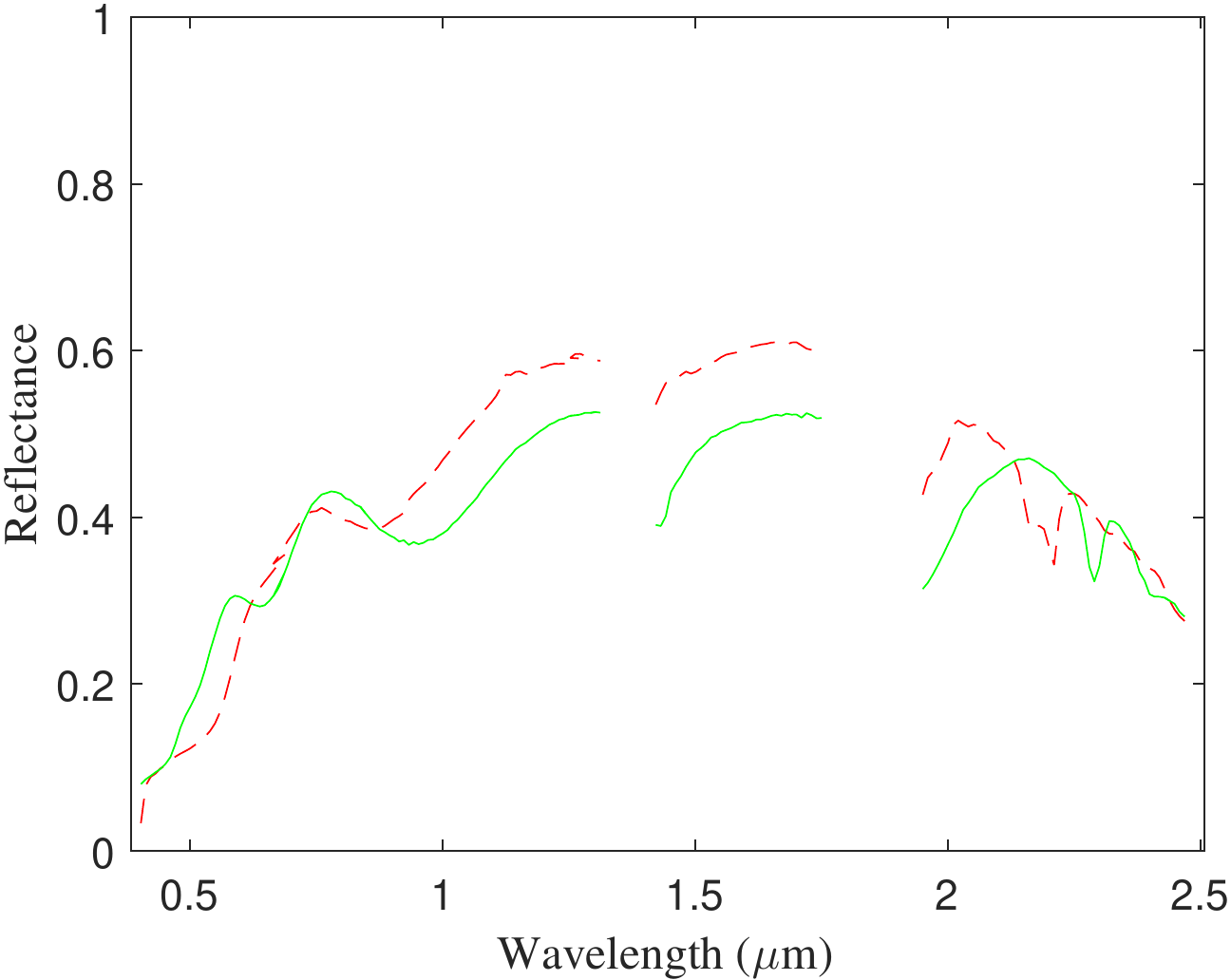}}
\subfigure[]{\includegraphics[width=2.45cm]{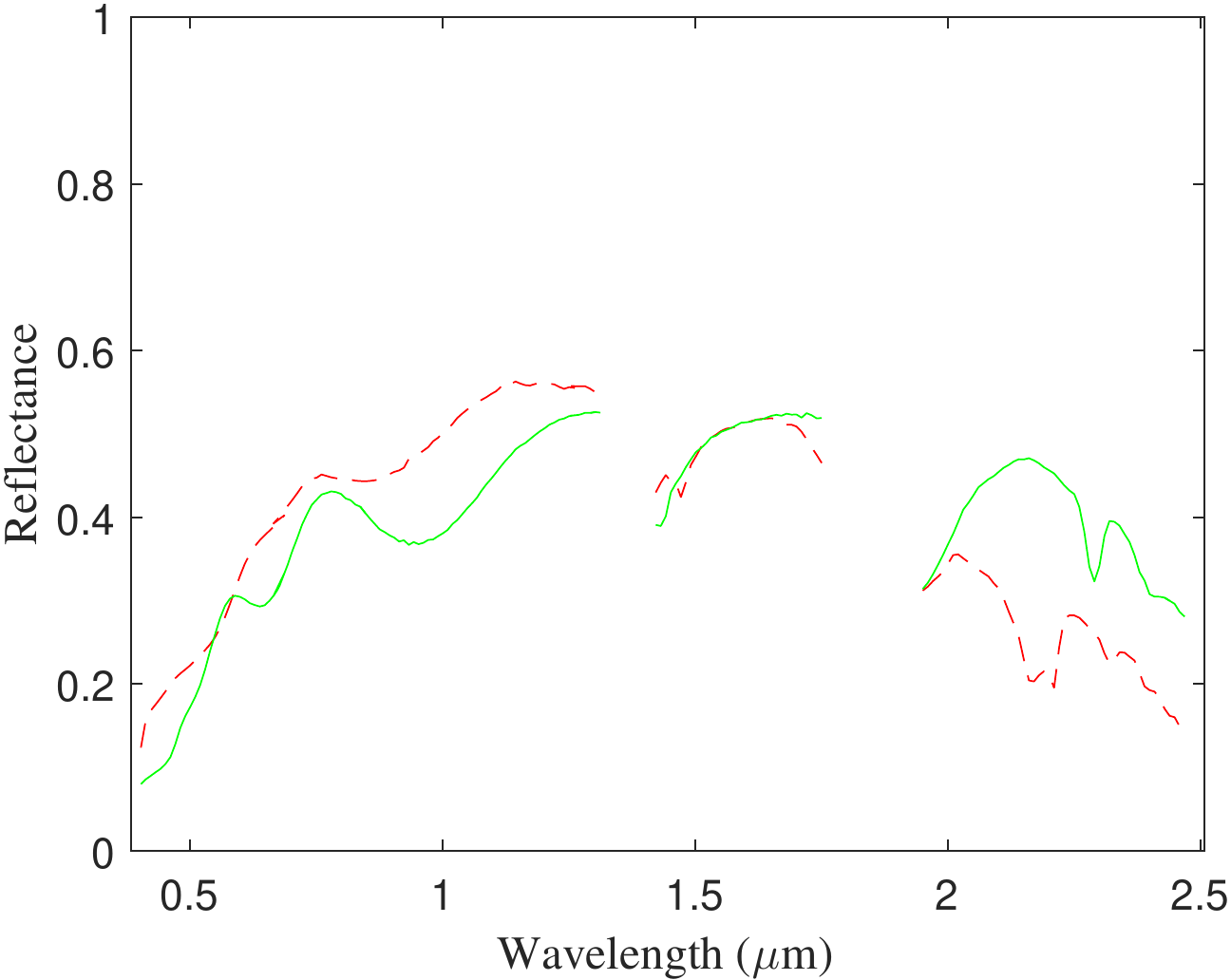}}
\subfigure[]{\includegraphics[width=2.45cm]{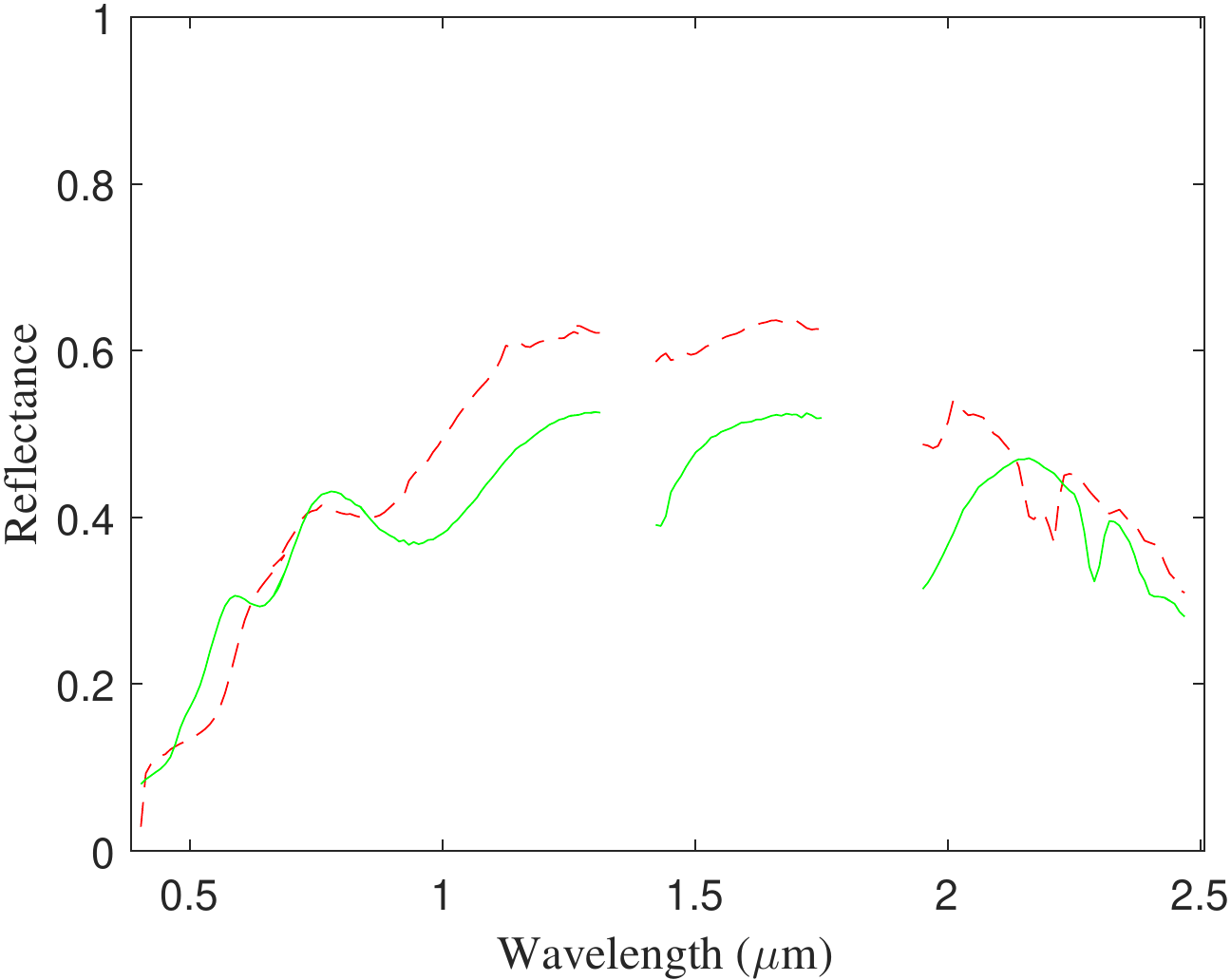}}
\subfigure[]{\includegraphics[width=2.45cm]{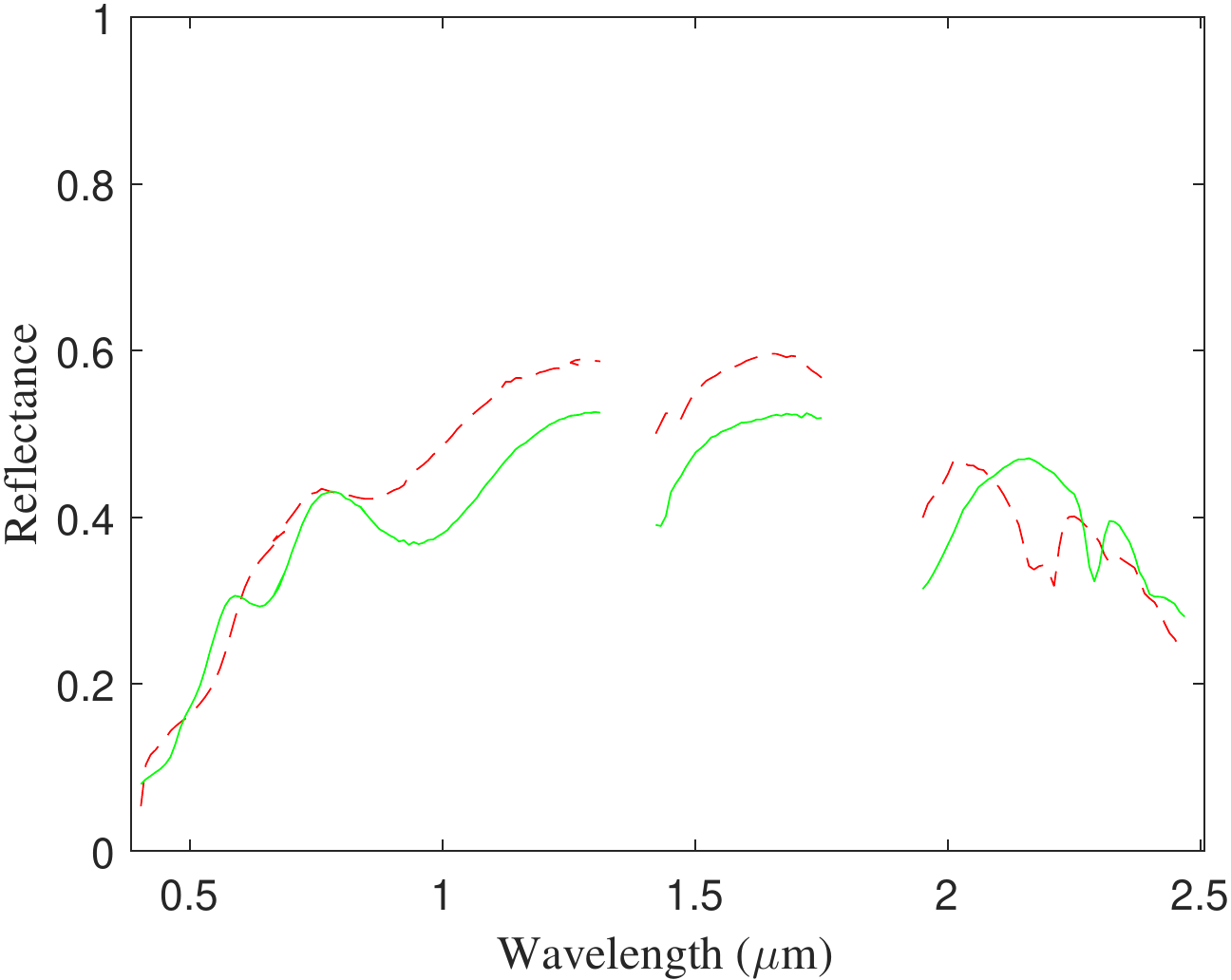}}}
\caption{Comparison of the USGS library spectra (green solid line) with those estimated by different methods (red dash line) of four endmembers on the AVIRIS Cuprite data set. From top to bottom: Buddingtonite GDS85 D-206. Kaolinite KGa-2. Montmorillonite+Illi CM37. Nontronite NG-1.a. From left to right: (a) $L_{1/2}$-NMF. (b) SGSNMF. (c) TV-RSNMF. (d) $L_{1/2}$-RNMF. (e) MV-NTF-TV. (f) MLNMF. (g) SSRDMF.}
\label{fig:5}
\end{figure*}

\begin{figure*}[!t]
\centering
\mbox{
{\includegraphics[width=2.45cm, height=2.74cm]{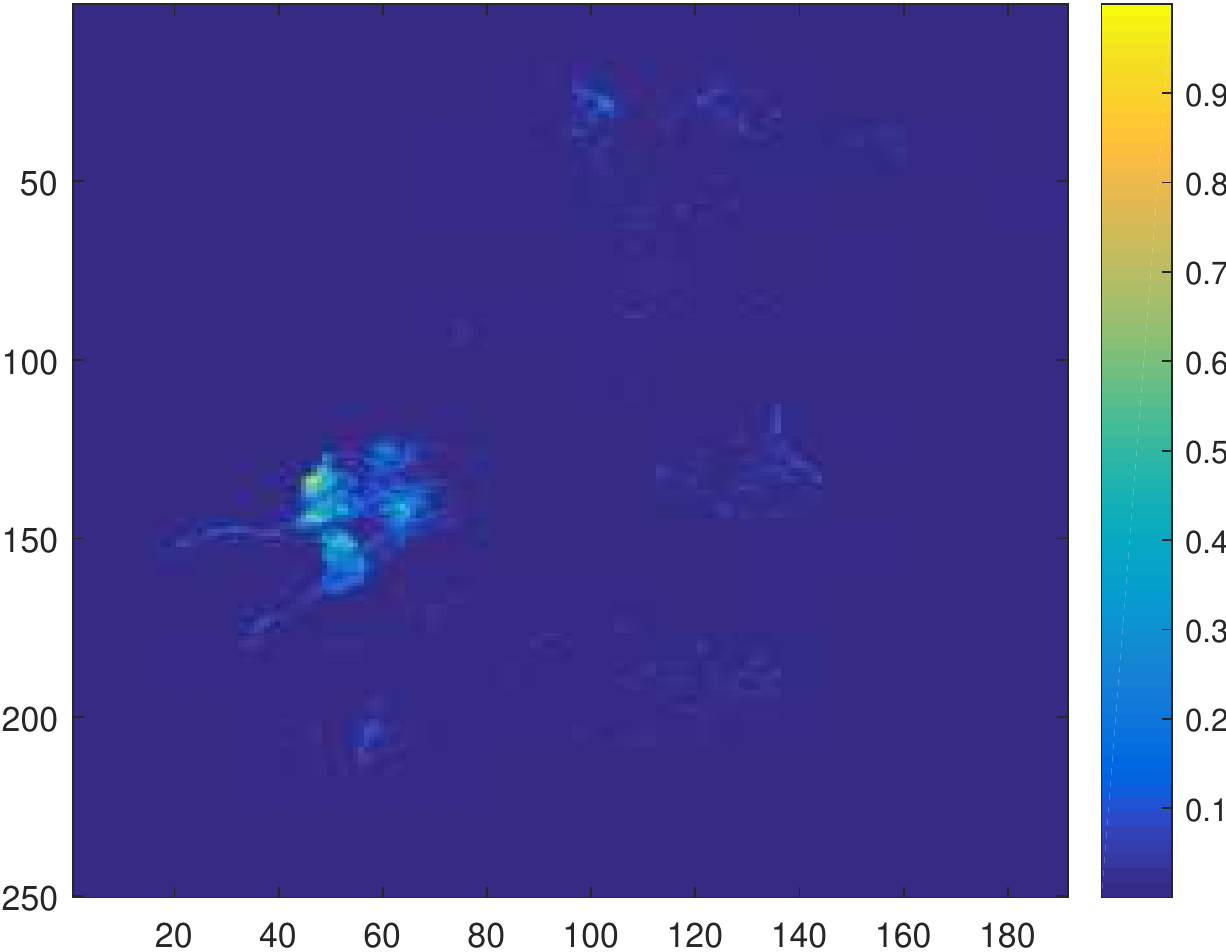}}
{\includegraphics[width=2.45cm, height=2.74cm]{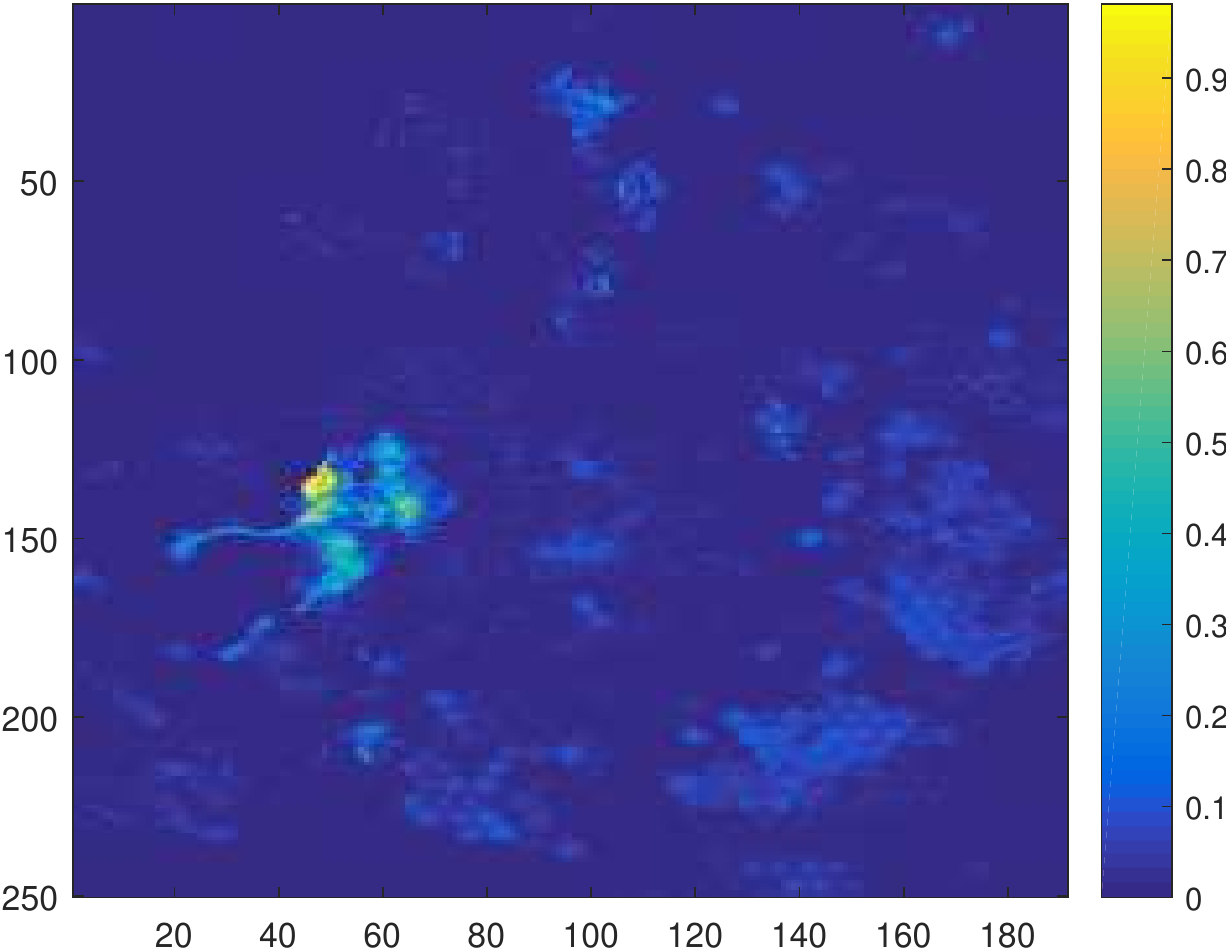}}
{\includegraphics[width=2.45cm, height=2.74cm]{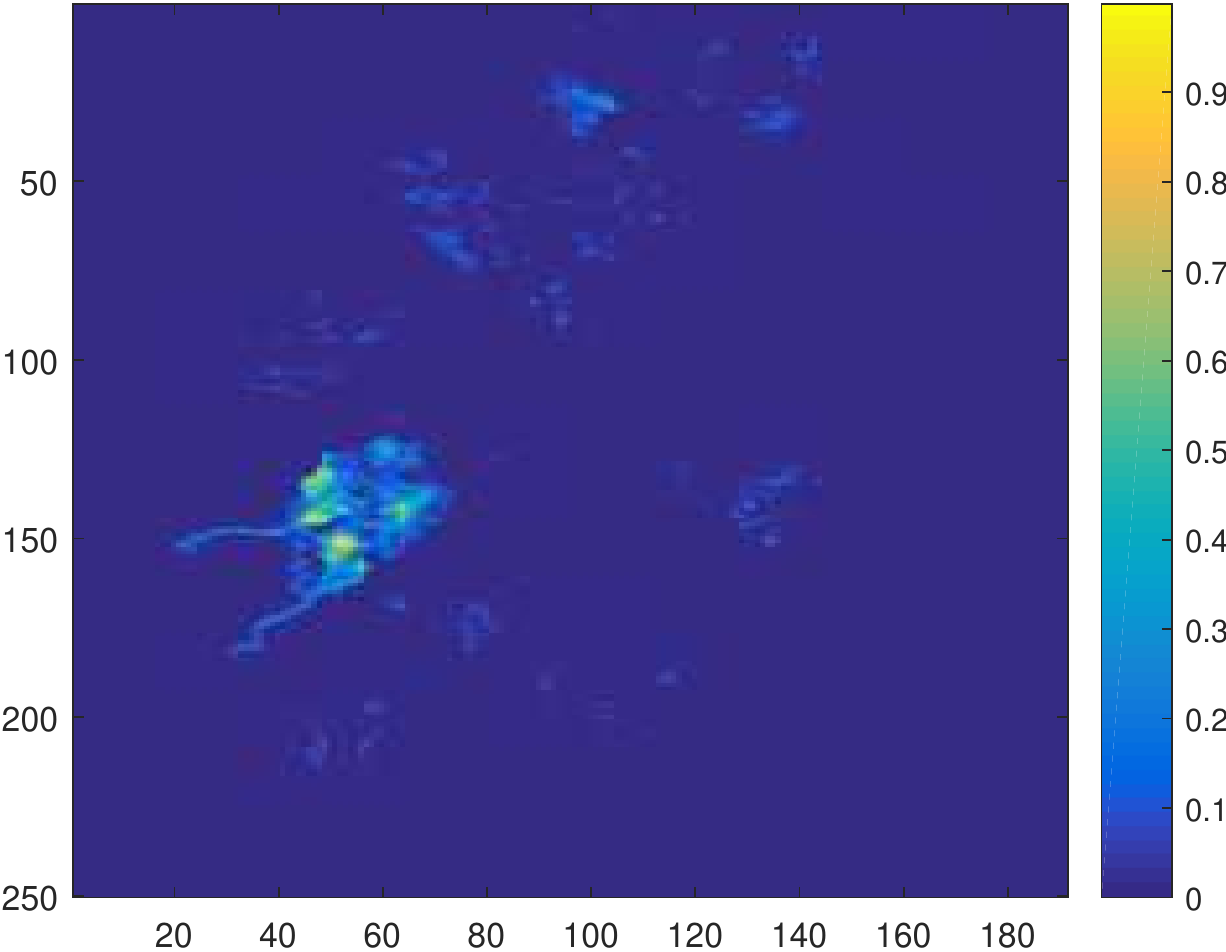}}
{\includegraphics[width=2.45cm, height=2.74cm]{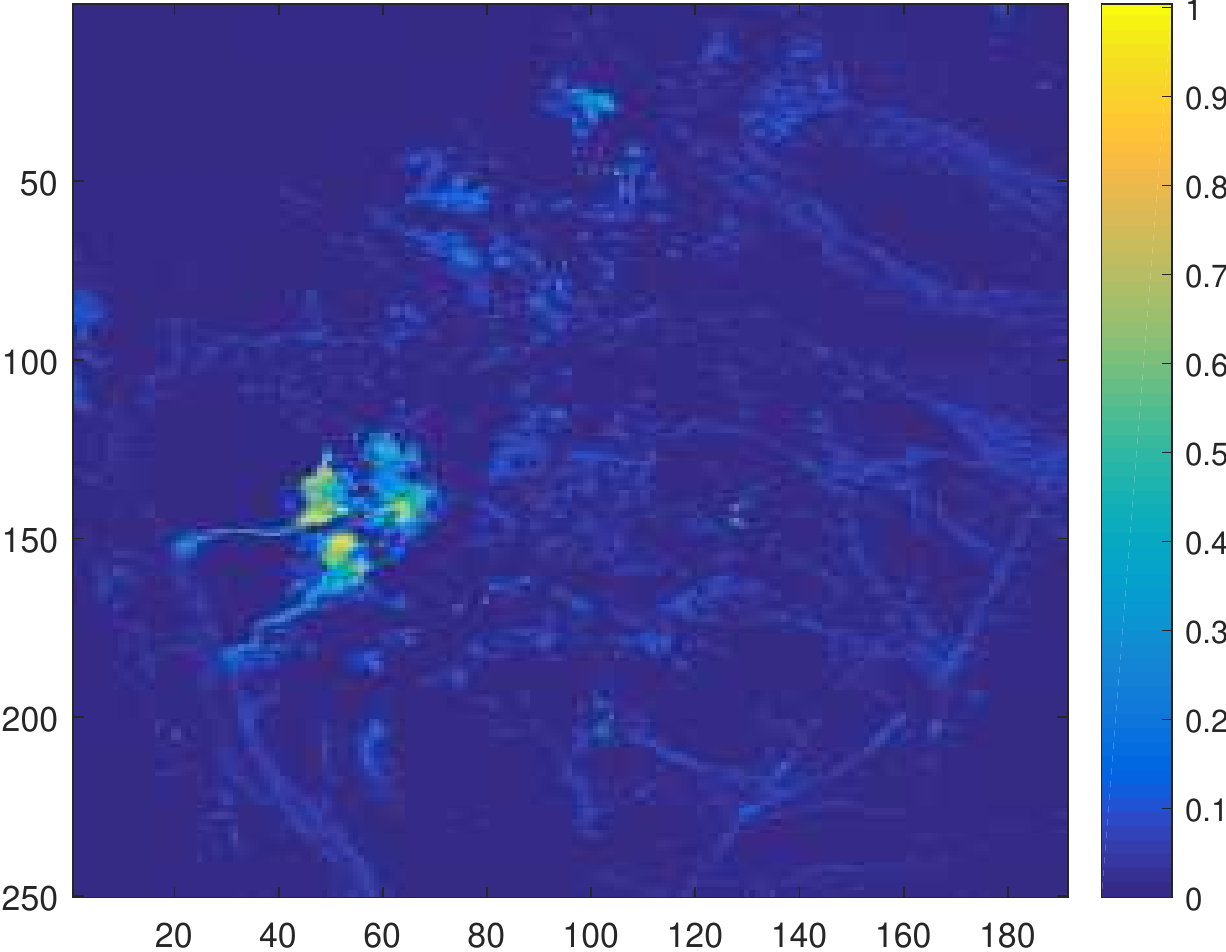}}
{\includegraphics[width=2.45cm, height=2.74cm]{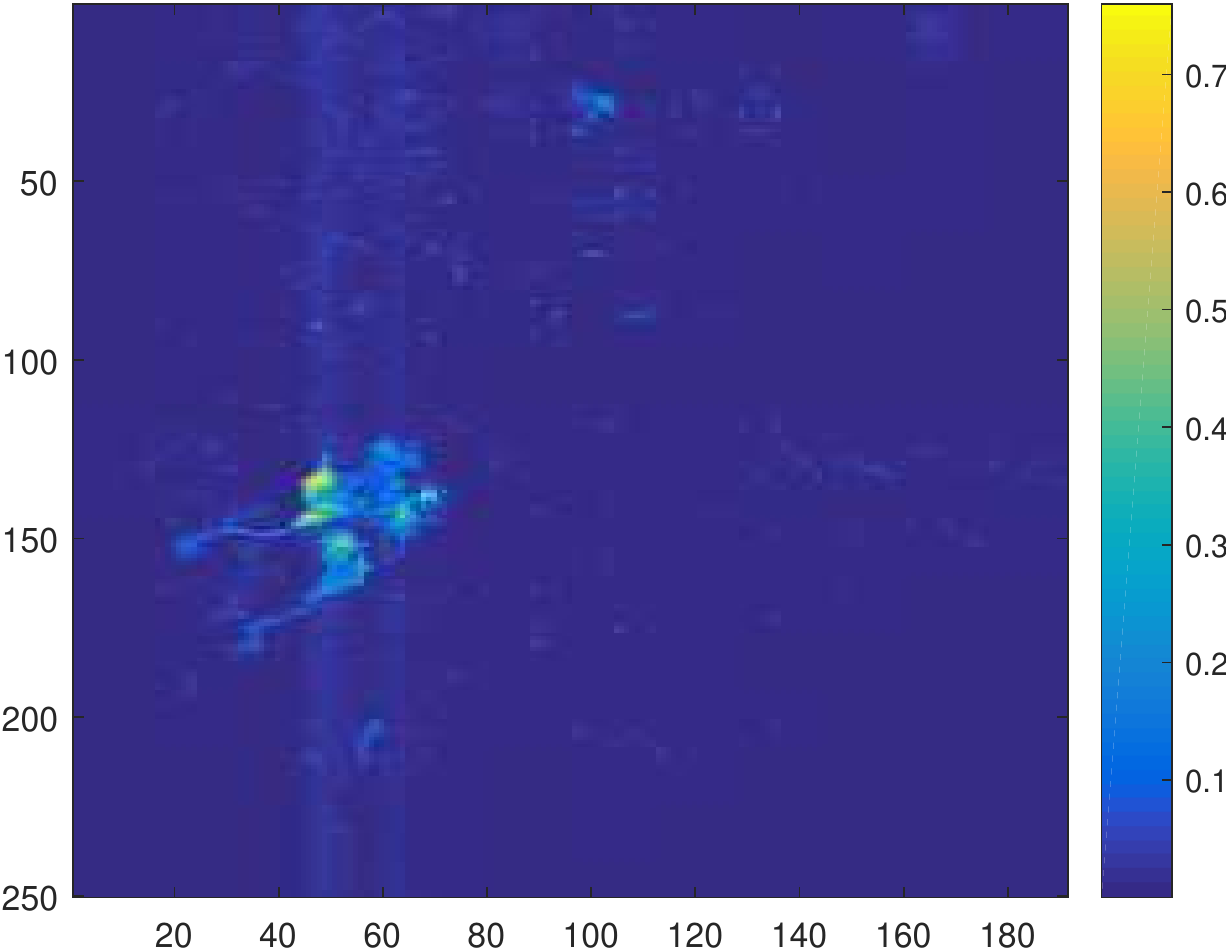}}
{\includegraphics[width=2.45cm, height=2.74cm]{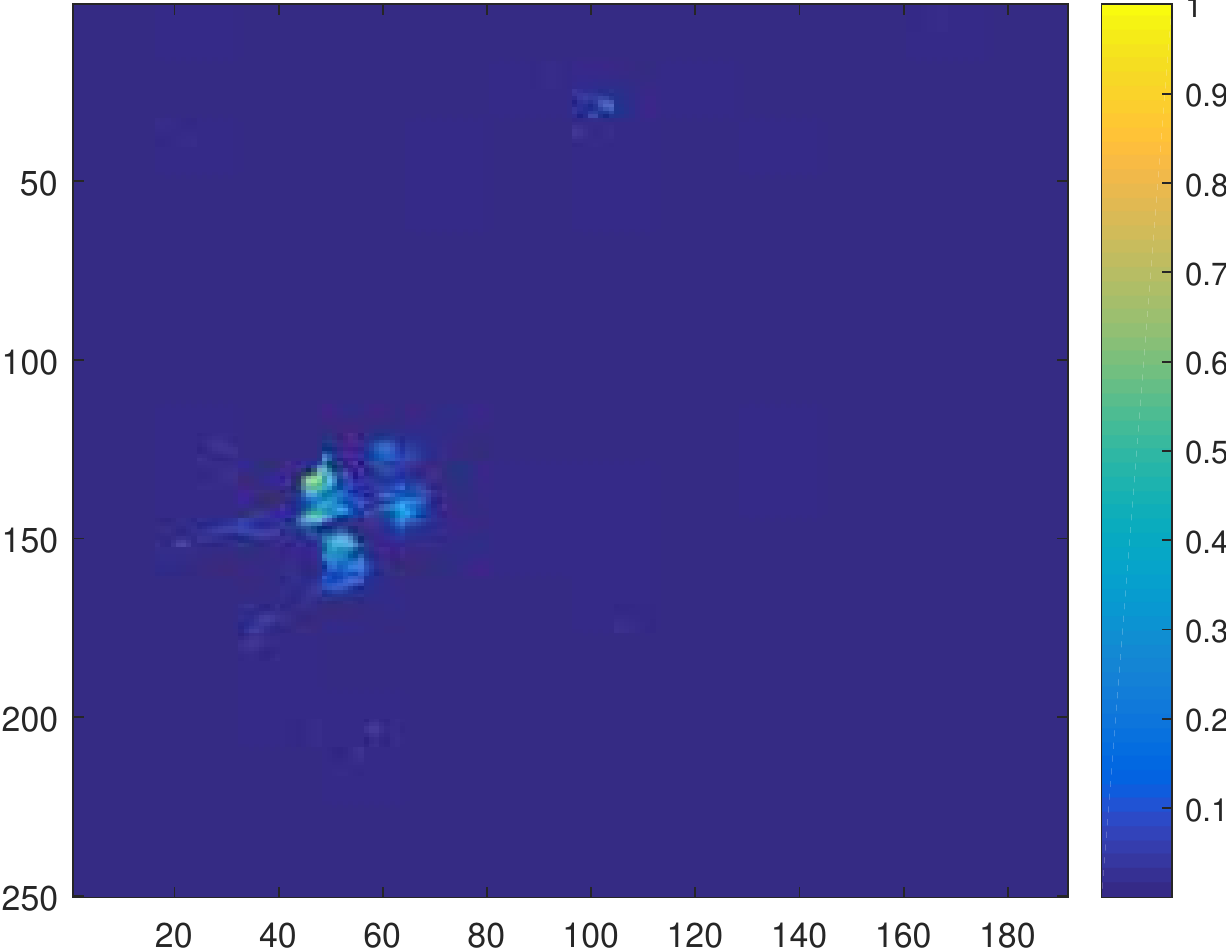}}
{\includegraphics[width=2.45cm, height=2.74cm]{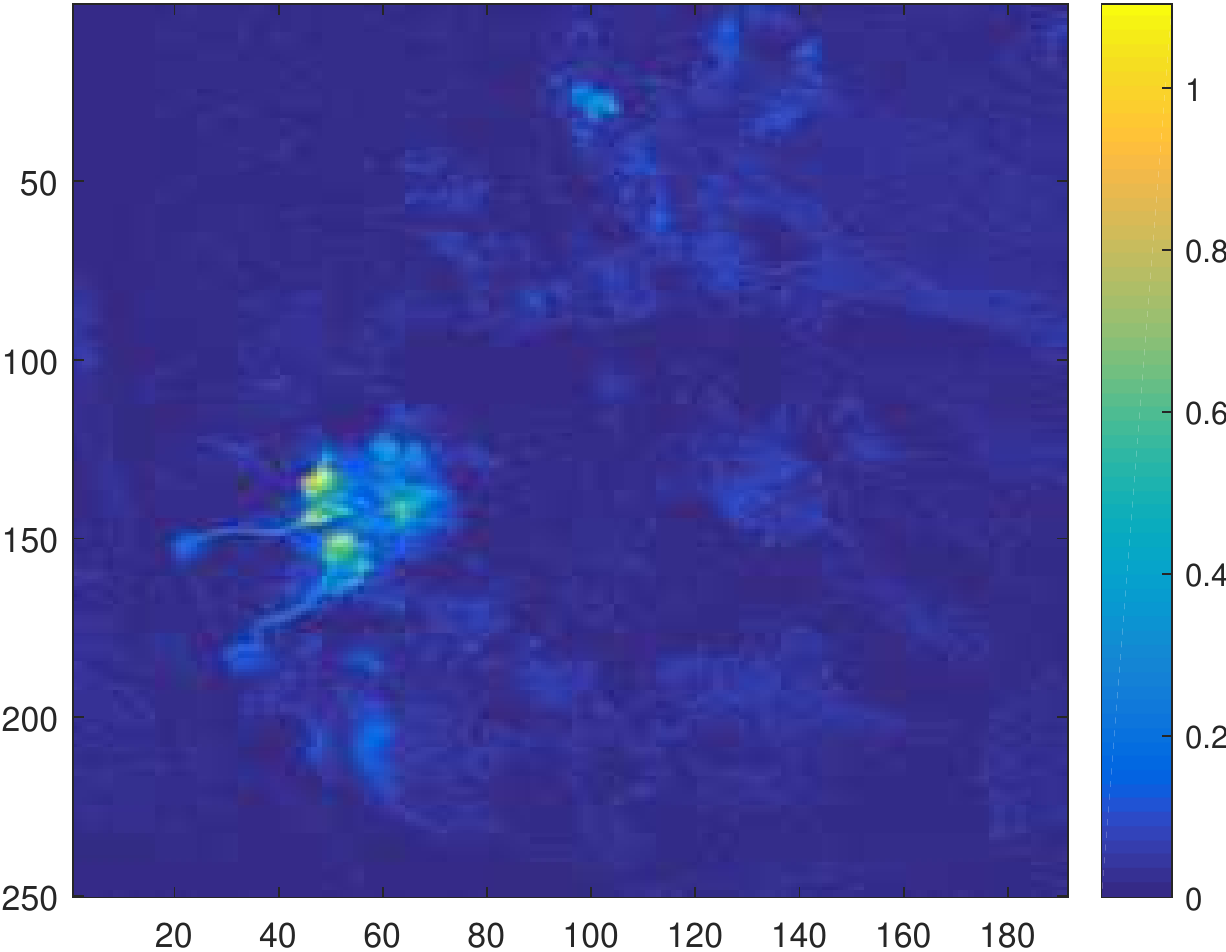}}}
\\
\mbox{
{\includegraphics[width=2.45cm, height=2.74cm]{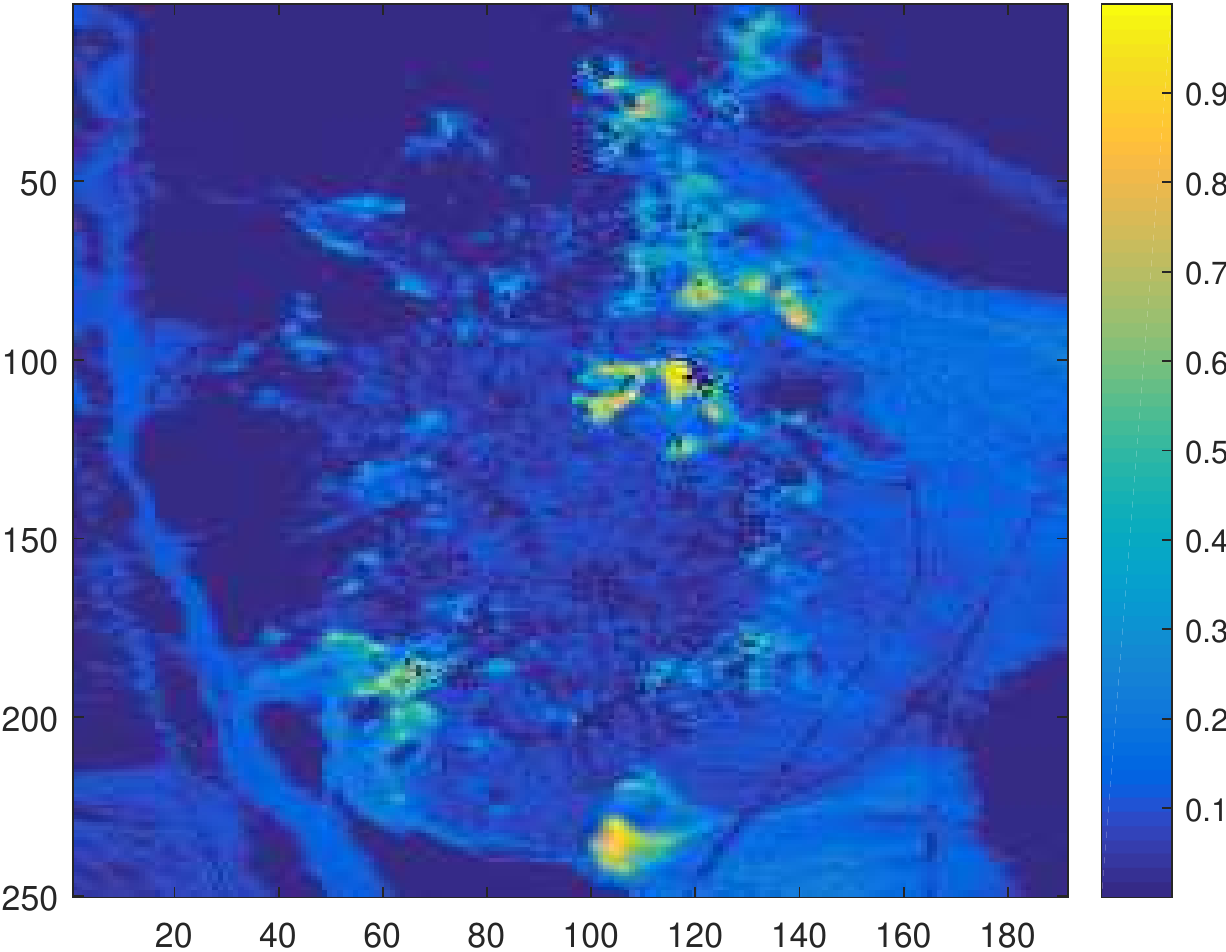}}
{\includegraphics[width=2.45cm, height=2.74cm]{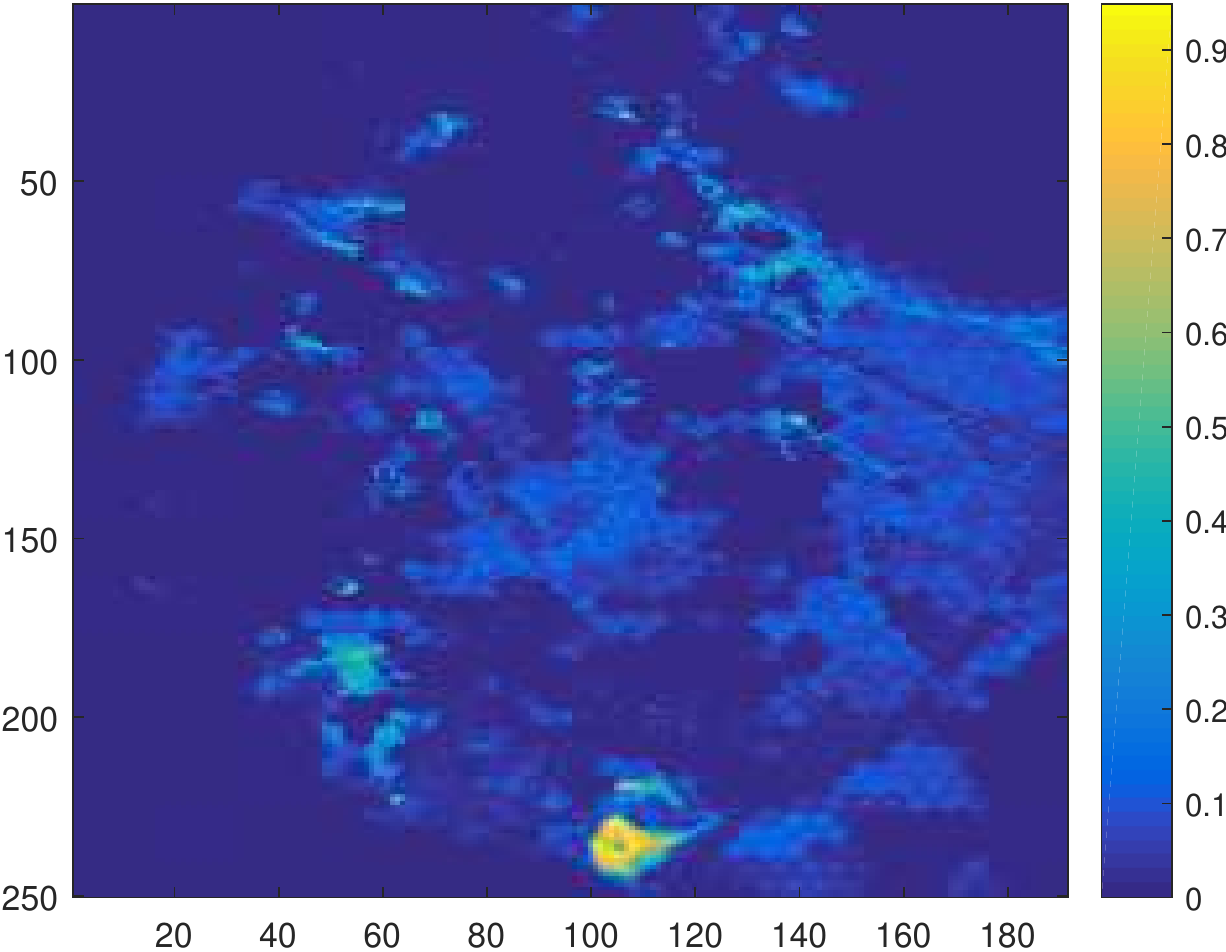}}
{\includegraphics[width=2.45cm, height=2.74cm]{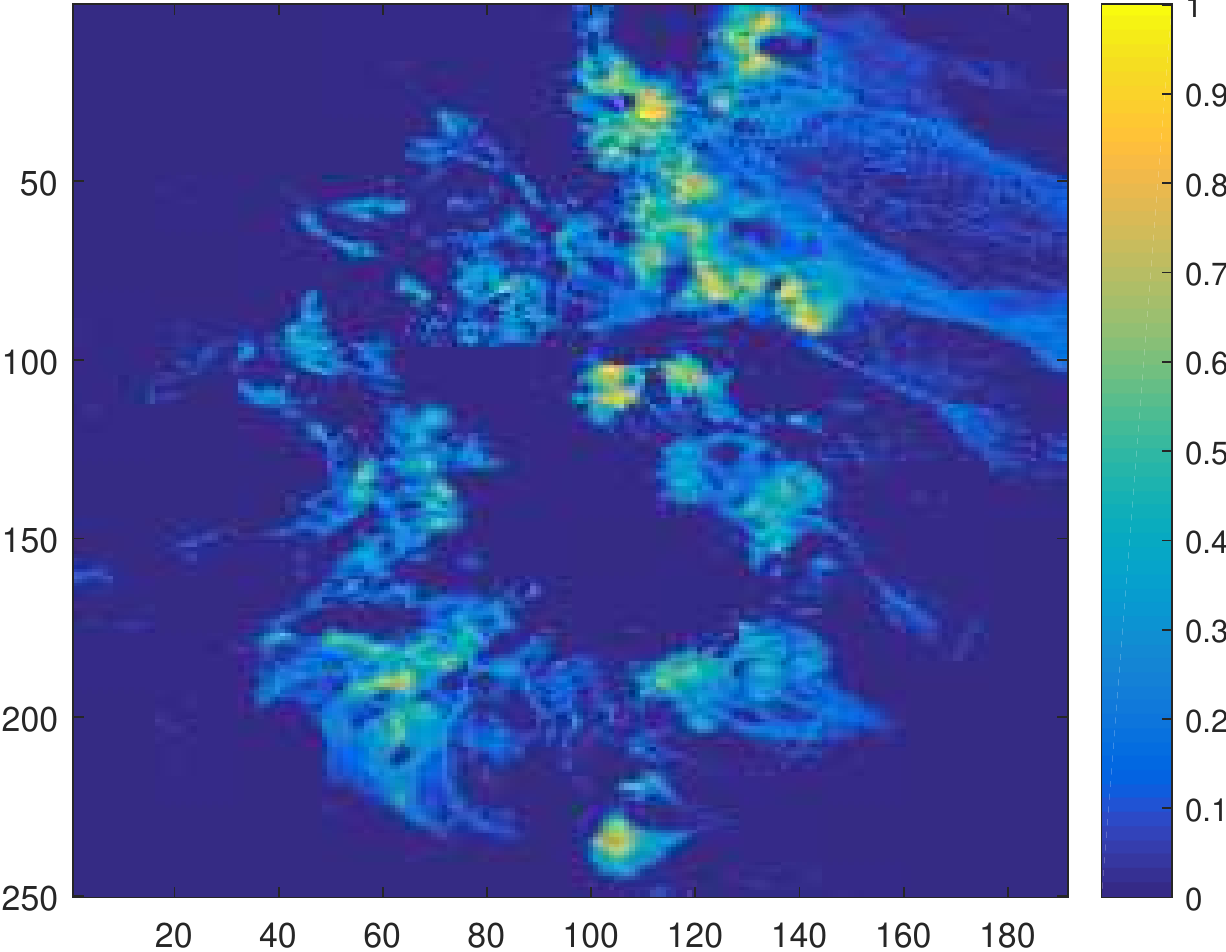}}
{\includegraphics[width=2.45cm, height=2.74cm]{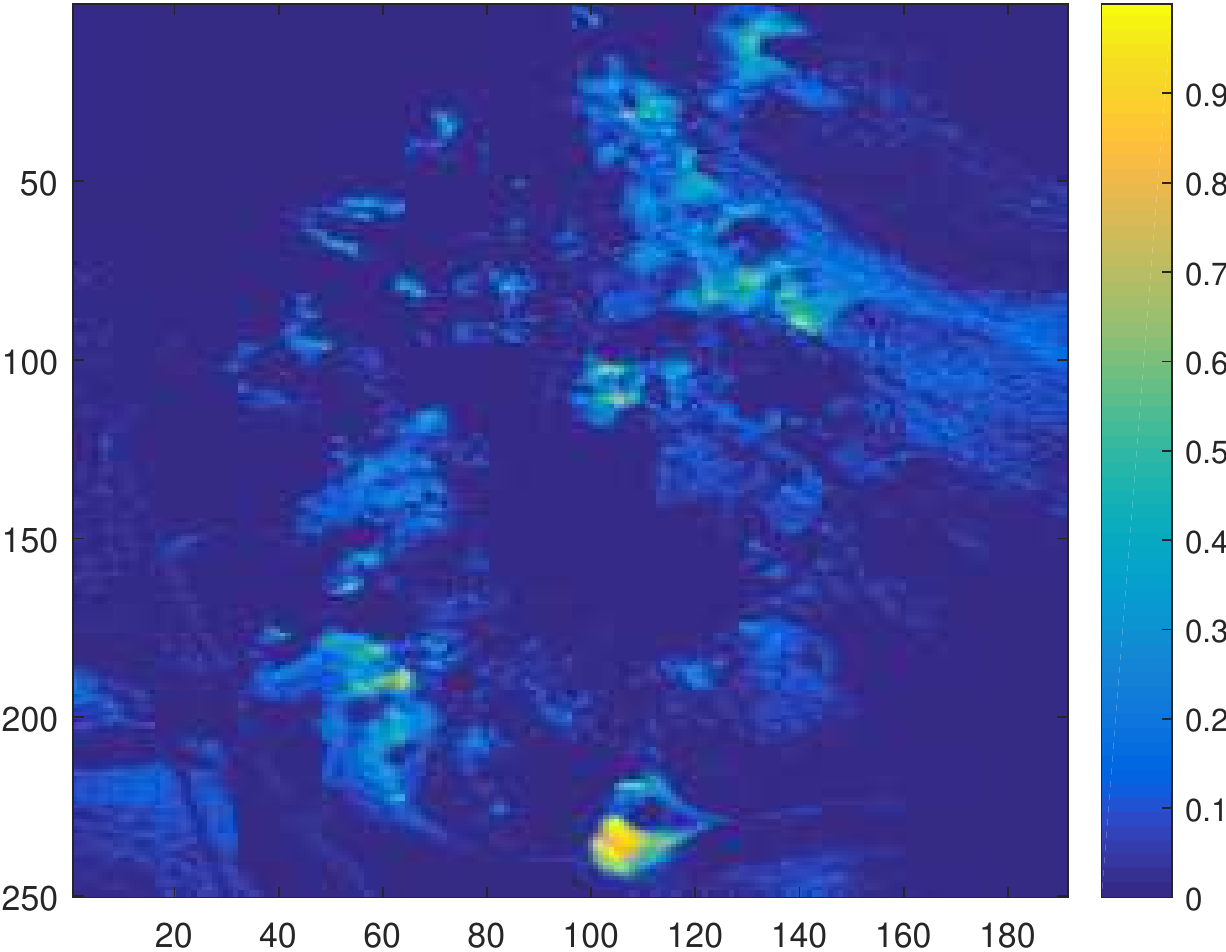}}
{\includegraphics[width=2.45cm, height=2.74cm]{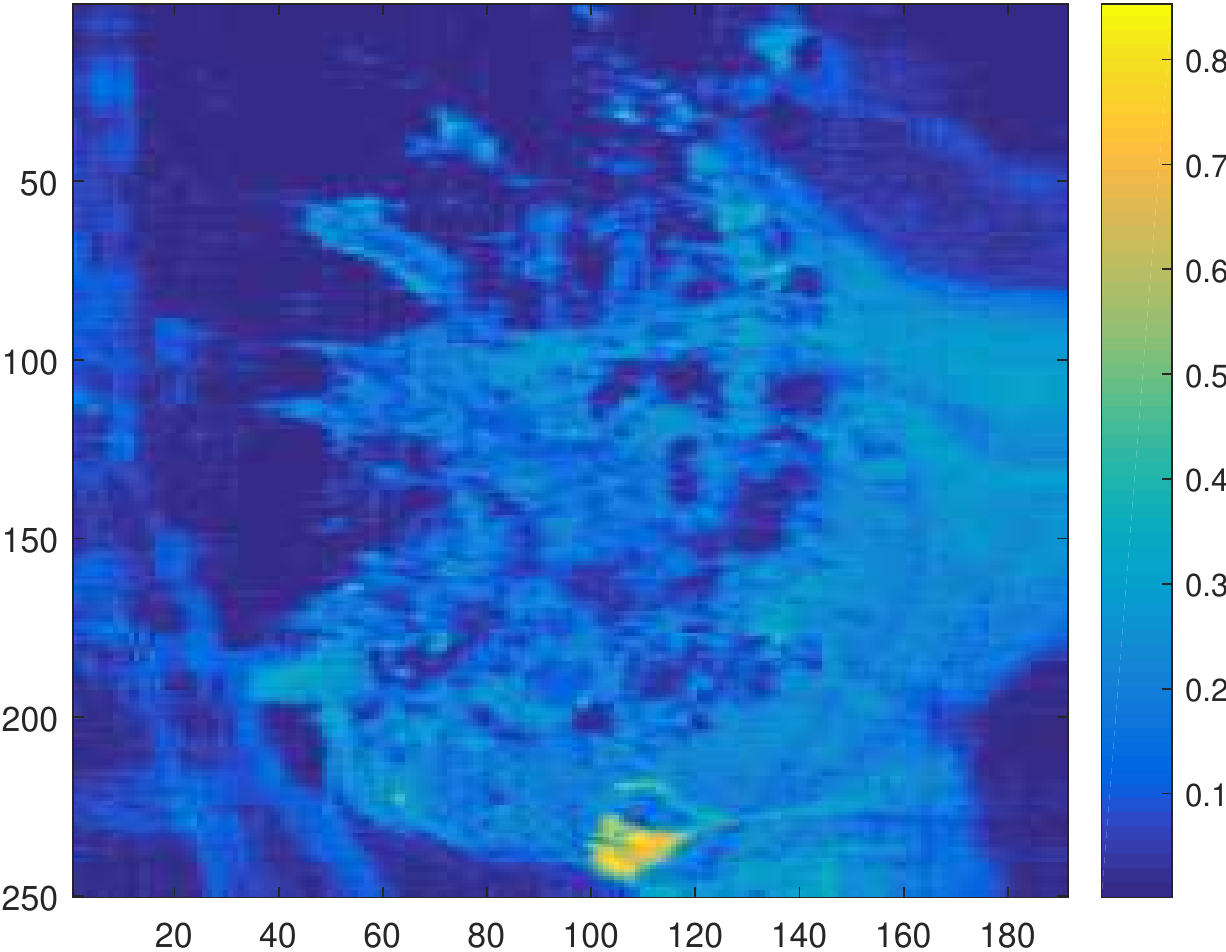}}
{\includegraphics[width=2.45cm, height=2.74cm]{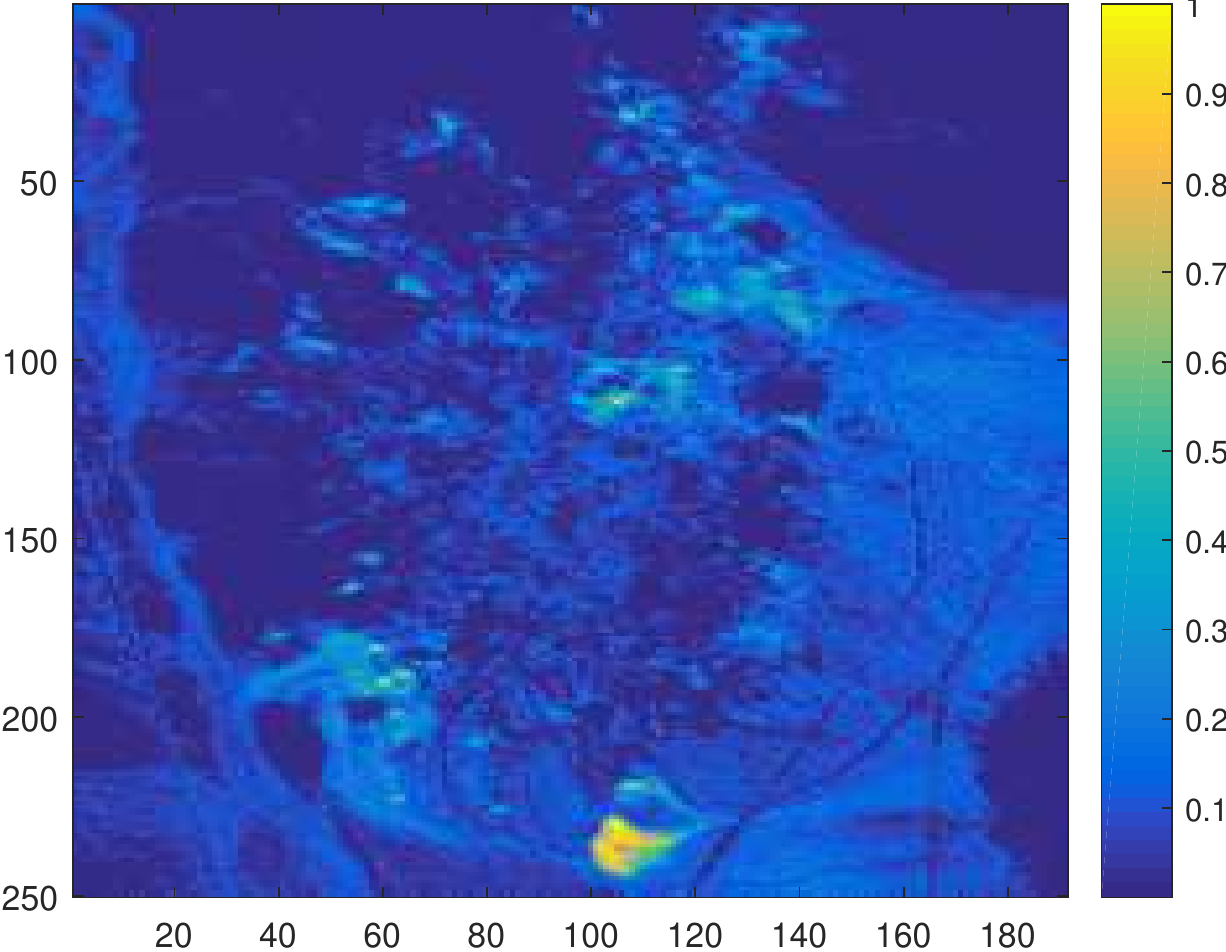}}
{\includegraphics[width=2.45cm, height=2.74cm]{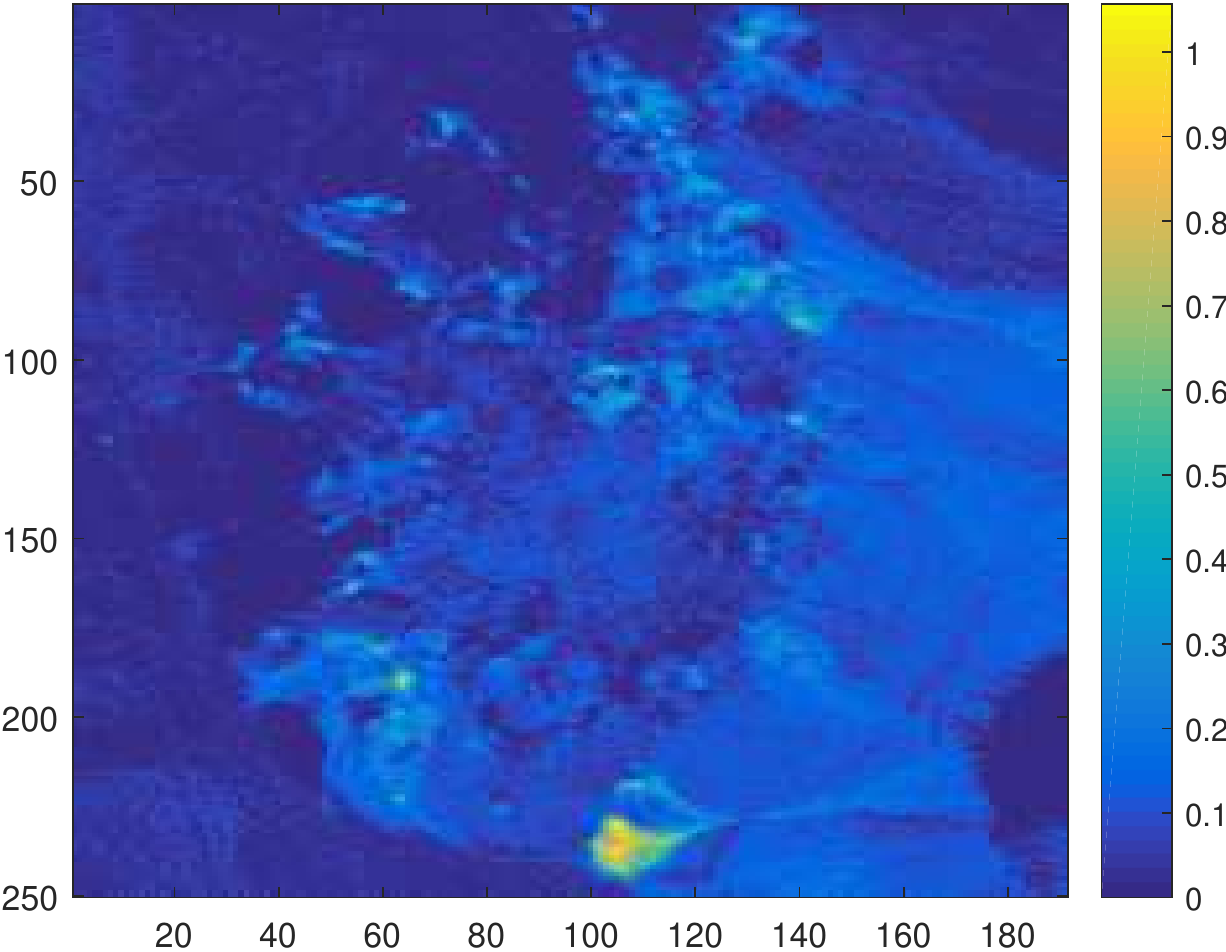}}}
\\
\mbox{
{\includegraphics[width=2.45cm, height=2.74cm]{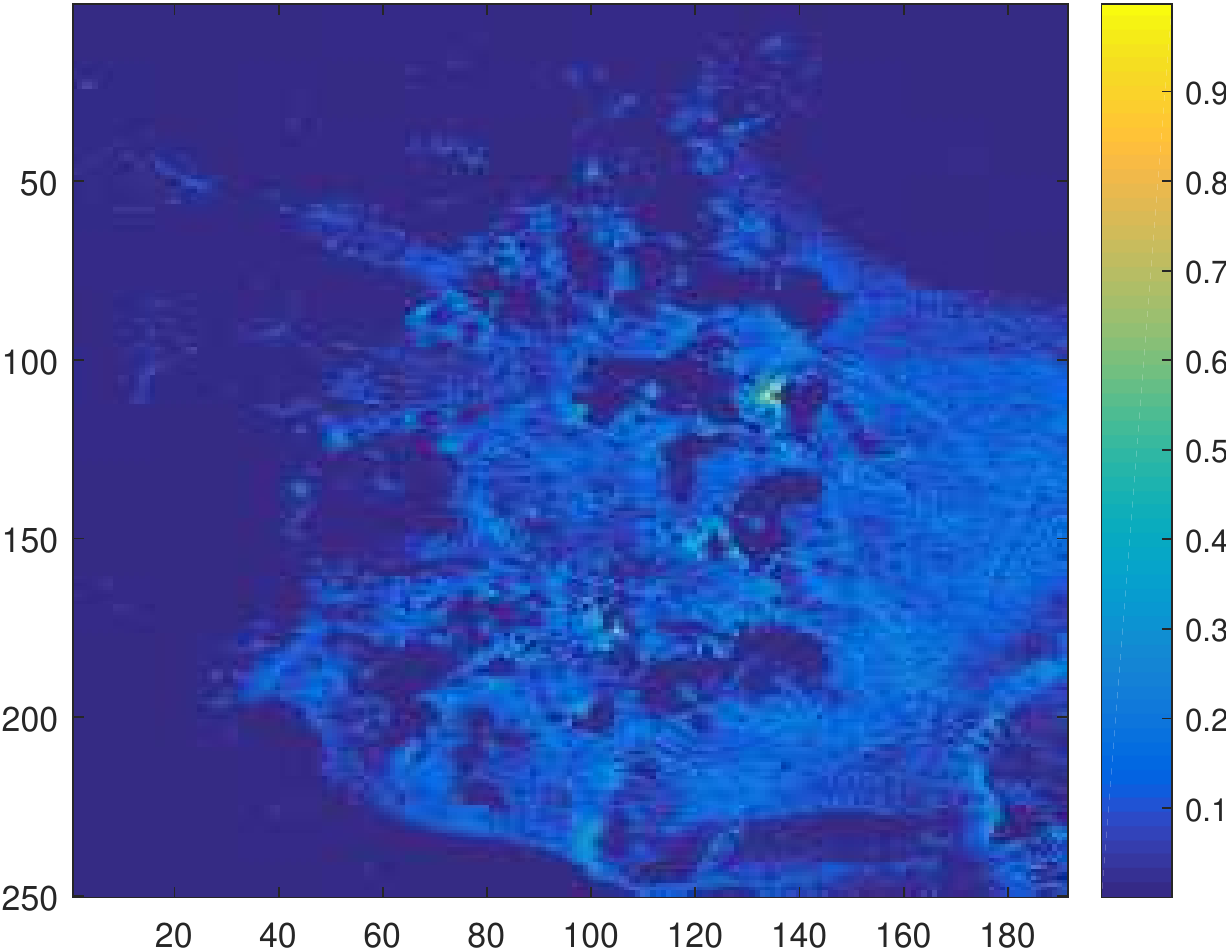}}
{\includegraphics[width=2.45cm, height=2.74cm]{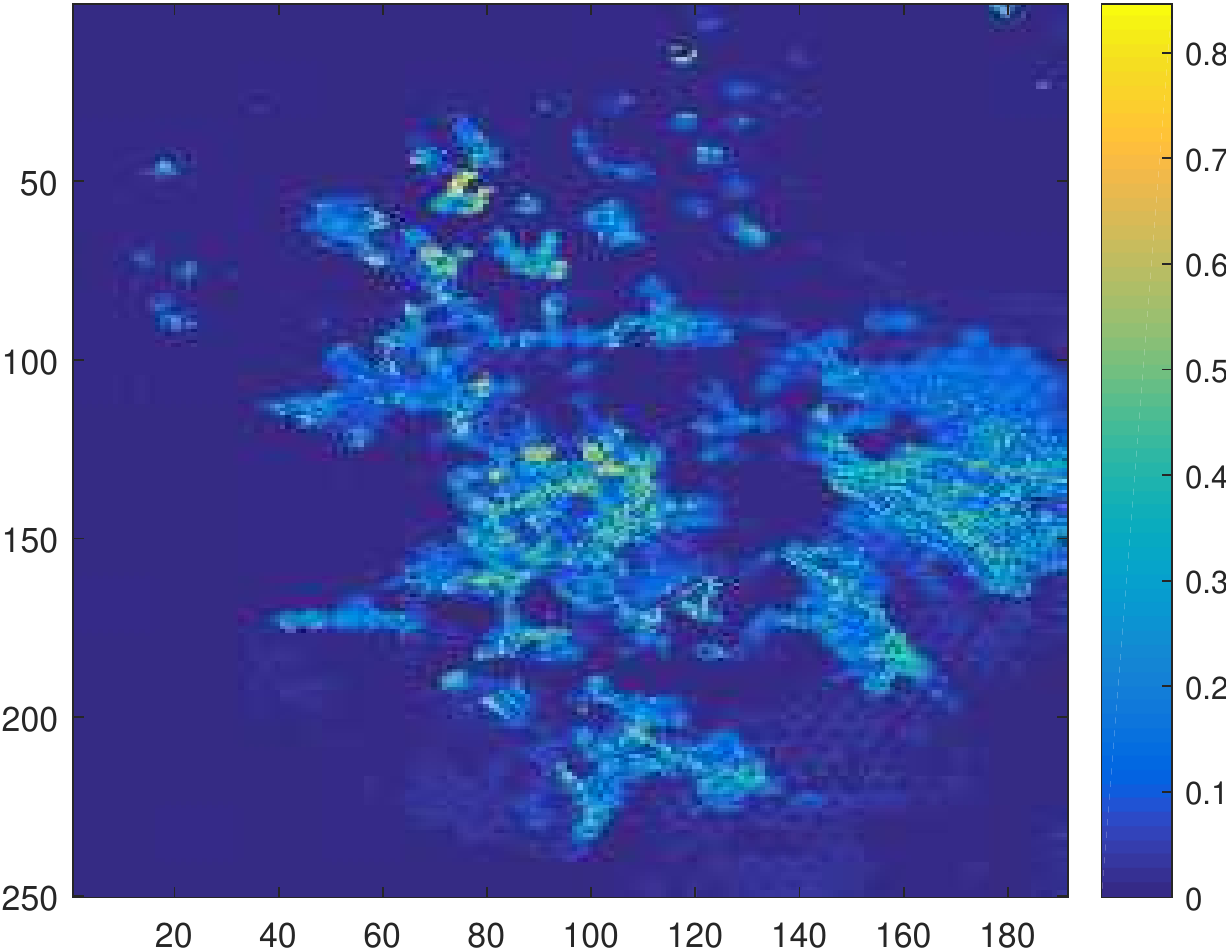}}
{\includegraphics[width=2.45cm, height=2.74cm]{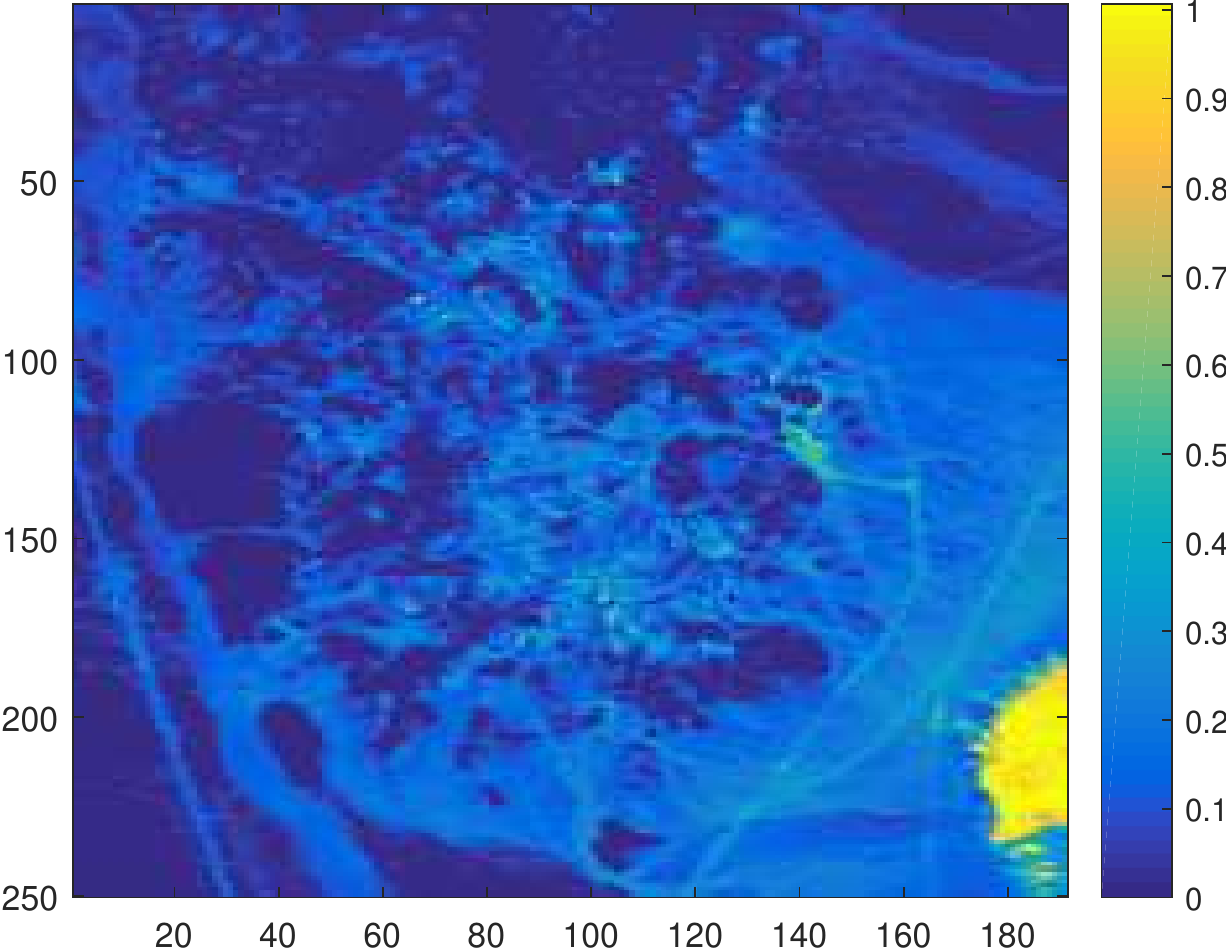}}
{\includegraphics[width=2.45cm, height=2.74cm]{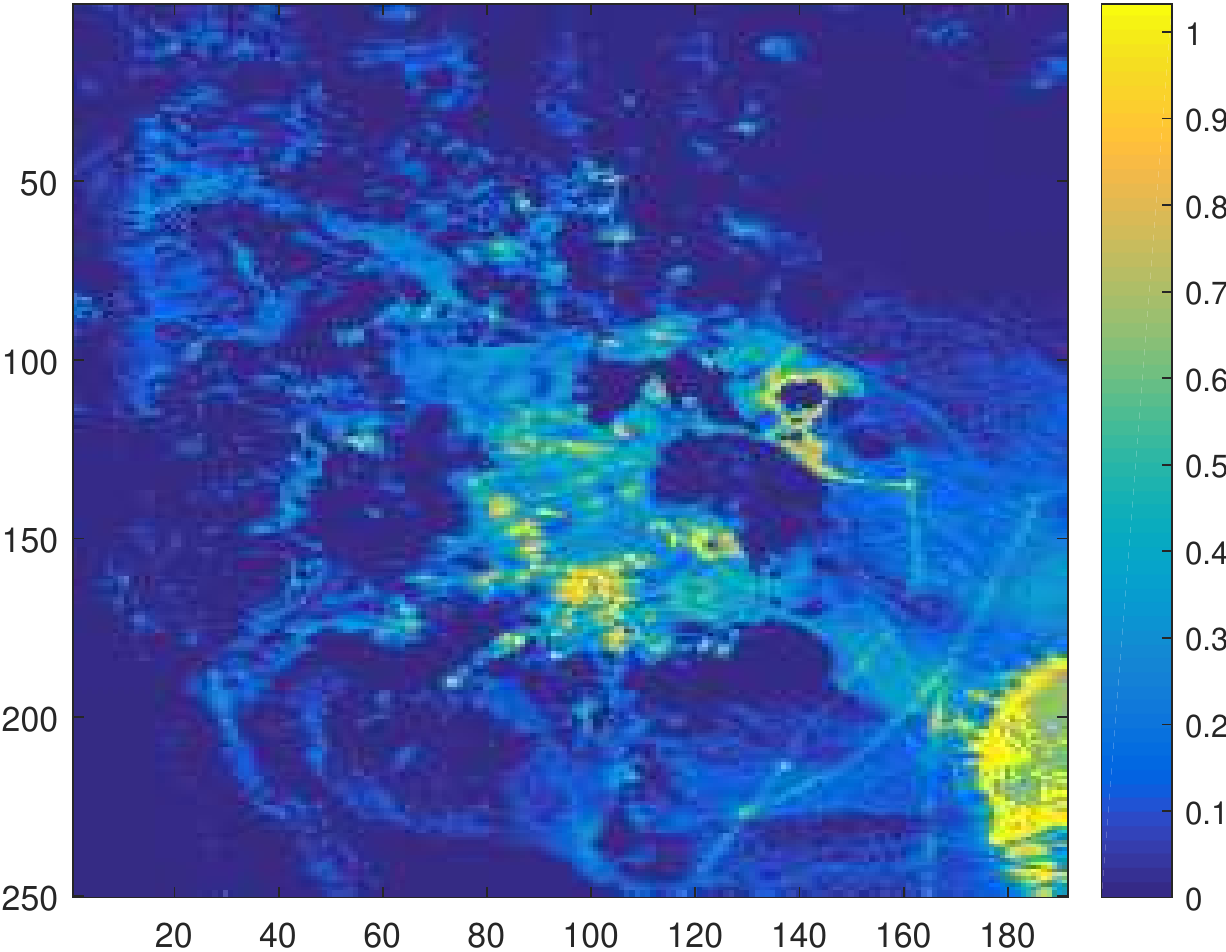}}
{\includegraphics[width=2.45cm, height=2.74cm]{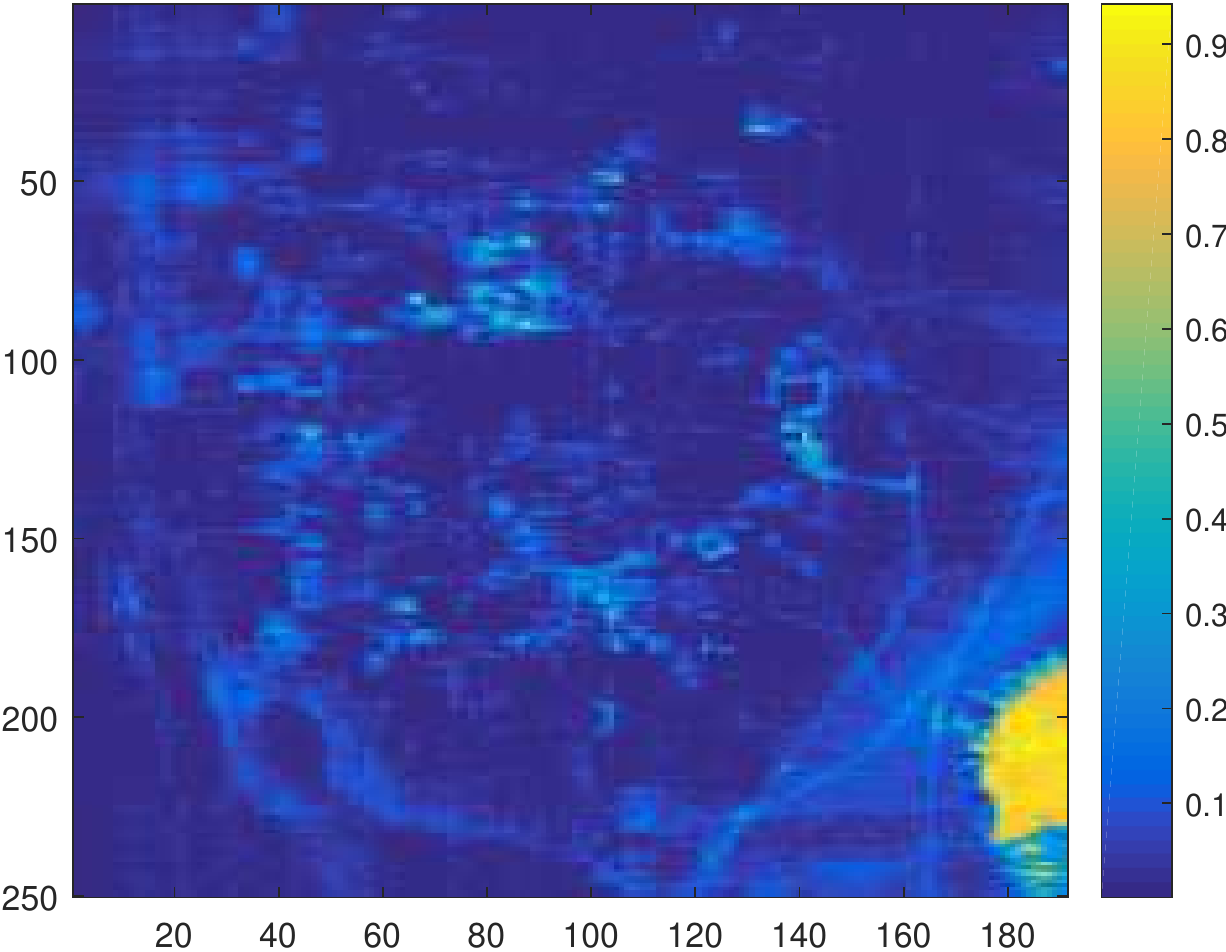}}
{\includegraphics[width=2.45cm, height=2.74cm]{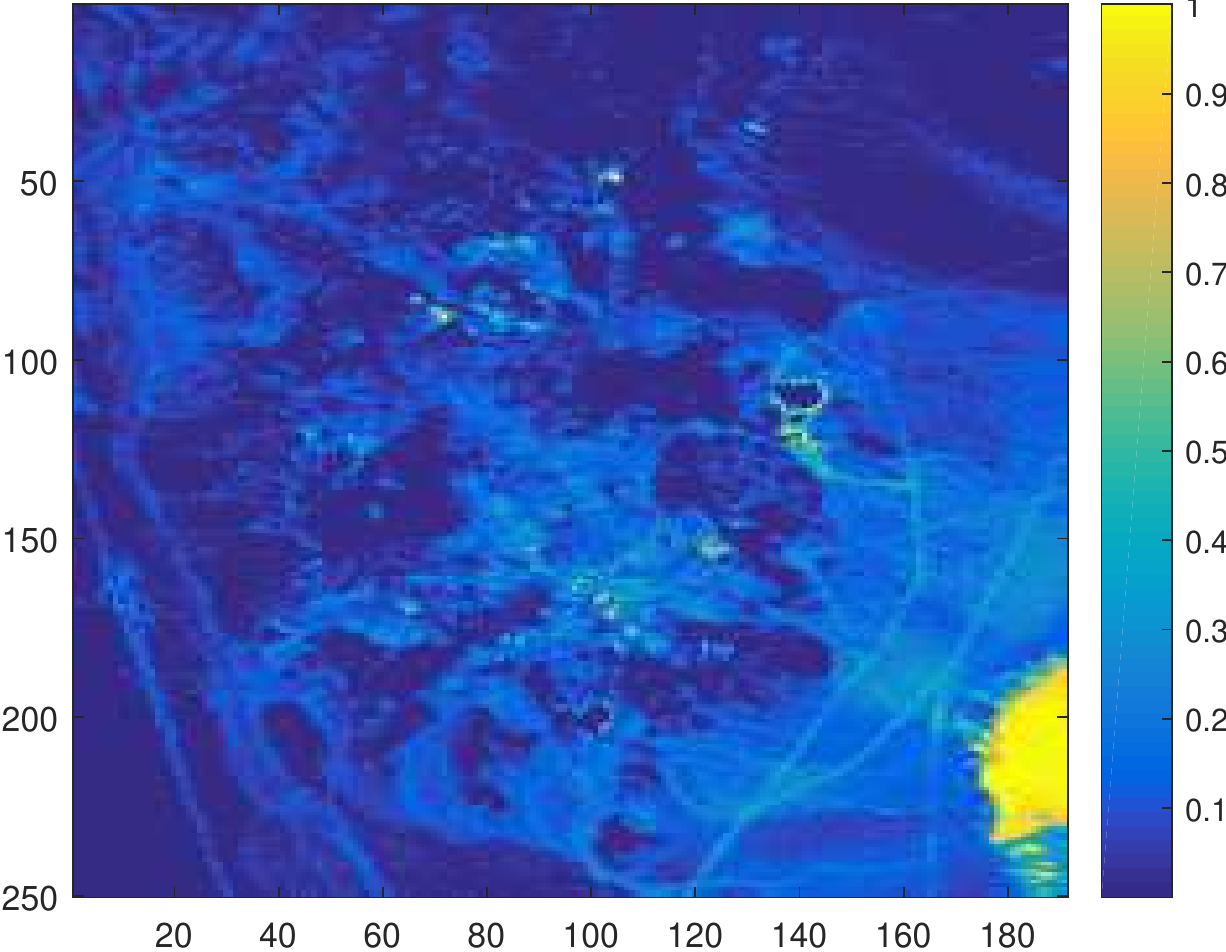}}
{\includegraphics[width=2.45cm, height=2.74cm]{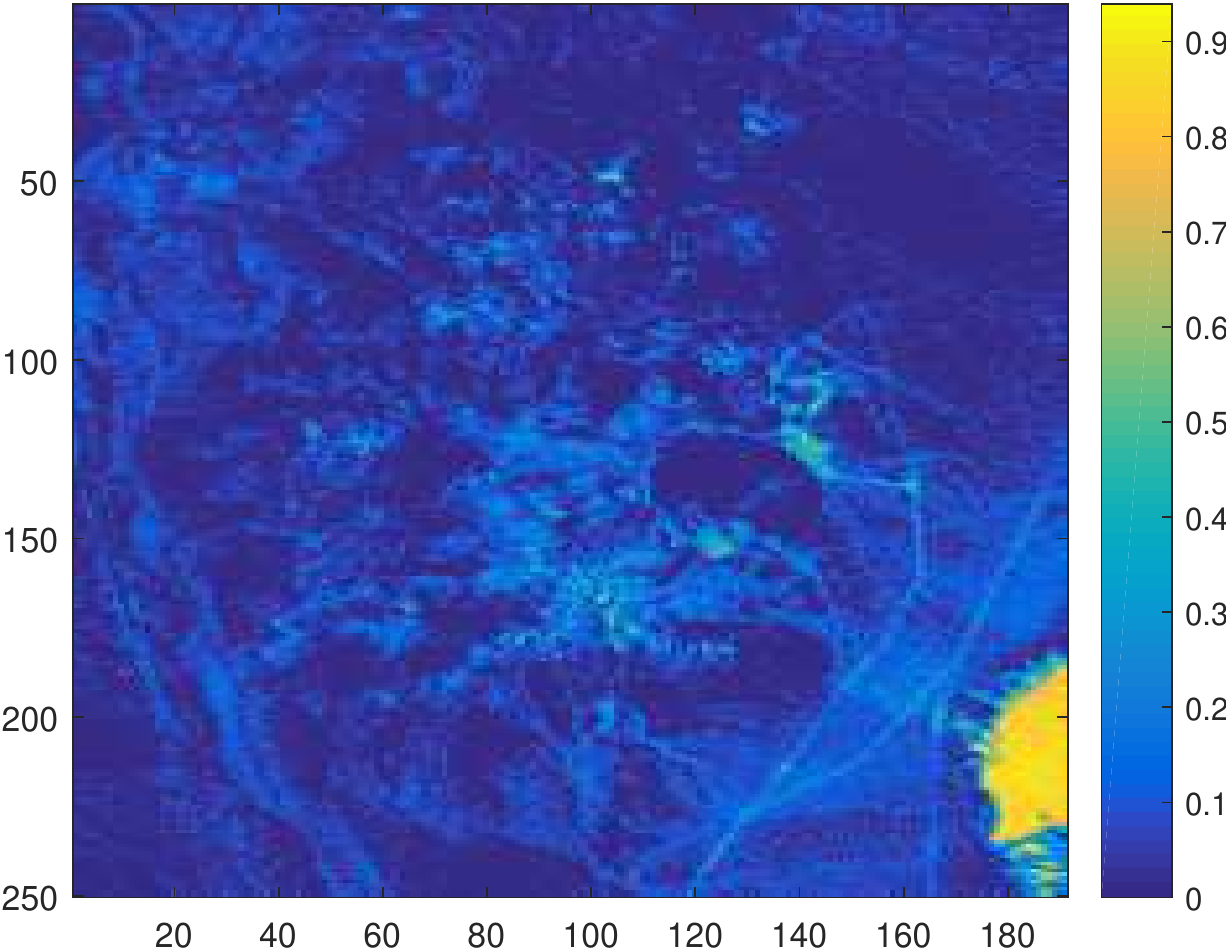}}}
\\
\mbox{
\subfigure[]{\includegraphics[width=2.45cm, height=2.74cm]{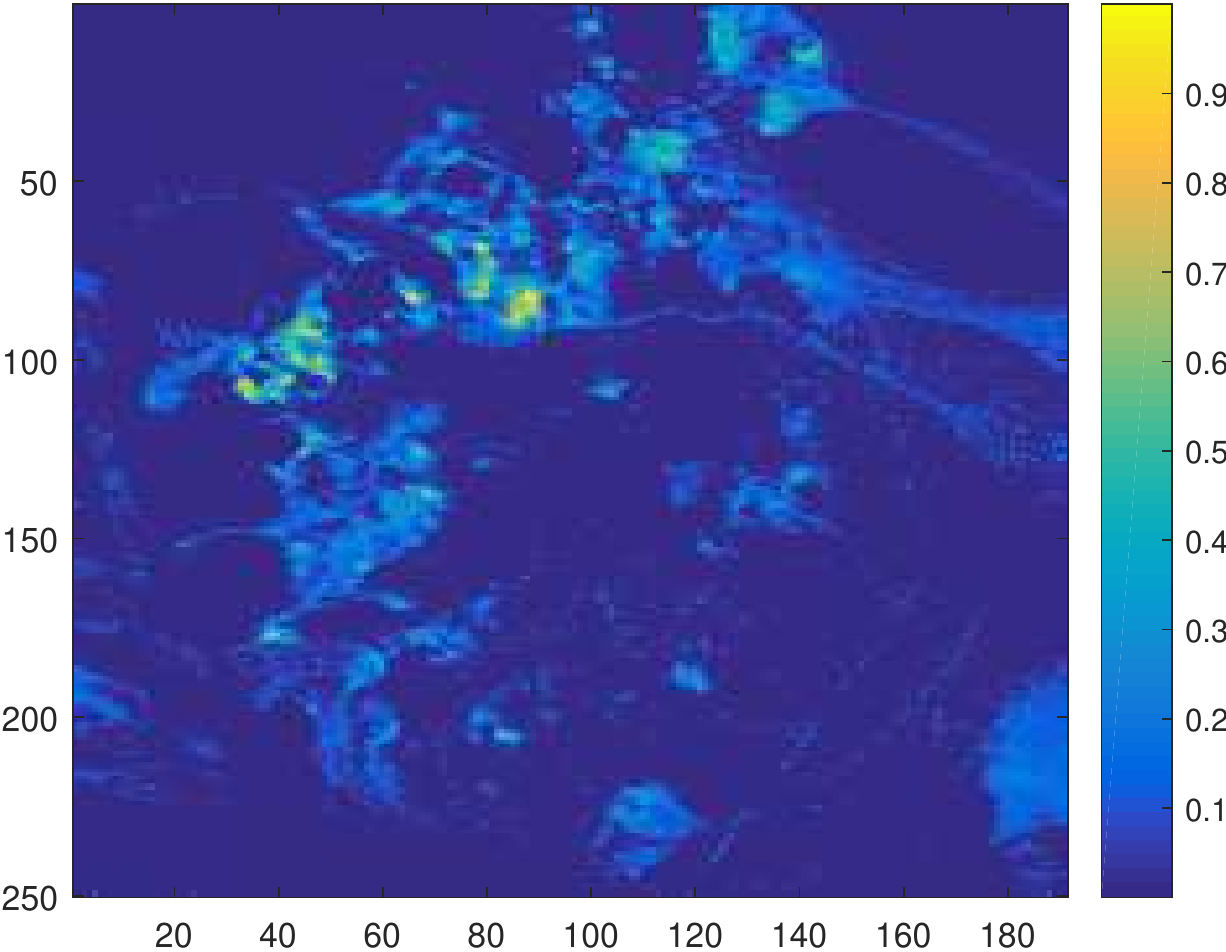}}
\subfigure[]{\includegraphics[width=2.45cm, height=2.74cm]{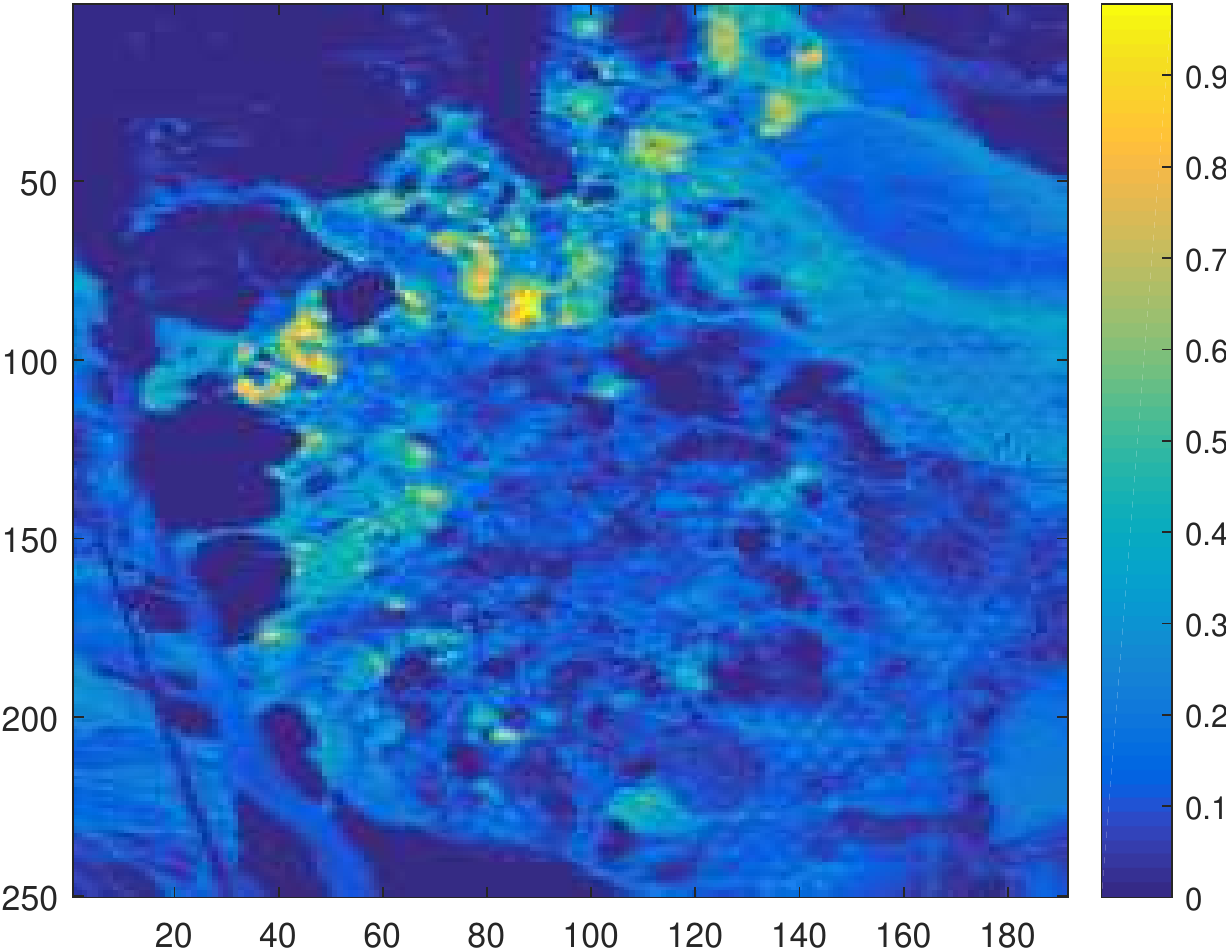}}
\subfigure[]{\includegraphics[width=2.45cm, height=2.74cm]{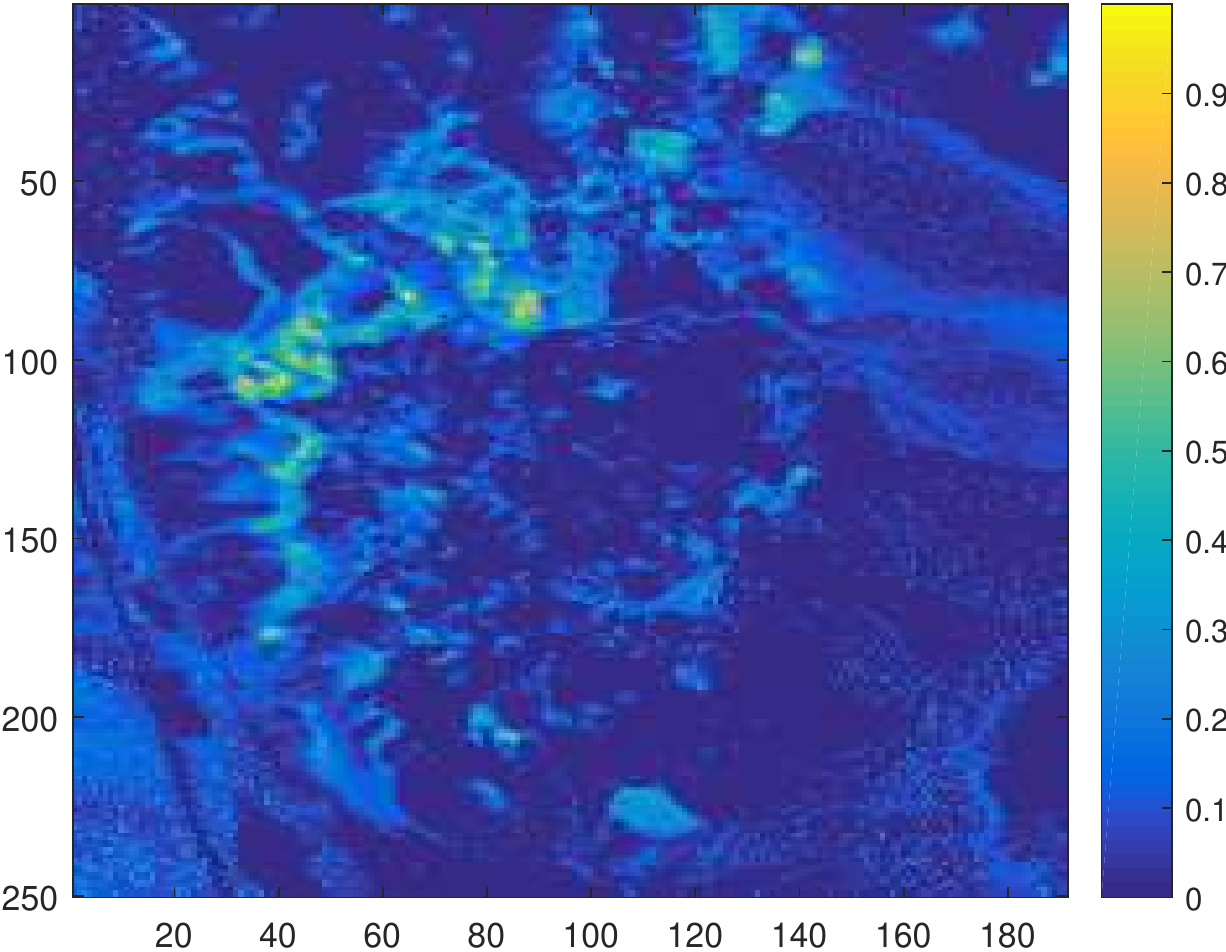}}
\subfigure[]{\includegraphics[width=2.45cm, height=2.74cm]{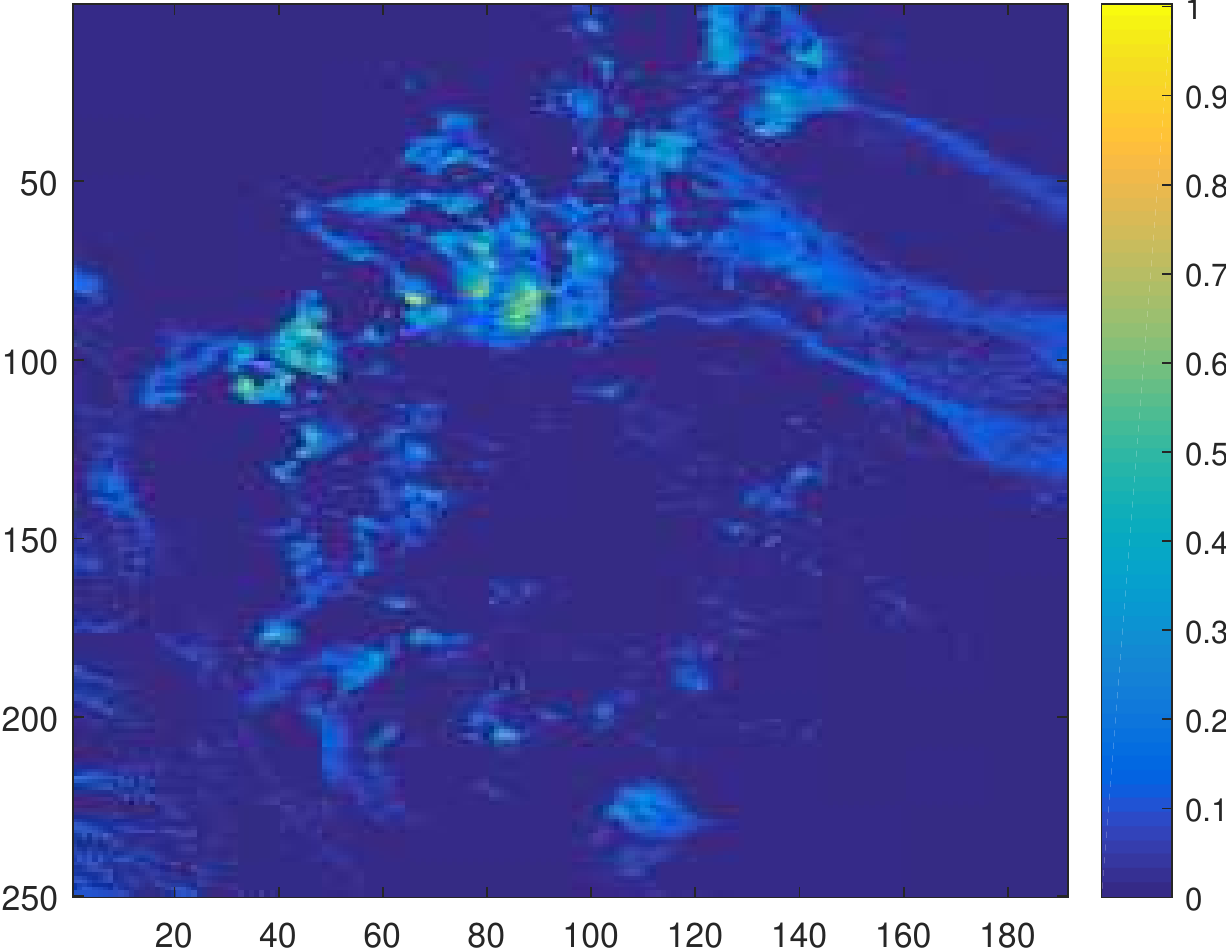}}
\subfigure[]{\includegraphics[width=2.45cm, height=2.74cm]{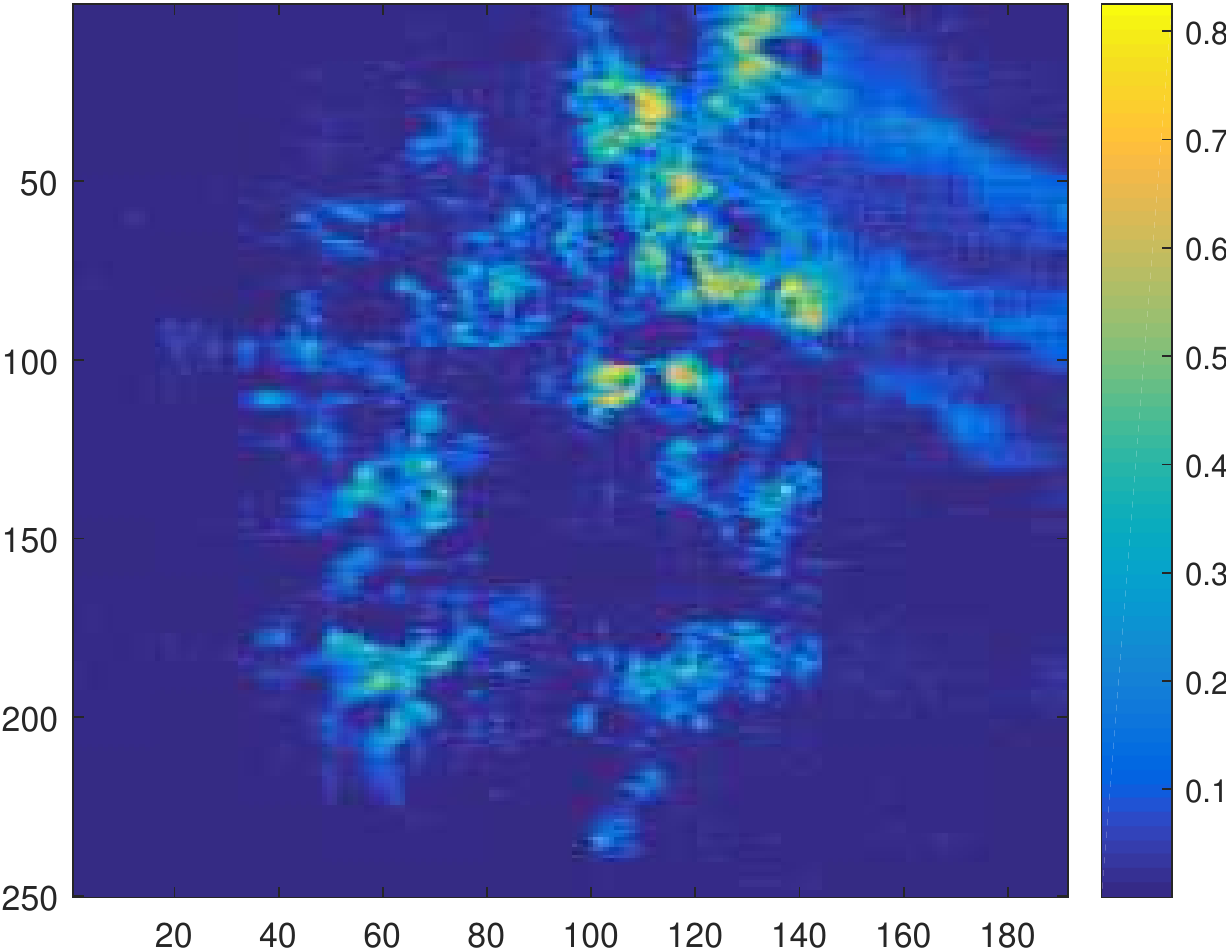}}
\subfigure[]{\includegraphics[width=2.45cm, height=2.74cm]{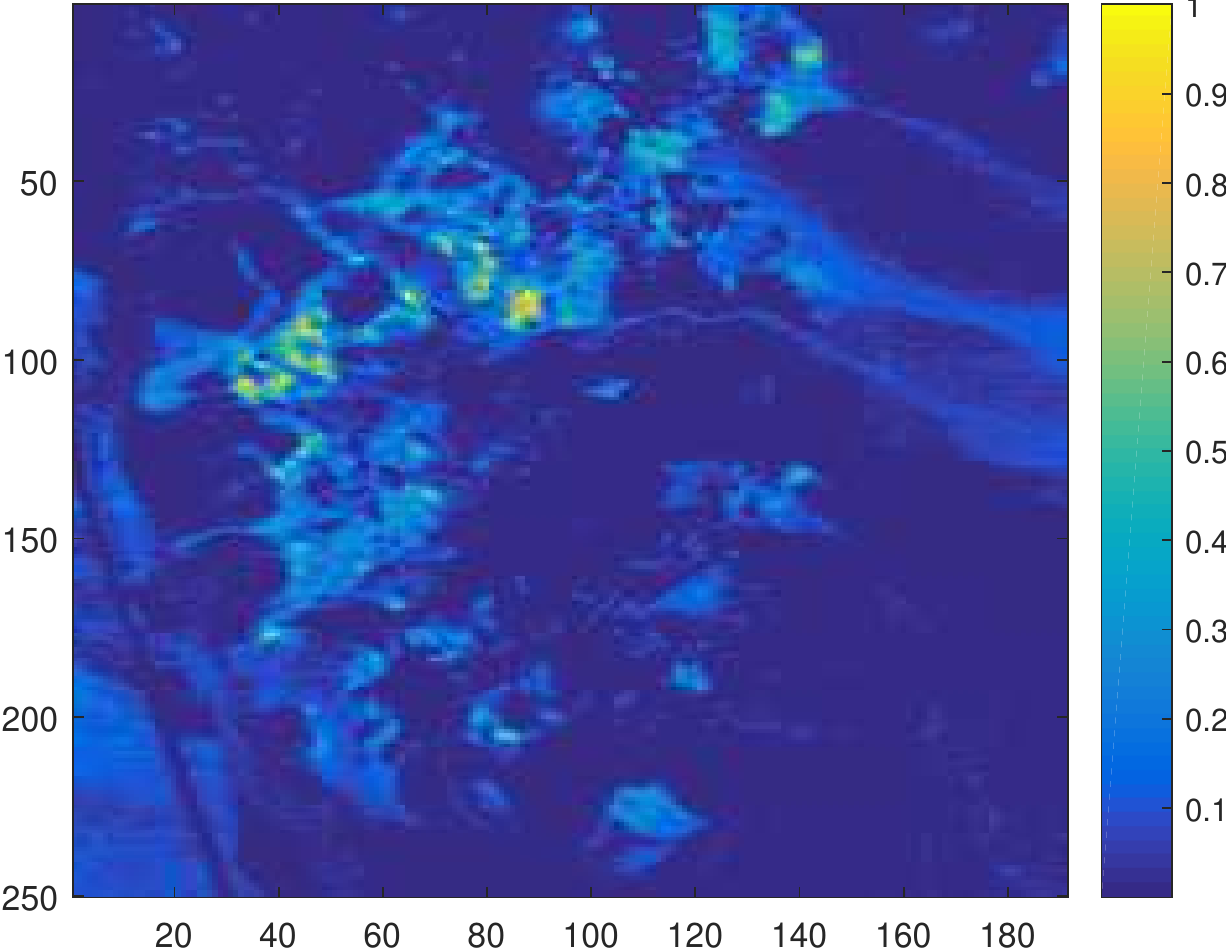}}
\subfigure[]{\includegraphics[width=2.45cm, height=2.74cm]{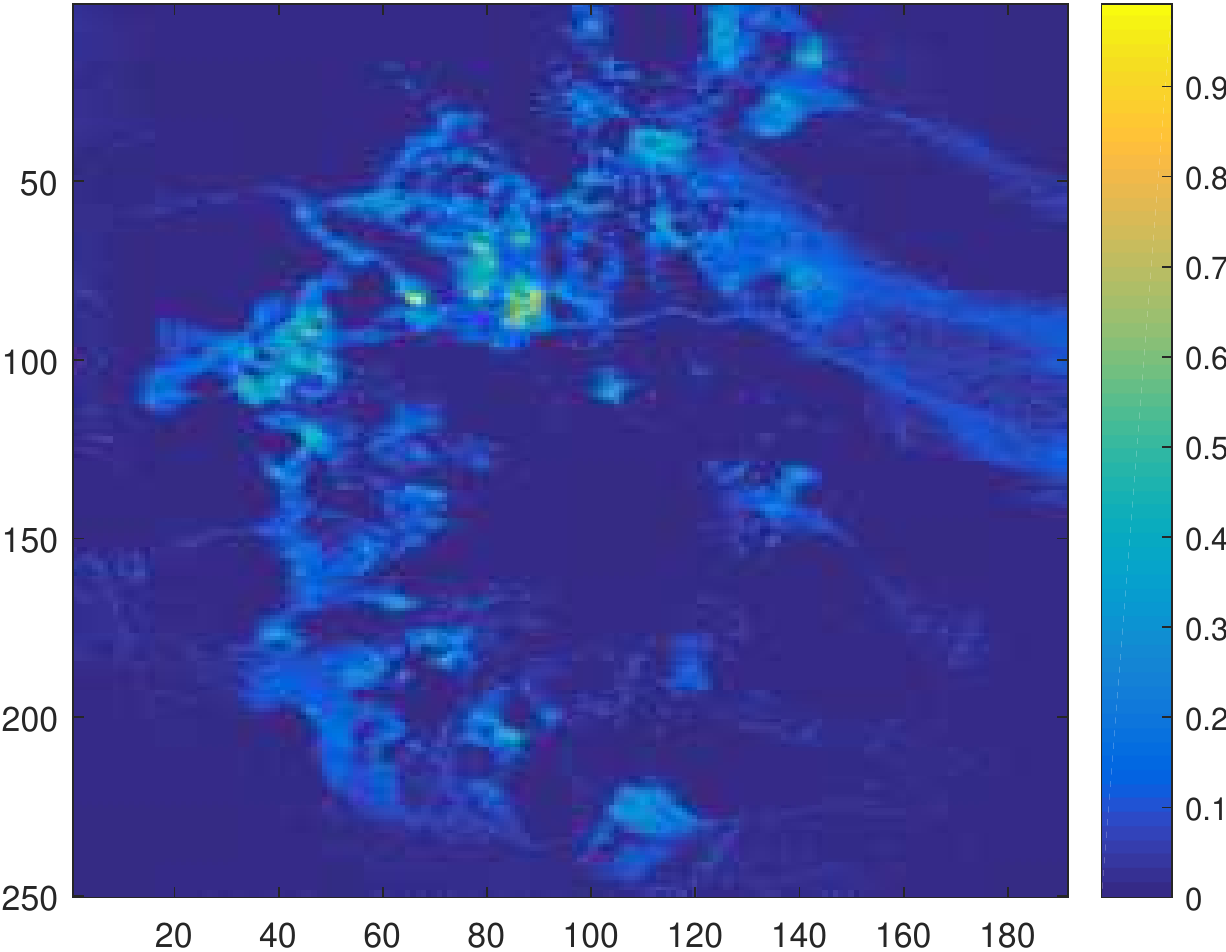}}}
\caption{Fractional abundance maps estimated by different methods of four endmembers on the AVIRIS Cuprite data set. From top to bottom: Buddingtonite GDS85 D-206. Kaolinite KGa-2. Montmorillonite+Illi CM37. Nontronite NG-1.a. From left to right: (a) $L_{1/2}$-NMF. (b) SGSNMF. (c) TV-RSNMF. (d) $L_{1/2}$-RNMF. (e) MV-NTF-TV. (f) MLNMF. (g) SSRDMF.}
\label{fig:6}
\end{figure*}

We choose the most popular methods in different categories to conduct experiments, such as $L_{1/2}$-NMF \cite{YQian2011}, SGSNMF \cite{XWang2017}\footnote{https{://}github.com{/}Xinyu-Wang{/}SGSNMF{\_}TGRS}, TV-RSNMF \cite{WHe2017}, $L_{1/2}$-RNMF\cite{WHe2016Sep}, MV-NTF-TV \cite{FXiong2019}, MLNMF \cite{RRajabi2015}\footnote{https{://}github.com{/}roozbehrajabi{/}mlnmf}, and SSRDMF \cite{HCLi2022}. It should be noted that all the results are averaged after ten independent runs. All experiments are conducted under the environment of MATLAB R2015b software and computer configuration Intel Core i5 CPU at $2.80$ GHz and $8.00$ GB RAM. For hyperspectral unmixing, the number of endmembers is a crucial factor, which can be set manually or estimated through an effective method, such as virtual dimensionality \cite{CIChang2018, VijayashekharSS2020} and HySime \cite{JMBioucasDias2008}.

Three hyperspectral scene have been utilized in the tests, shown in Fig. \ref{fig:4}. The first data set is the notable Cuprite data, which was obtained by the Airborne Visible Infrared Imaging Spectrometer (AVIRIS). The second data set is Samson image, which is the first public real hyperspectral data set. The third data set is Jasper Ridge scene. Since these real HSI data sets have been studied for unmixing in many articles, we refer to the experimental results reported in some published references.

1) \emph{AVIRIS Cuprite Data set}:  As shown in Fig. \ref{fig:4}(a), it is typically used to verify the effectiveness of hyperspectral unmixing methods. It contains $250\times191$ pixels, and each pixel contains $188$ bands (\emph{i.e.}, $3-103$, $114-147$, and $168-220$) after removal of noisy bands. The covered wavelength range comprises $0.4-2.5\mu$m. There are mainly $12$ minerals in the subscene: Alunite GDS82 Na82, Andradite WS487, Buddingtonite GDS85 D-206, Chalcedony CU91-6A, Kaolin/Smect H89-FR-5 30K, Kaolin/Smect KLF508 85\%K, Kaolinite KGa-2, Montmorillonite + Illi CM37, Muscovite IL107, Nontronite NG-1.a, Pyrope WS474, and Sphene HS189.3B. The reference signatures are from the U.S. Geological Survey (USGS) spectral library\footnote{http{://}speclab.cr.usgs.gov{/}spectral.lib06}, and a mineral map is often used for illustrative purposes\footnote{https{://}www.usgs.gov{/}labs{/}spectroscopy-lab}.

2) \emph{Samson Data set}: As shown in Fig. \ref{fig:4}(b), it contains $95\times95$ pixels, and each pixel has $156$ bands ranging from $0.401-0.889\mu$m. The number of endmembers is set to $3$, including Soil, Tree, and Water.

3) \emph{Jasper Ridge Data set}: As shown in Fig. \ref{fig:4}(c), it contains $100\times100$ pixels, and each pixel has $198$ bands (\emph{i.e.}, $4-107$, $113-153$, and $167-219$) after removal of noisy bands ranging from $0.38-2.5\mu$m. The number of endmembers is set to $4$, including Tree, Water, Soil, and Road.

\begin{figure*}[!t]
\centering
\mbox{
{\includegraphics[width=2.45cm]{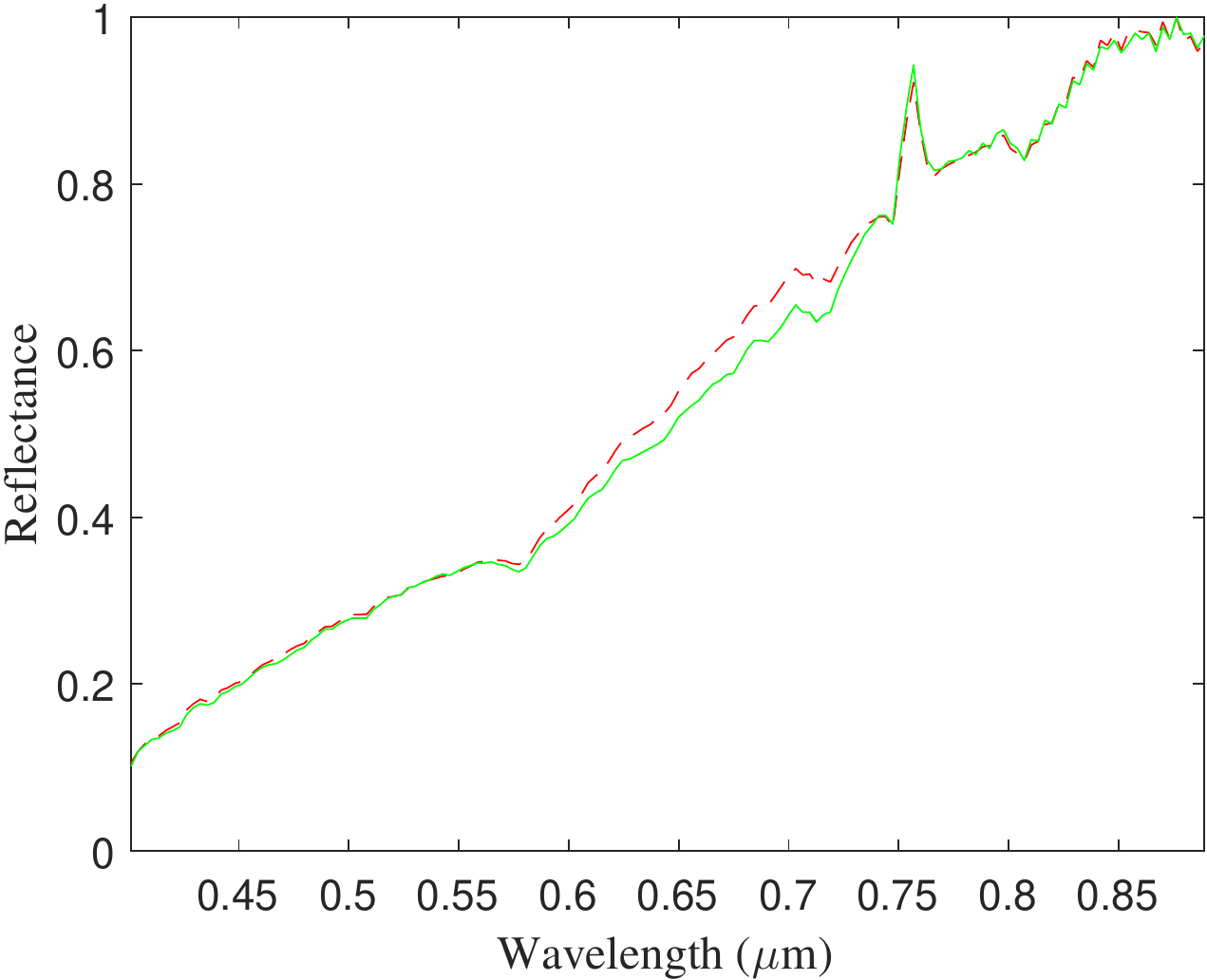}}
{\includegraphics[width=2.45cm]{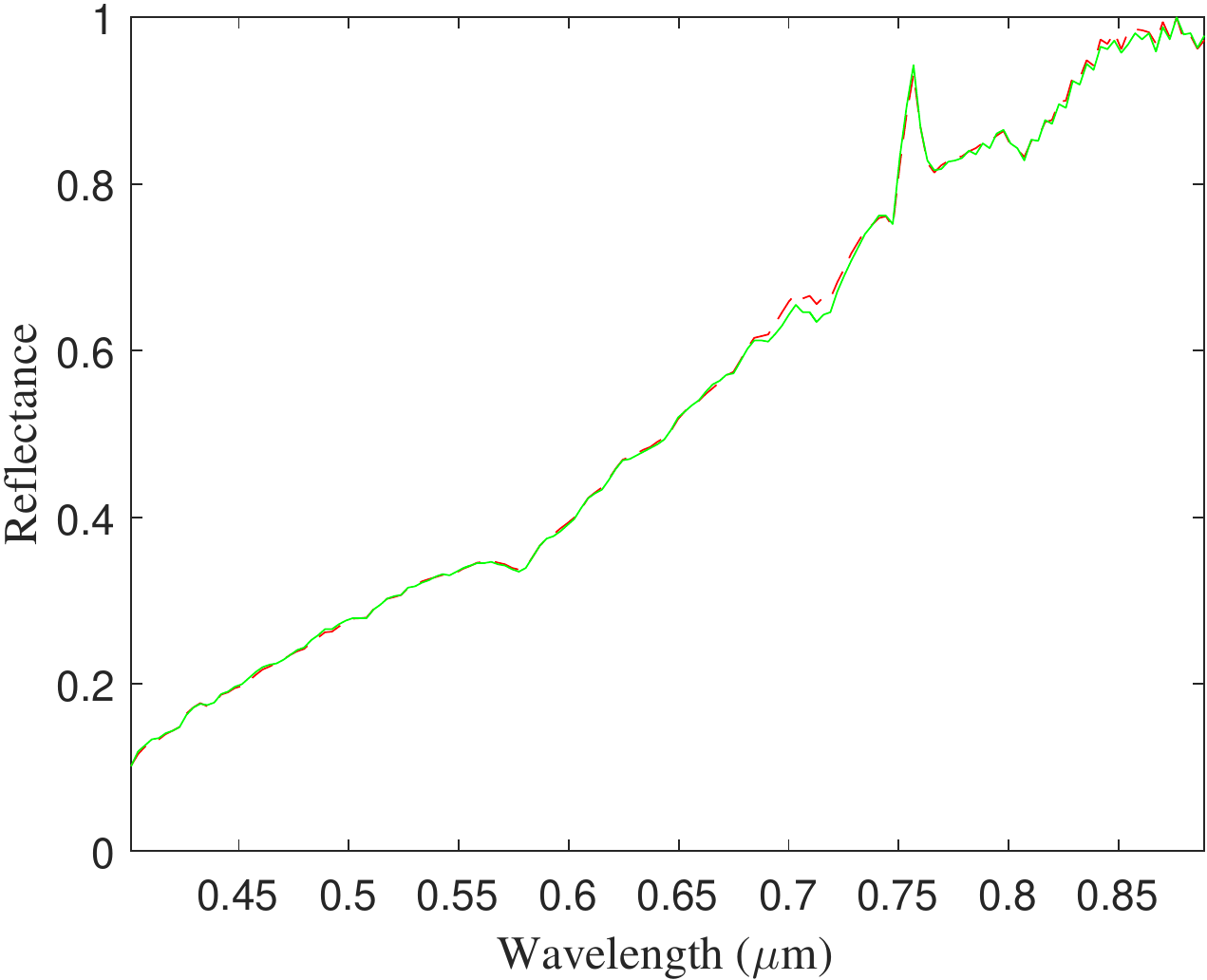}}
{\includegraphics[width=2.45cm]{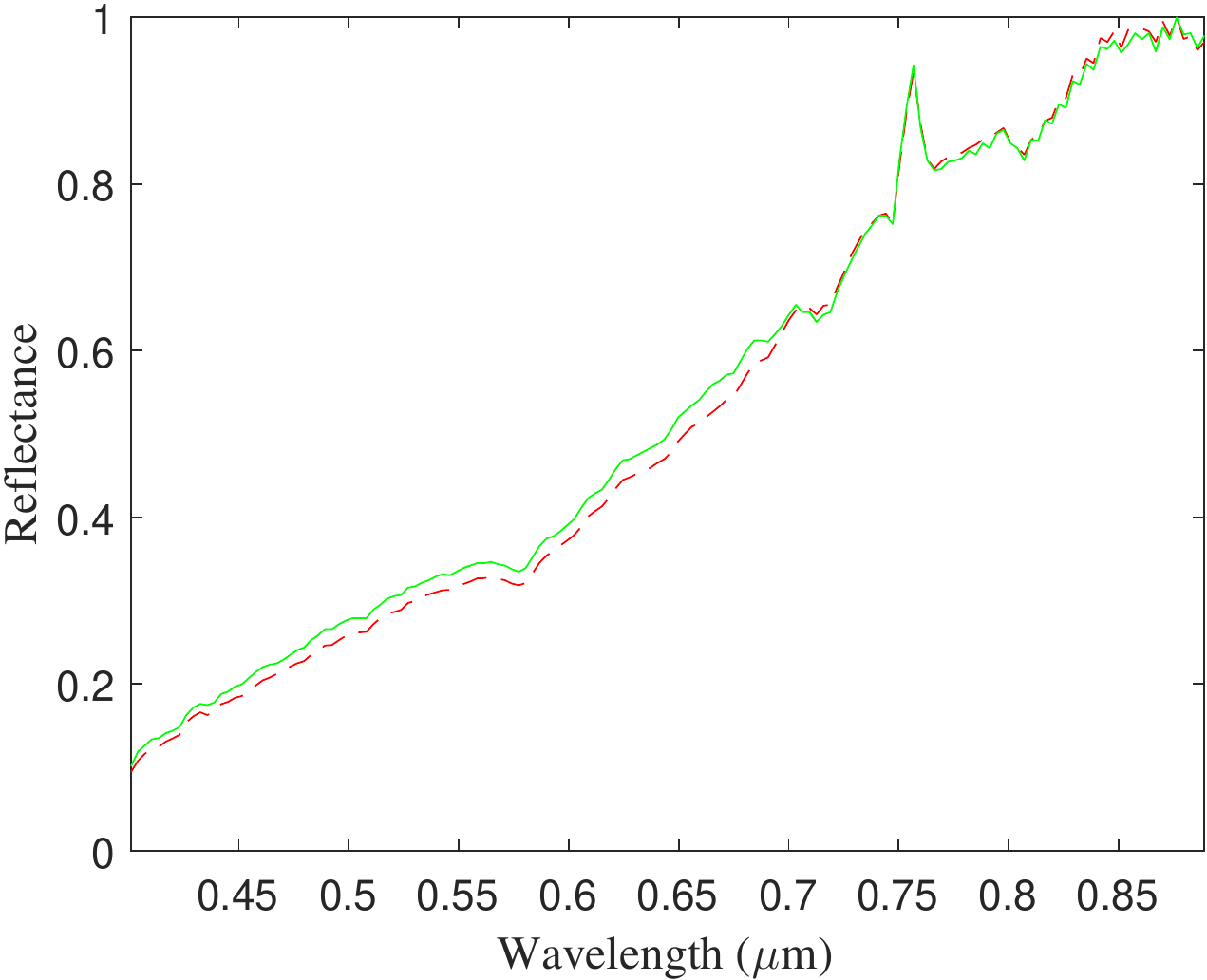}}
{\includegraphics[width=2.45cm]{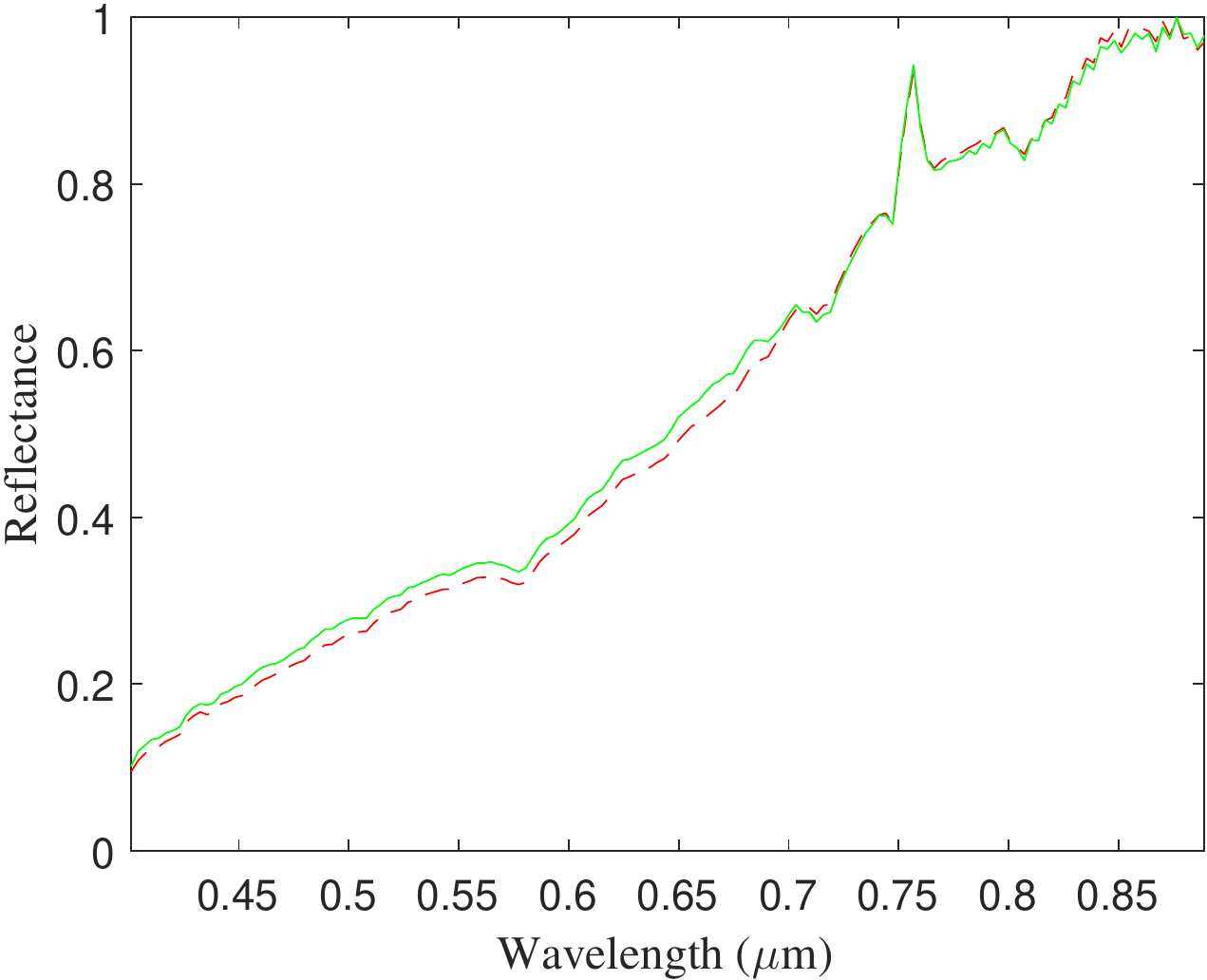}}
{\includegraphics[width=2.45cm]{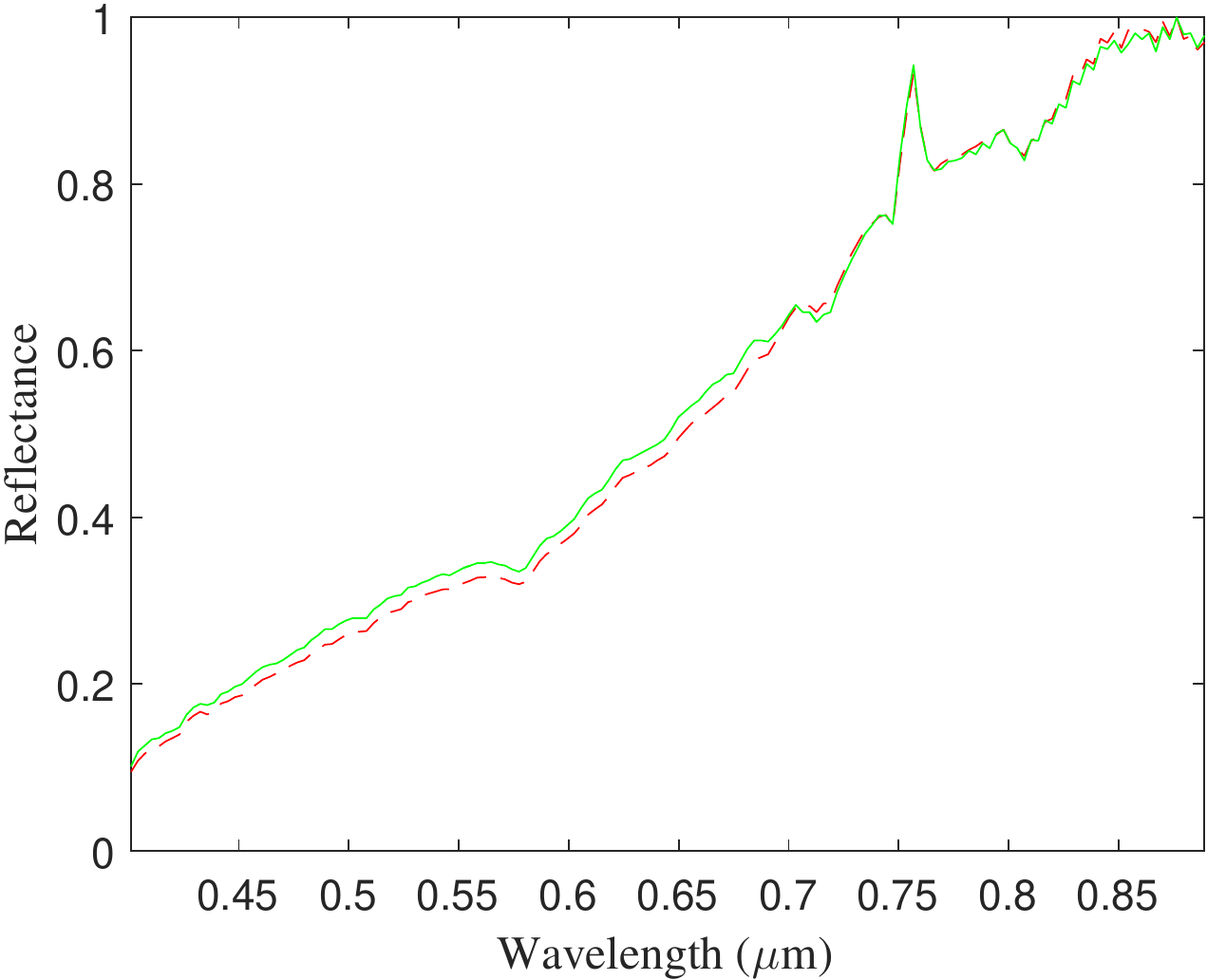}}
{\includegraphics[width=2.45cm]{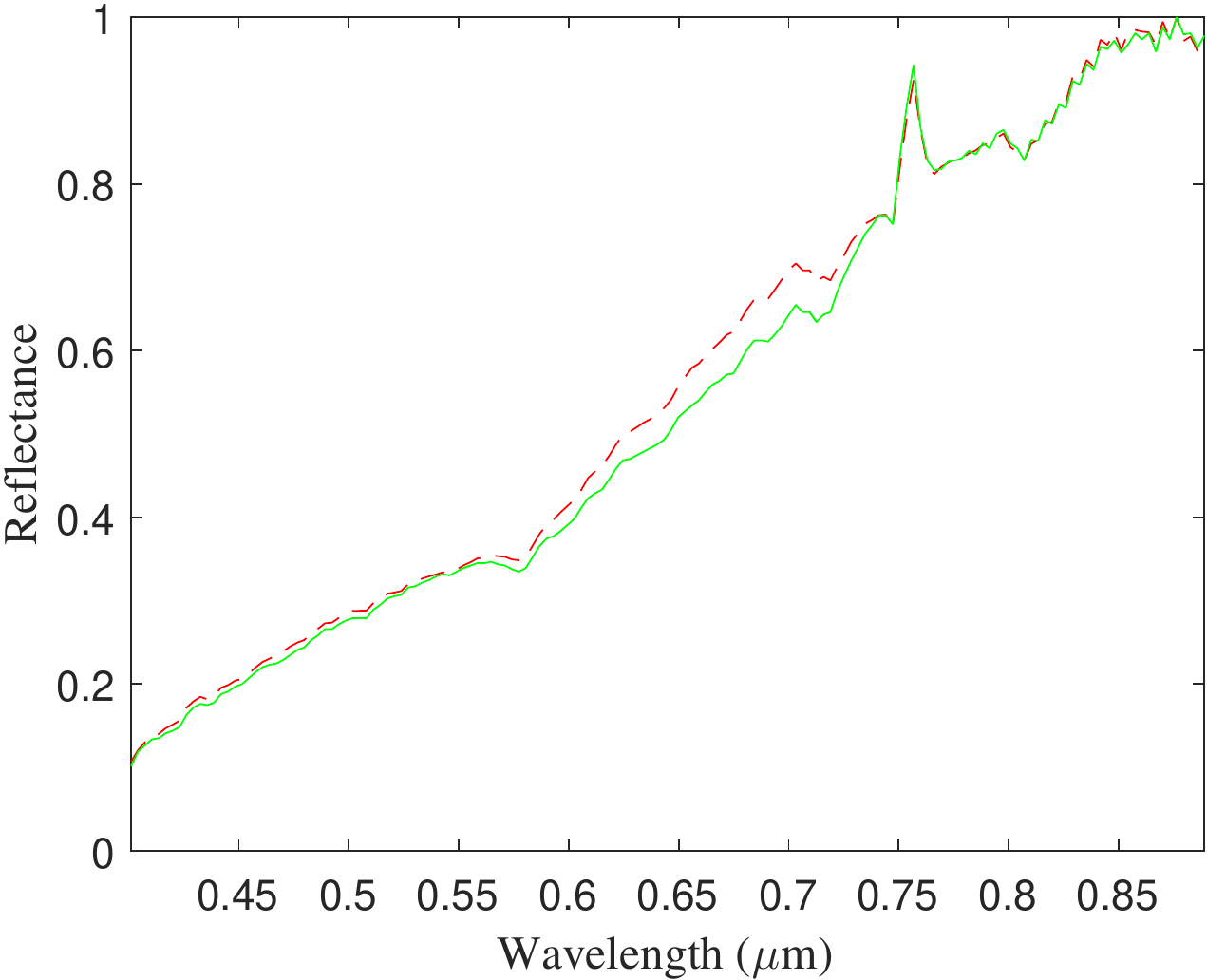}}
{\includegraphics[width=2.45cm]{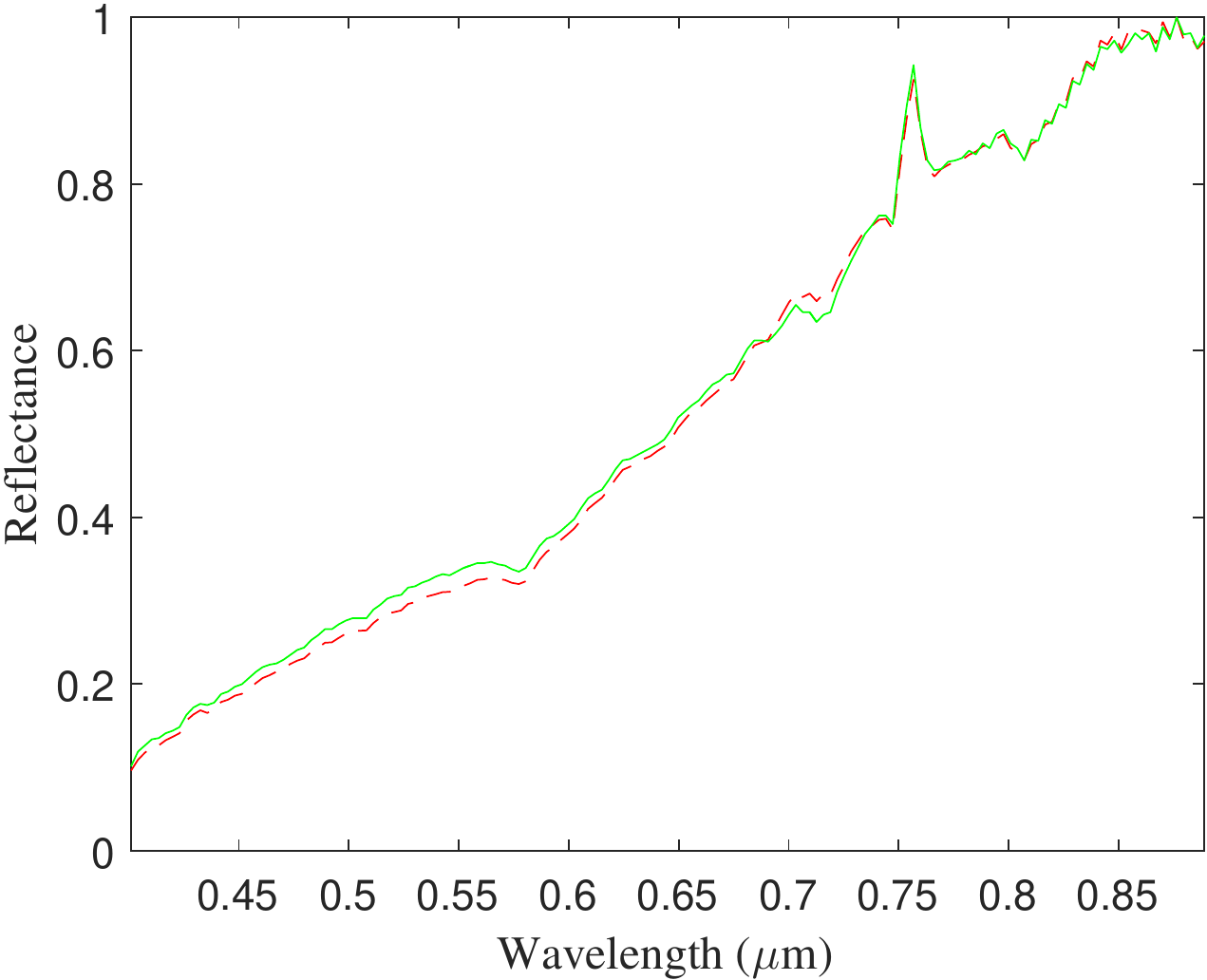}}}
\\
\mbox{
{\includegraphics[width=2.45cm]{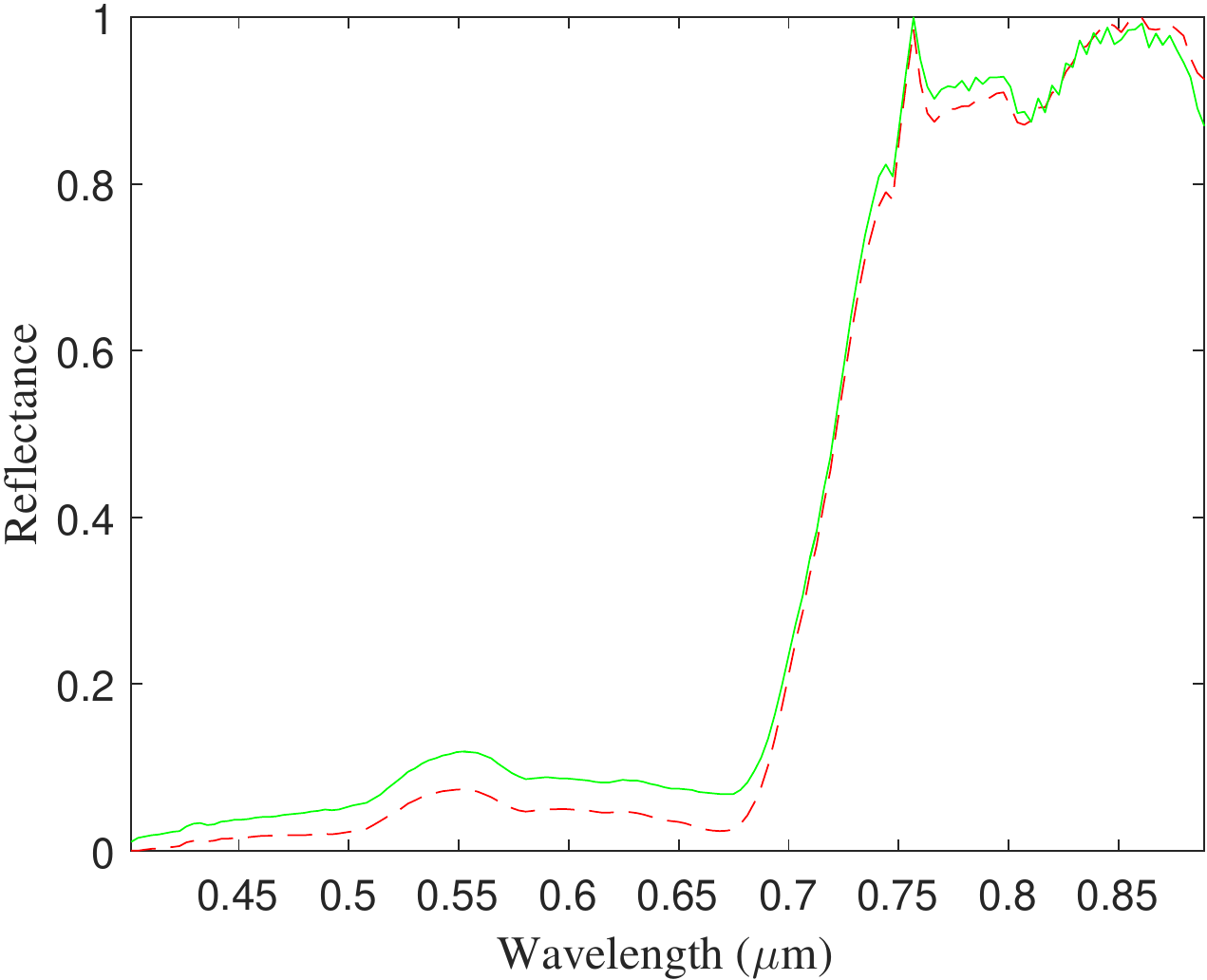}}
{\includegraphics[width=2.45cm]{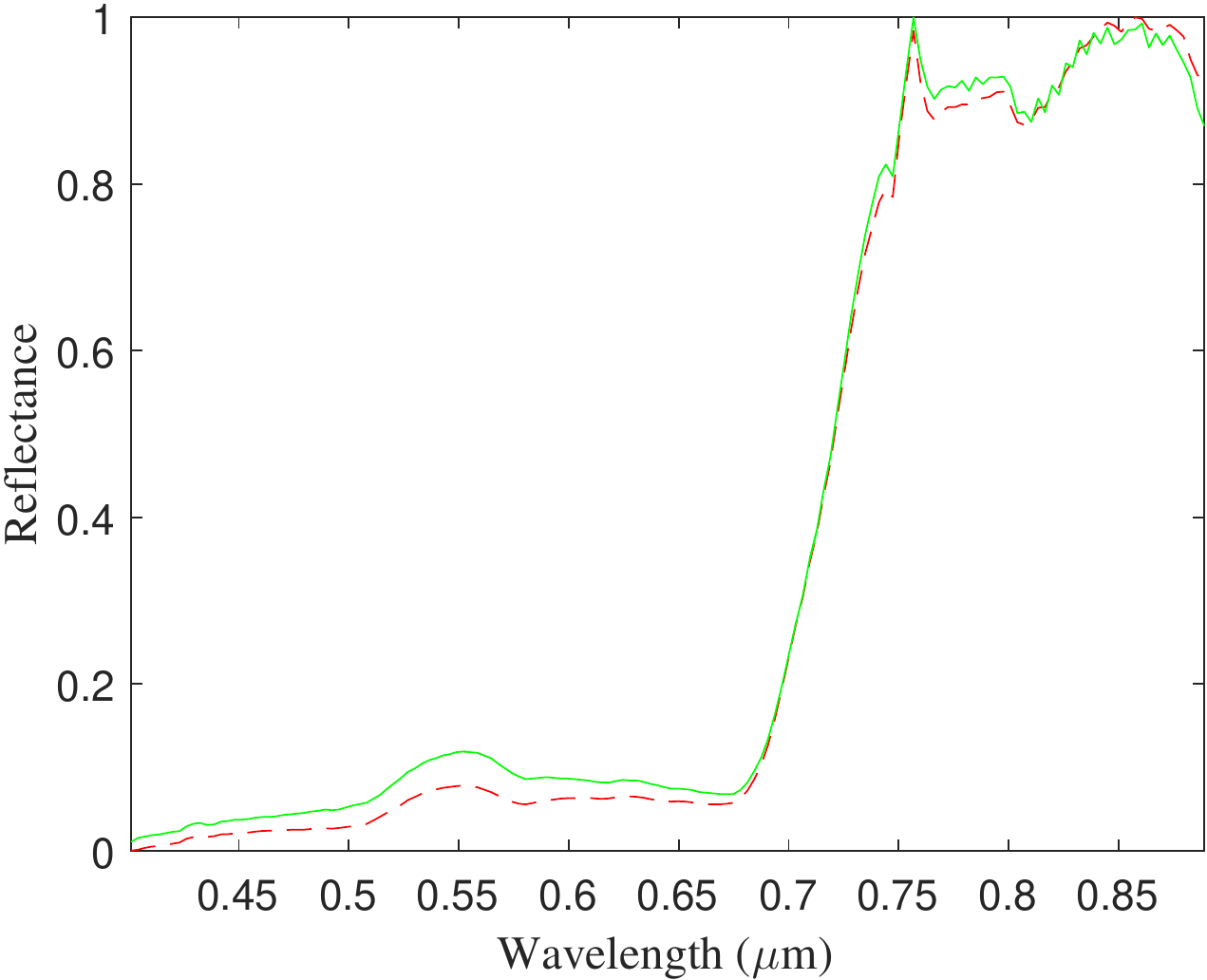}}
{\includegraphics[width=2.45cm]{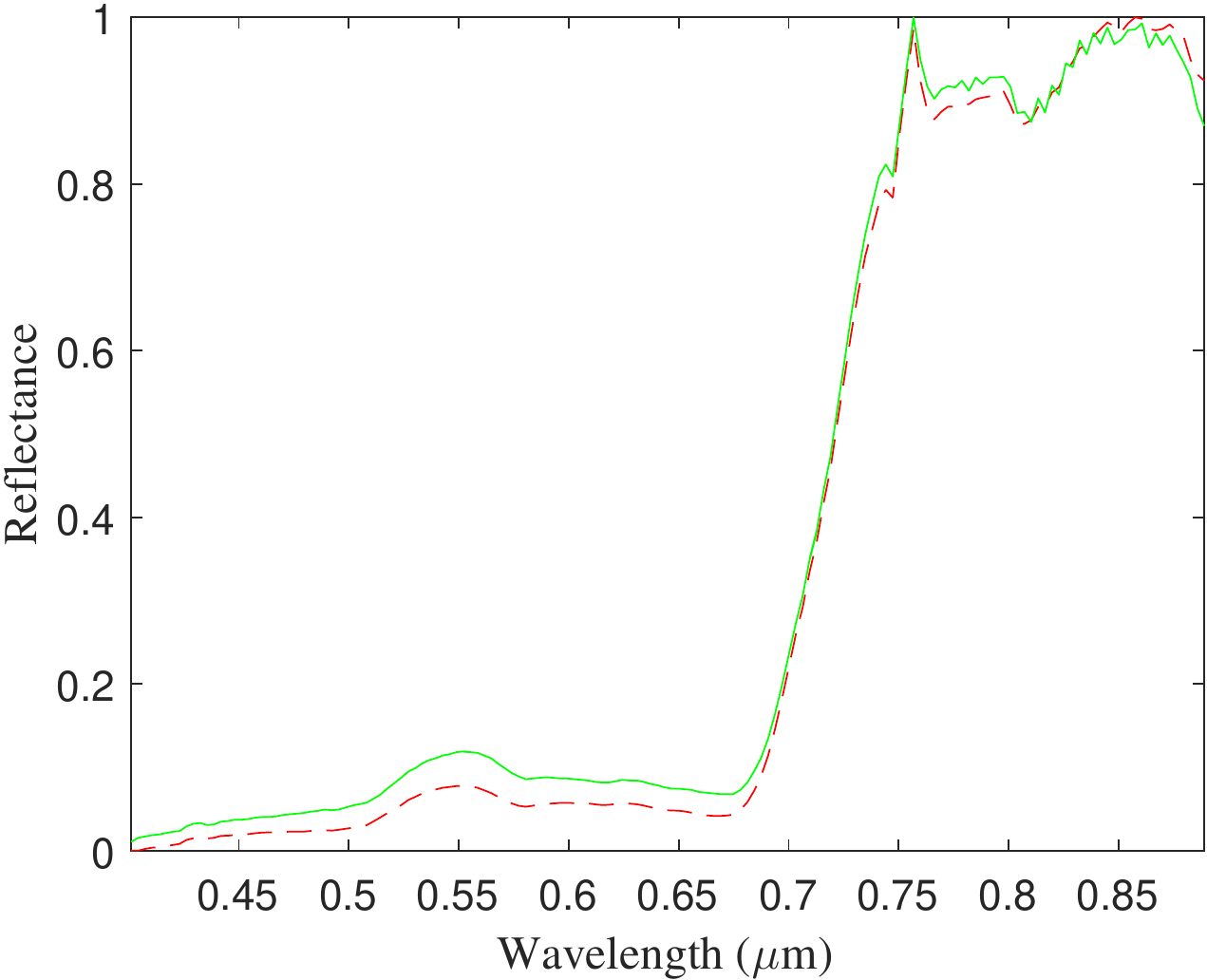}}
{\includegraphics[width=2.45cm]{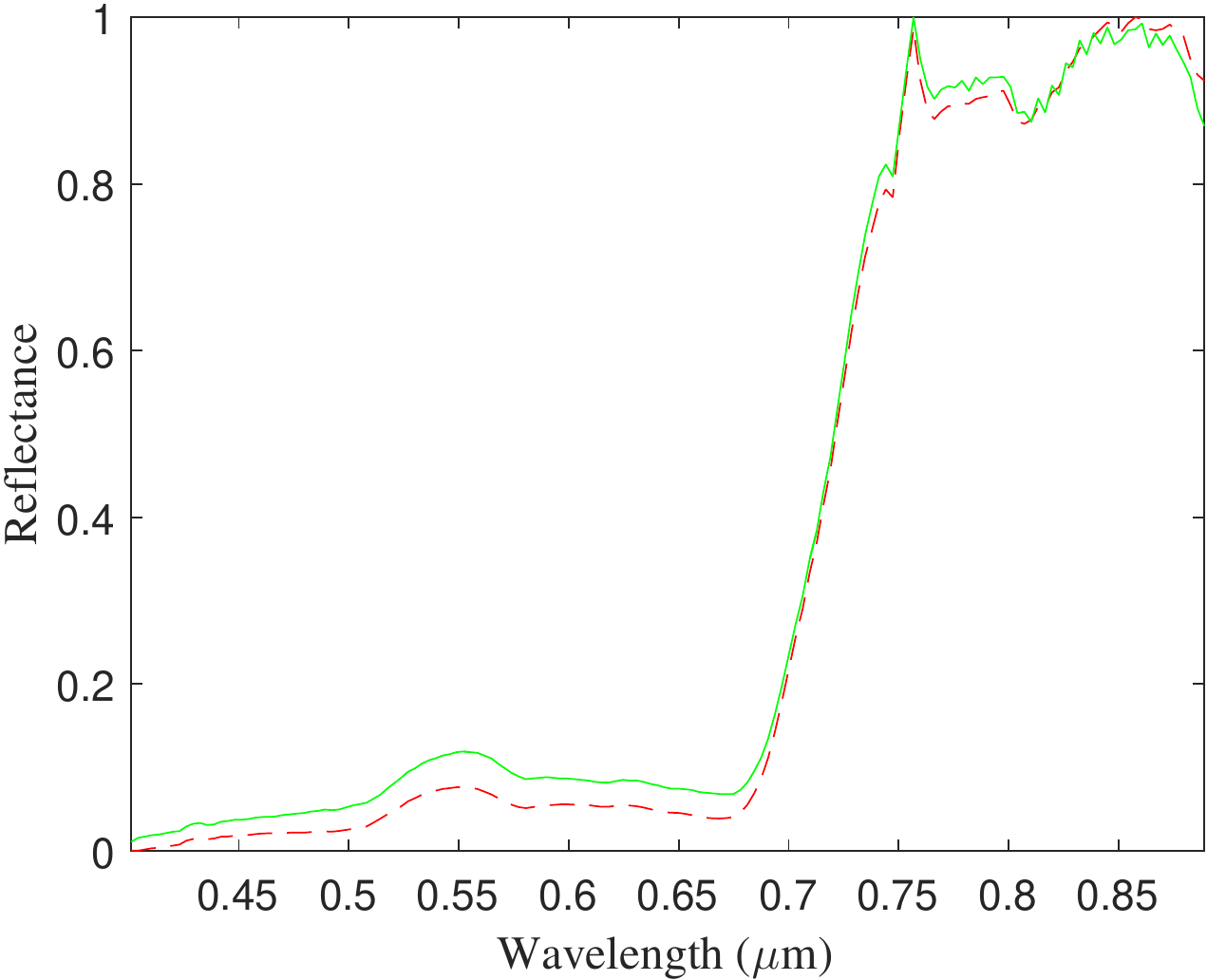}}
{\includegraphics[width=2.45cm]{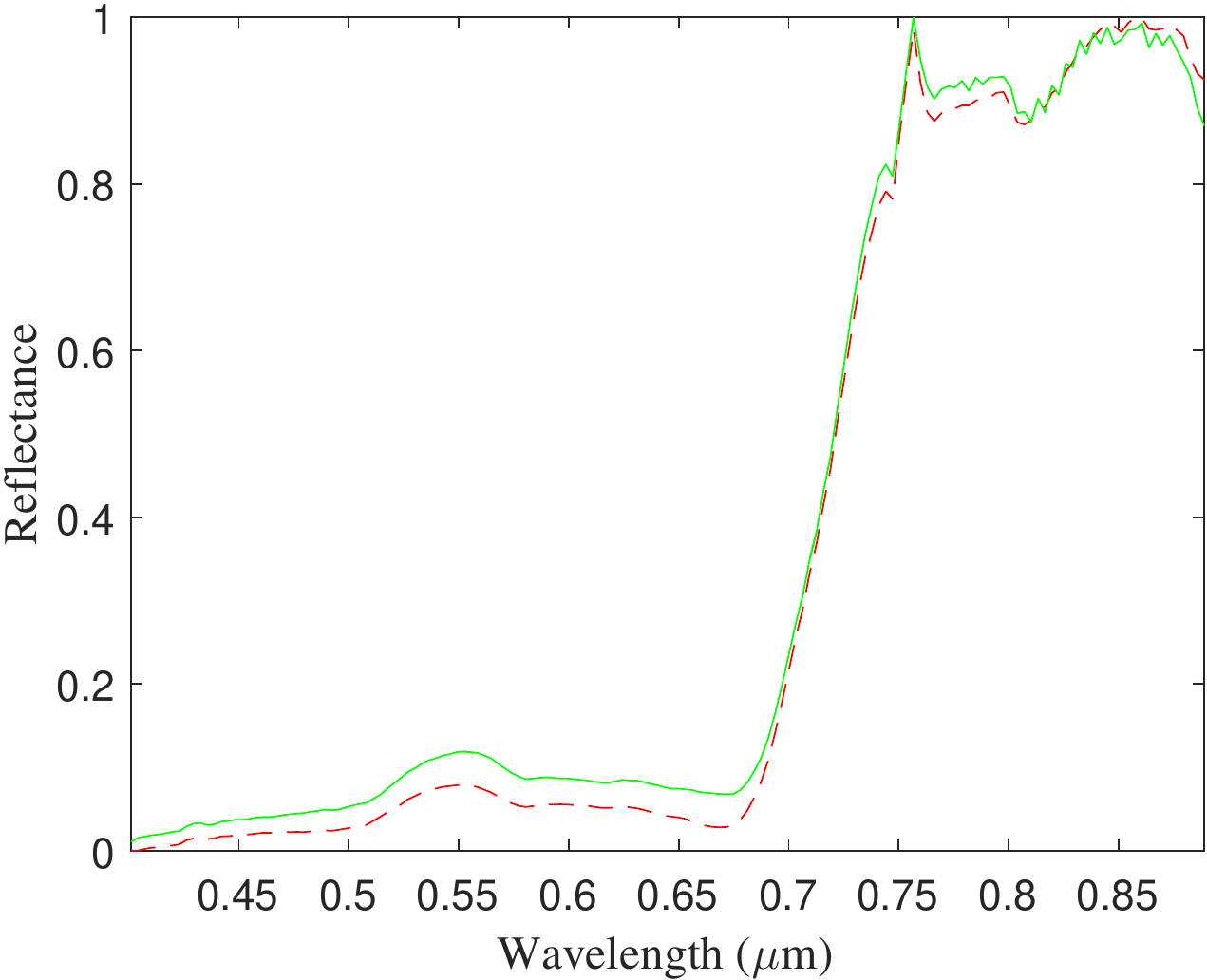}}
{\includegraphics[width=2.45cm]{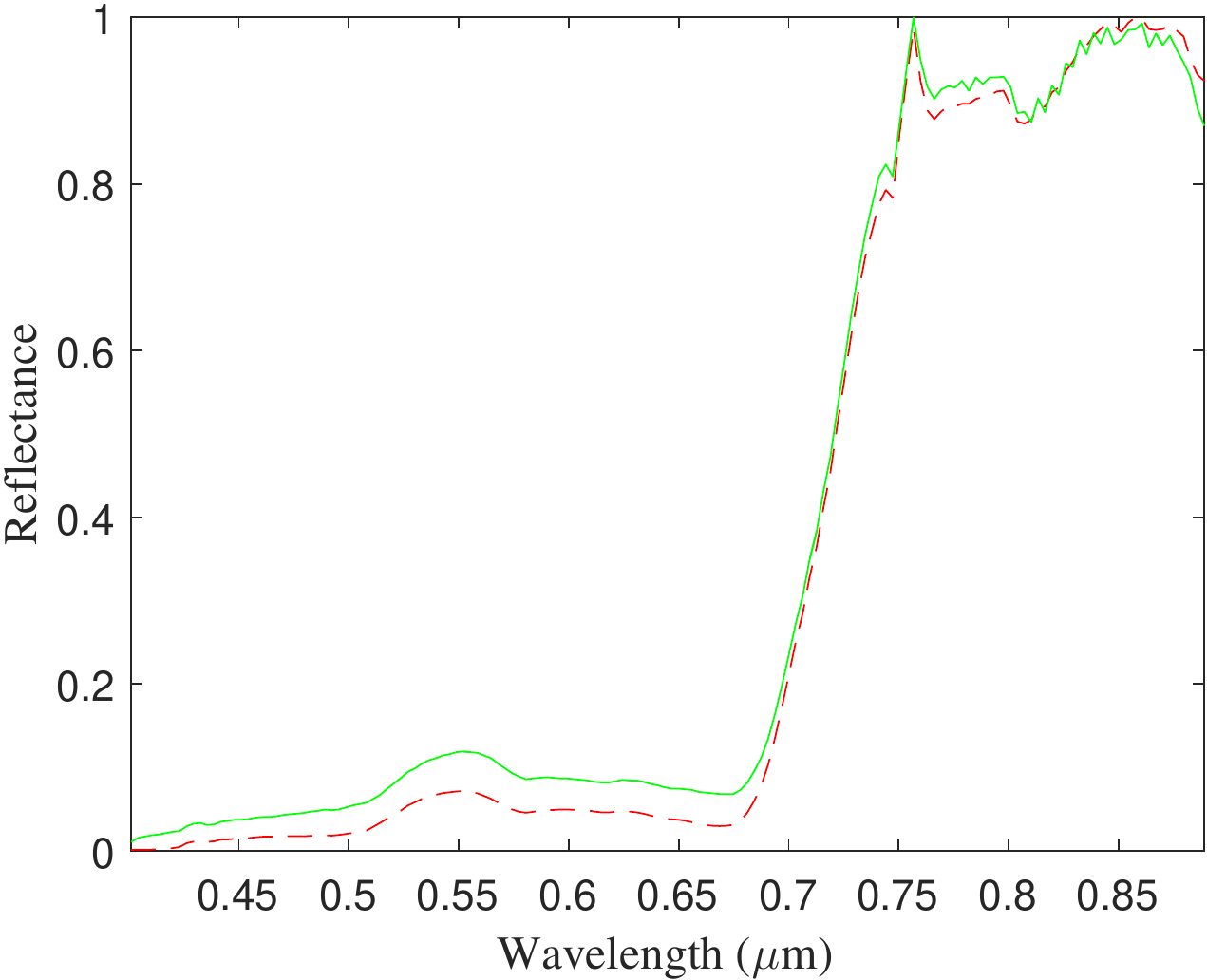}}
{\includegraphics[width=2.45cm]{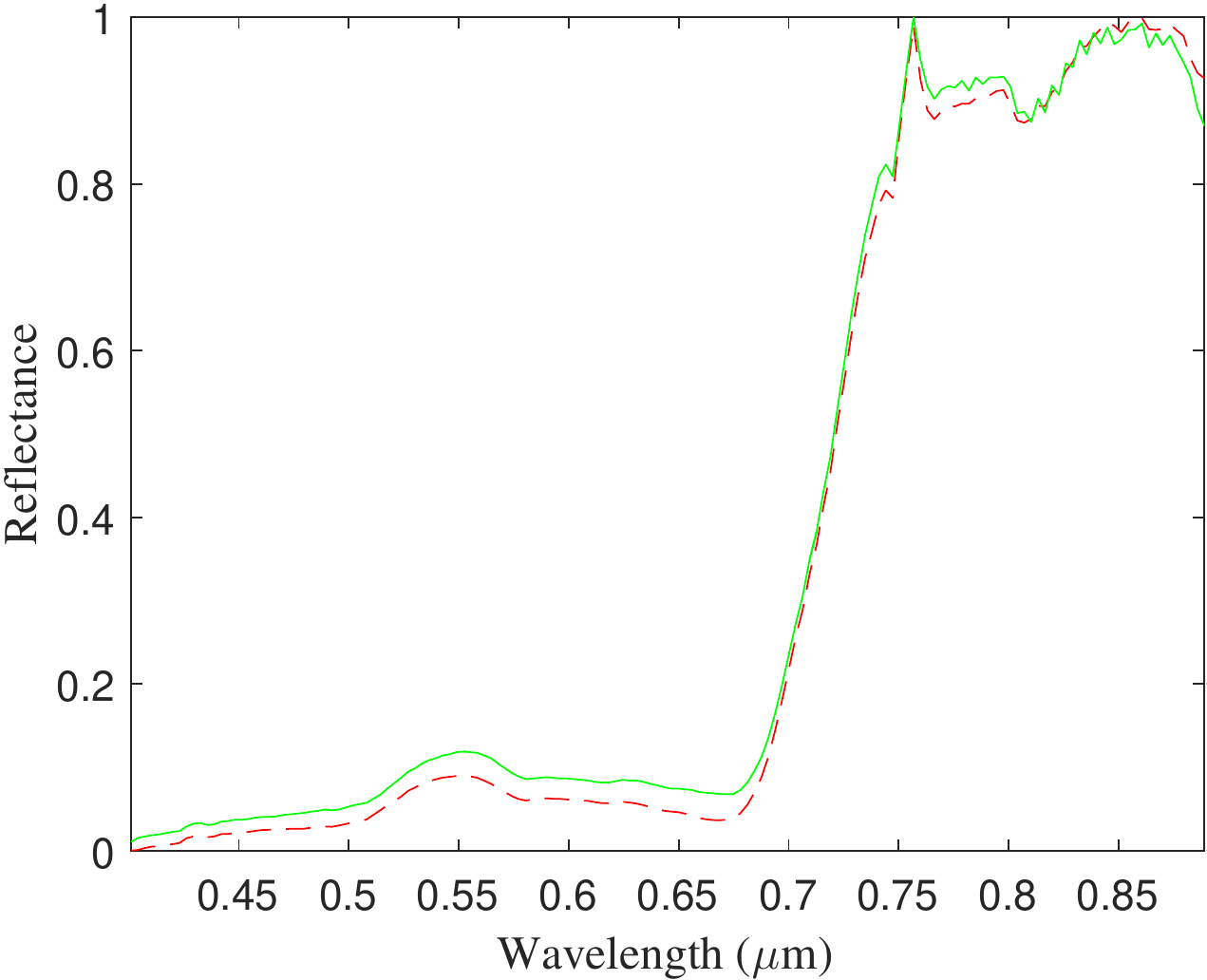}}}
\\
\mbox{
\subfigure[]{\includegraphics[width=2.45cm]{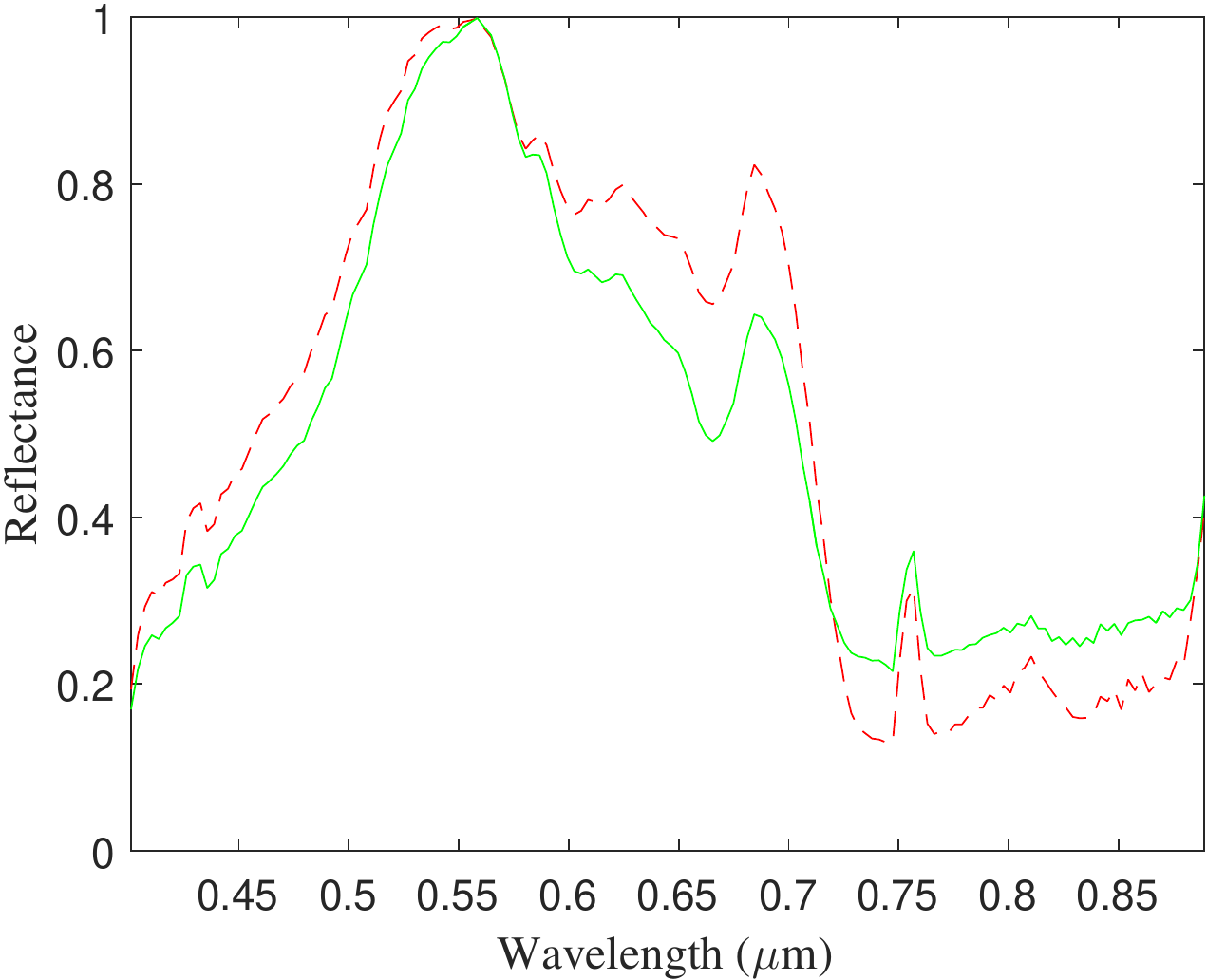}}
\subfigure[]{\includegraphics[width=2.45cm]{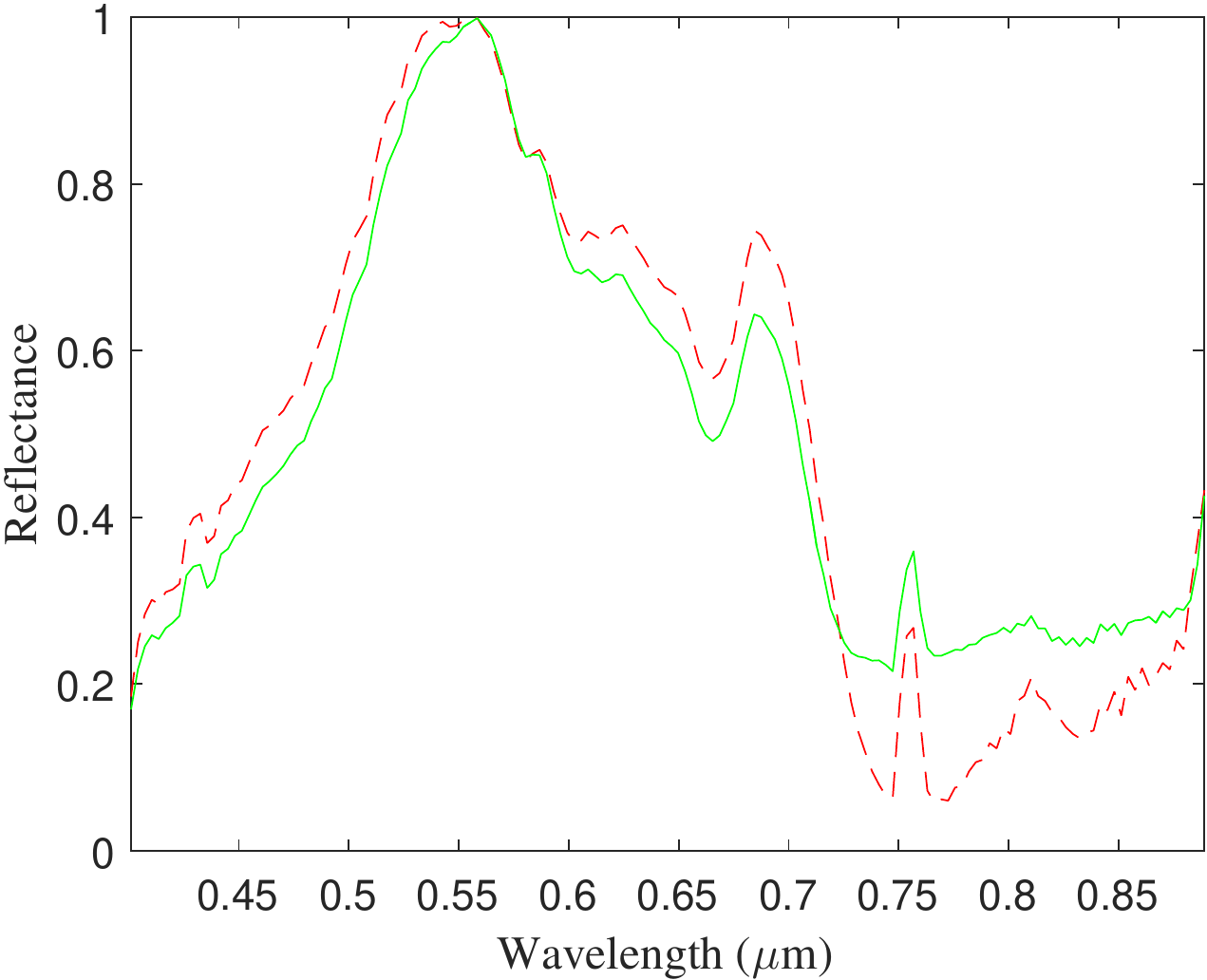}}
\subfigure[]{\includegraphics[width=2.45cm]{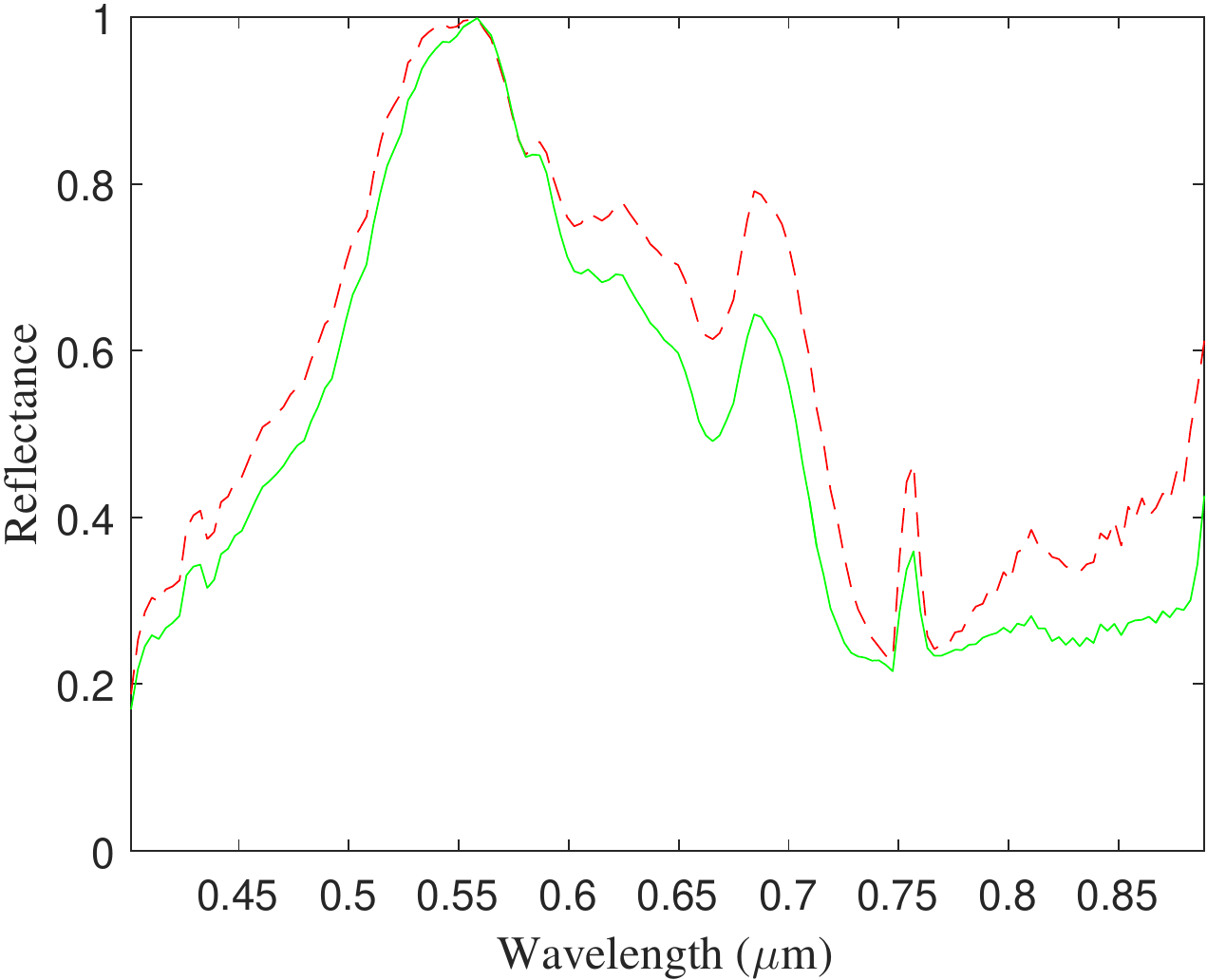}}
\subfigure[]{\includegraphics[width=2.45cm]{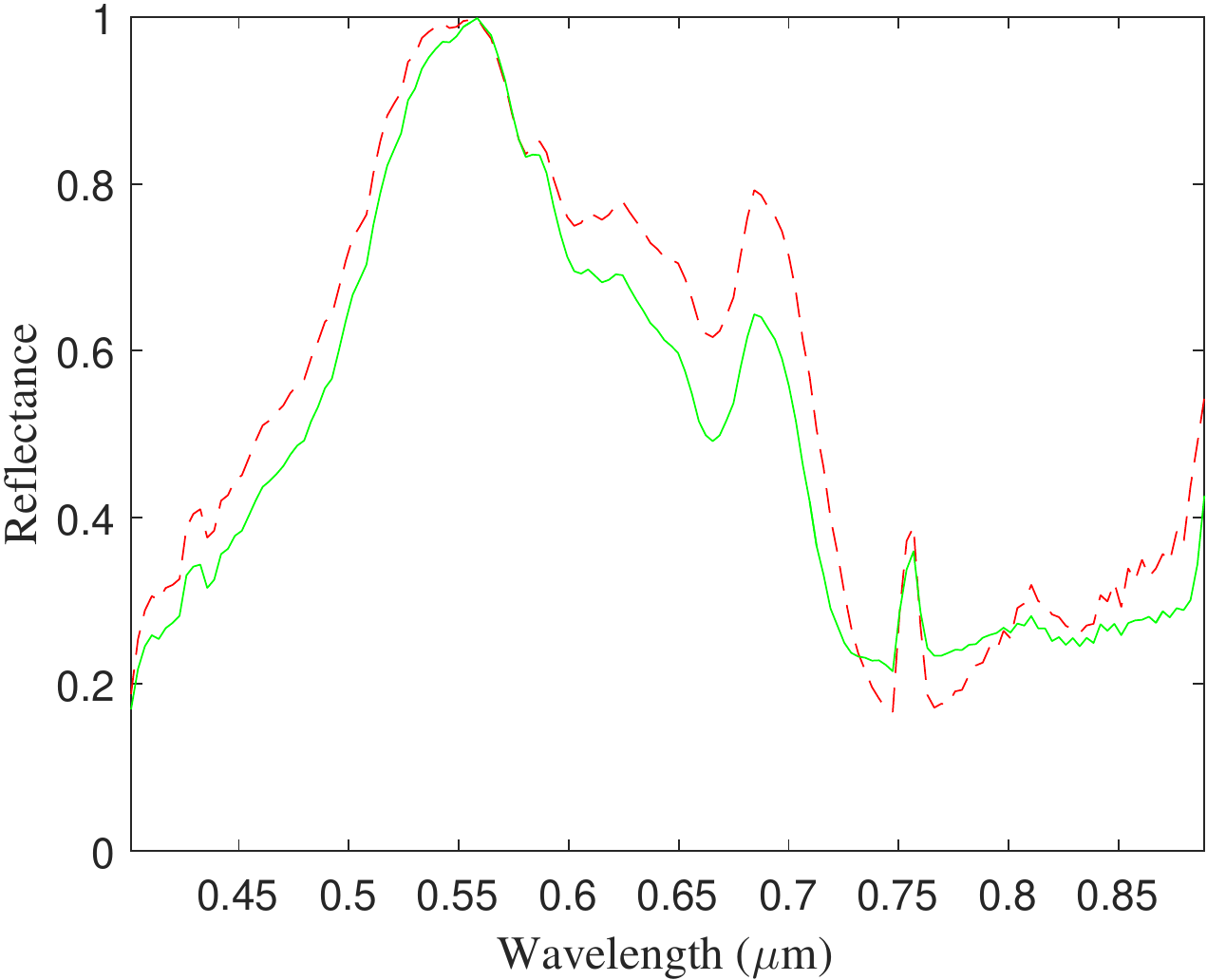}}
\subfigure[]{\includegraphics[width=2.45cm]{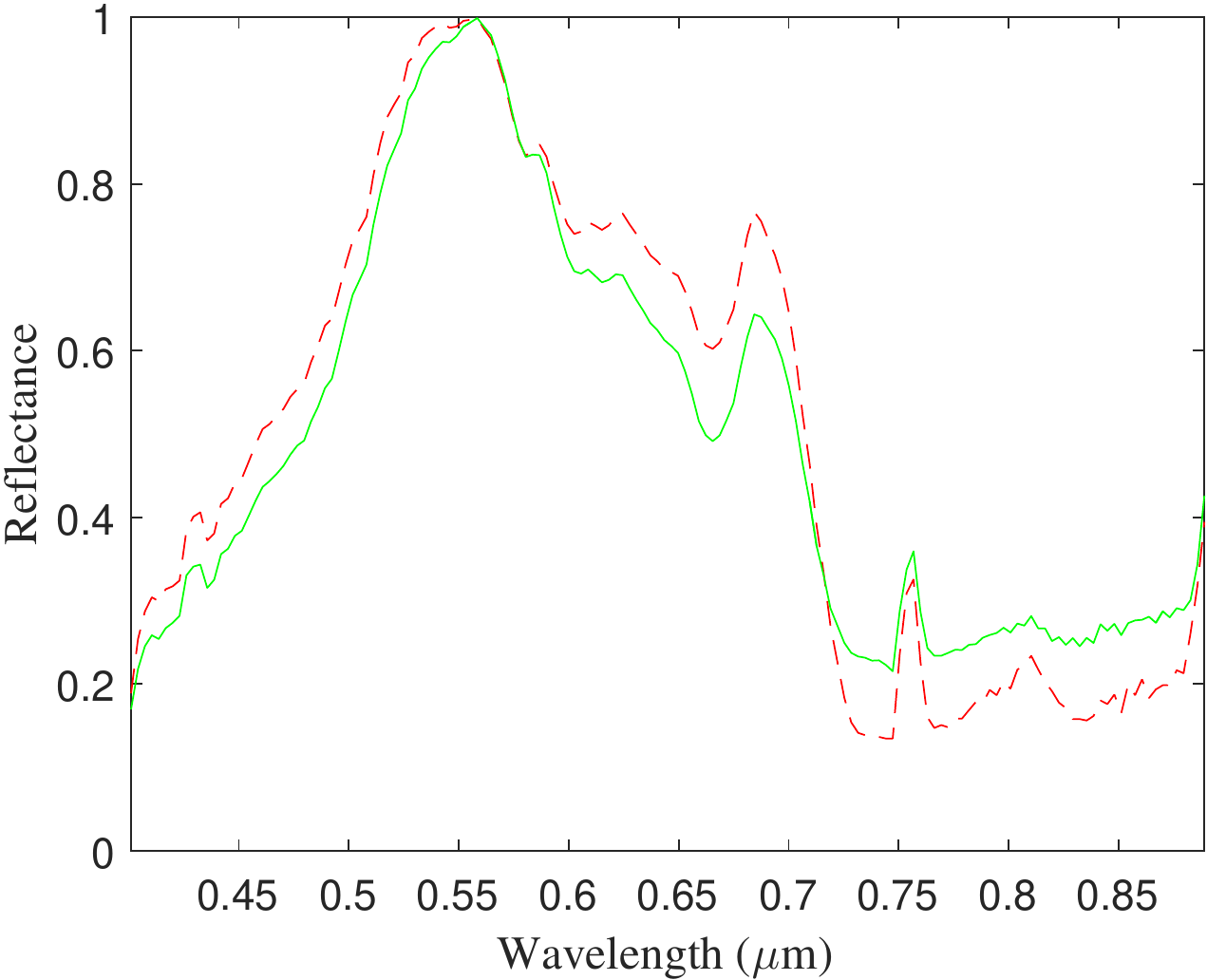}}
\subfigure[]{\includegraphics[width=2.45cm]{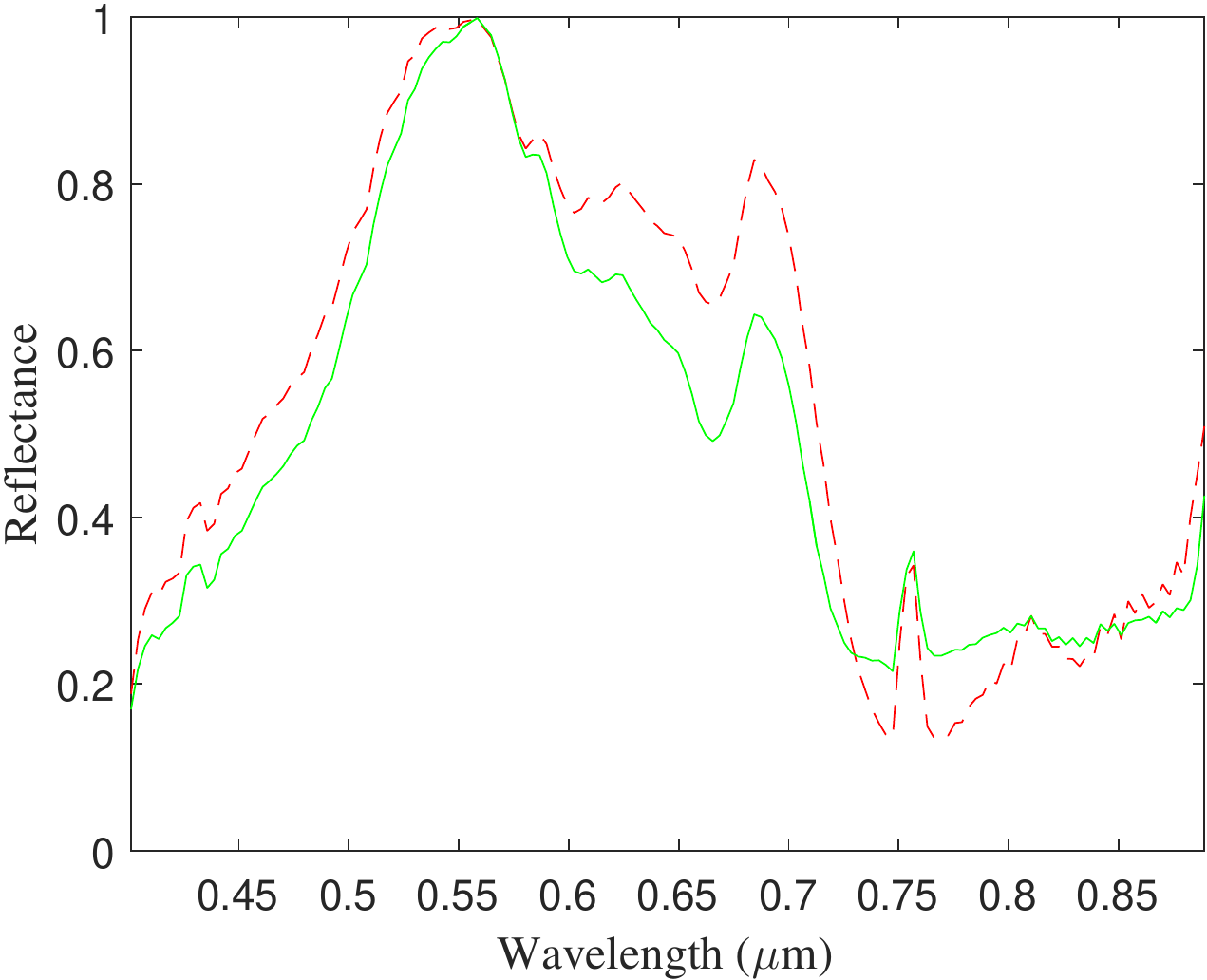}}
\subfigure[]{\includegraphics[width=2.45cm]{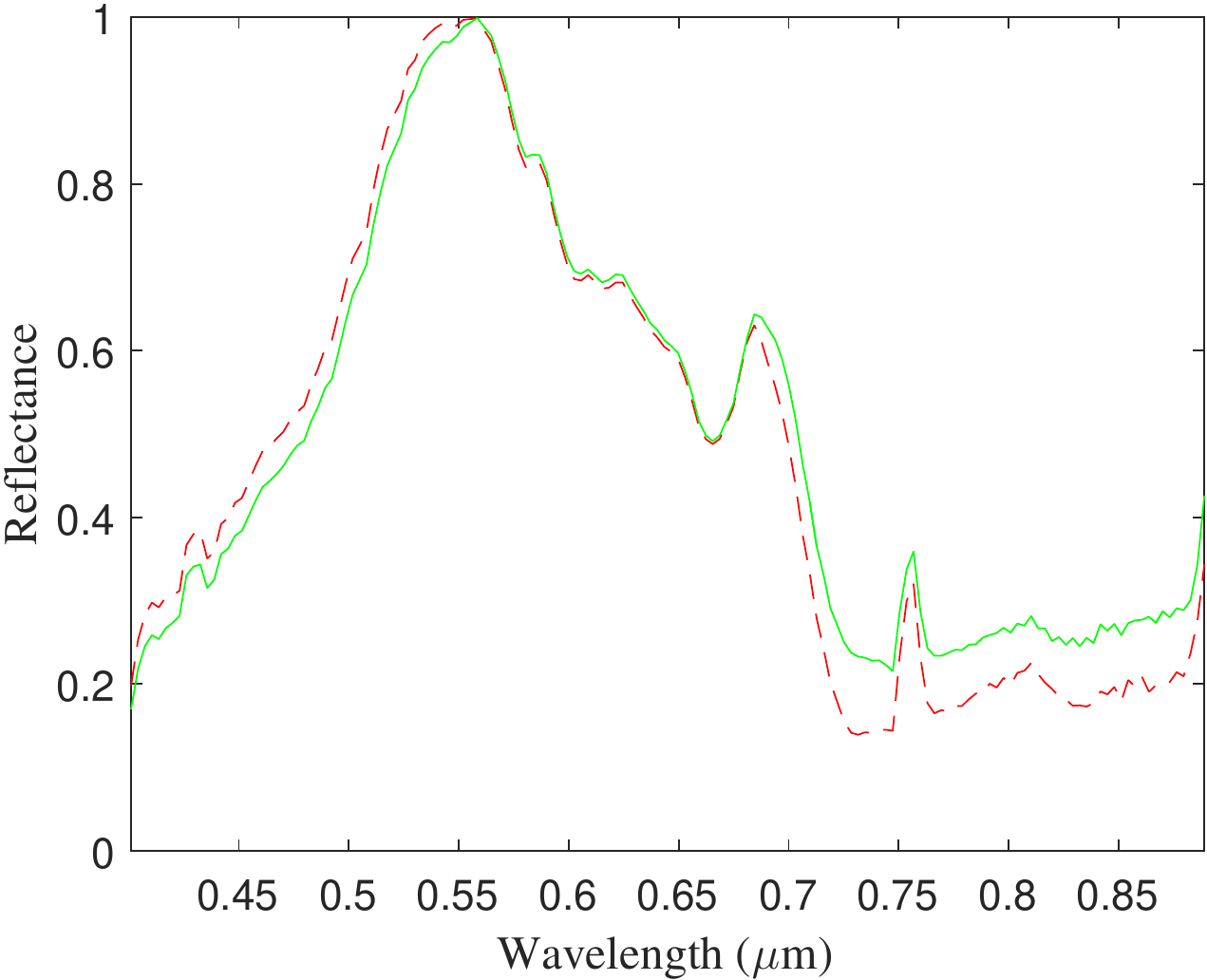}}}
\caption{Comparison of the reference spectra (green solid line) with those estimated by different methods (red dash line) of three endmembers on the Samson data set. From top to bottom: Soil. Tree. Water. From left to right: (a) $L_{1/2}$-NMF. (b) SGSNMF. (c) TV-RSNMF. (d) $L_{1/2}$-RNMF. (e) MV-NTF-TV. (f) MLNMF. (g) SSRDMF.}
\label{fig:7}
\end{figure*}

\begin{figure*}[!t]
\centering
\mbox{
{\includegraphics[width=2.45cm]{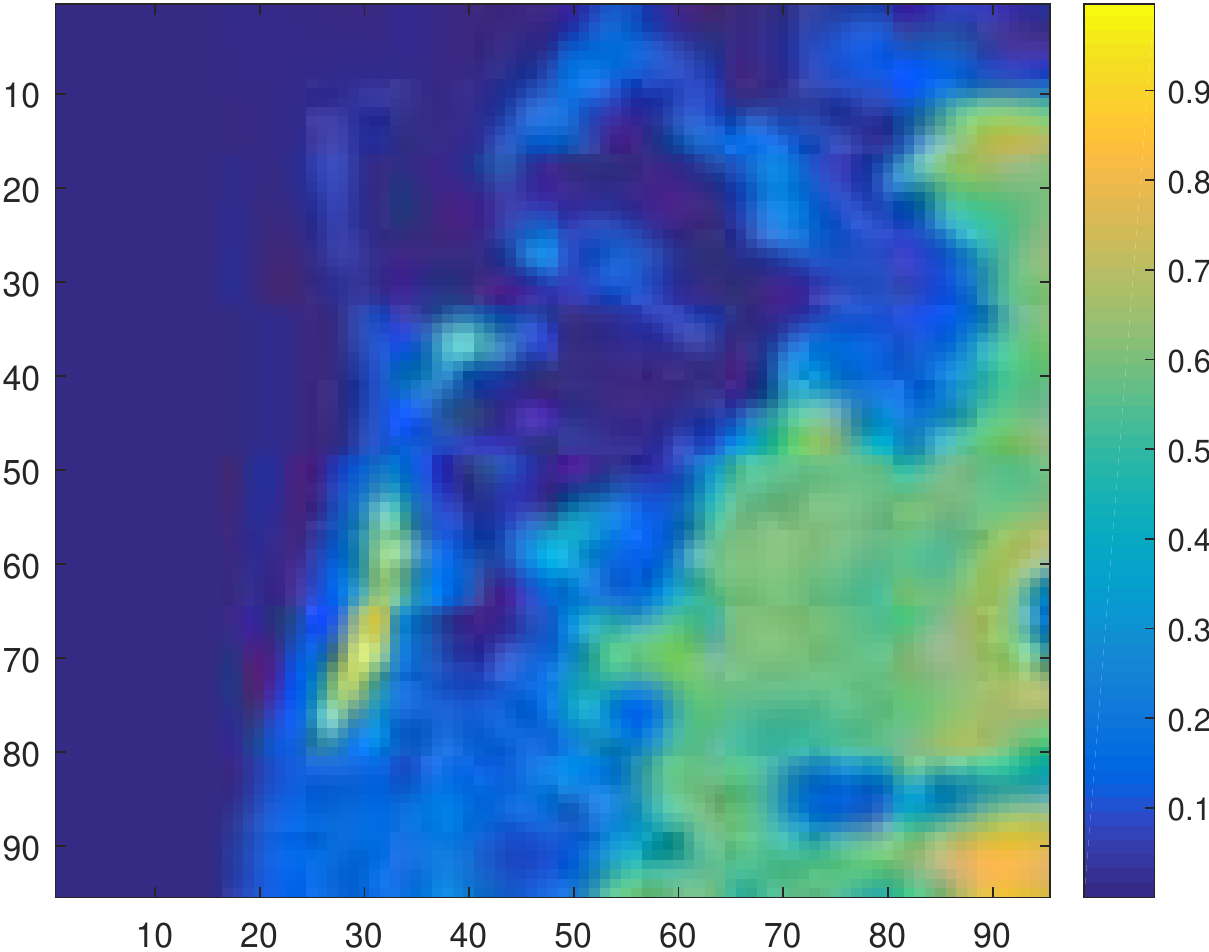}}
{\includegraphics[width=2.45cm]{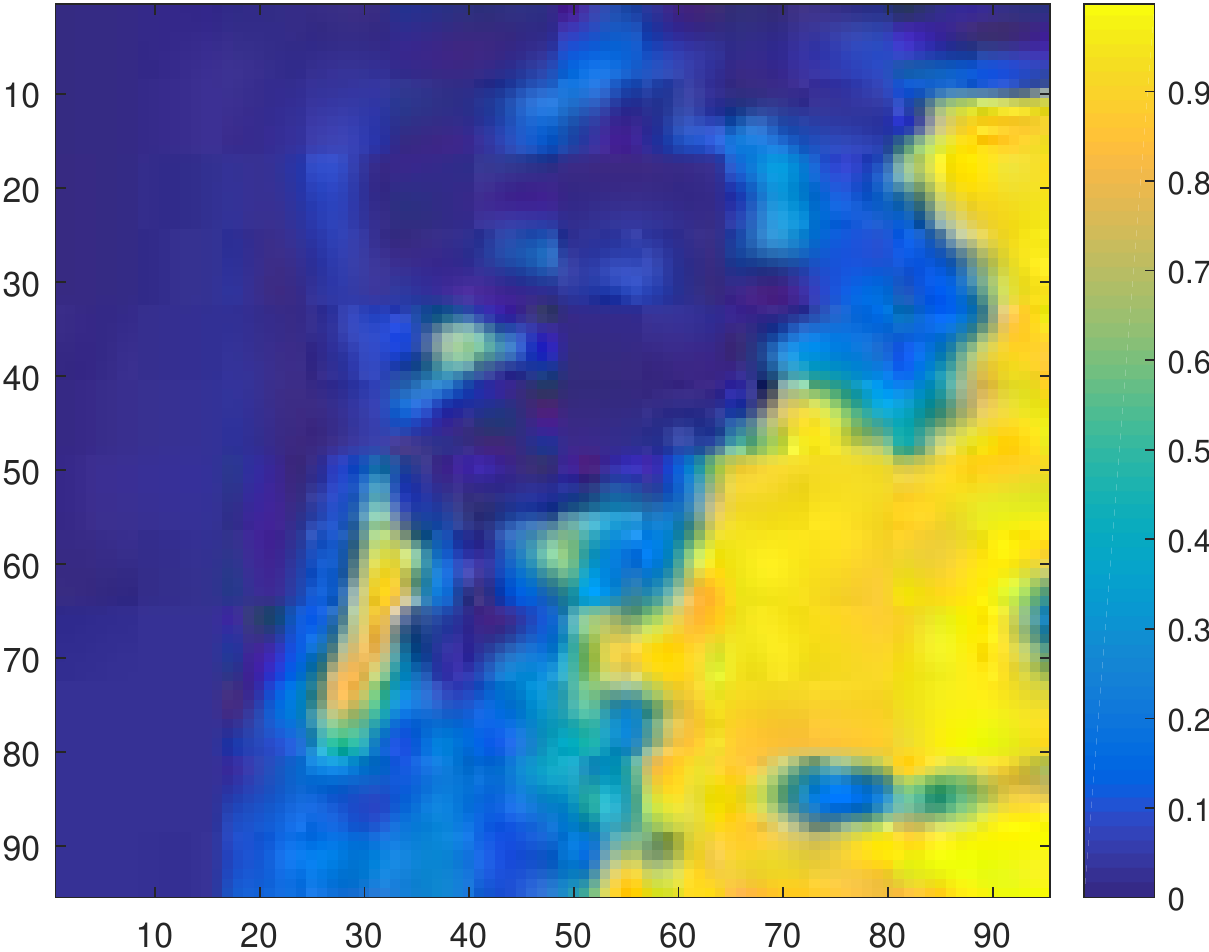}}
{\includegraphics[width=2.45cm]{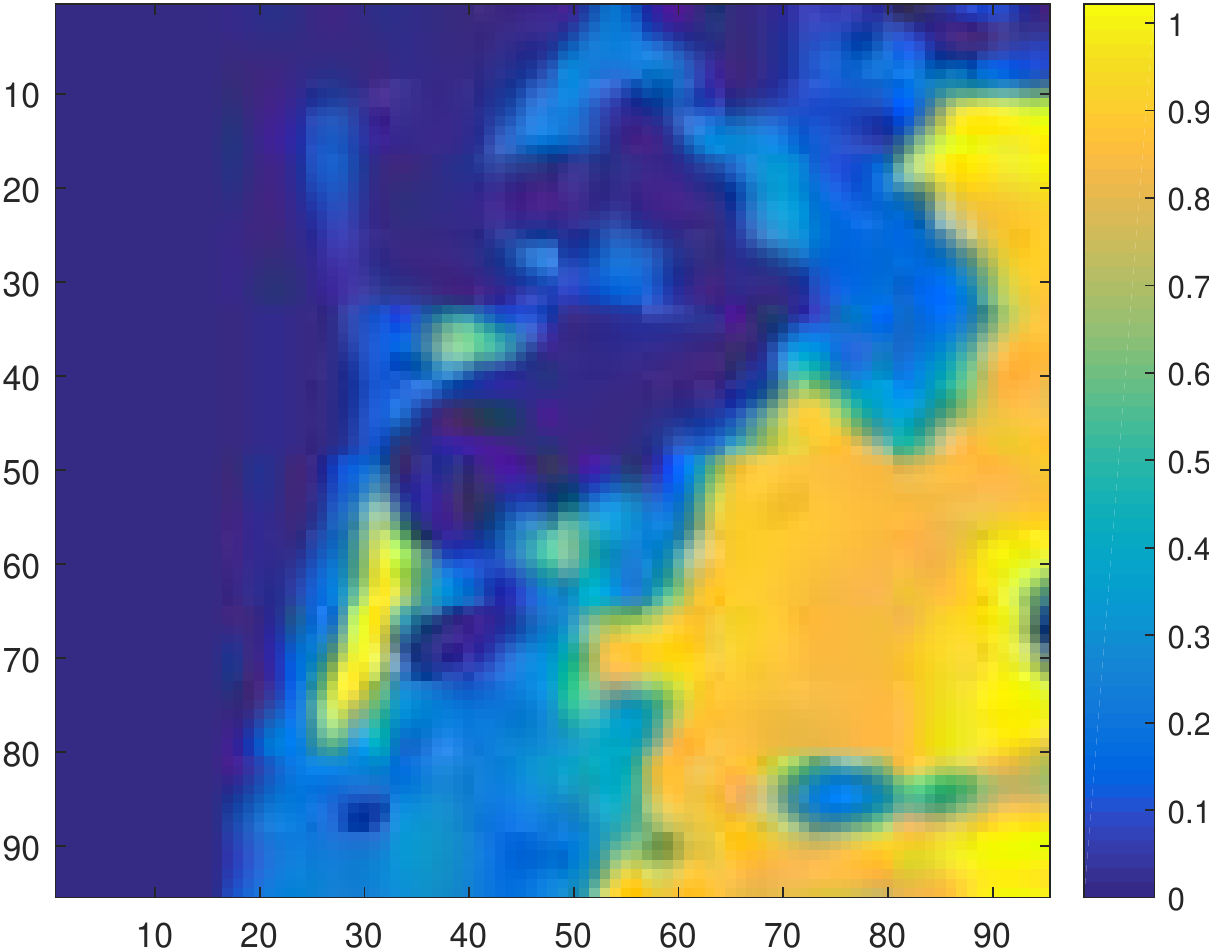}}
{\includegraphics[width=2.45cm]{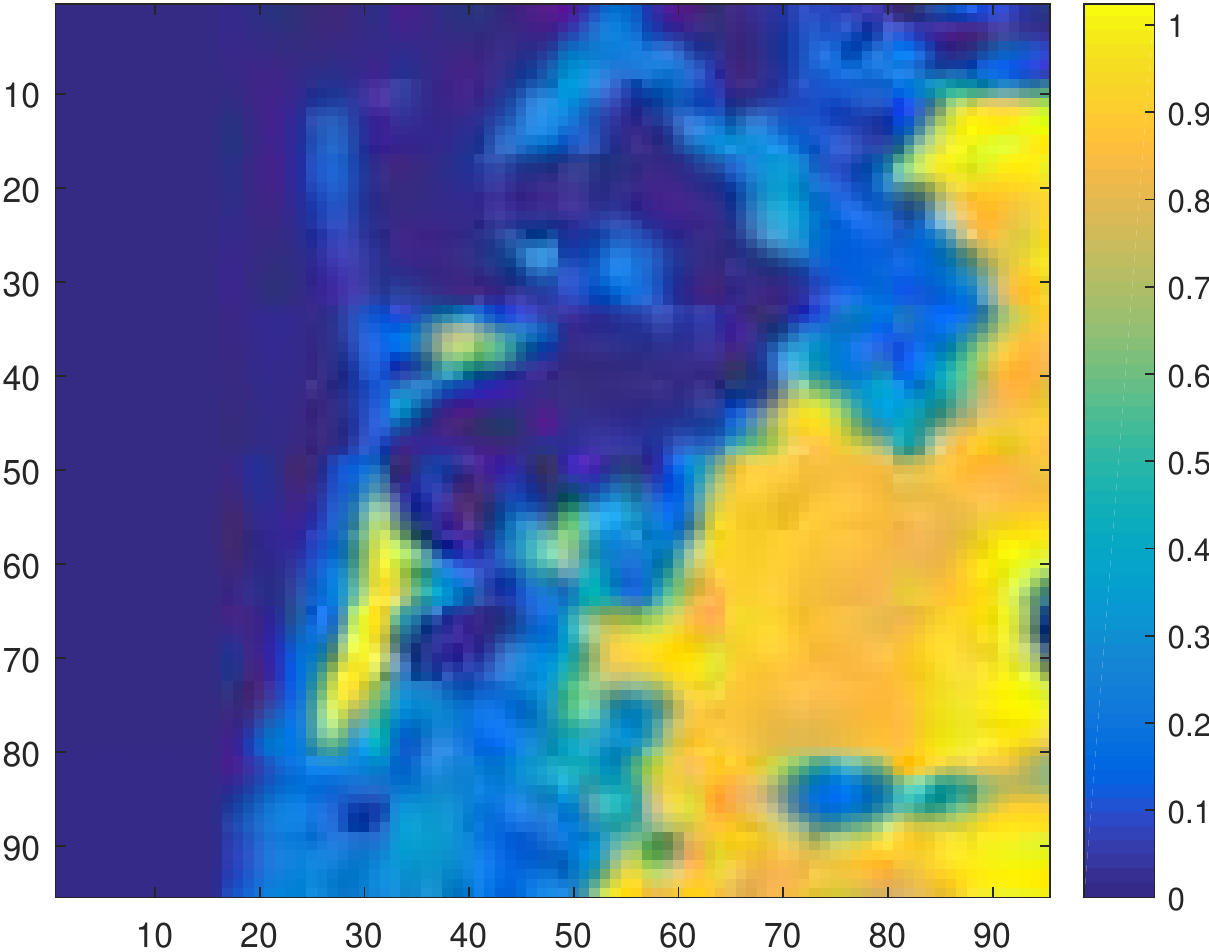}}
{\includegraphics[width=2.45cm]{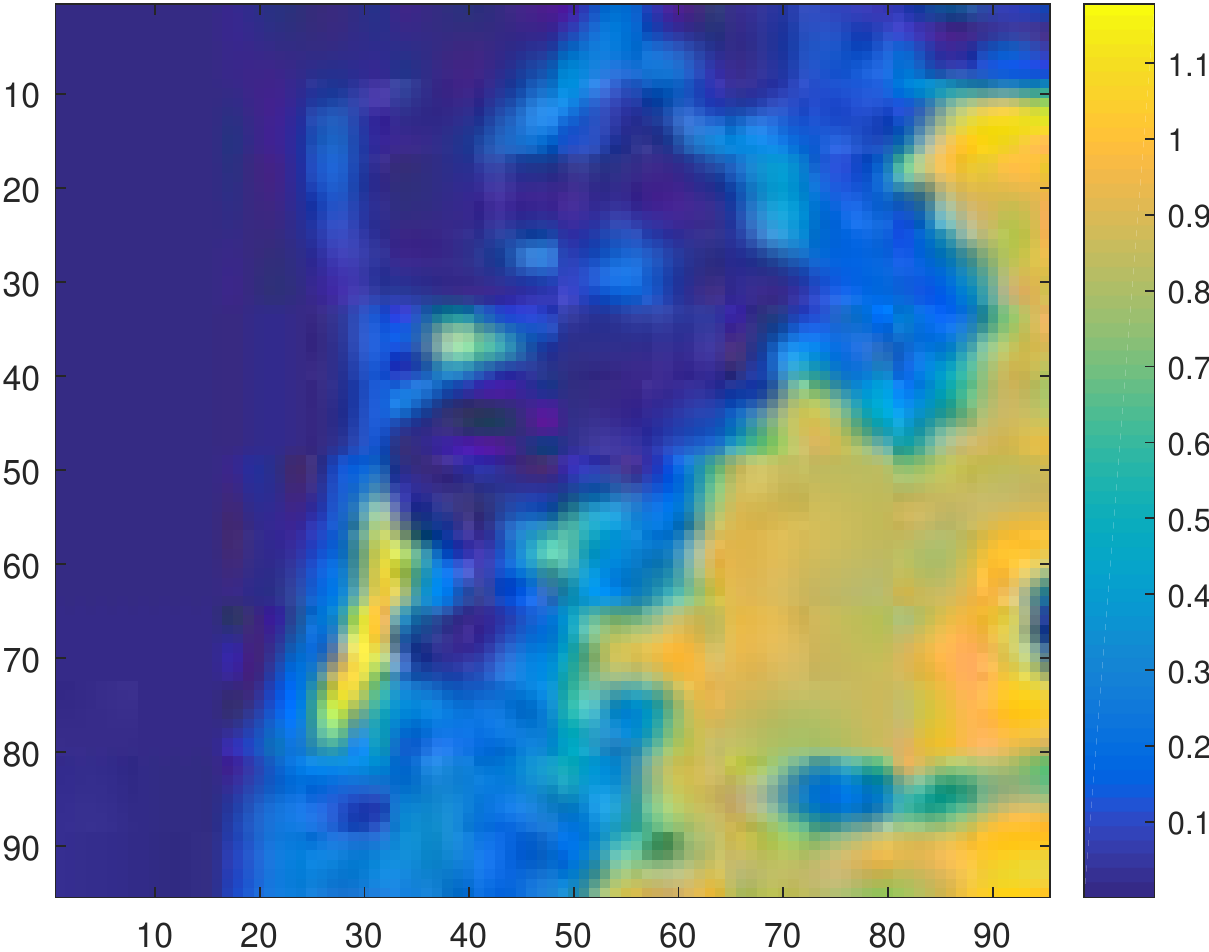}}
{\includegraphics[width=2.45cm]{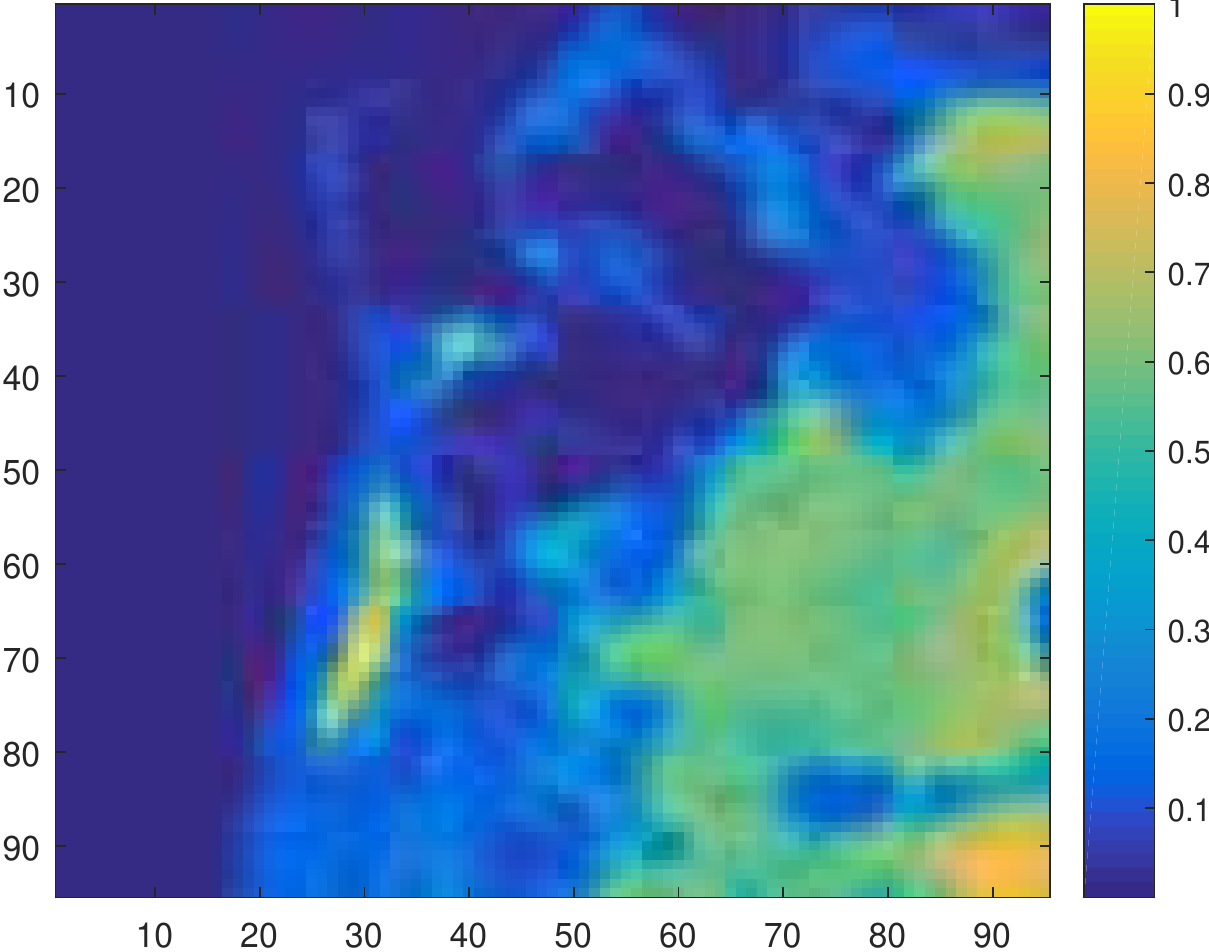}}
{\includegraphics[width=2.45cm]{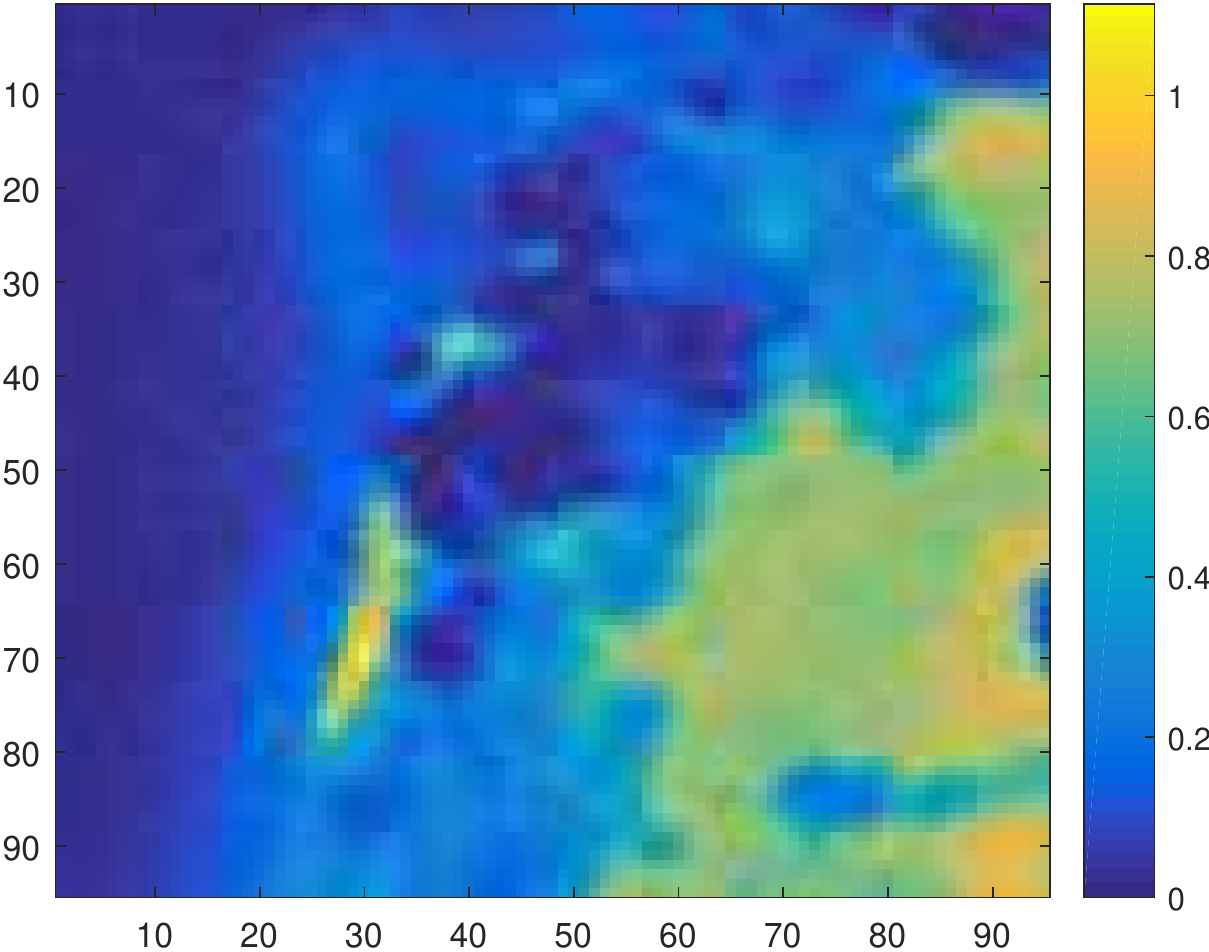}}}
\\
\mbox{
{\includegraphics[width=2.45cm]{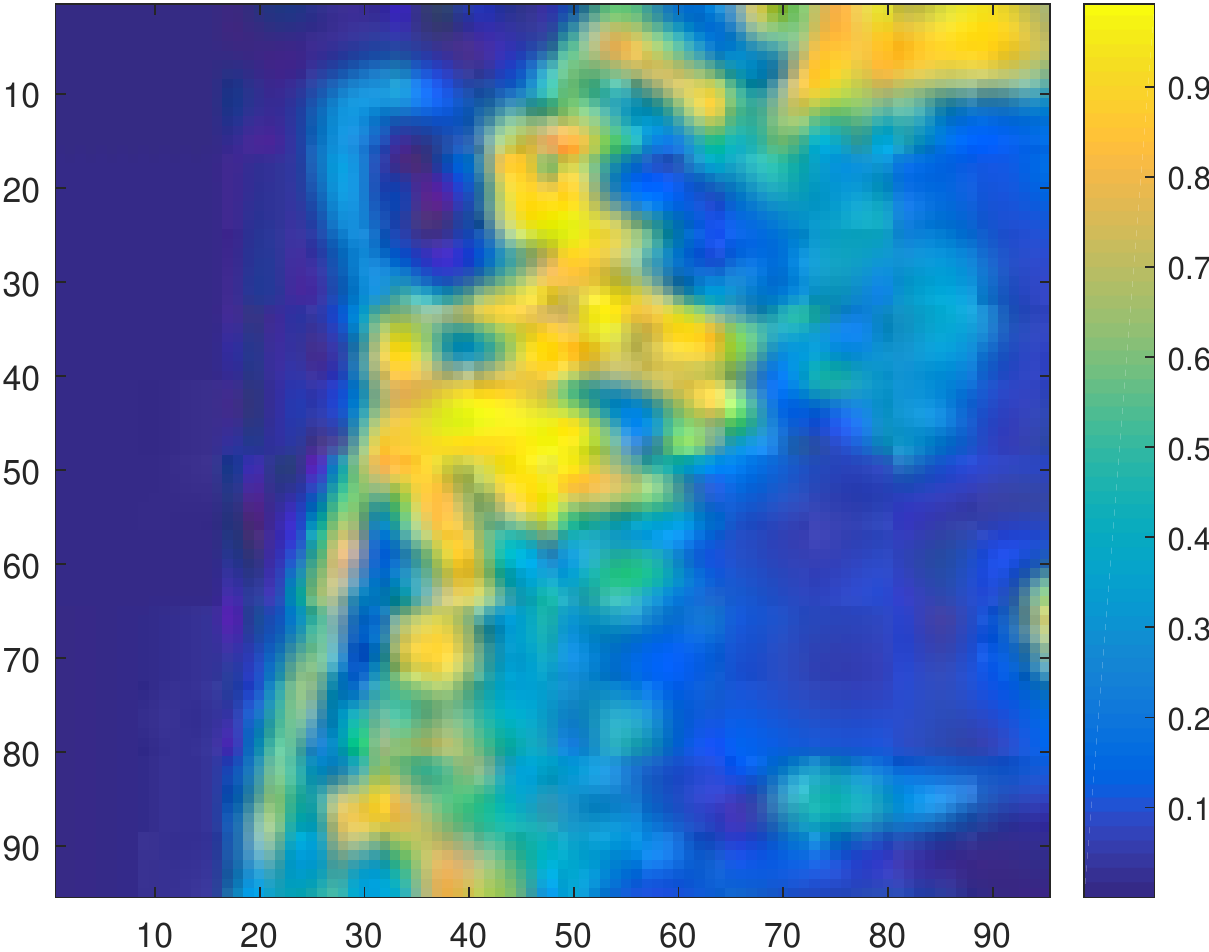}}
{\includegraphics[width=2.45cm]{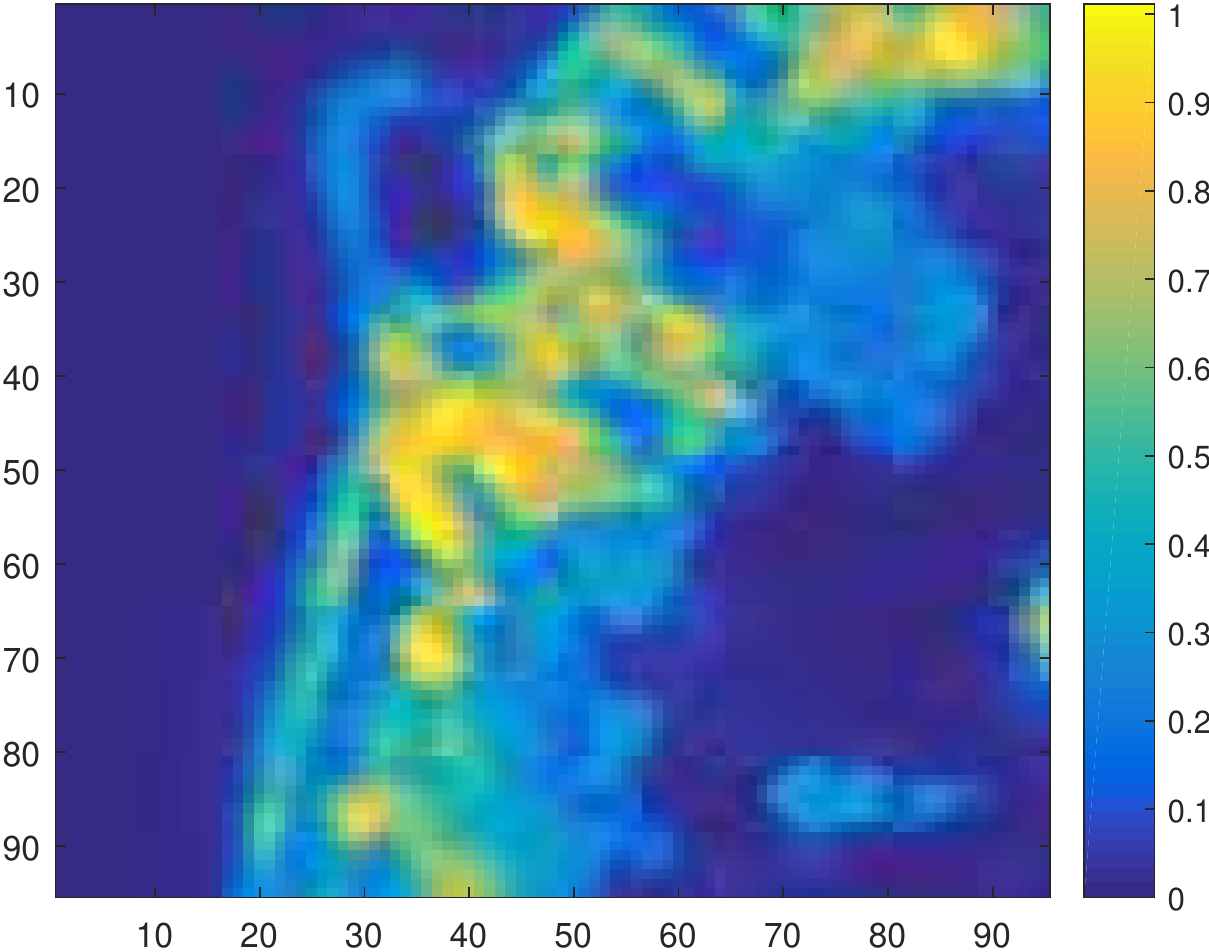}}
{\includegraphics[width=2.45cm]{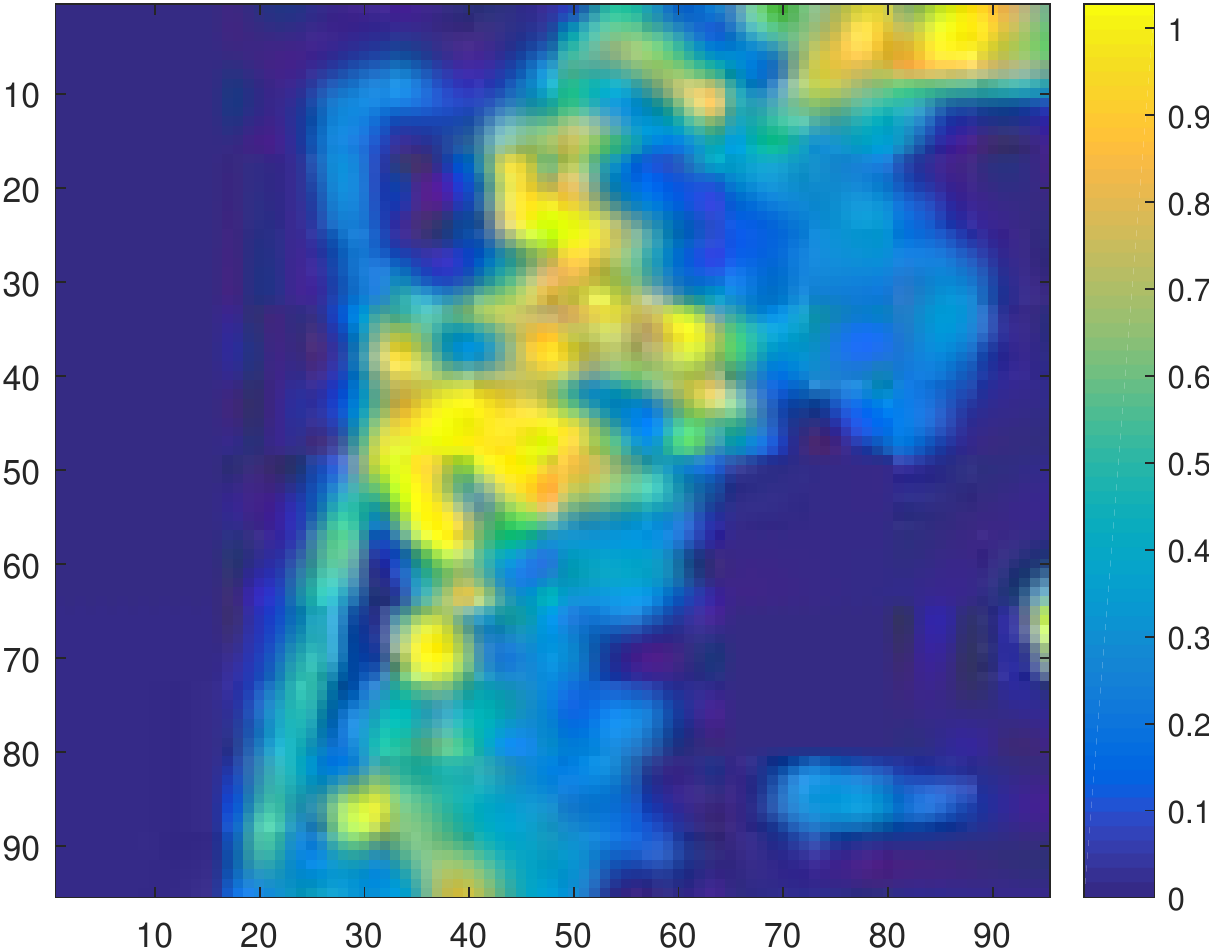}}
{\includegraphics[width=2.45cm]{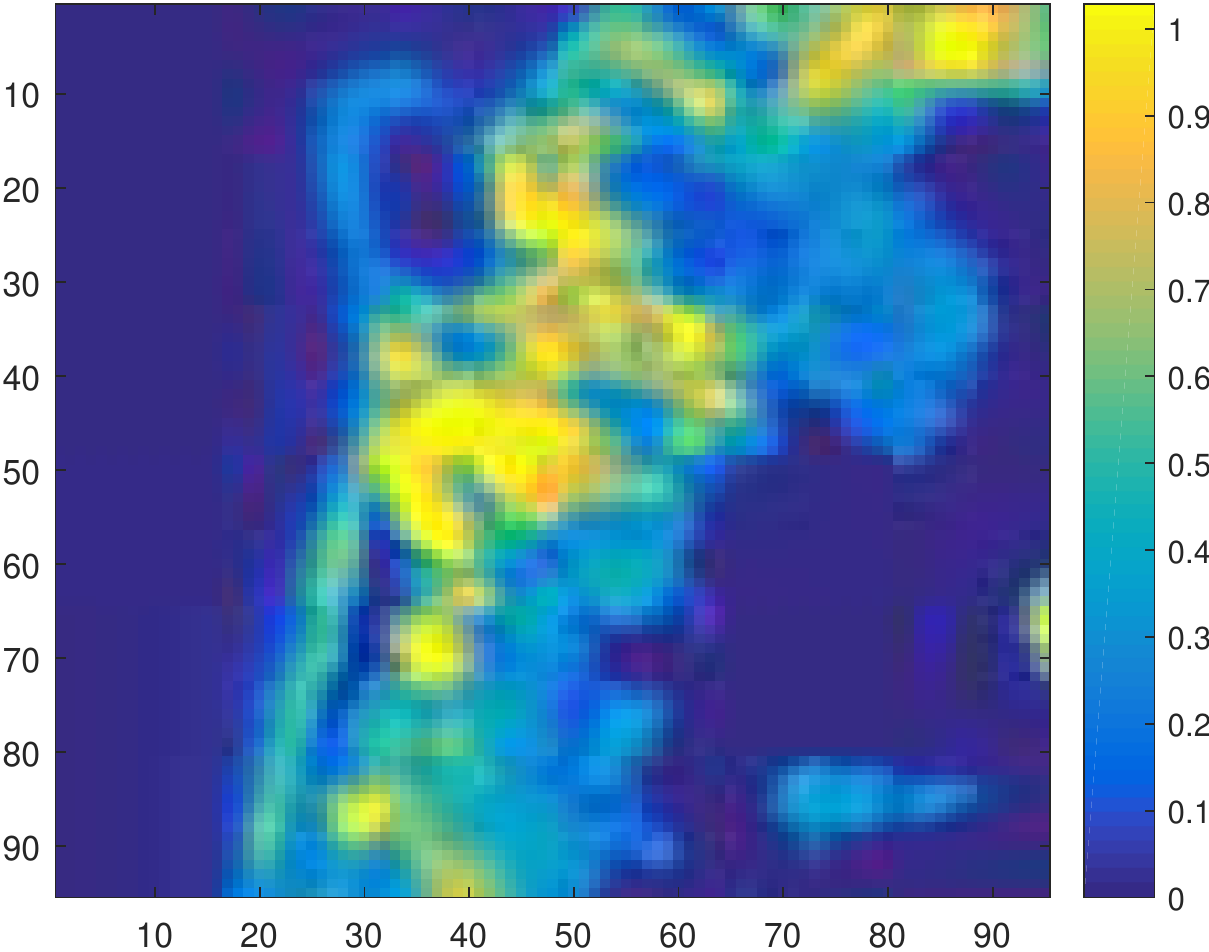}}
{\includegraphics[width=2.45cm]{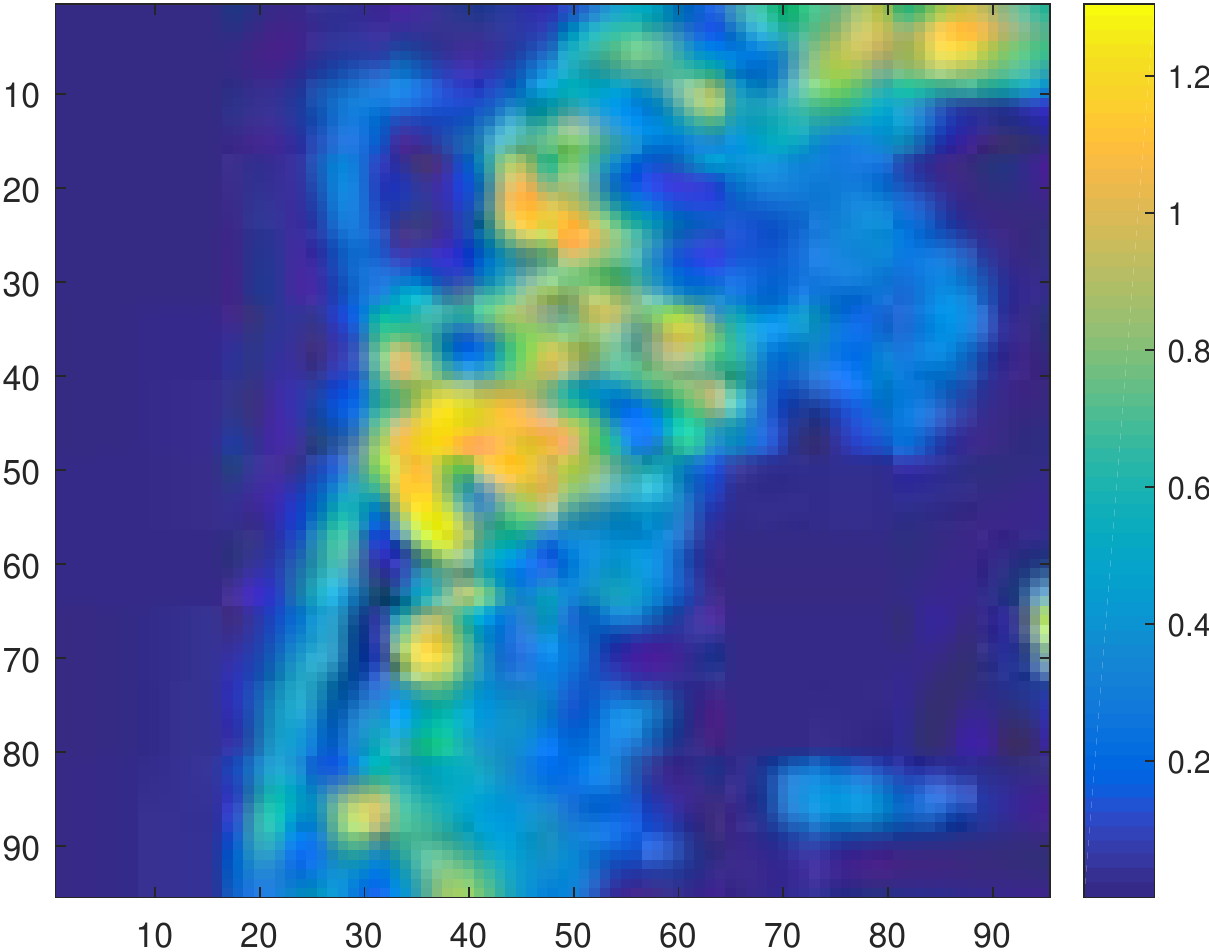}}
{\includegraphics[width=2.45cm]{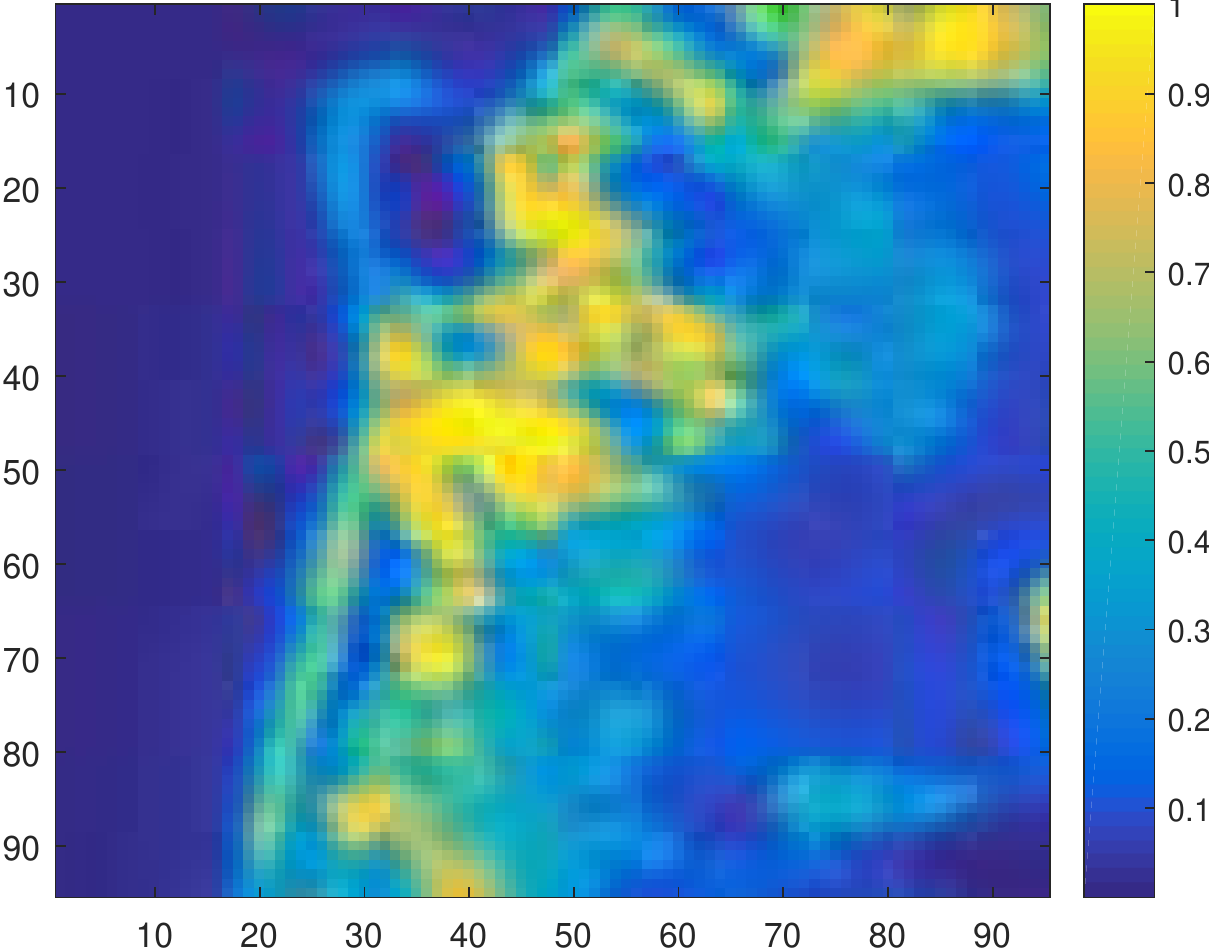}}
{\includegraphics[width=2.45cm]{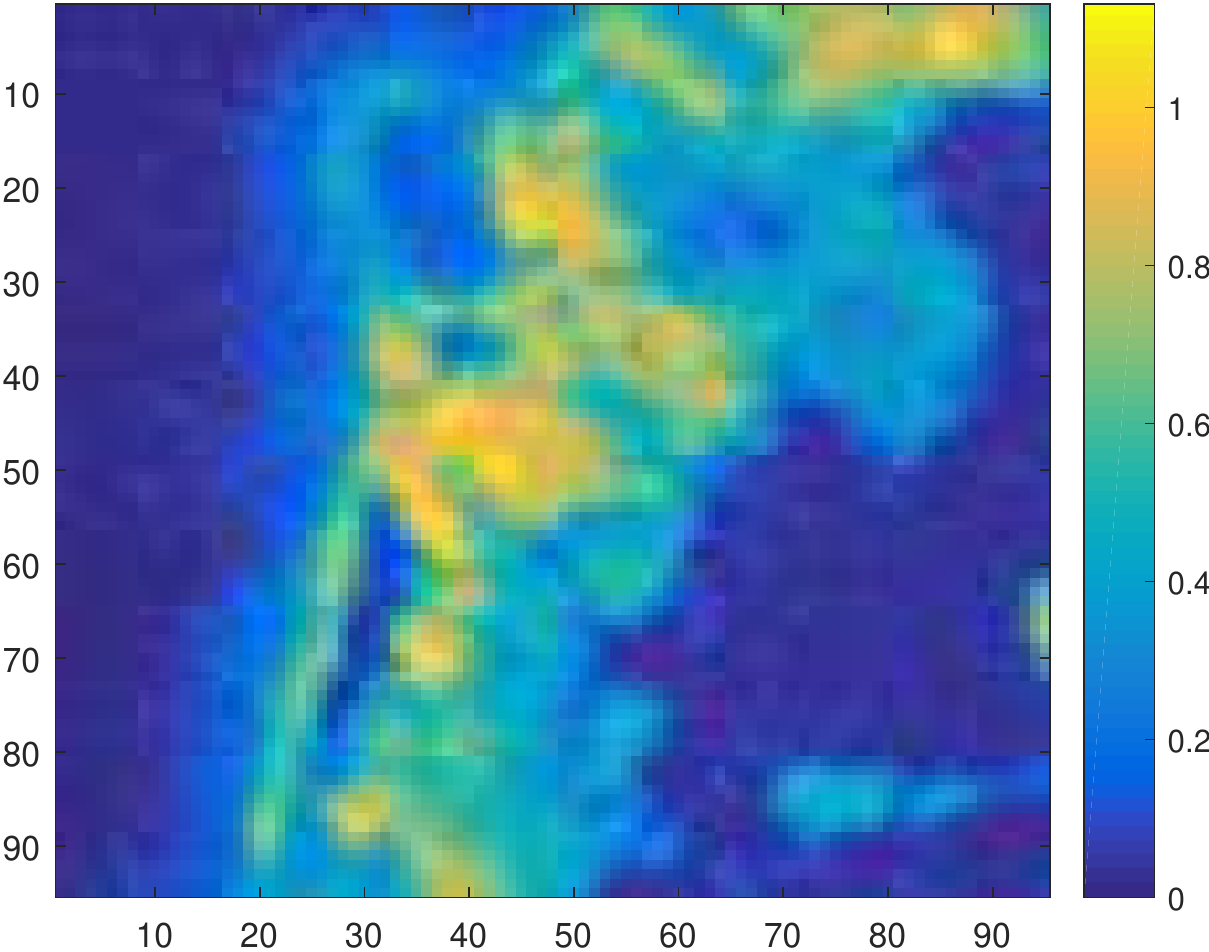}}}
\\
\mbox{
\subfigure[]{\includegraphics[width=2.45cm]{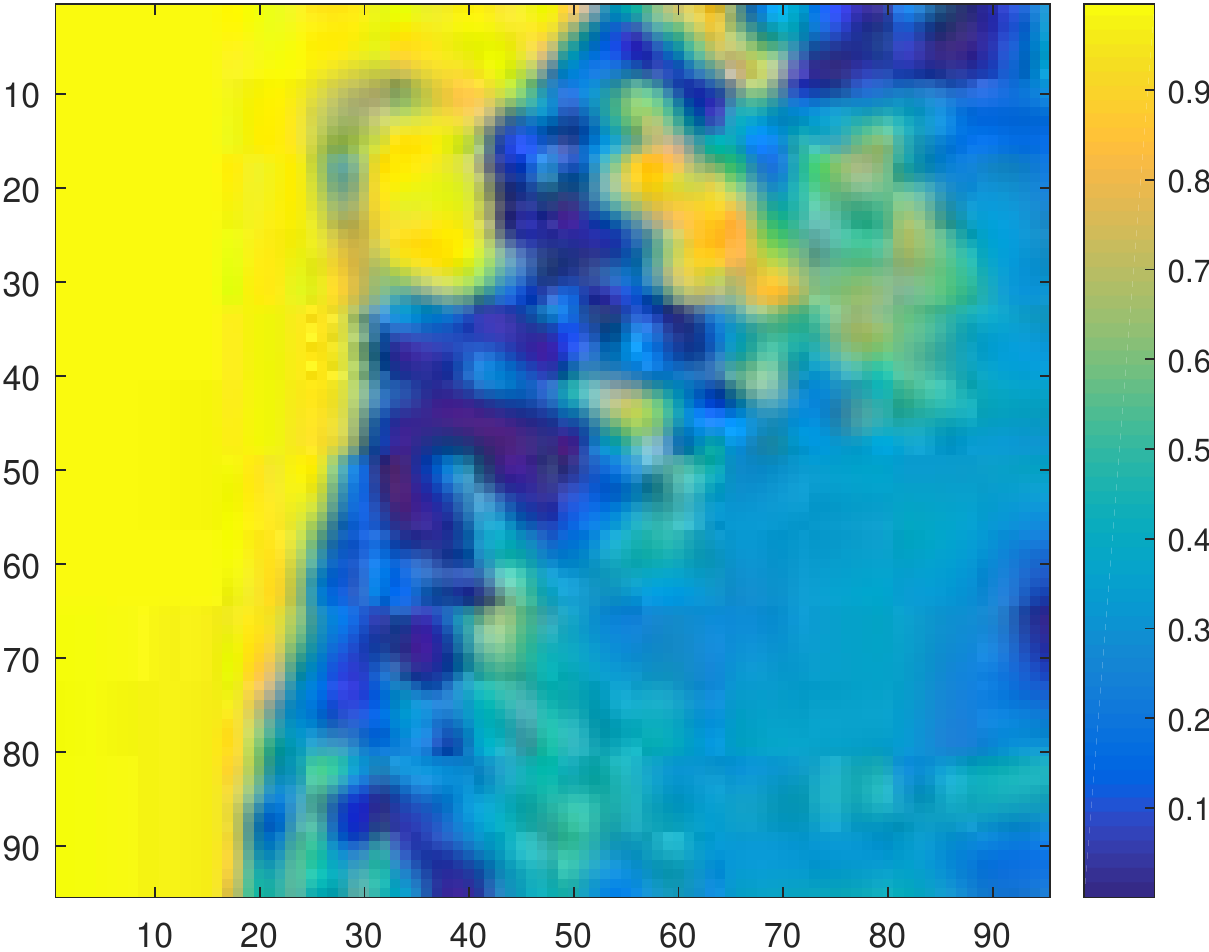}}
\subfigure[]{\includegraphics[width=2.45cm]{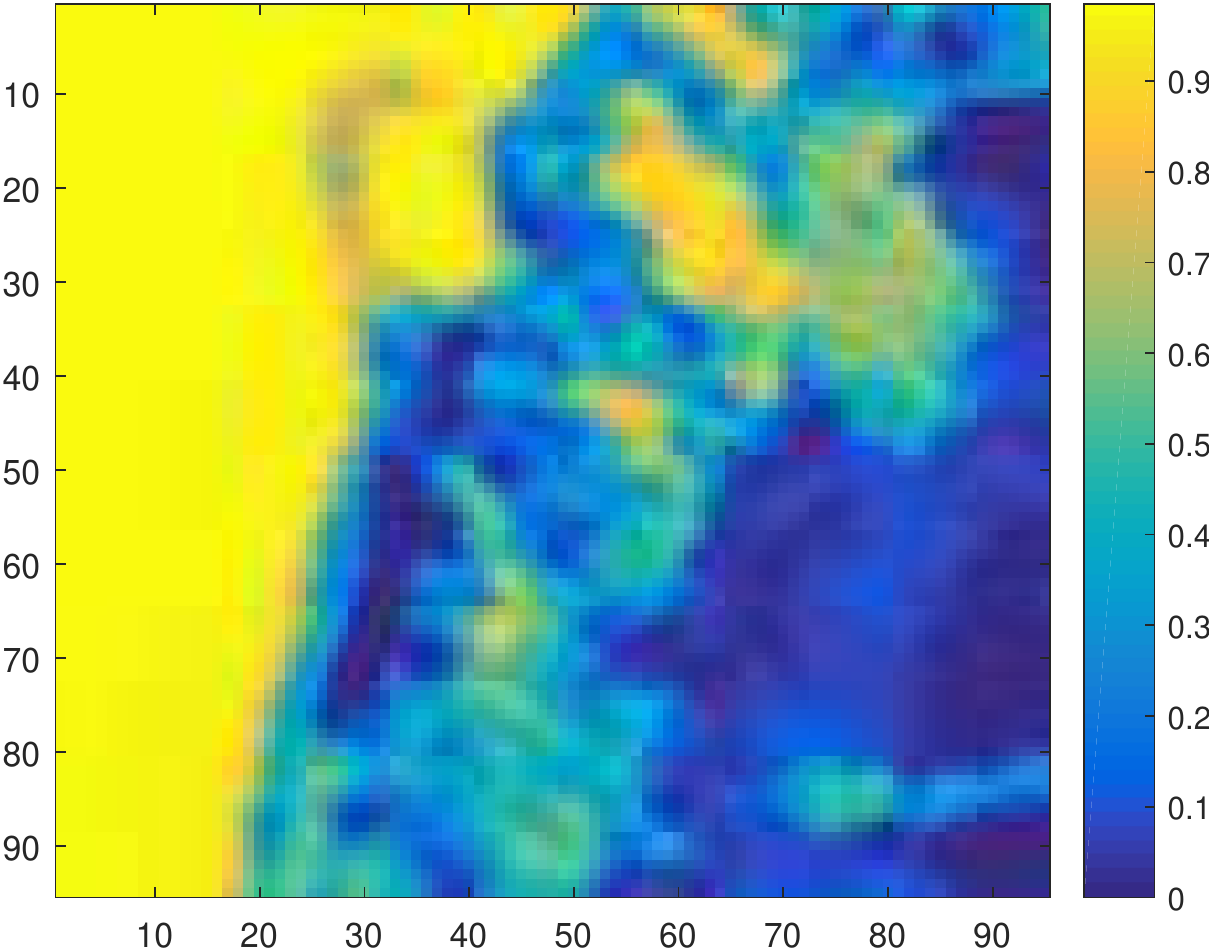}}
\subfigure[]{\includegraphics[width=2.45cm]{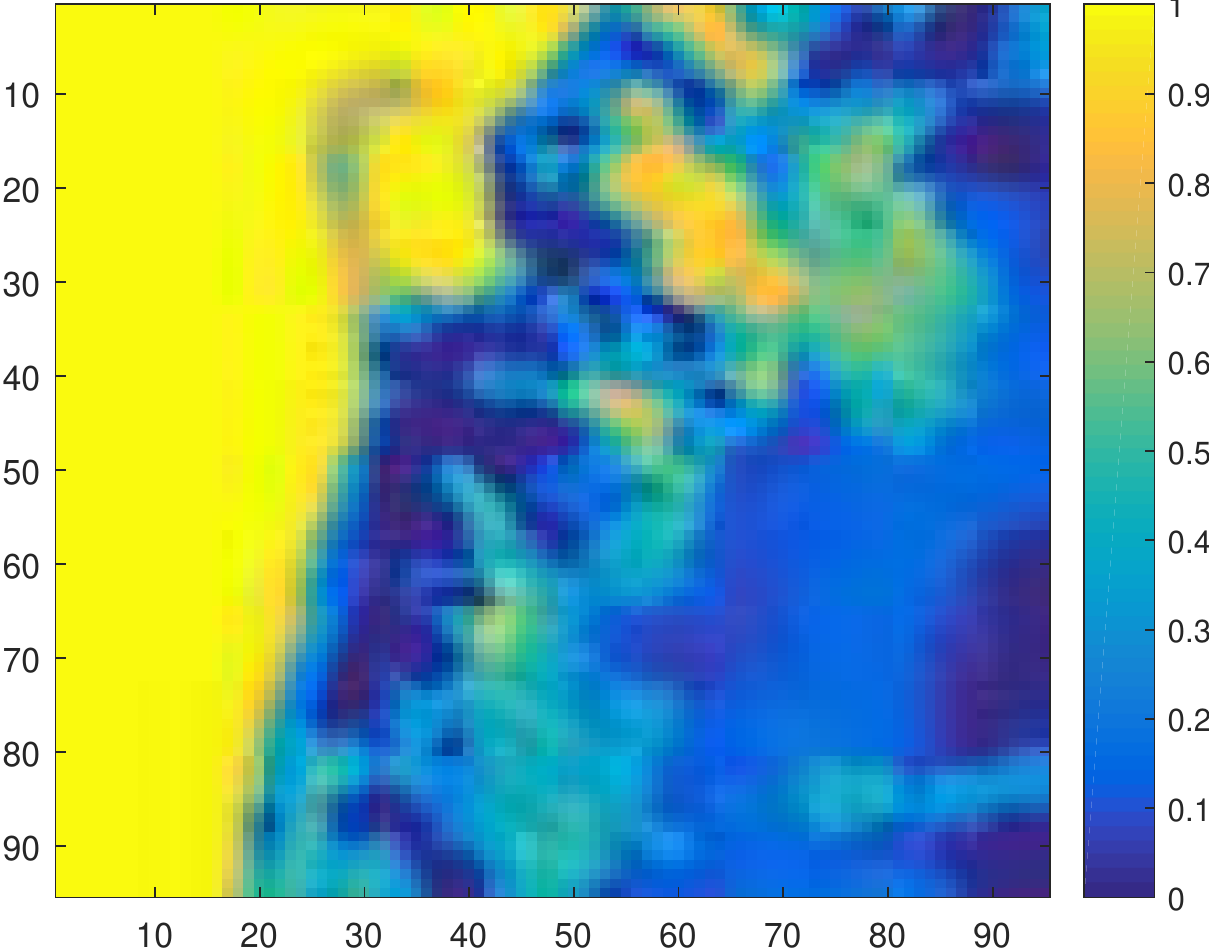}}
\subfigure[]{\includegraphics[width=2.45cm]{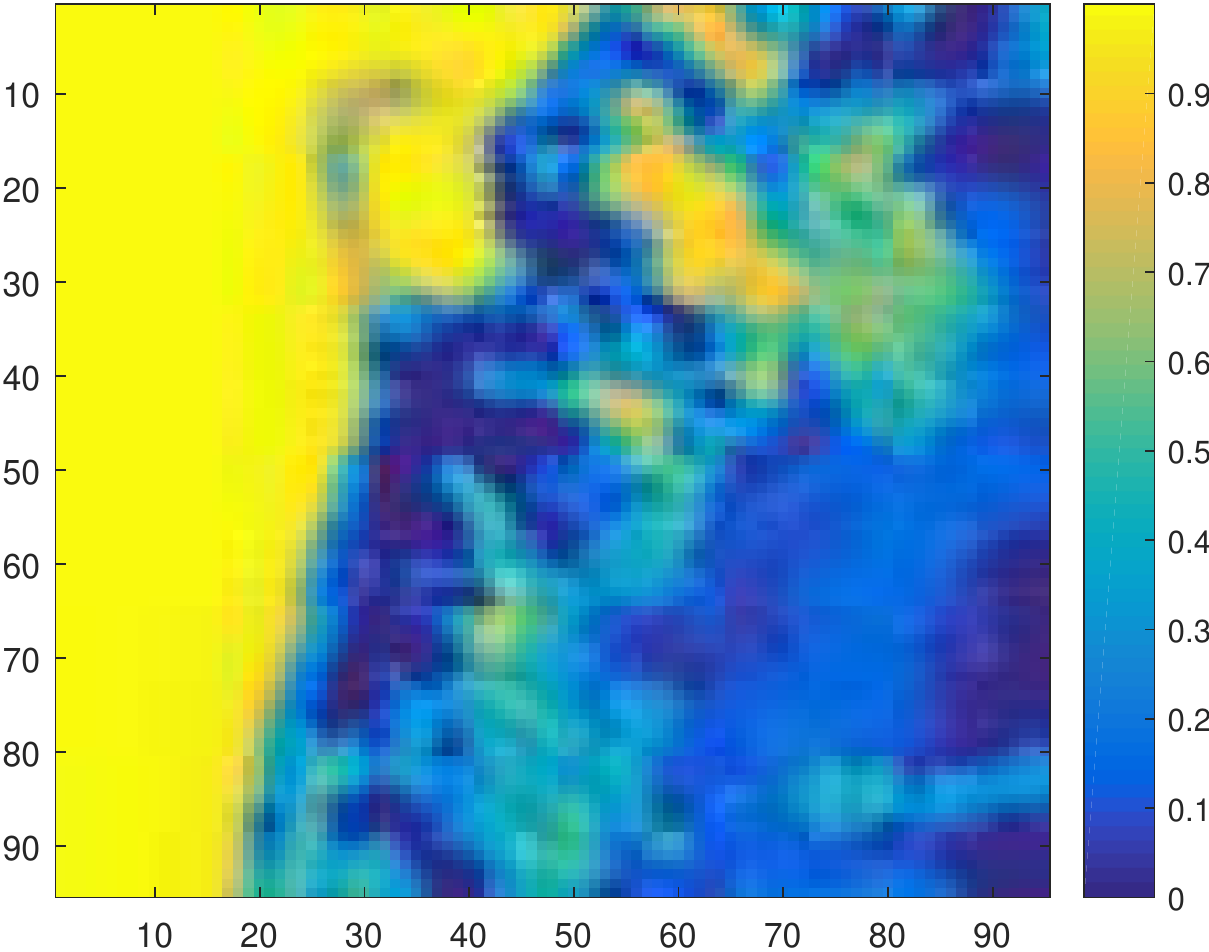}}
\subfigure[]{\includegraphics[width=2.45cm]{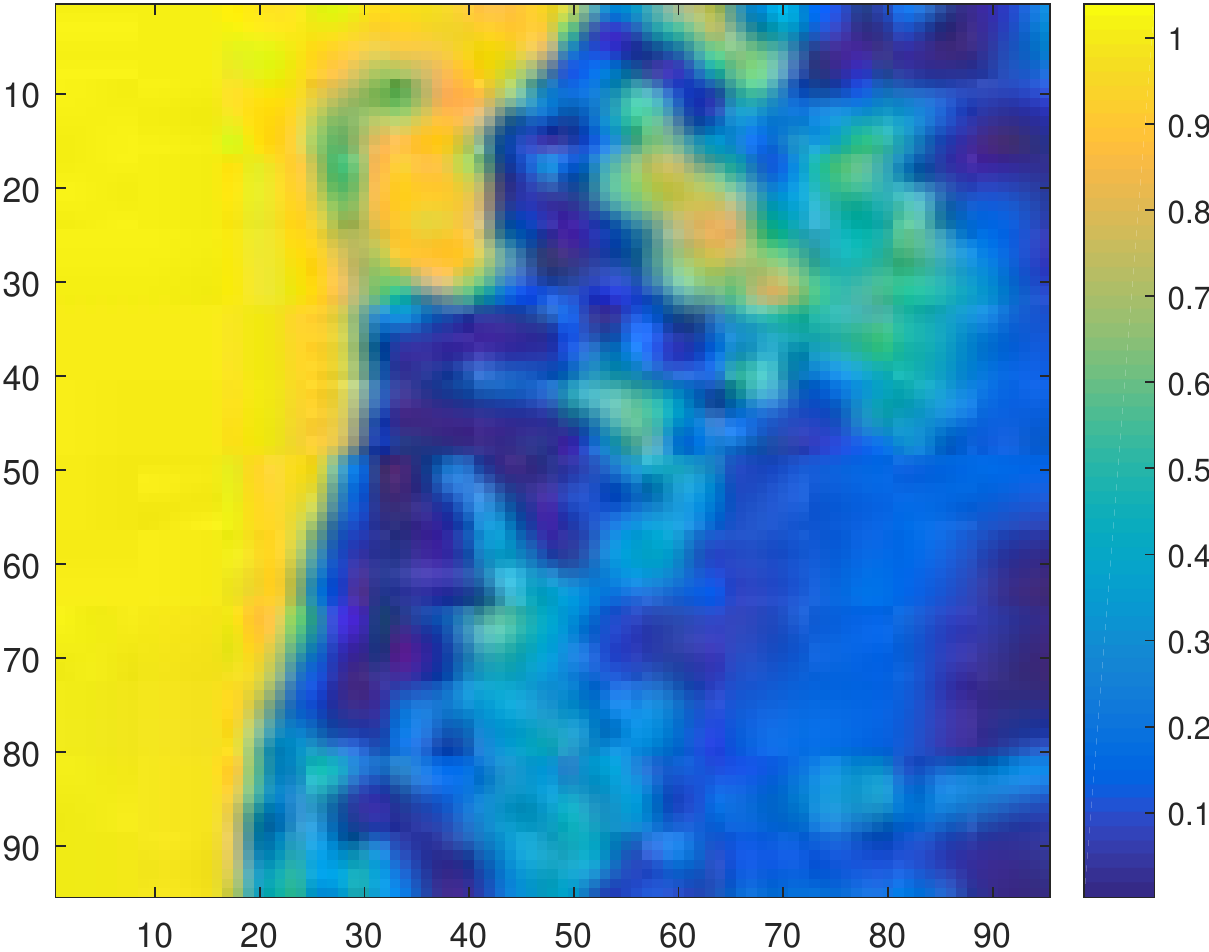}}
\subfigure[]{\includegraphics[width=2.45cm]{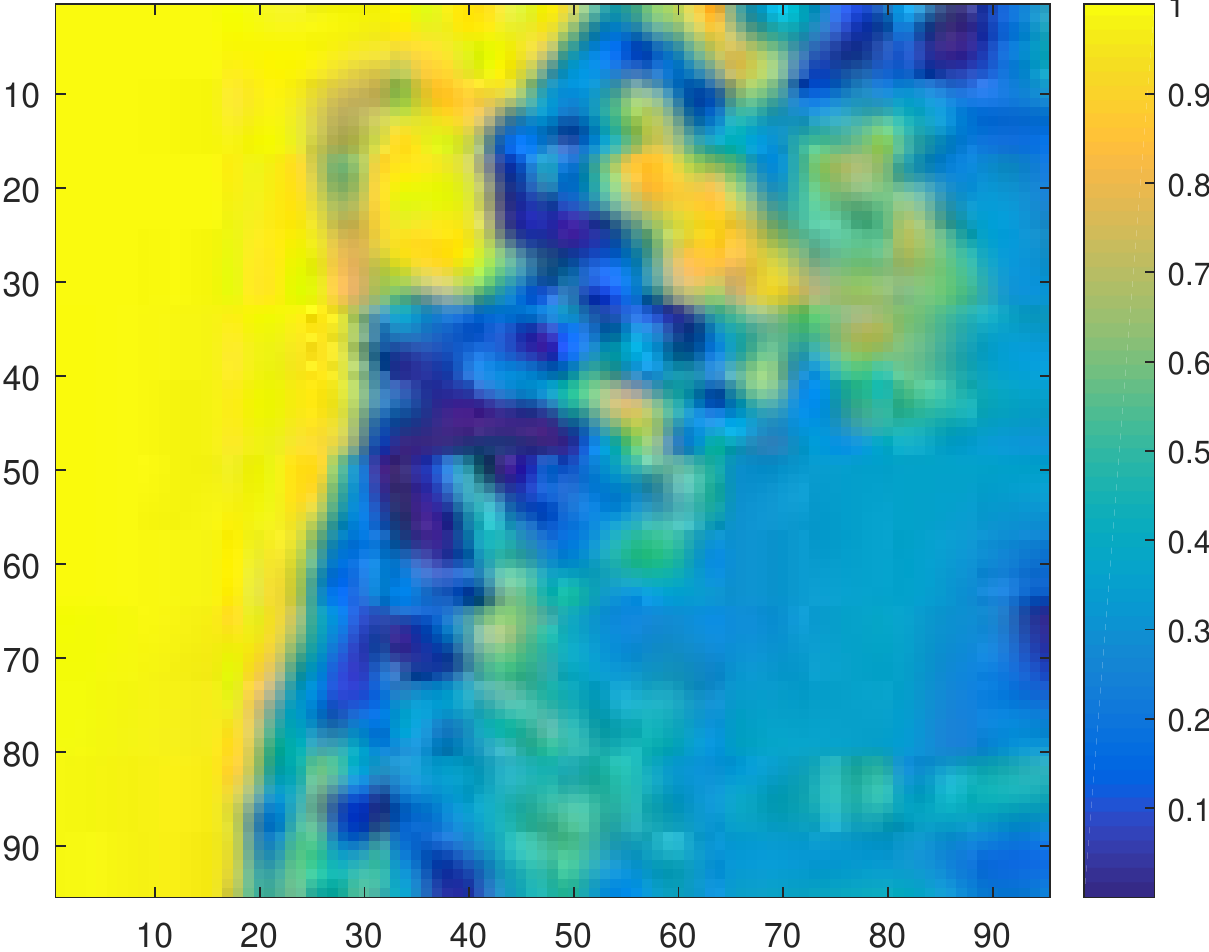}}
\subfigure[]{\includegraphics[width=2.45cm]{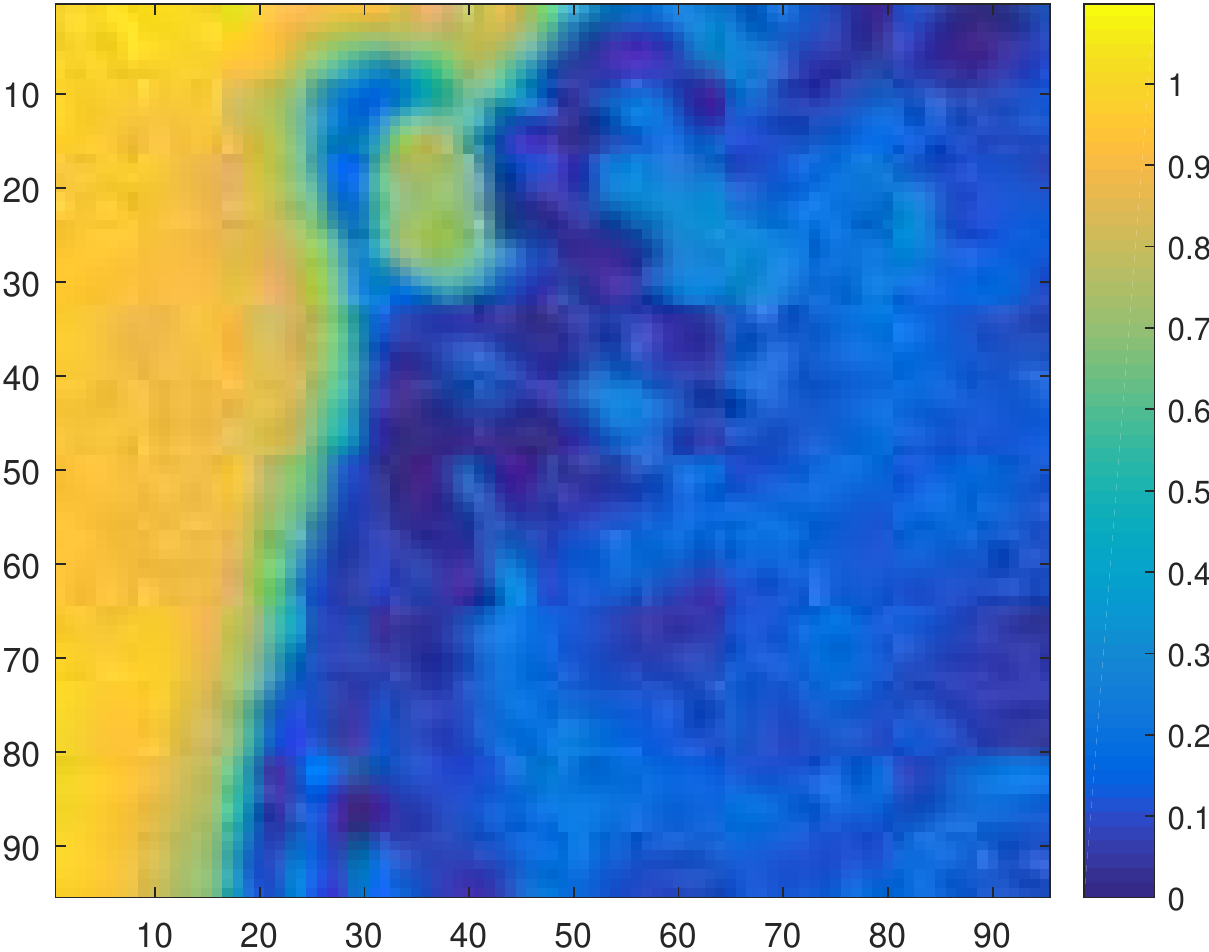}}}
\caption{Fractional abundance maps estimated by different methods of three endmembers on the Samson data set. From top to bottom: Soil. Tree. Water. From left to right: (a) $L_{1/2}$-NMF. (b) SGSNMF. (c) TV-RSNMF. (d) $L_{1/2}$-RNMF. (e) MV-NTF-TV. (f) MLNMF. (g) SSRDMF.}
\label{fig:8}
\end{figure*}

\begin{figure*}[!t]
\centering
\mbox{
{\includegraphics[width=2.45cm]{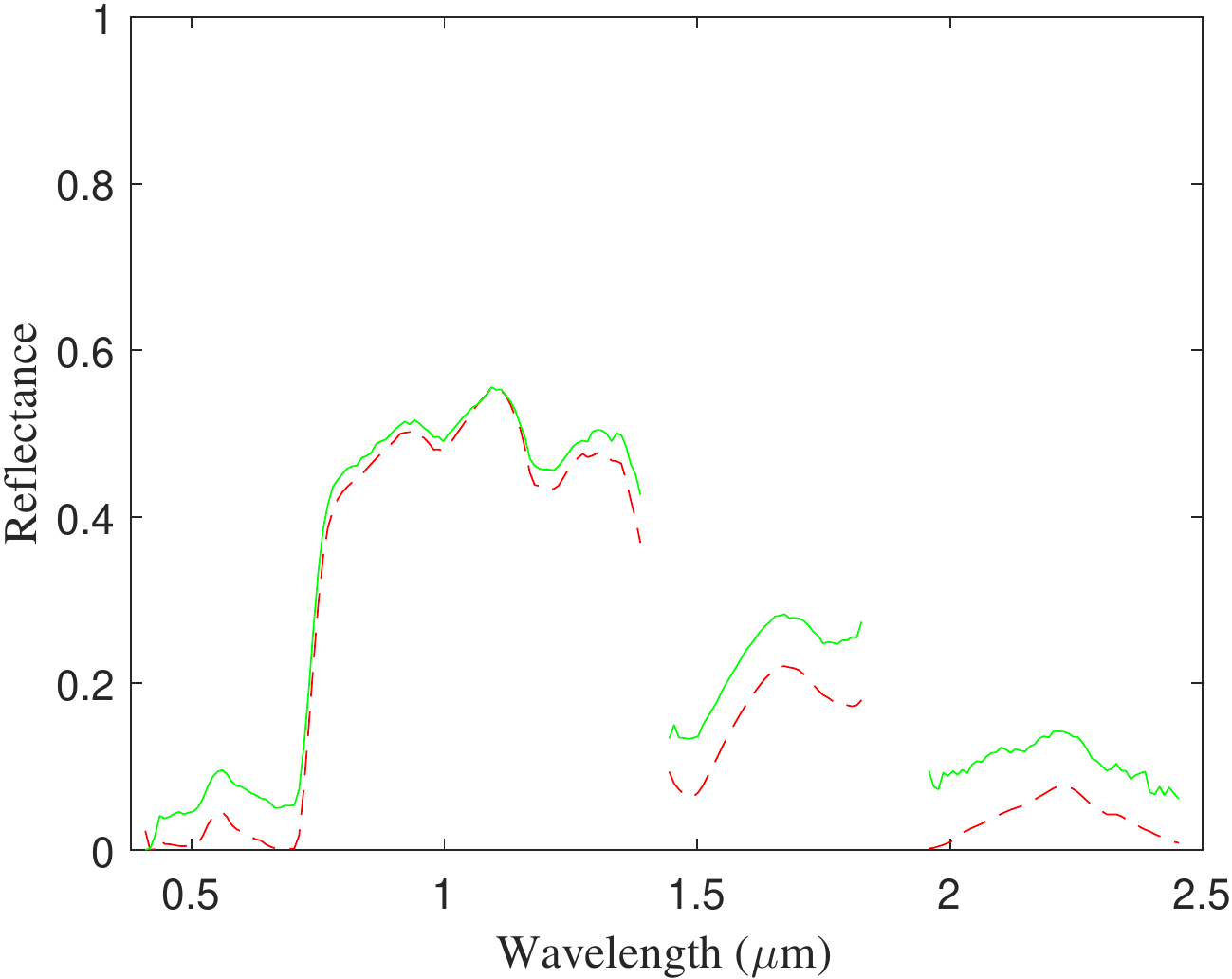}}
{\includegraphics[width=2.45cm]{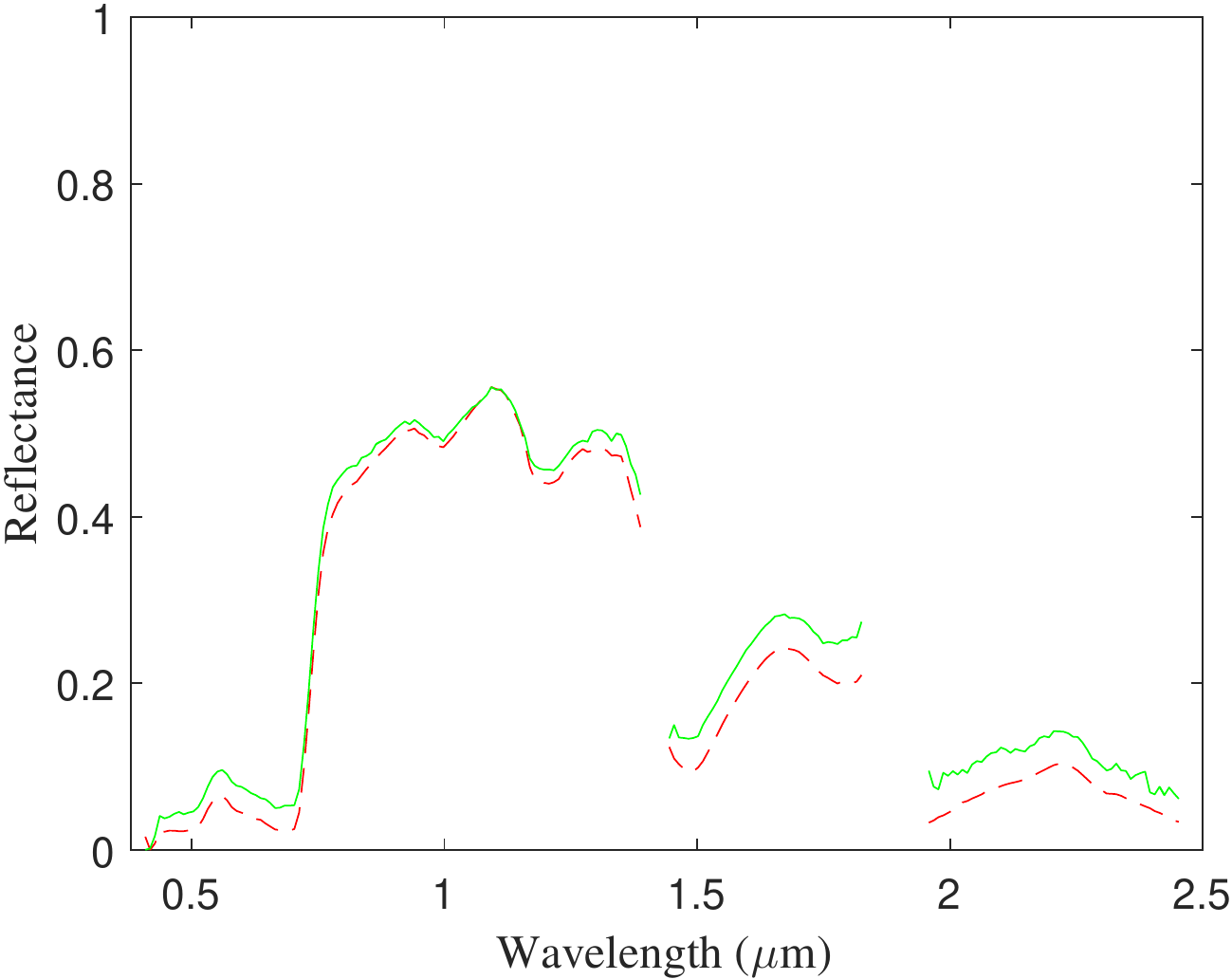}}
{\includegraphics[width=2.45cm]{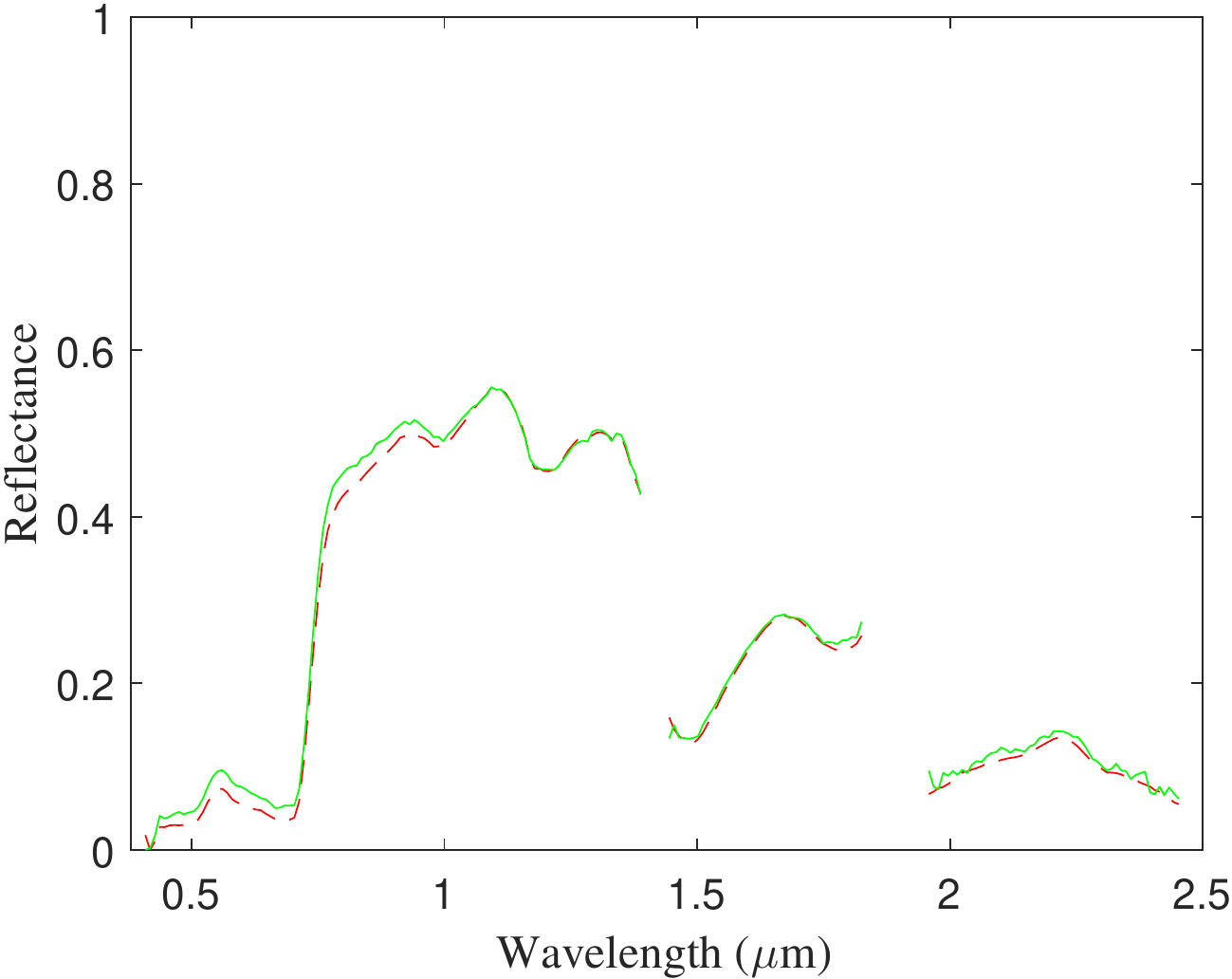}}
{\includegraphics[width=2.45cm]{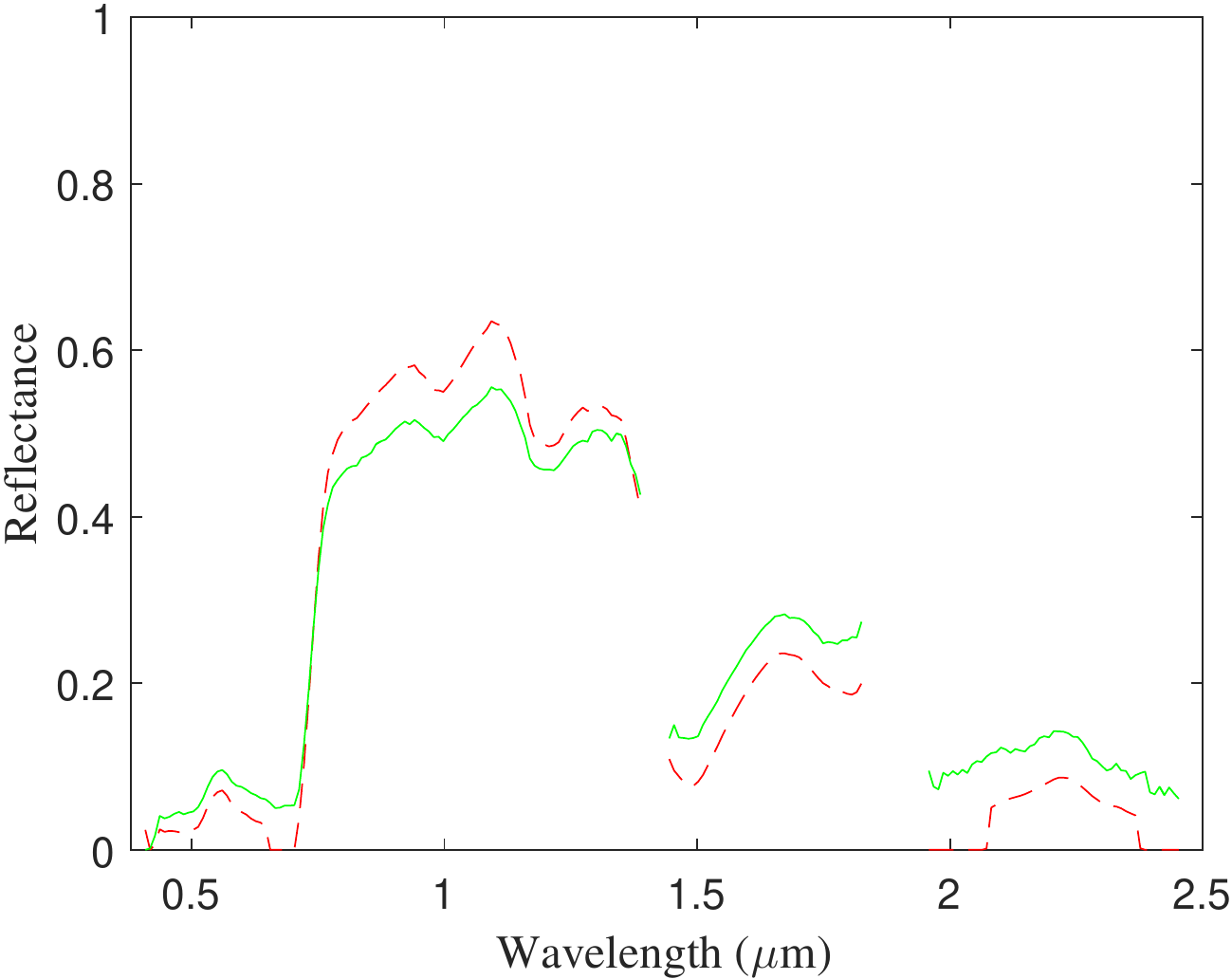}}
{\includegraphics[width=2.45cm]{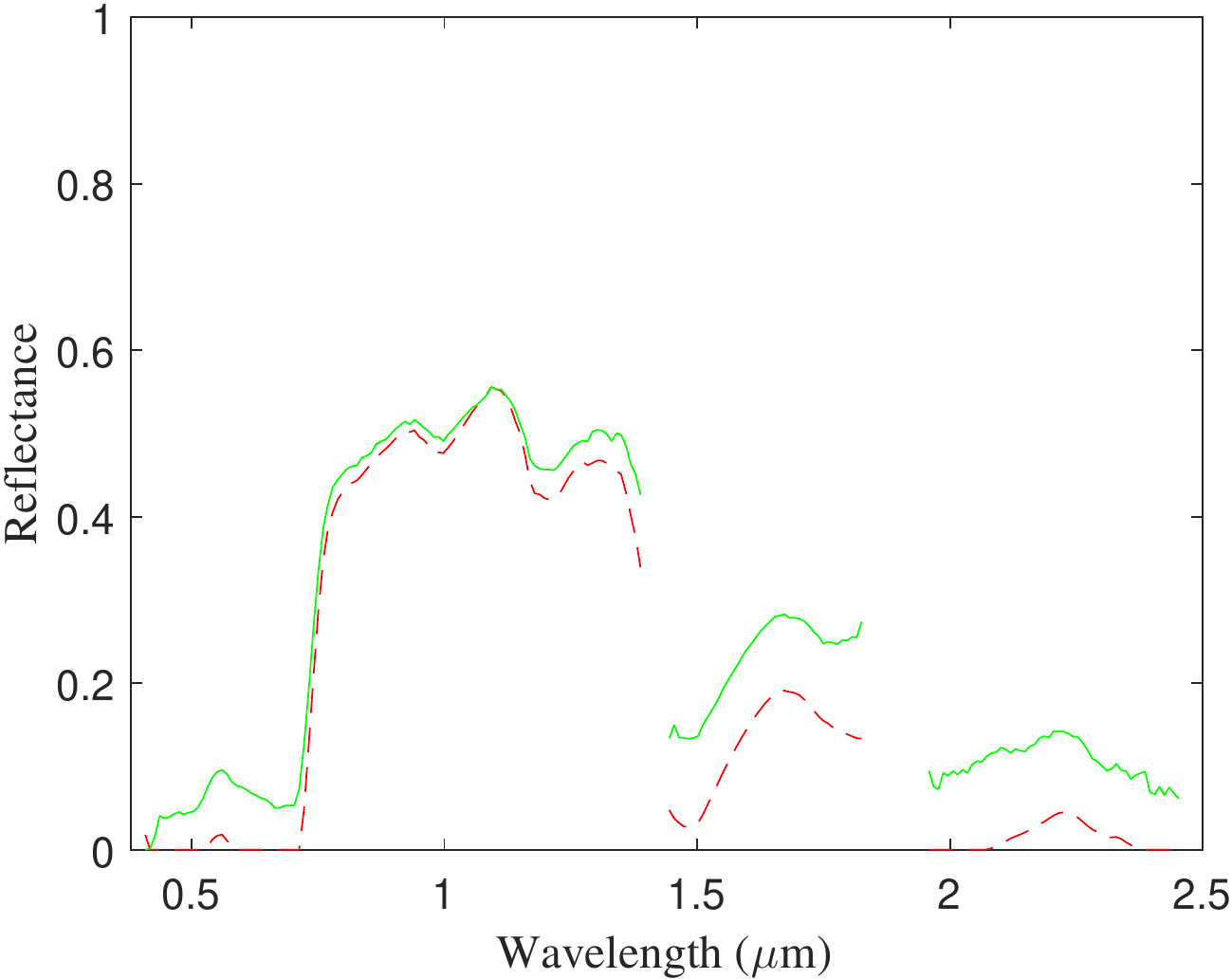}}
{\includegraphics[width=2.45cm]{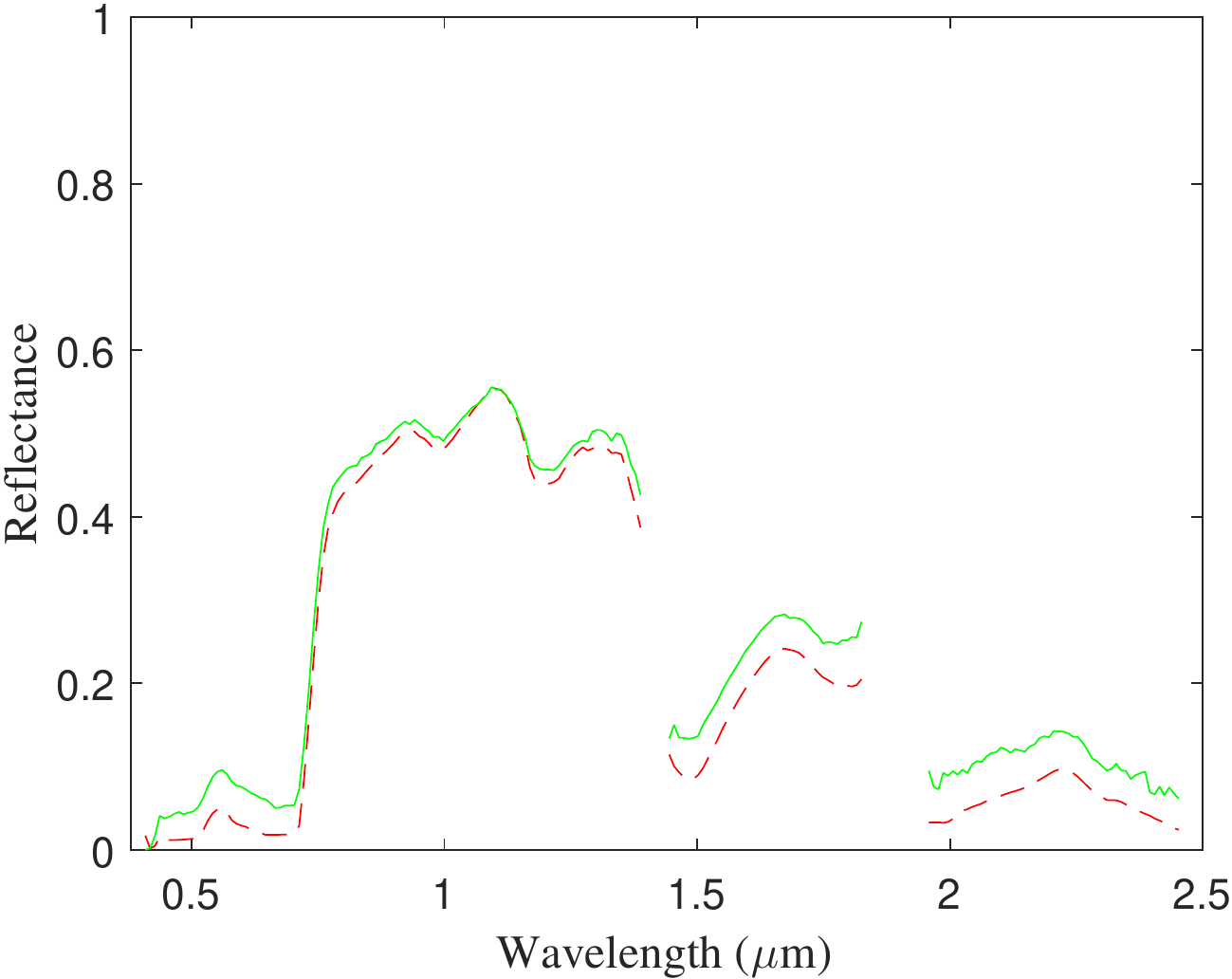}}
{\includegraphics[width=2.45cm]{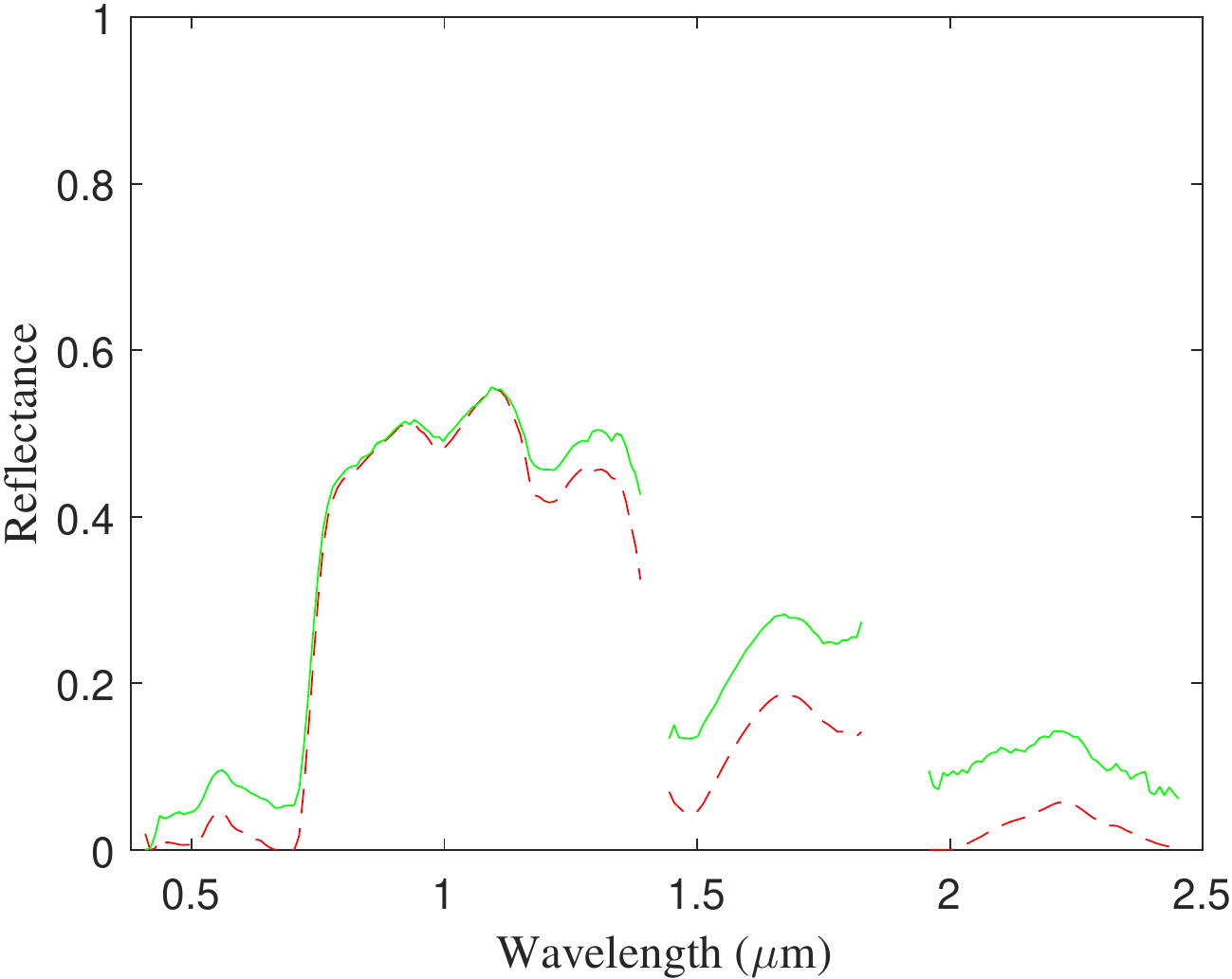}}}
\\
\mbox{
{\includegraphics[width=2.45cm]{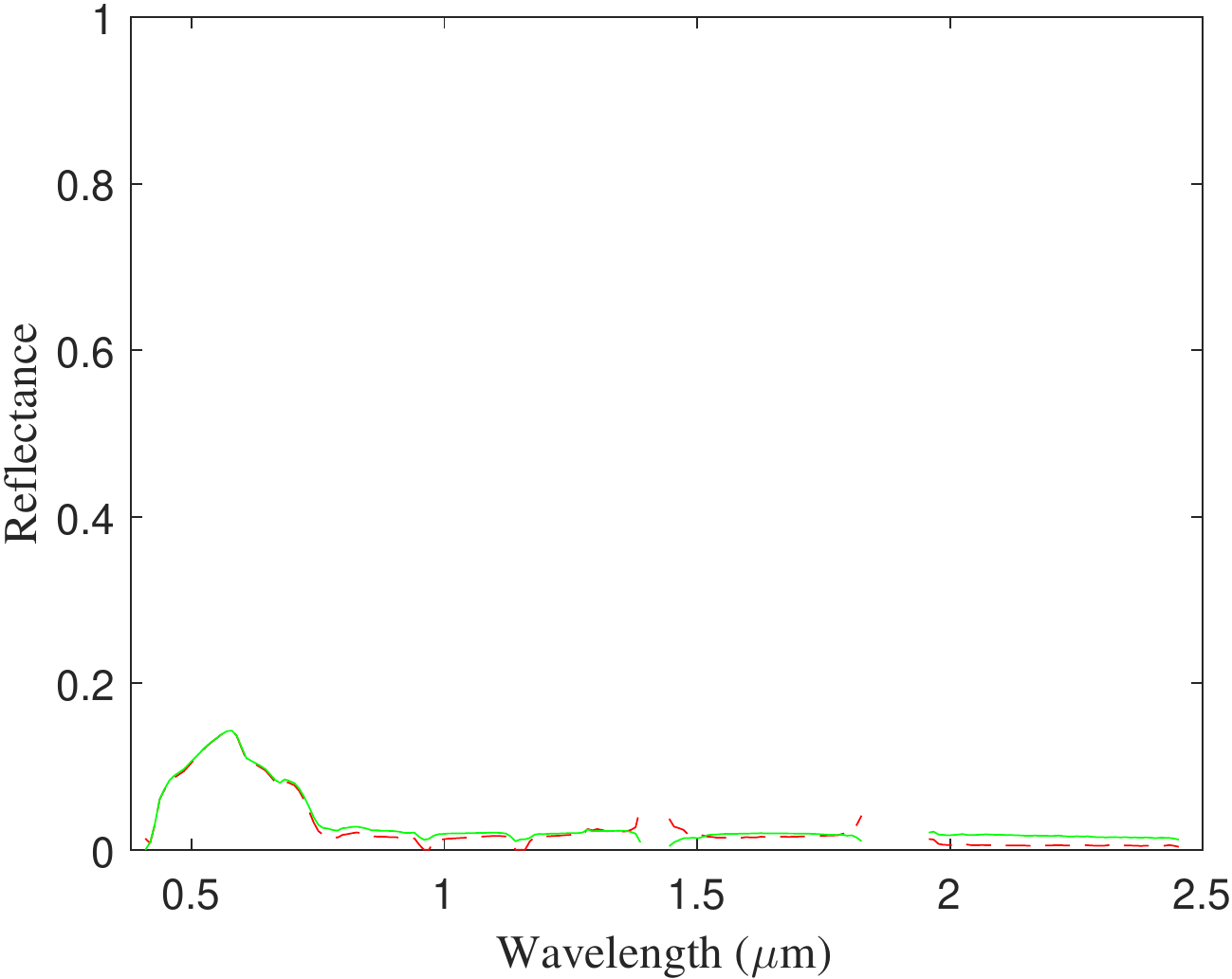}}
{\includegraphics[width=2.45cm]{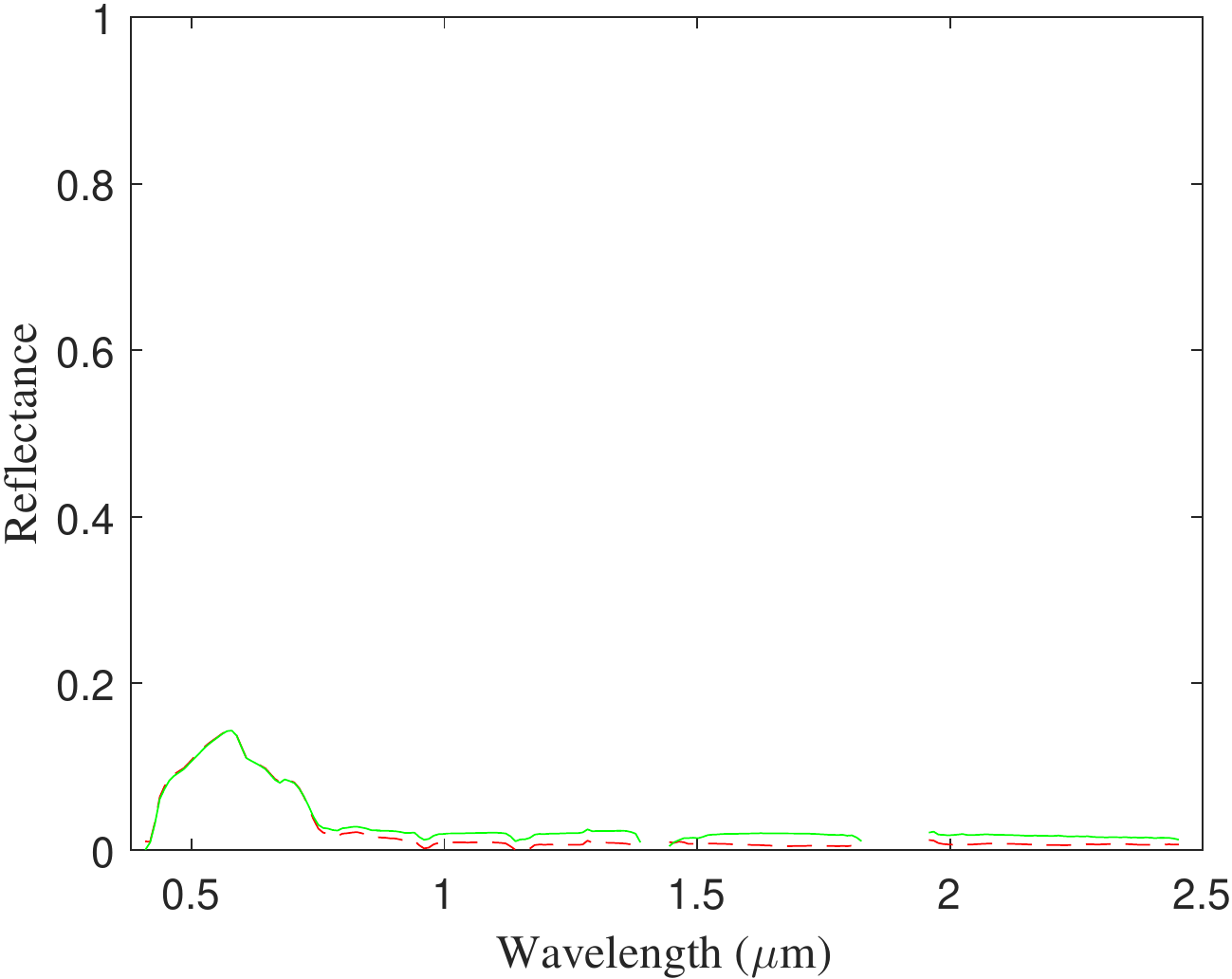}}
{\includegraphics[width=2.45cm]{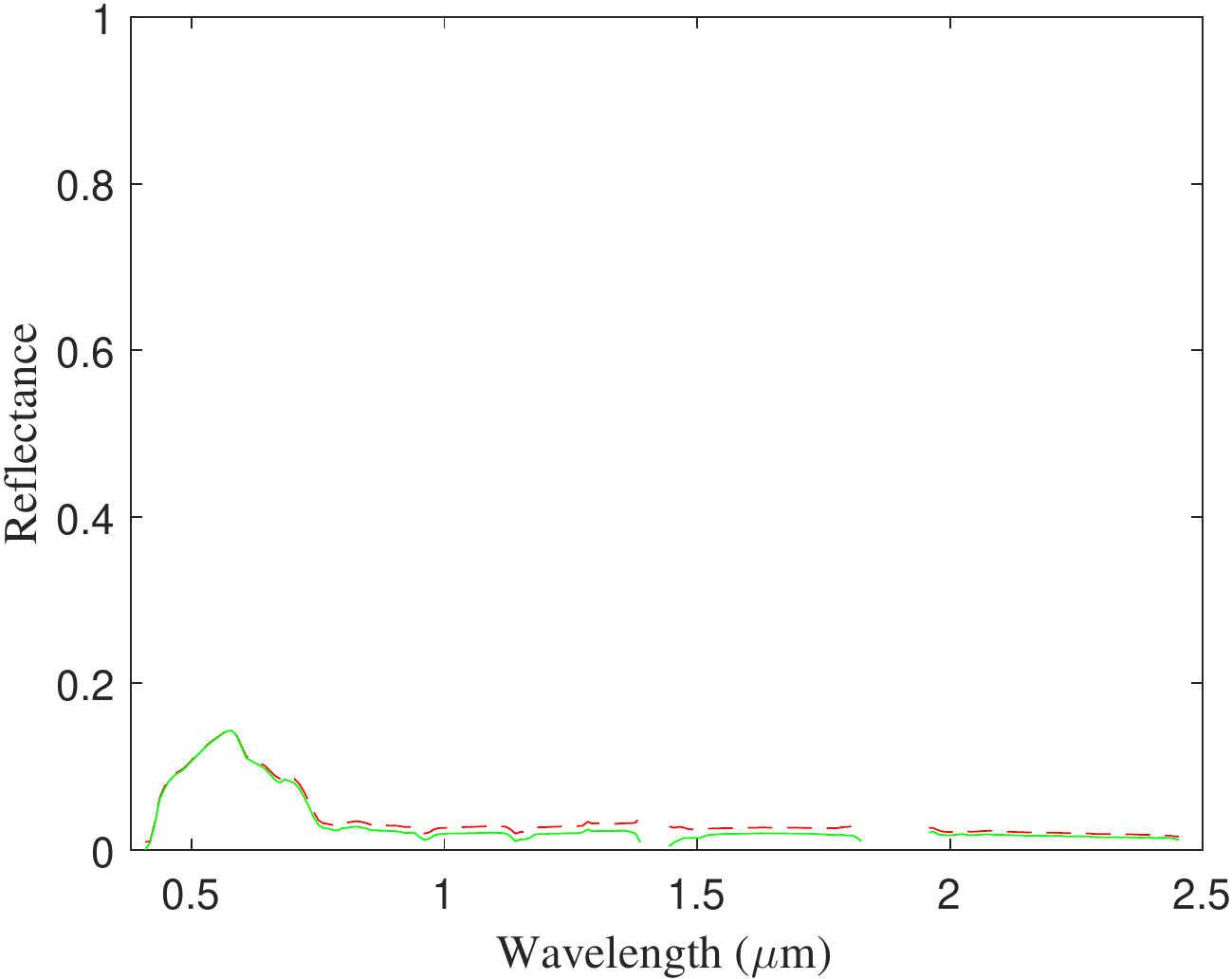}}
{\includegraphics[width=2.45cm]{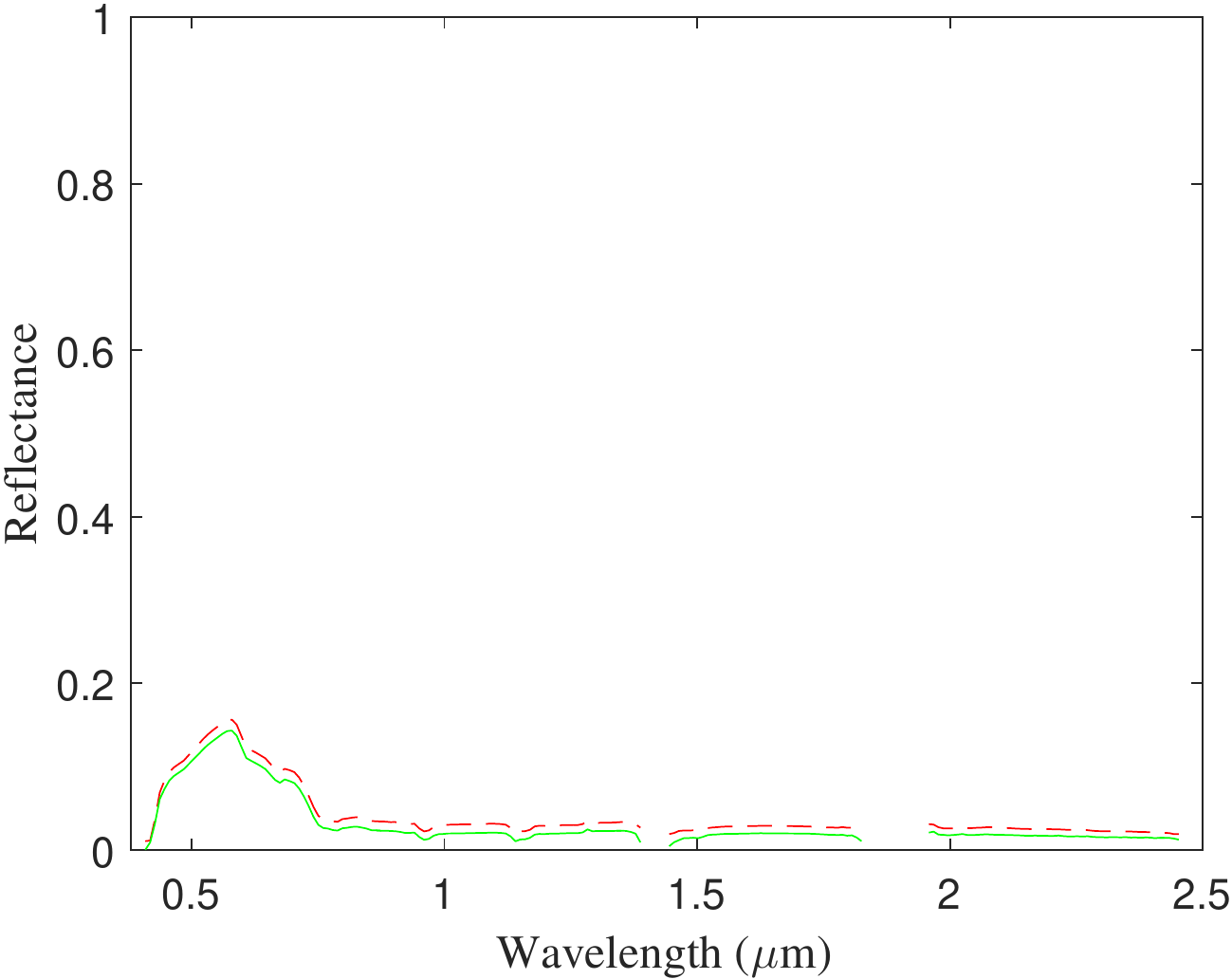}}
{\includegraphics[width=2.45cm]{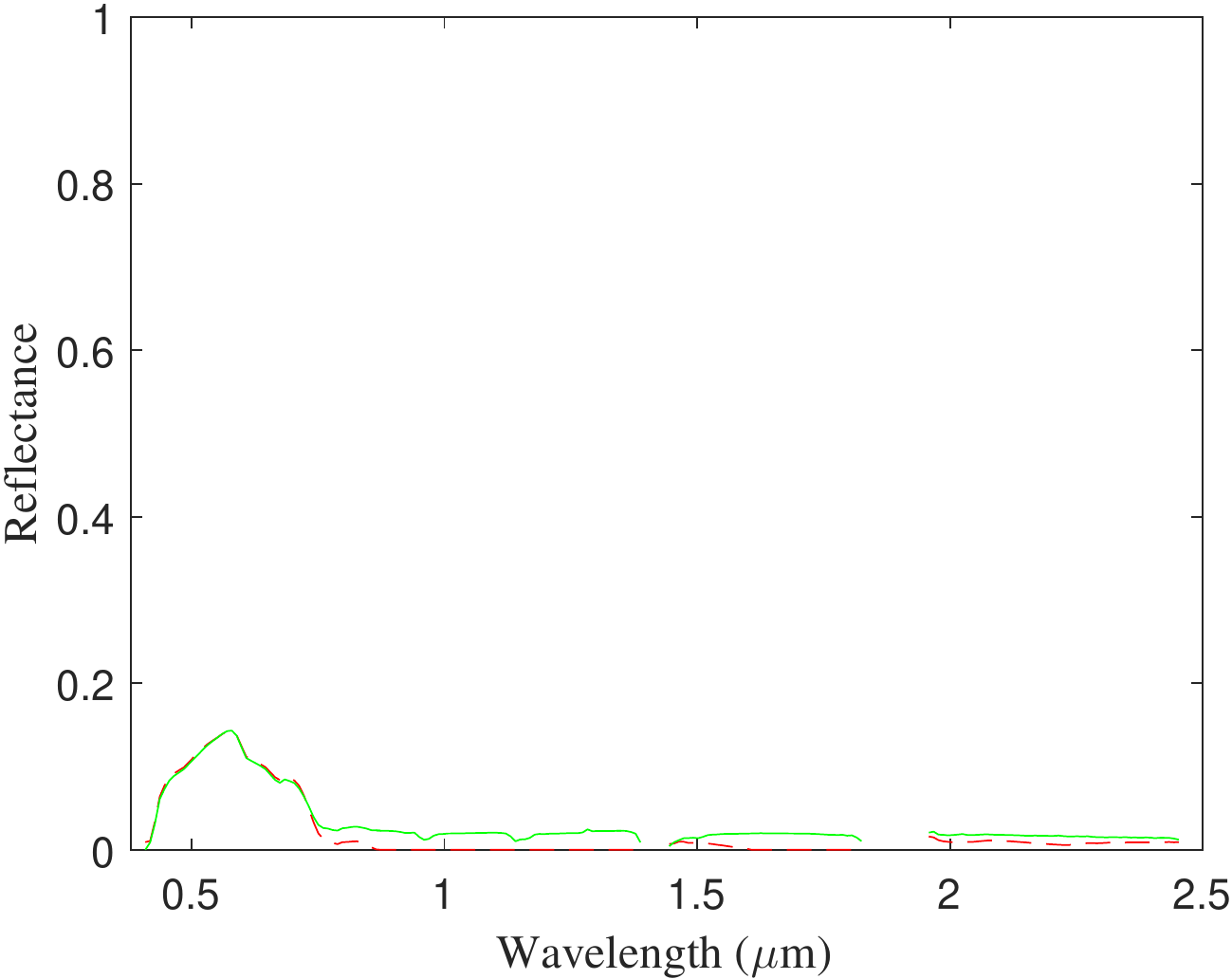}}
{\includegraphics[width=2.45cm]{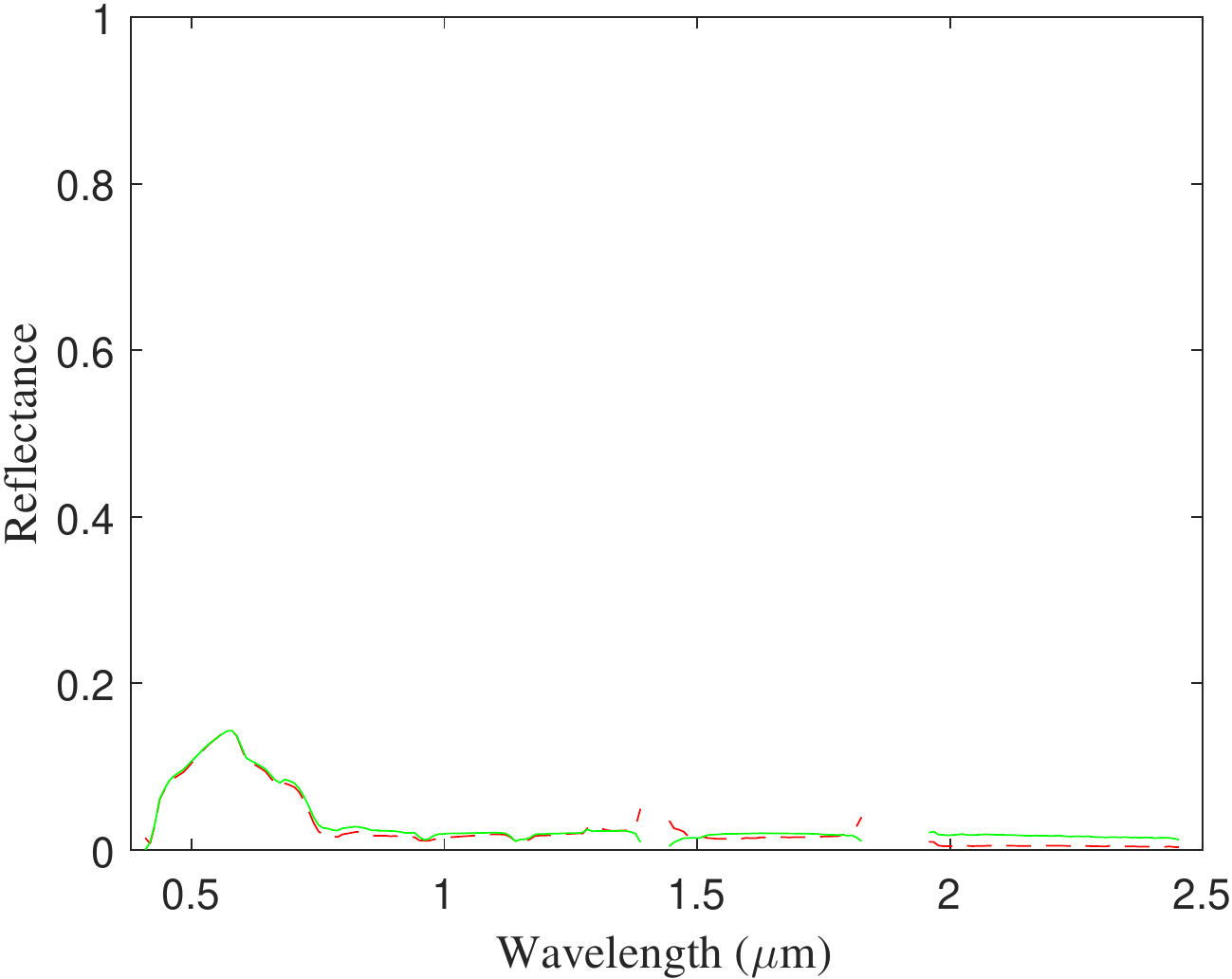}}
{\includegraphics[width=2.45cm]{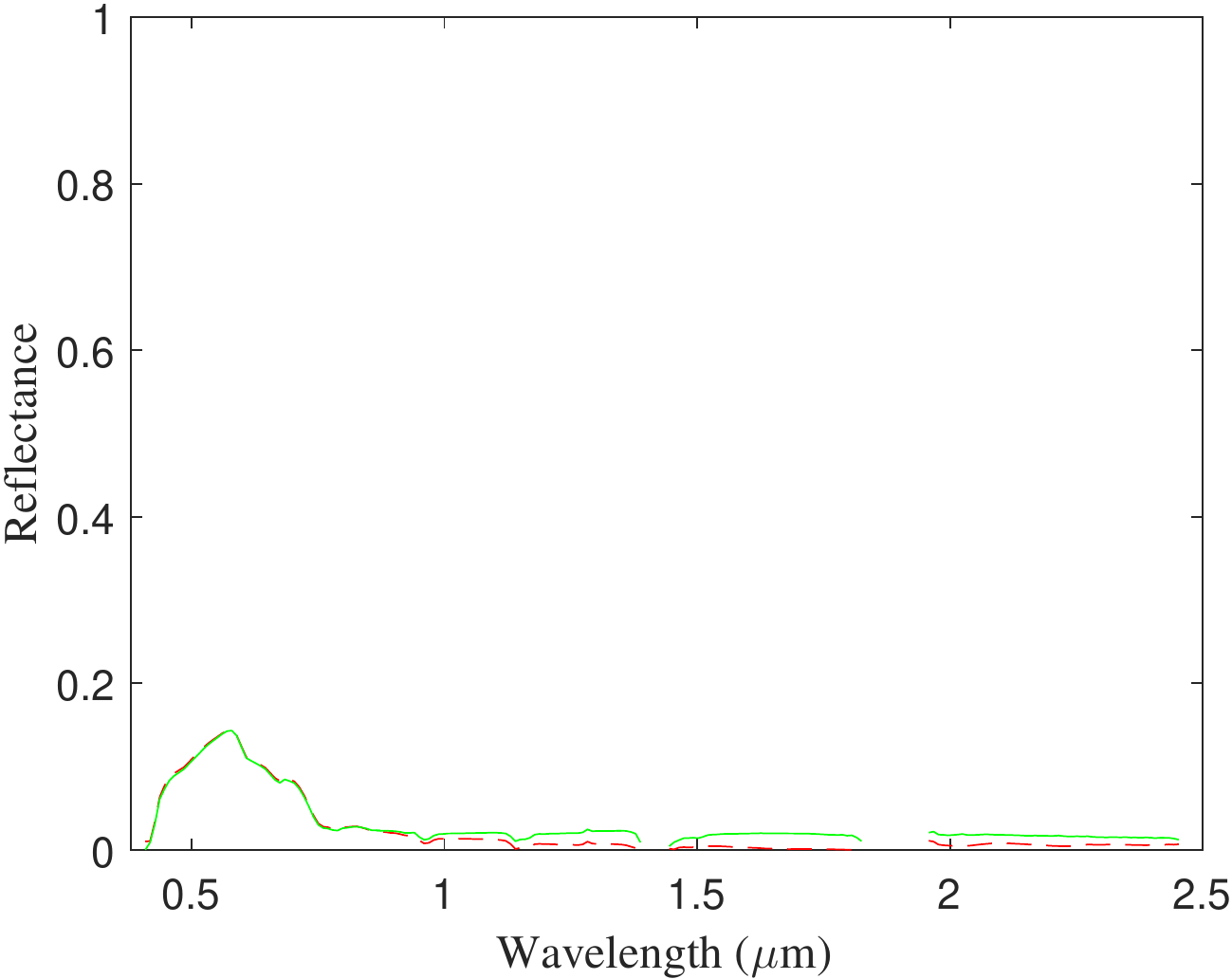}}}
\\
\mbox{
{\includegraphics[width=2.45cm]{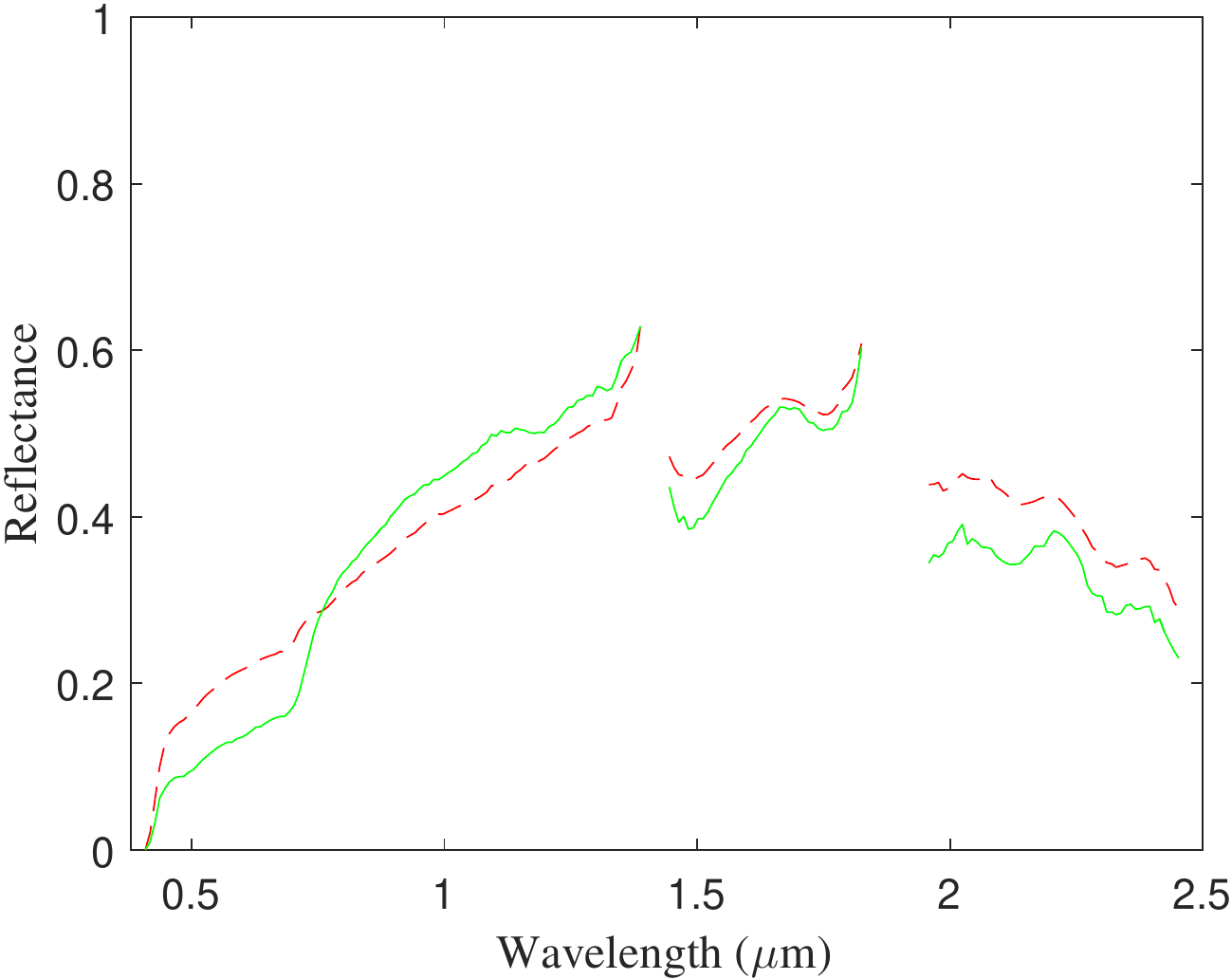}}
{\includegraphics[width=2.45cm]{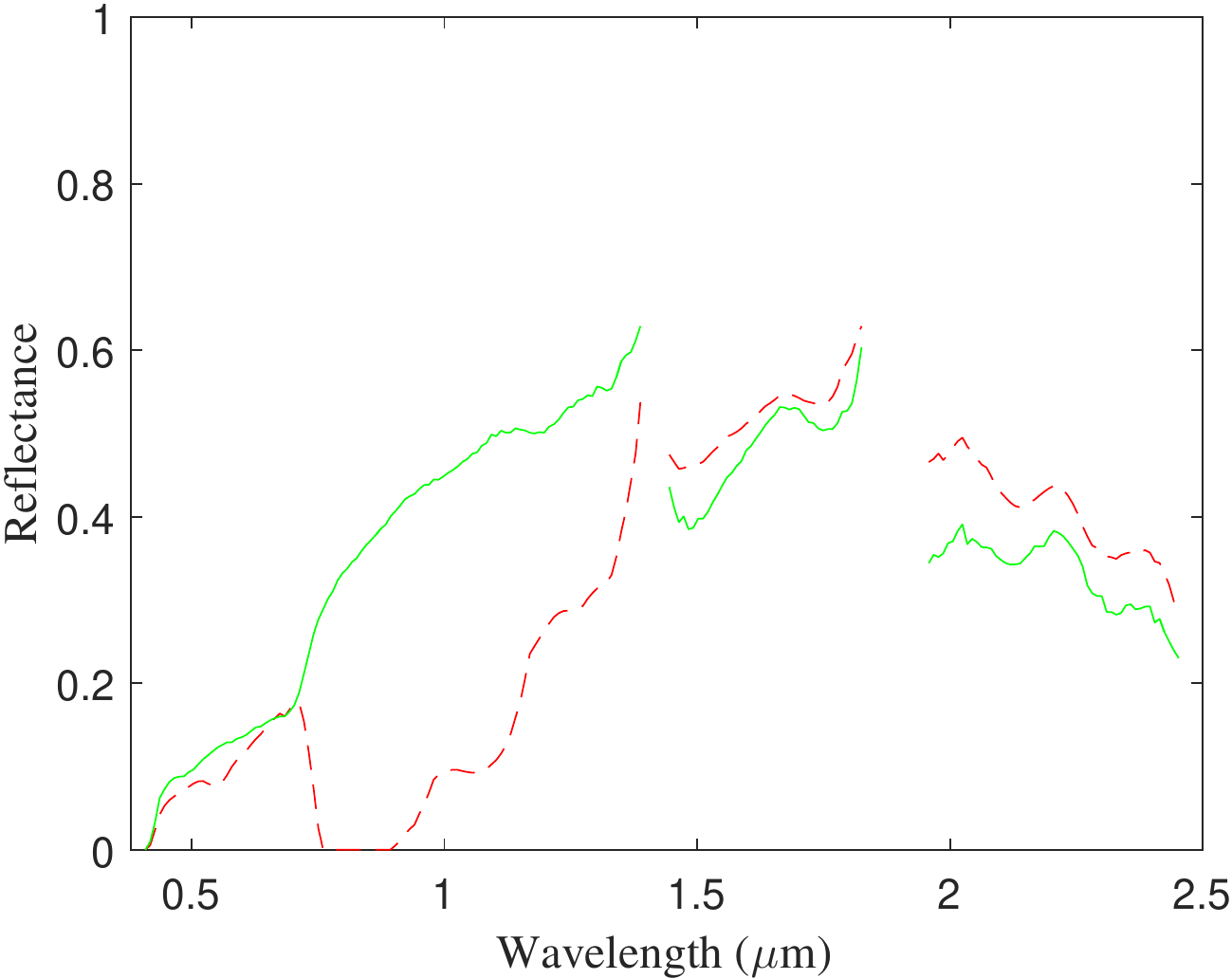}}
{\includegraphics[width=2.45cm]{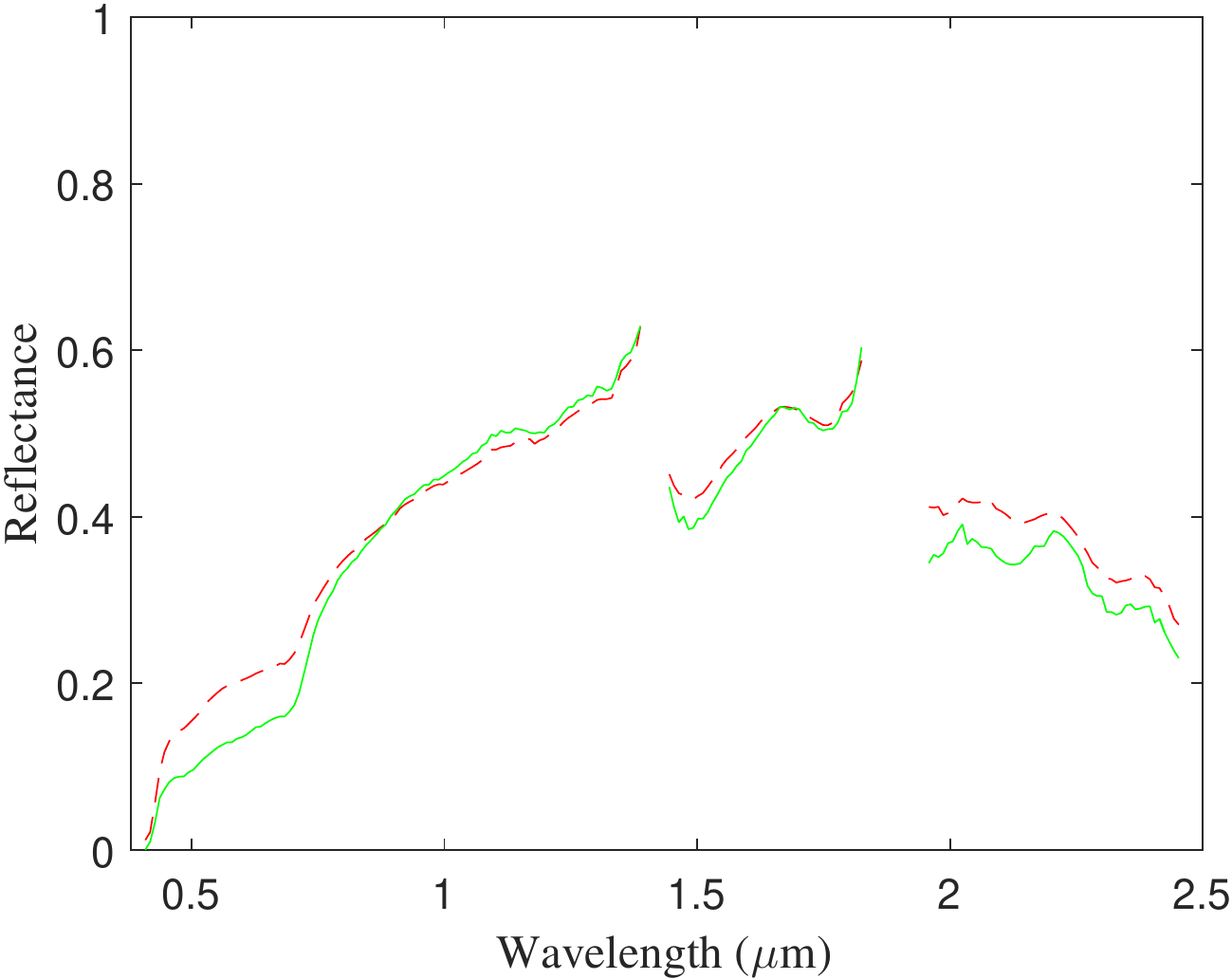}}
{\includegraphics[width=2.45cm]{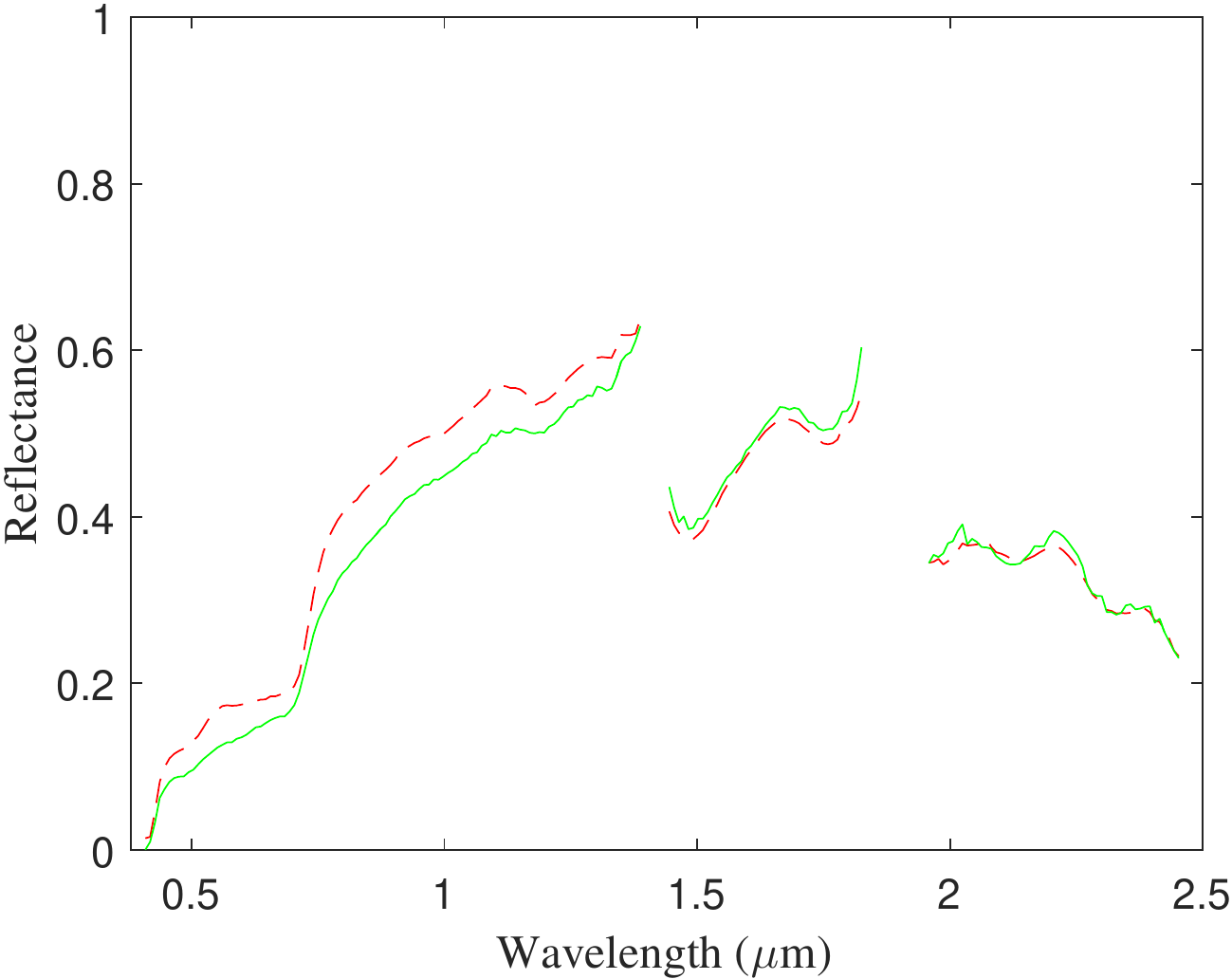}}
{\includegraphics[width=2.45cm]{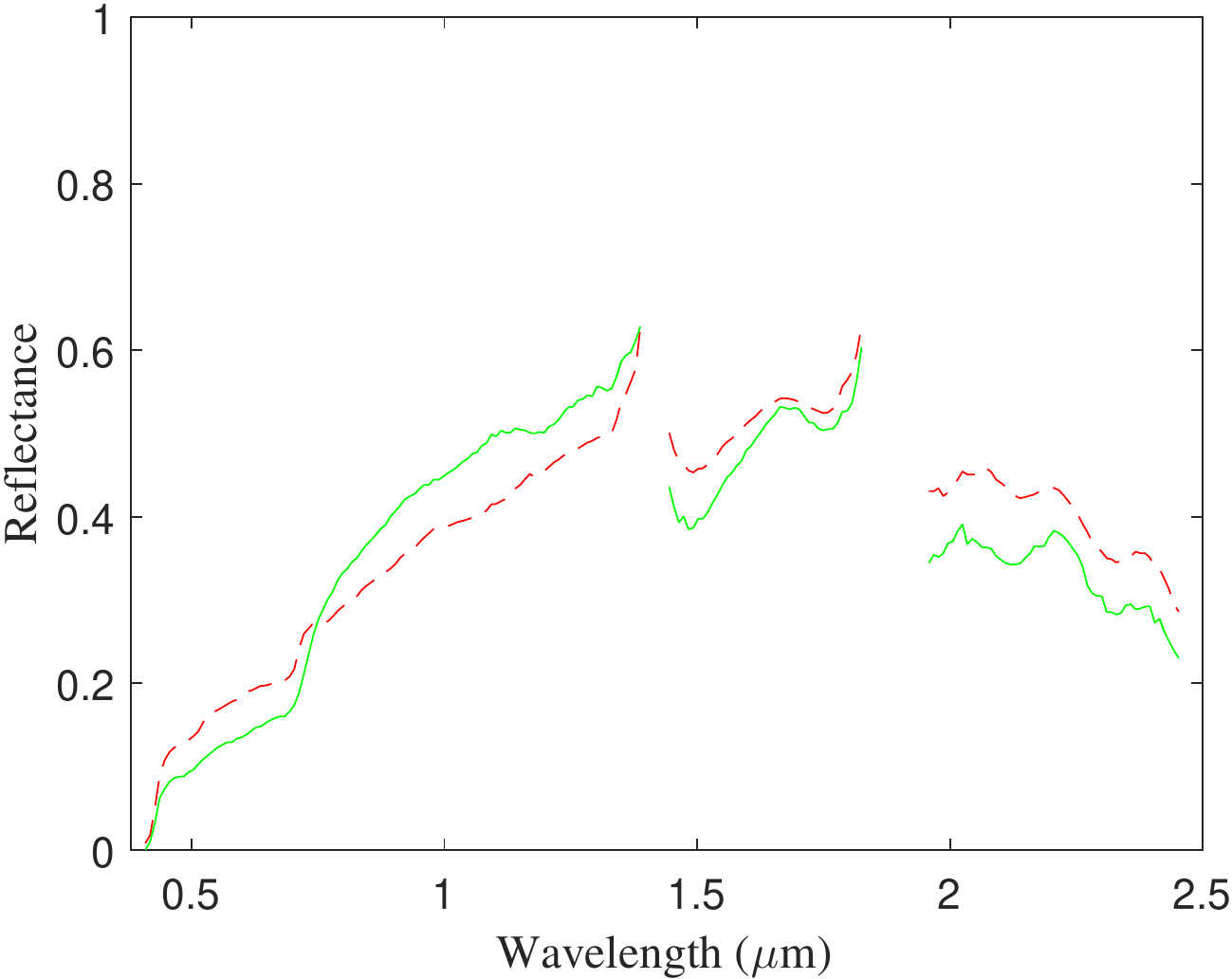}}
{\includegraphics[width=2.45cm]{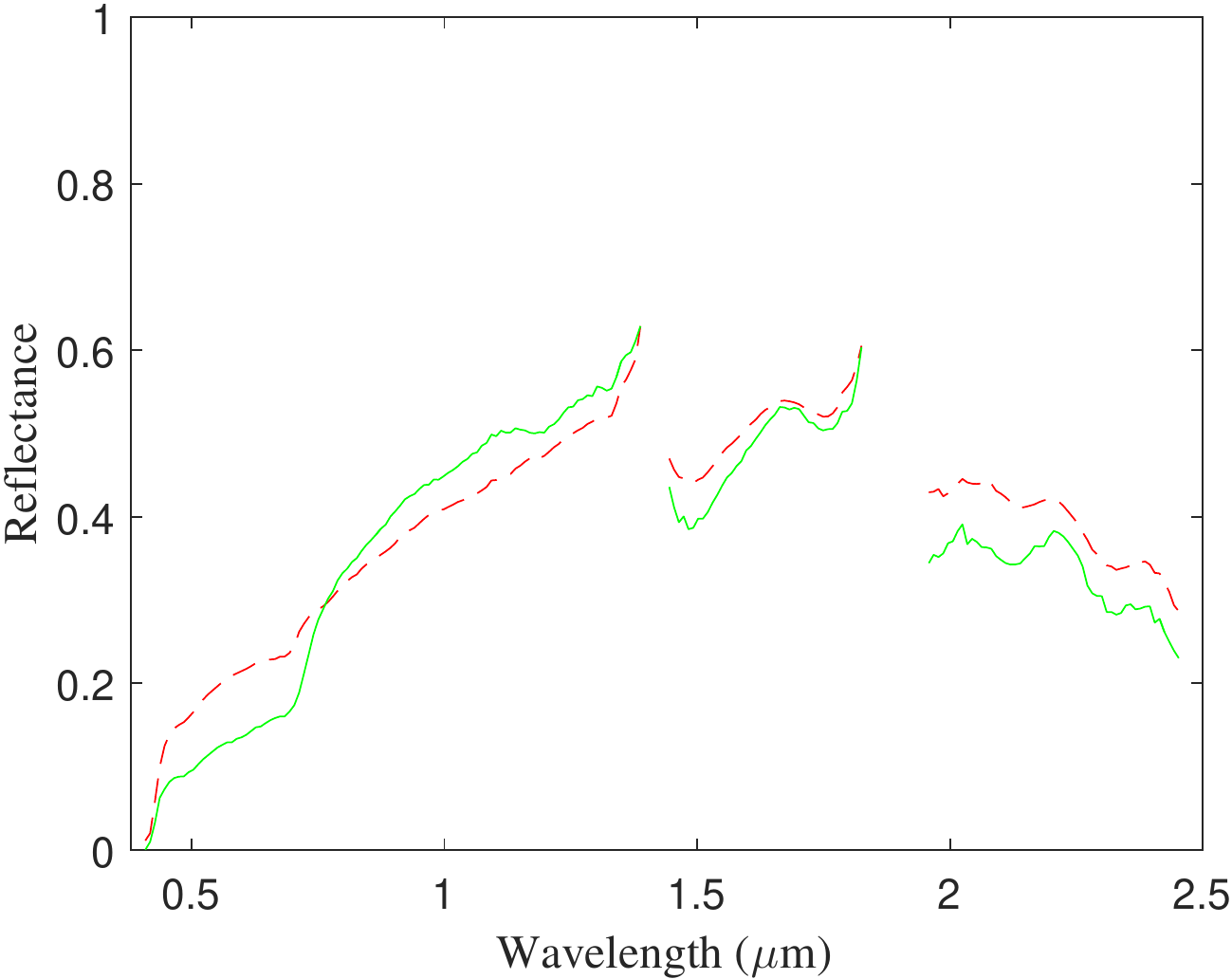}}
{\includegraphics[width=2.45cm]{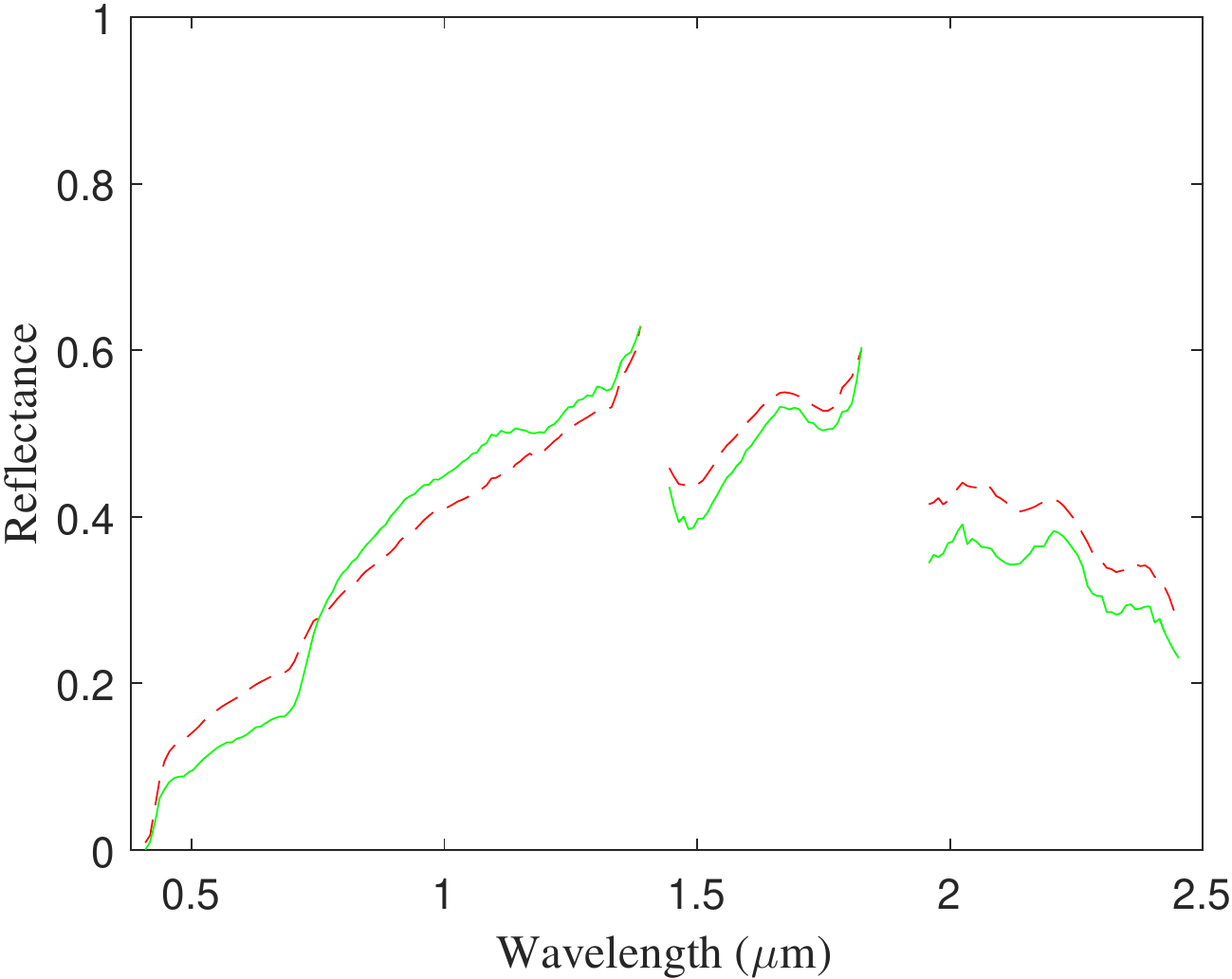}}}
\\
\mbox{
\subfigure[]{\includegraphics[width=2.45cm]{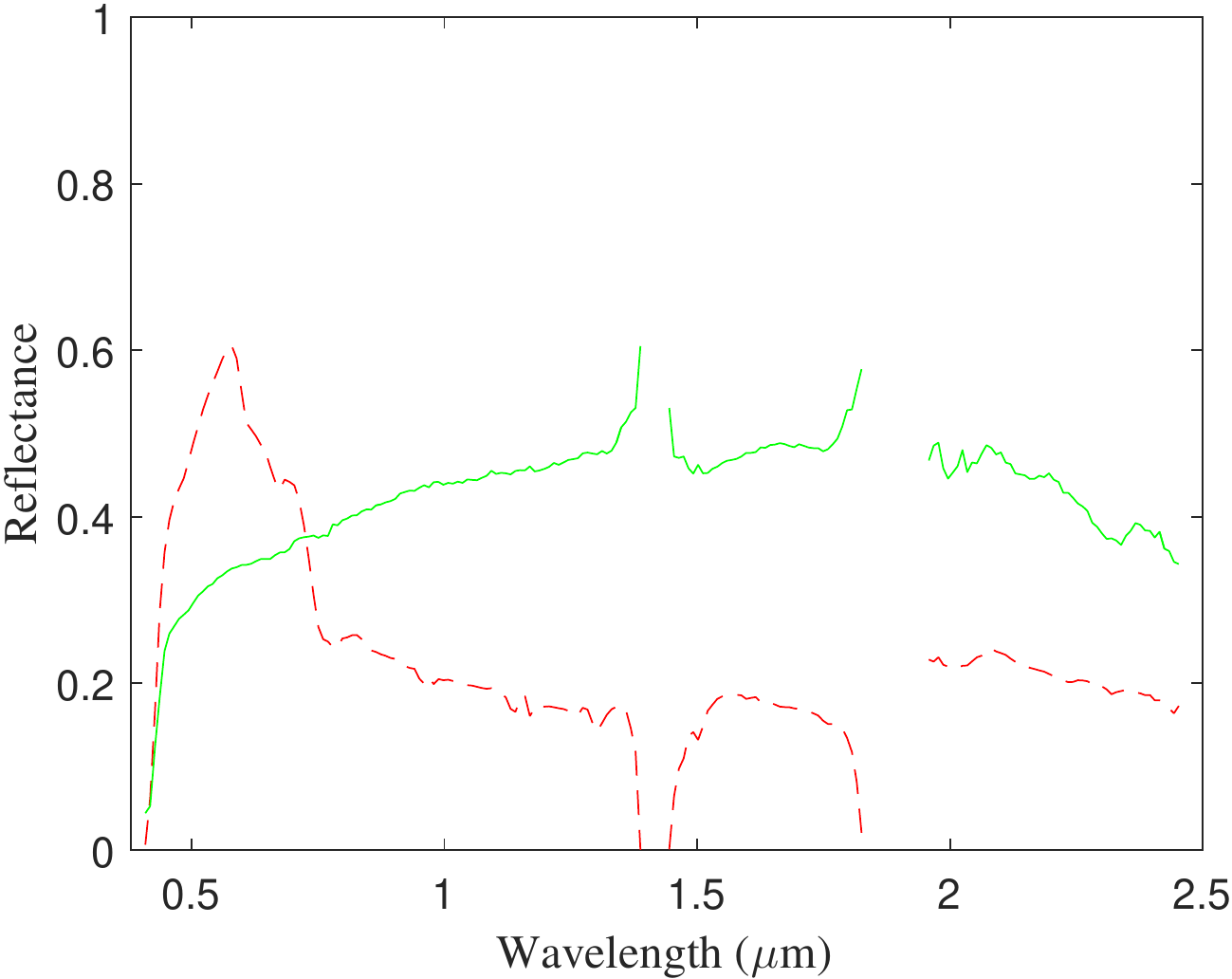}}
\subfigure[]{\includegraphics[width=2.45cm]{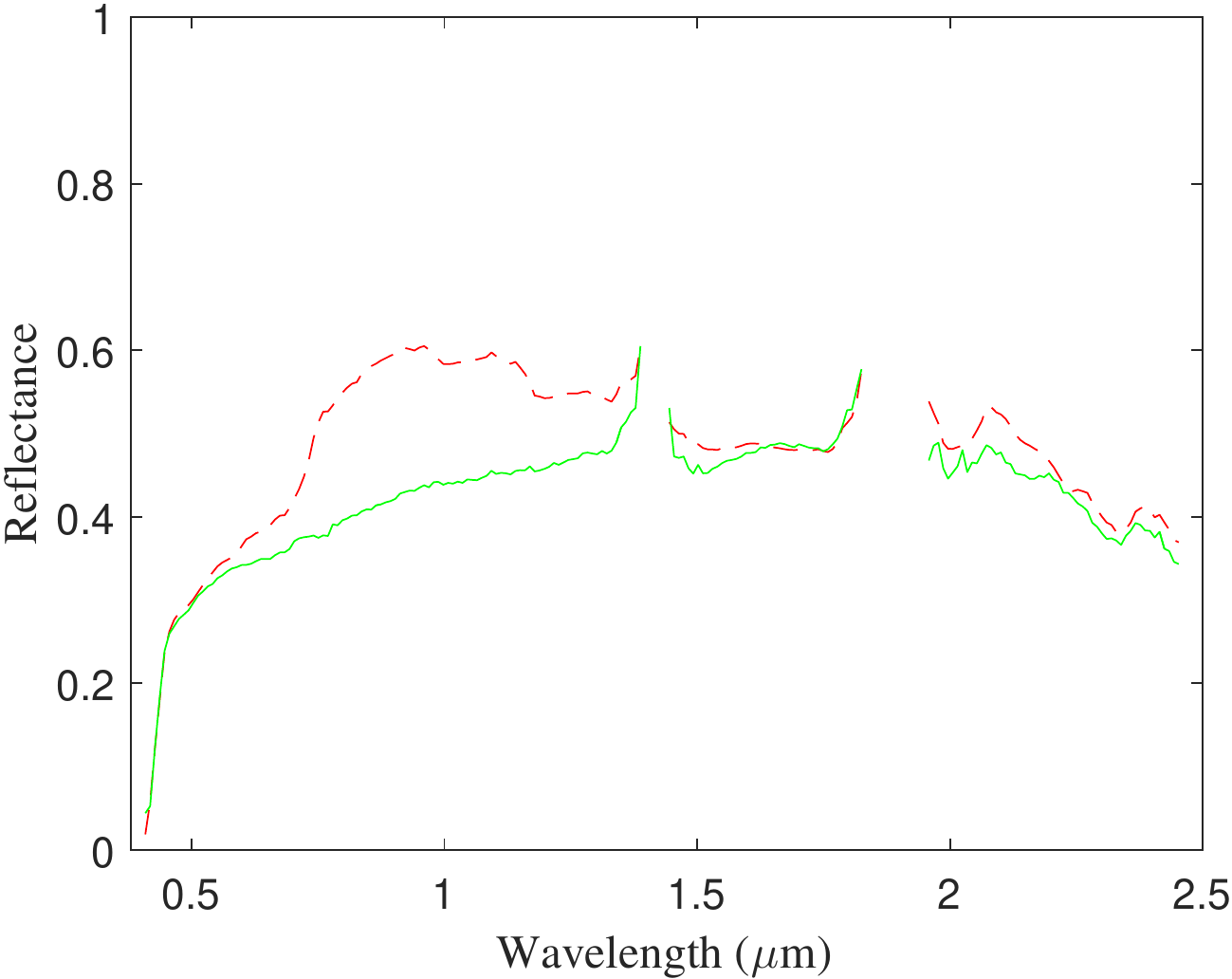}}
\subfigure[]{\includegraphics[width=2.45cm]{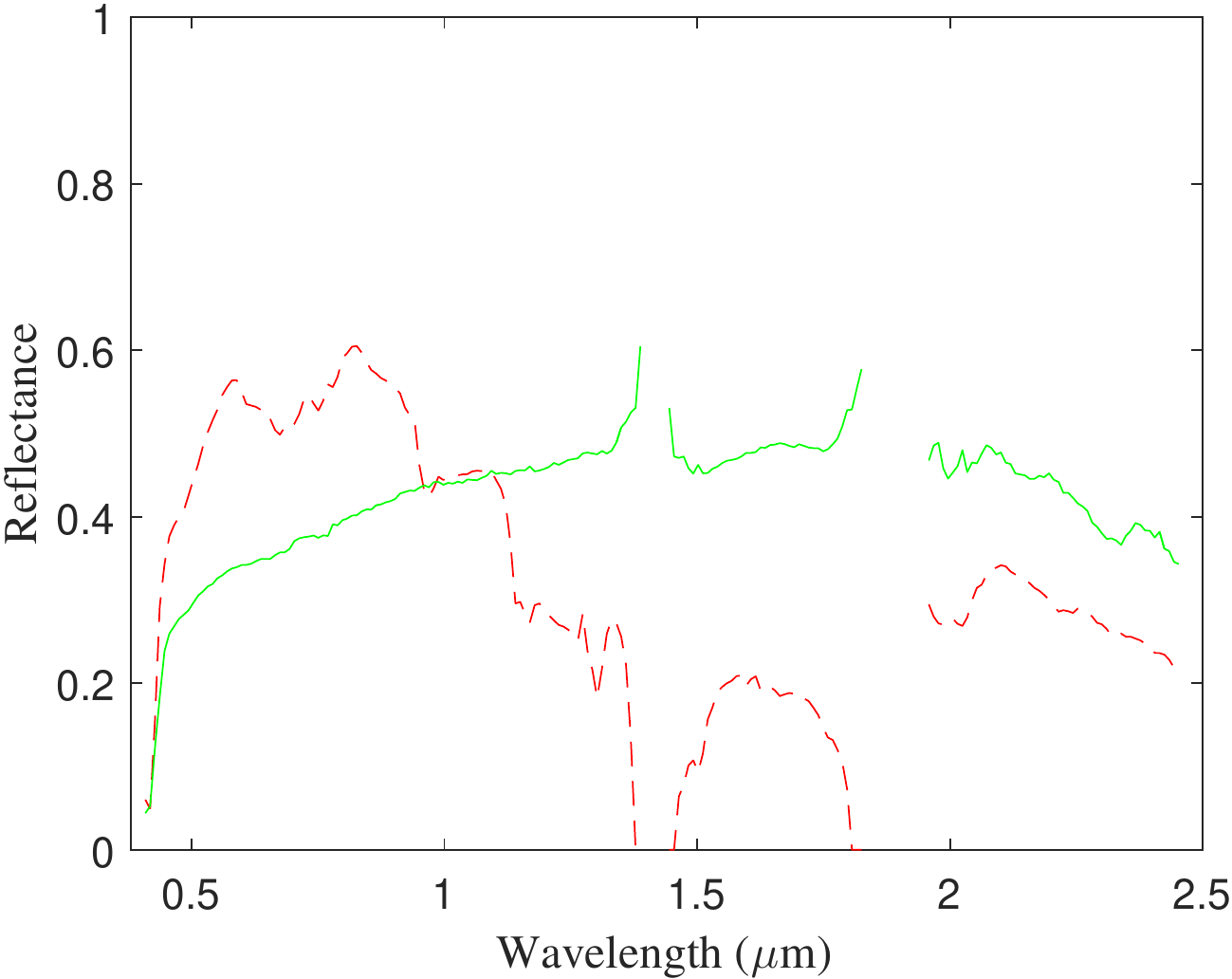}}
\subfigure[]{\includegraphics[width=2.45cm]{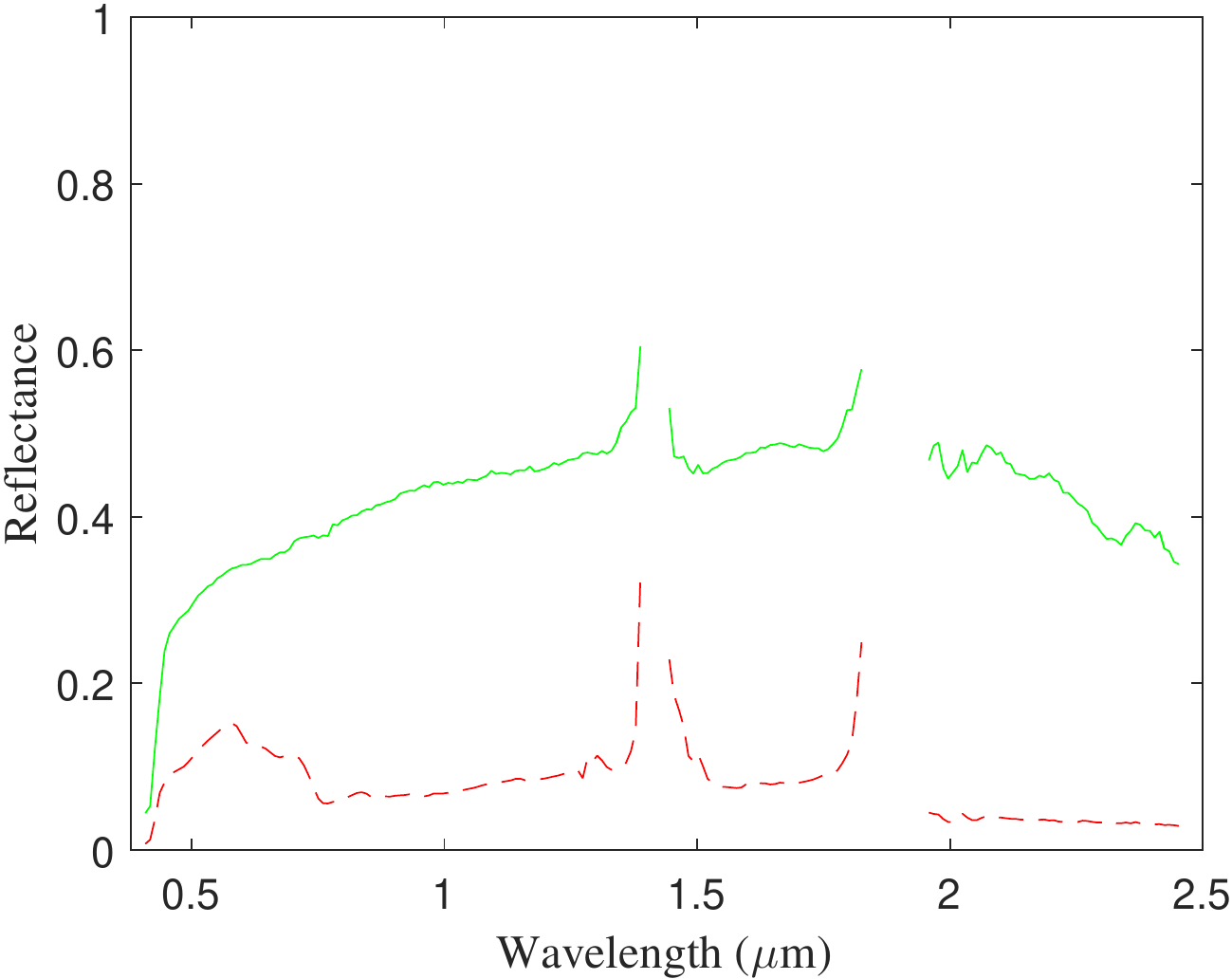}}
\subfigure[]{\includegraphics[width=2.45cm]{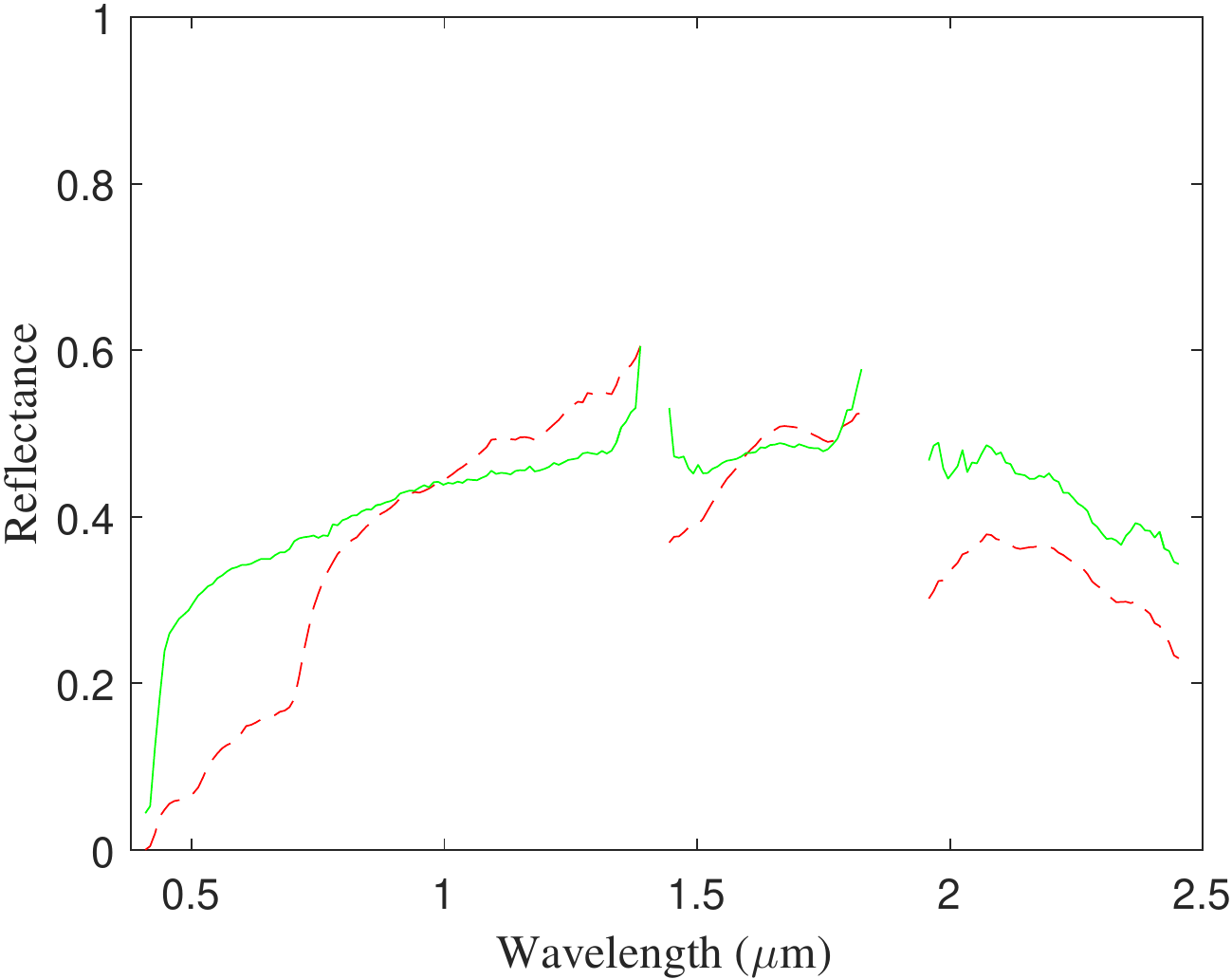}}
\subfigure[]{\includegraphics[width=2.45cm]{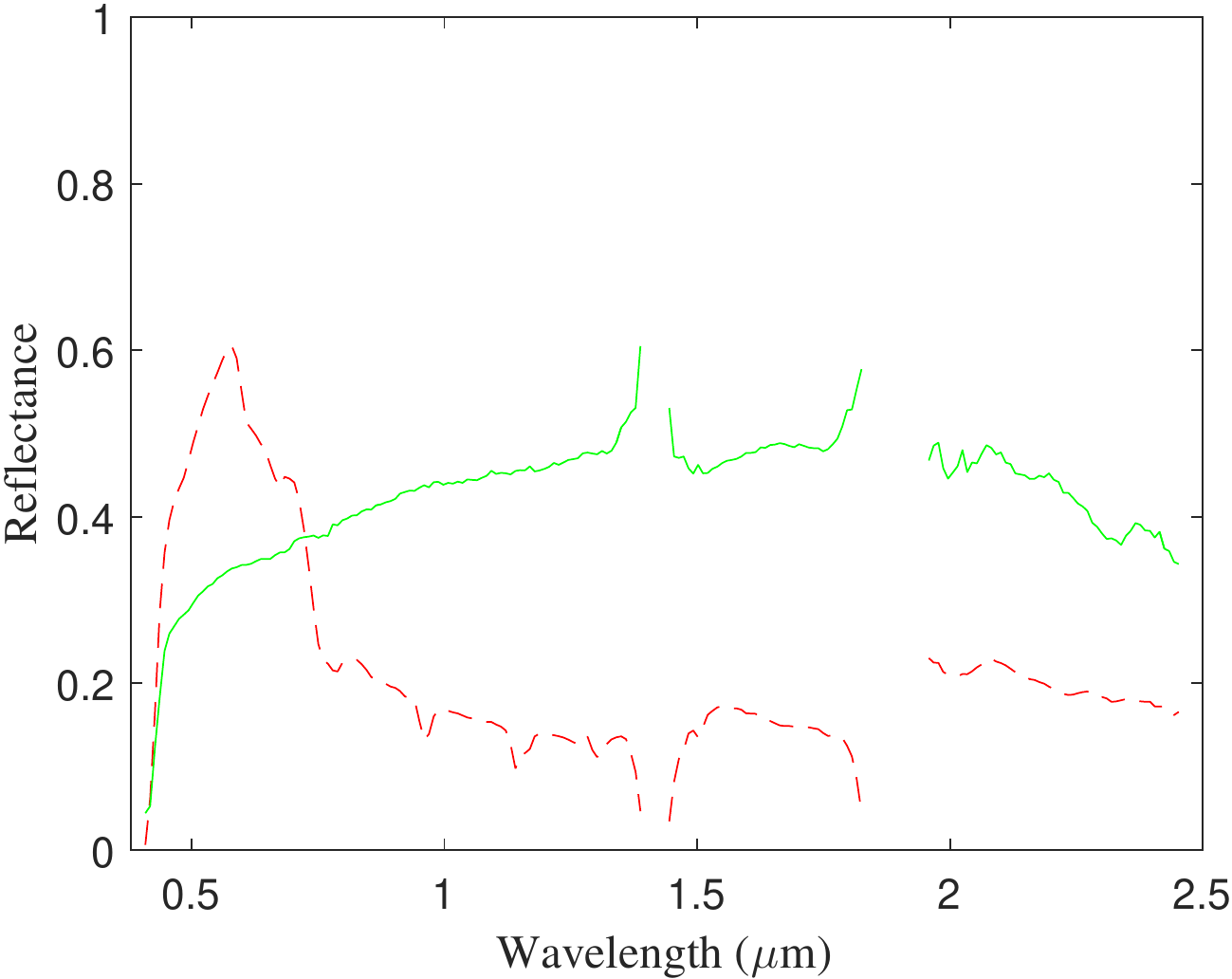}}
\subfigure[]{\includegraphics[width=2.45cm]{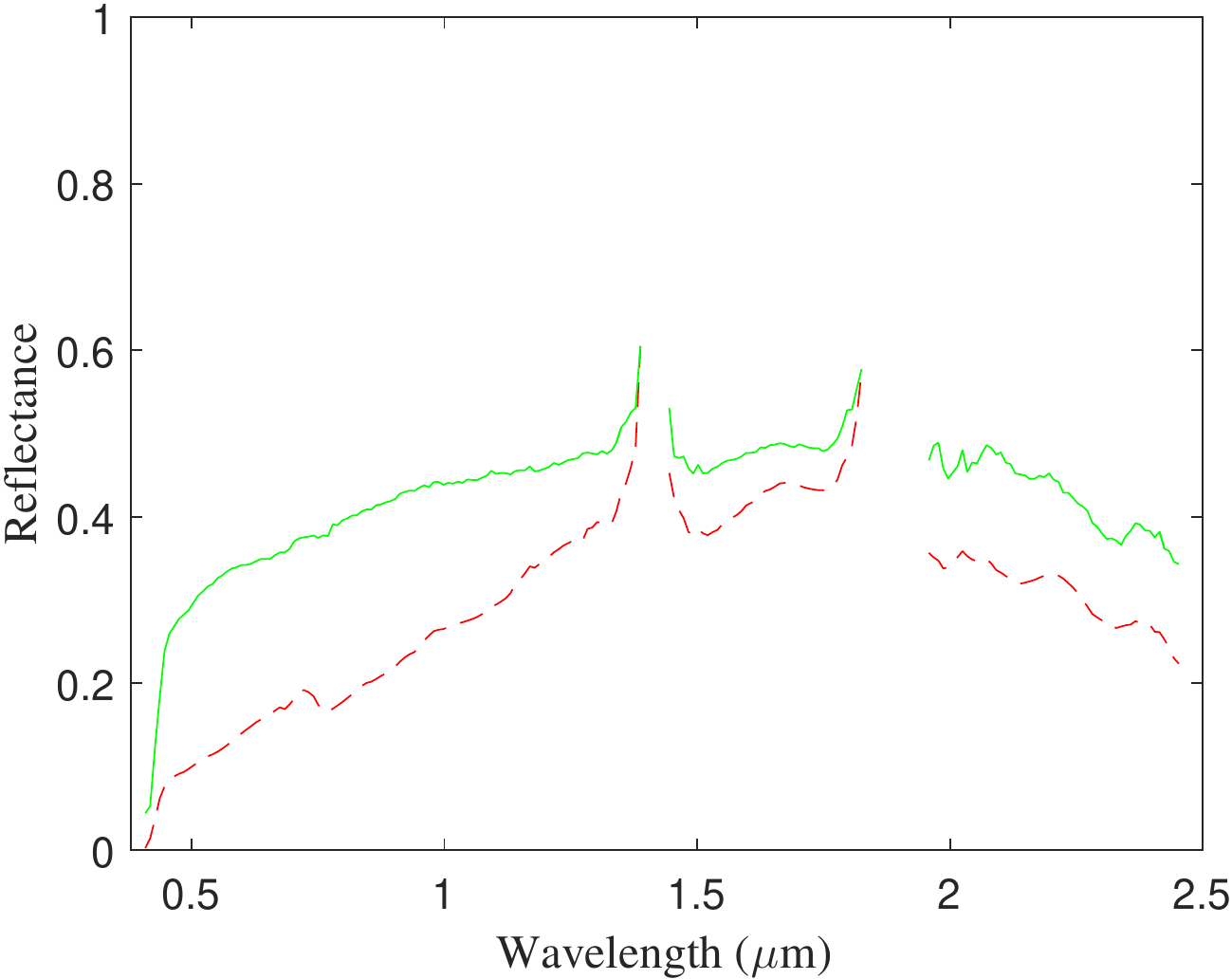}}}
\caption{Comparison of the reference spectra (green solid line) with those estimated by different methods (red dash line) of three endmembers on the Jasper Ridge data set. From top to bottom: Tree.  Water. Soil. Road. From left to right: (a) $L_{1/2}$-NMF. (b) SGSNMF. (c) TV-RSNMF. (d) $L_{1/2}$-RNMF. (e) MV-NTF-TV. (f) MLNMF. (g) SSRDMF.}
\label{fig:9}
\end{figure*}

\begin{figure*}[!t]
\centering
\mbox{
{\includegraphics[width=2.45cm]{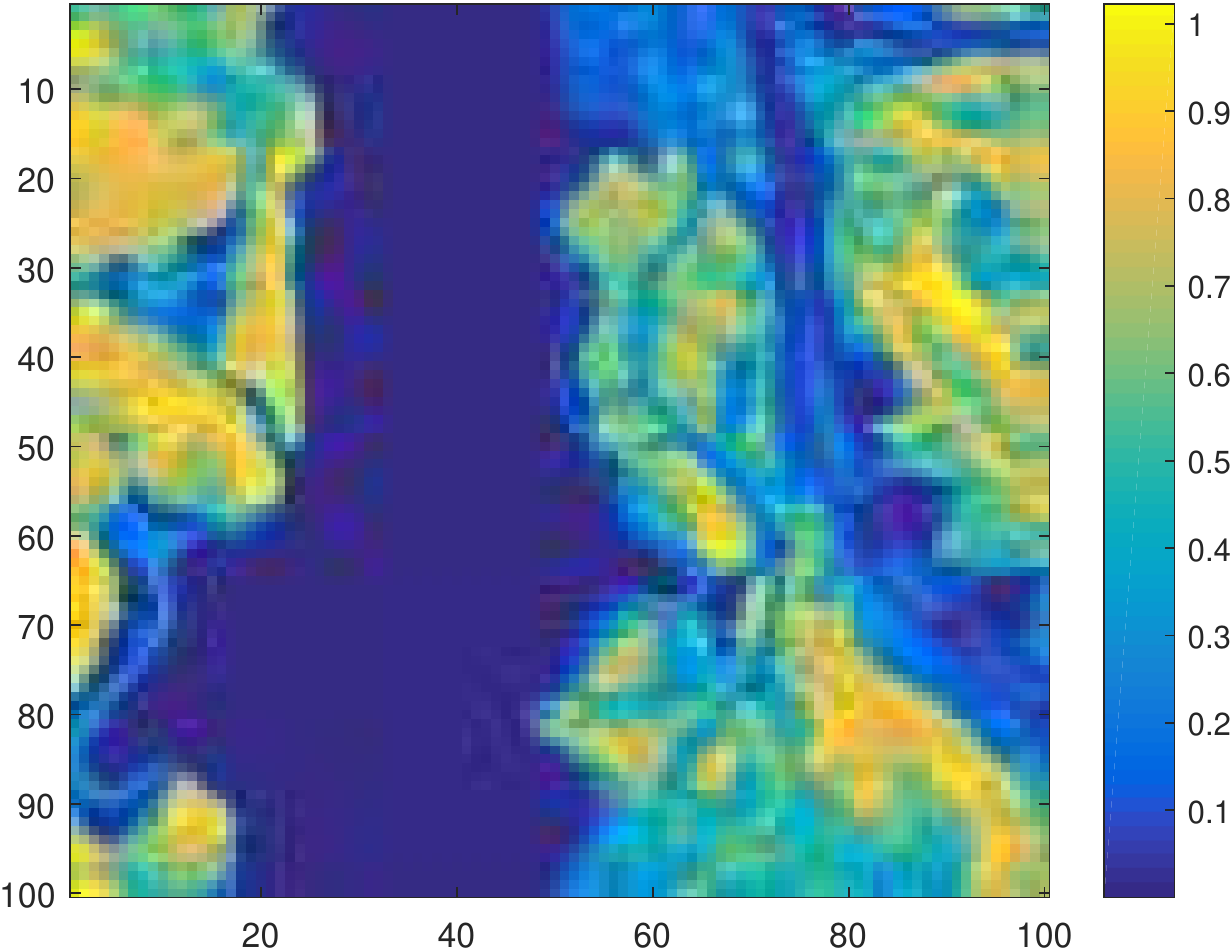}}
{\includegraphics[width=2.45cm]{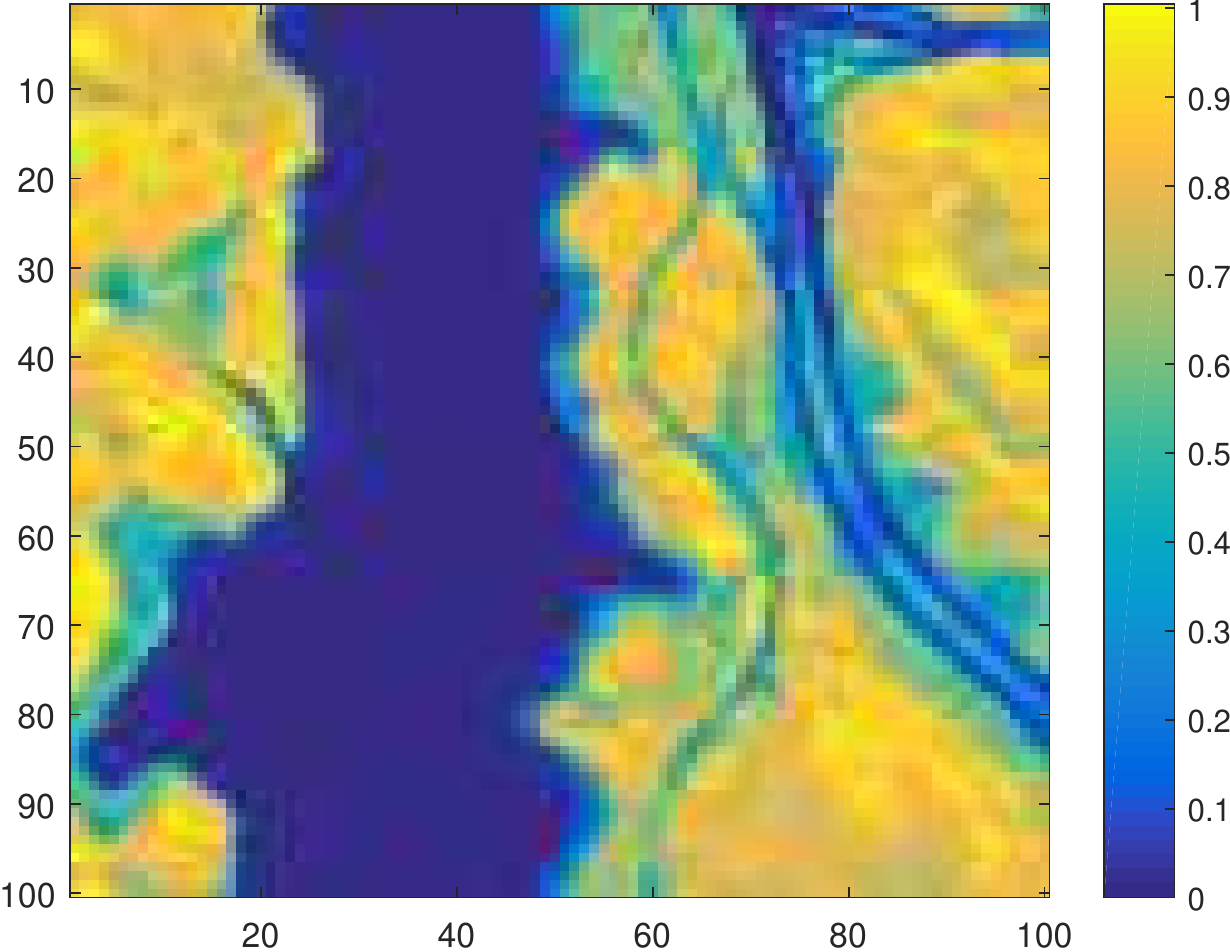}}
{\includegraphics[width=2.45cm]{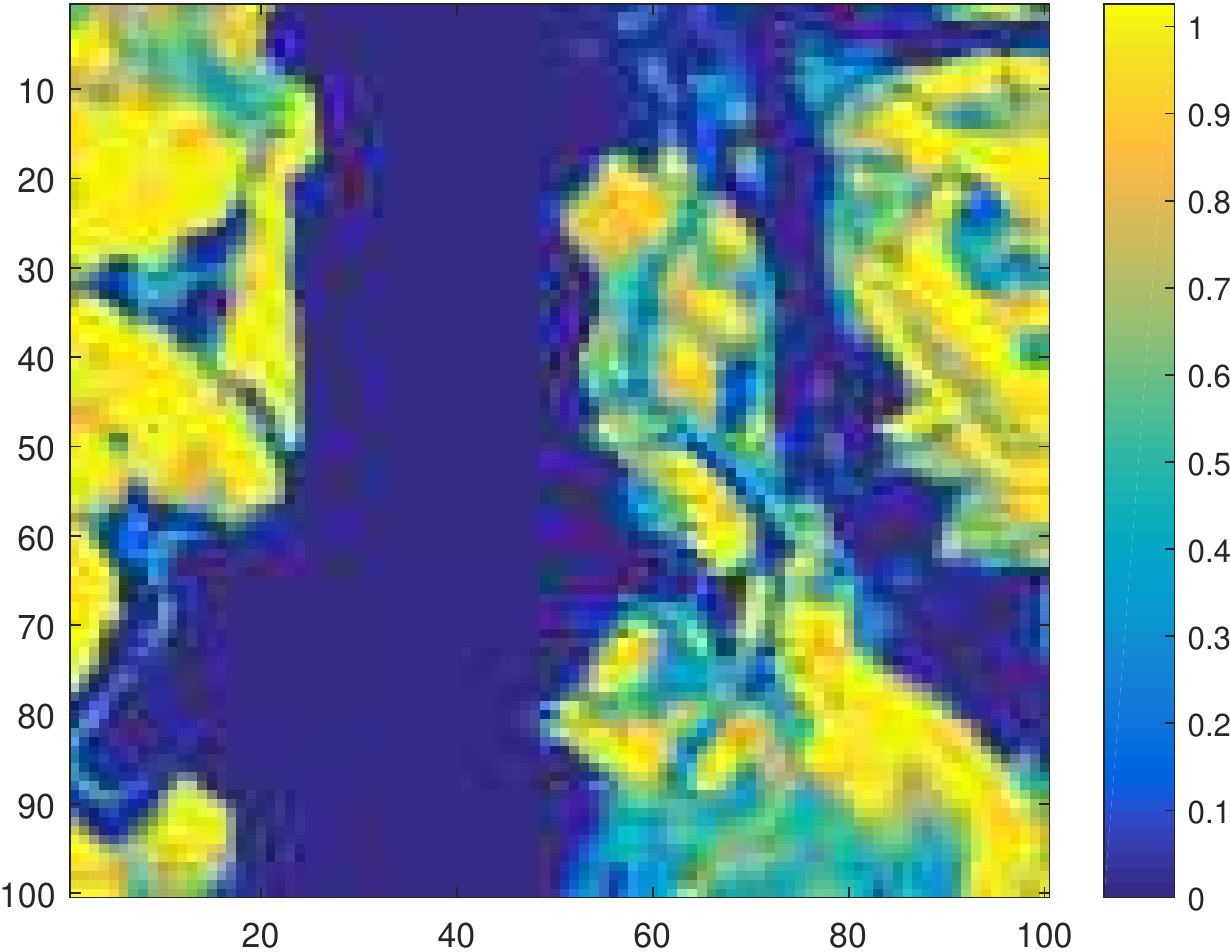}}
{\includegraphics[width=2.45cm]{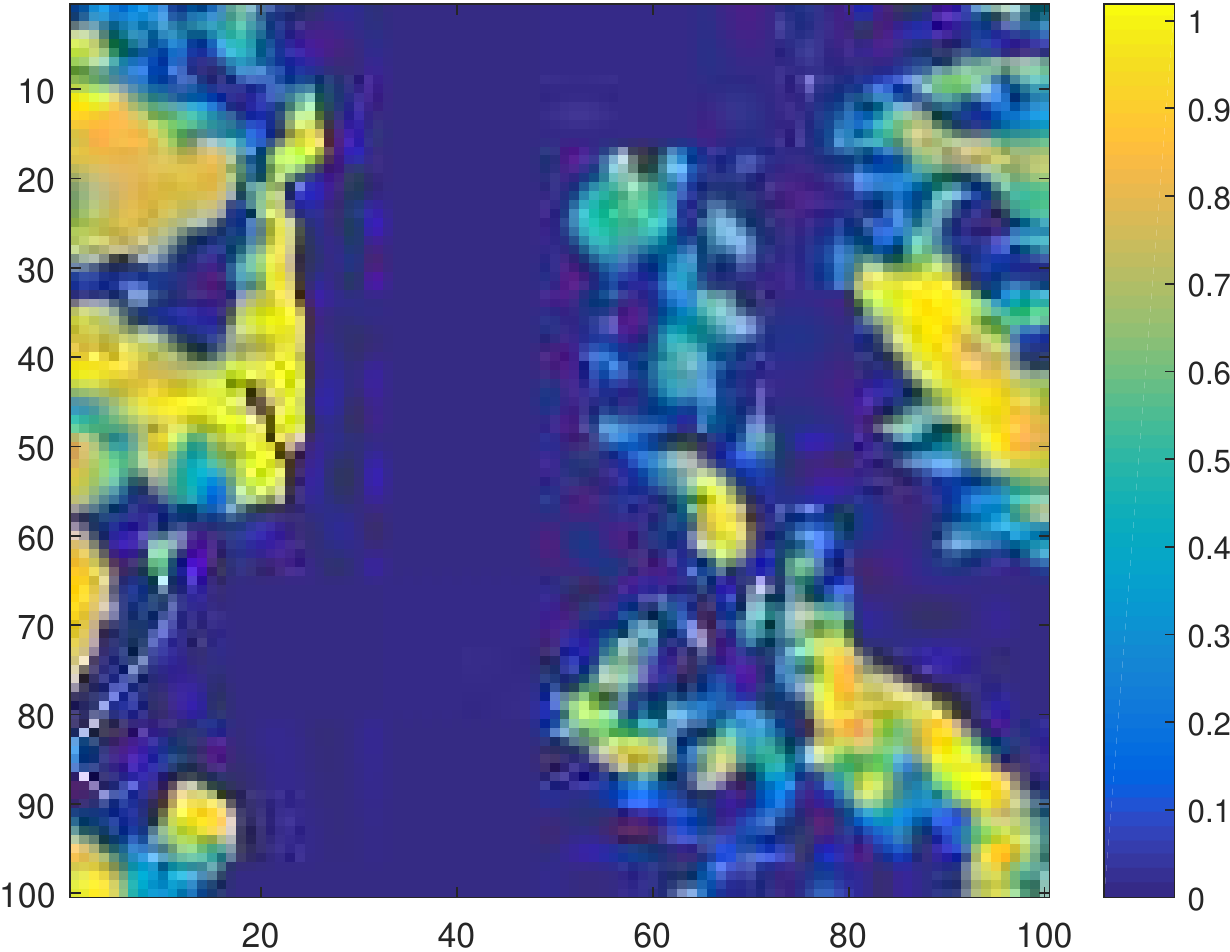}}
{\includegraphics[width=2.45cm]{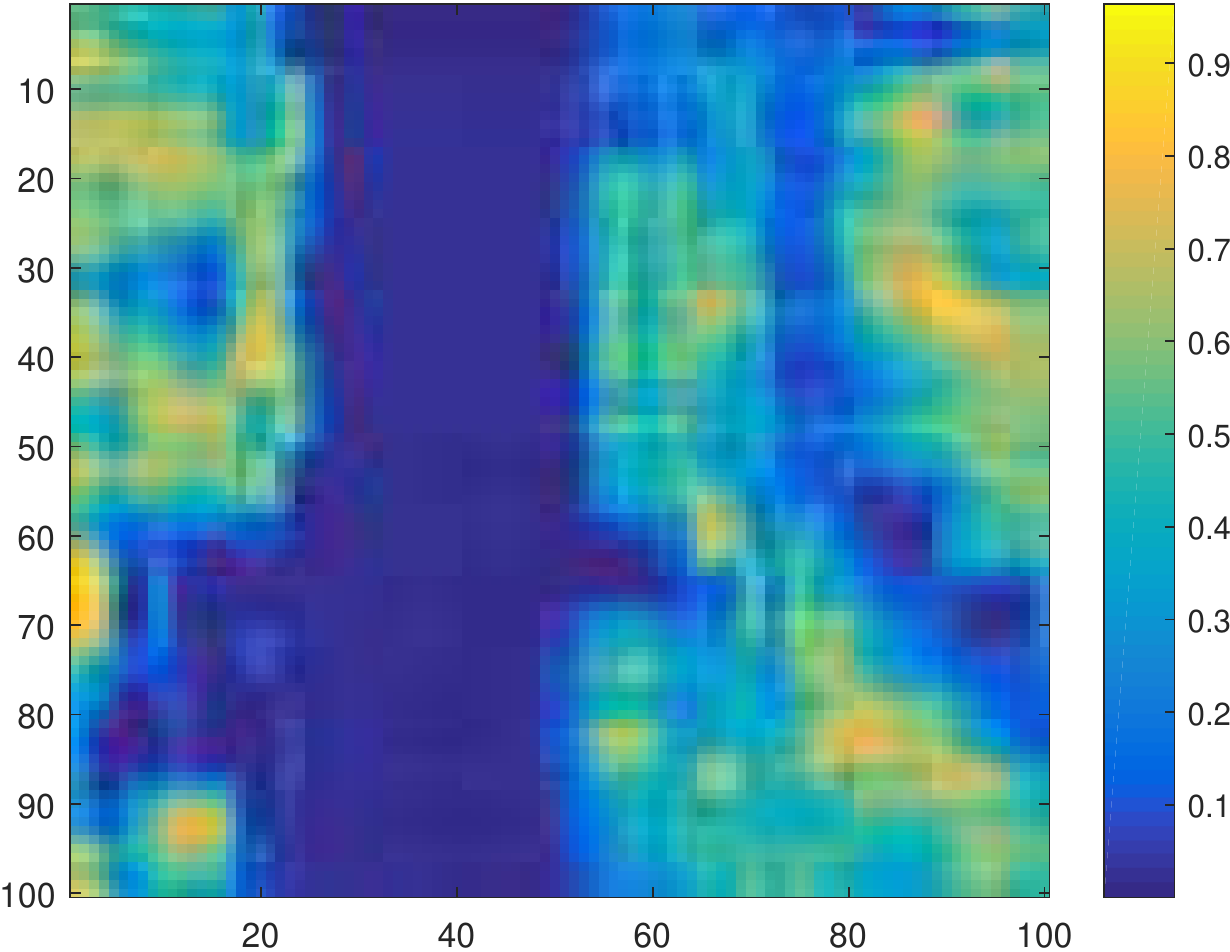}}
{\includegraphics[width=2.45cm]{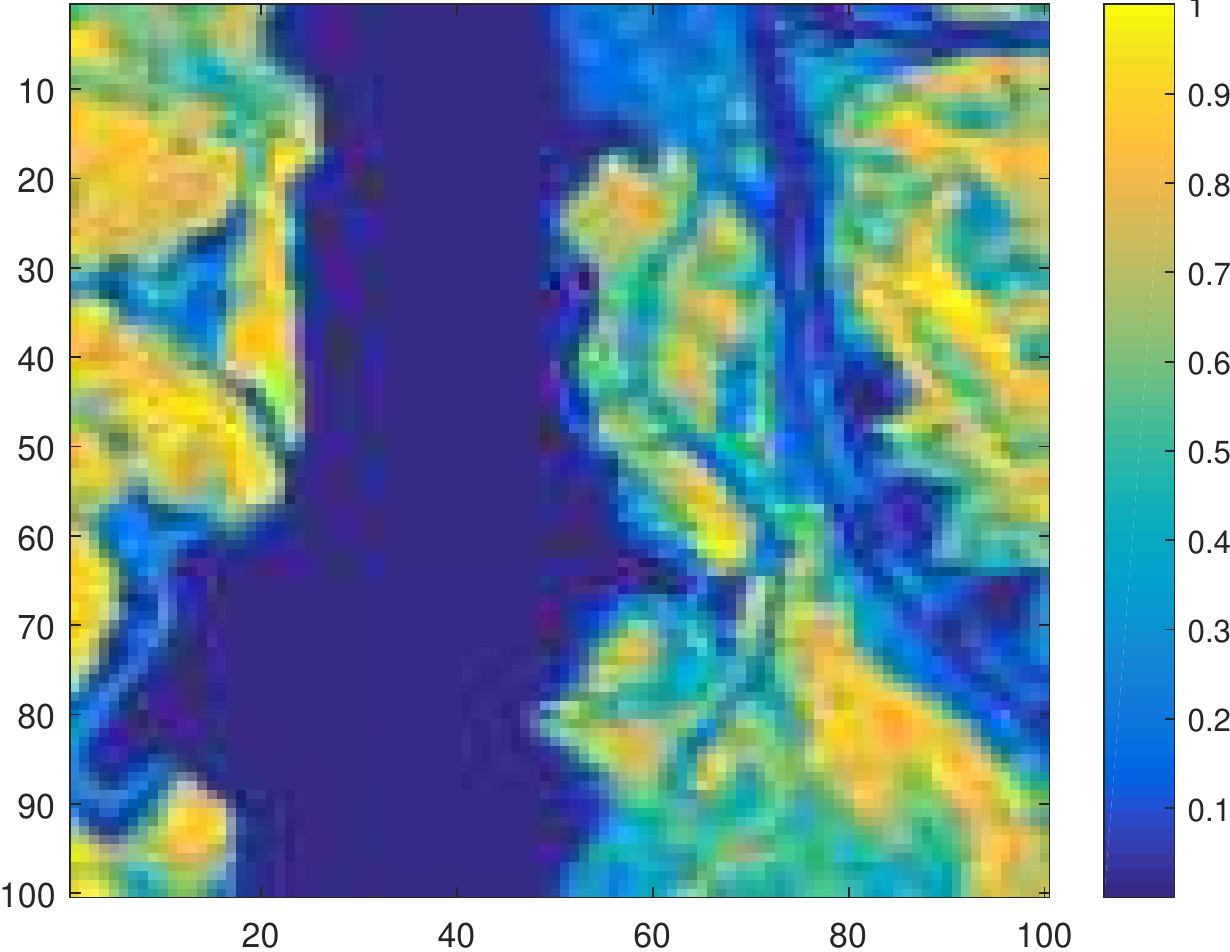}}
{\includegraphics[width=2.45cm]{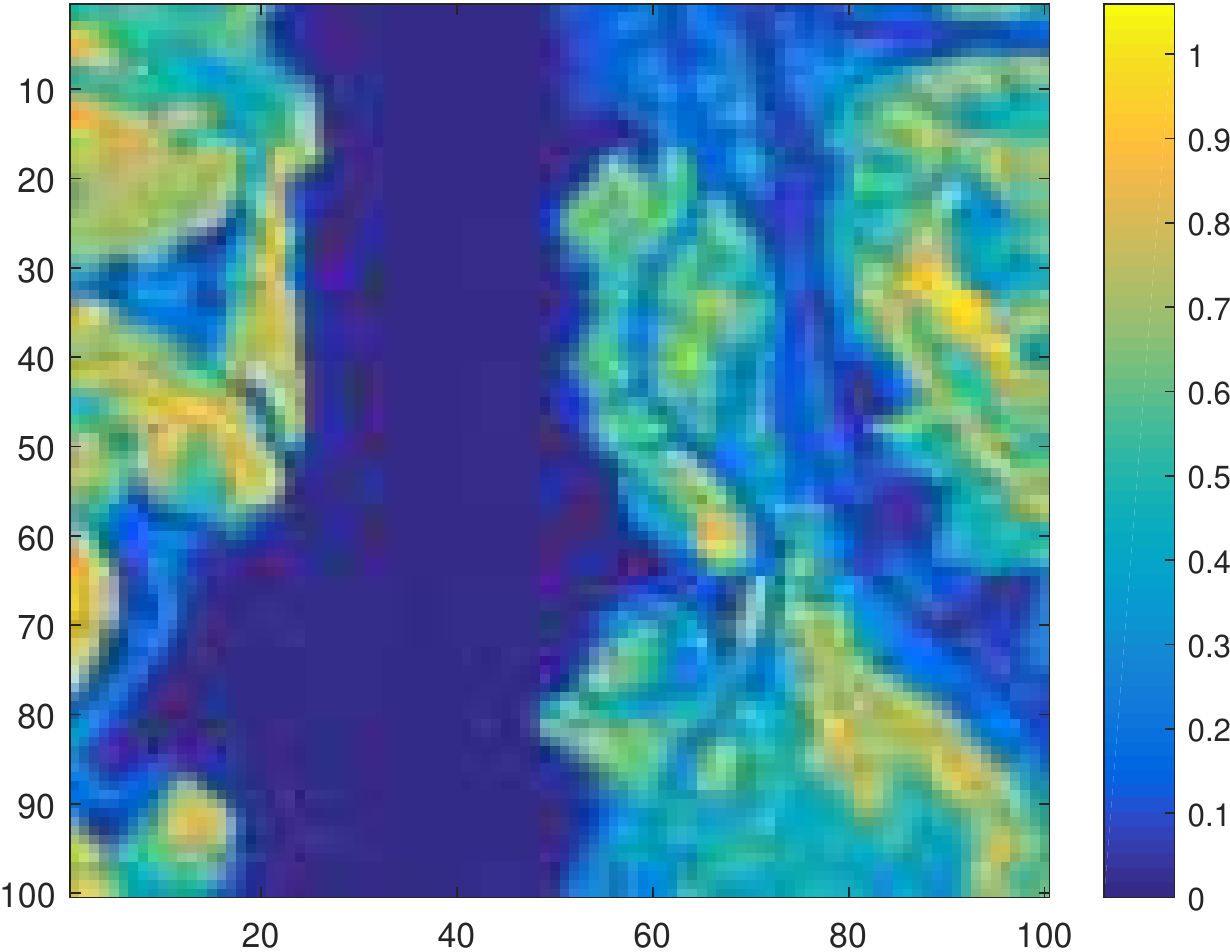}}}
\\
\mbox{
{\includegraphics[width=2.45cm]{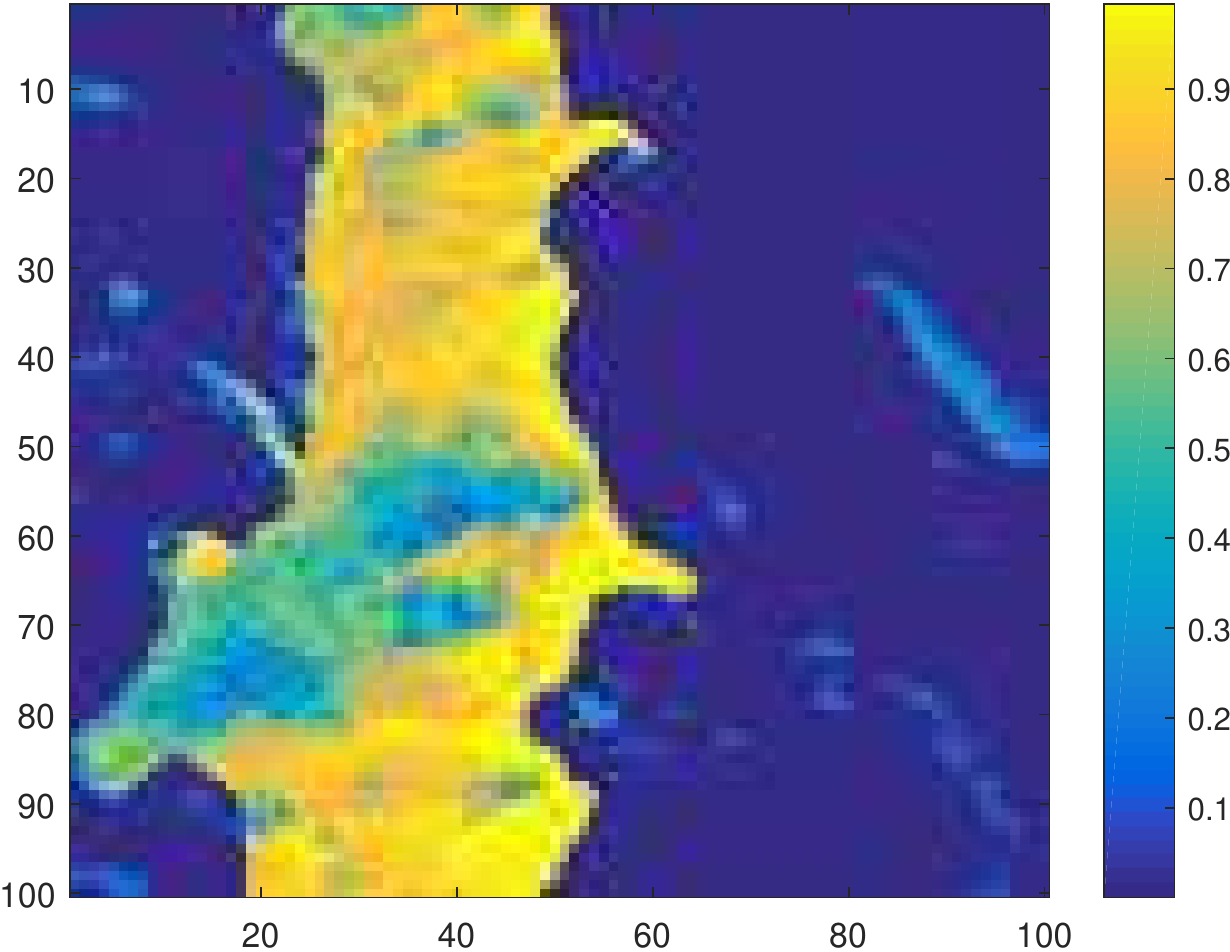}}
{\includegraphics[width=2.45cm]{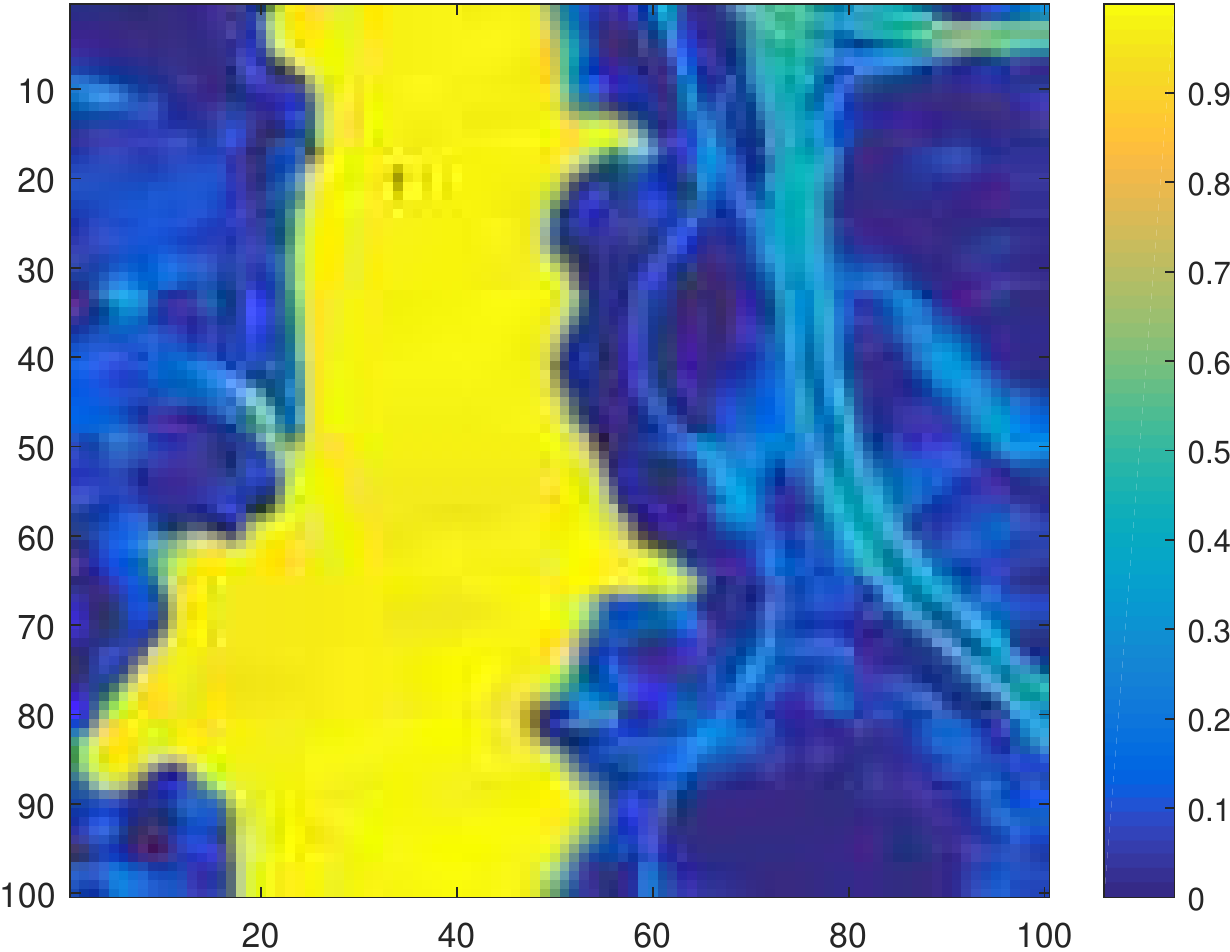}}
{\includegraphics[width=2.45cm]{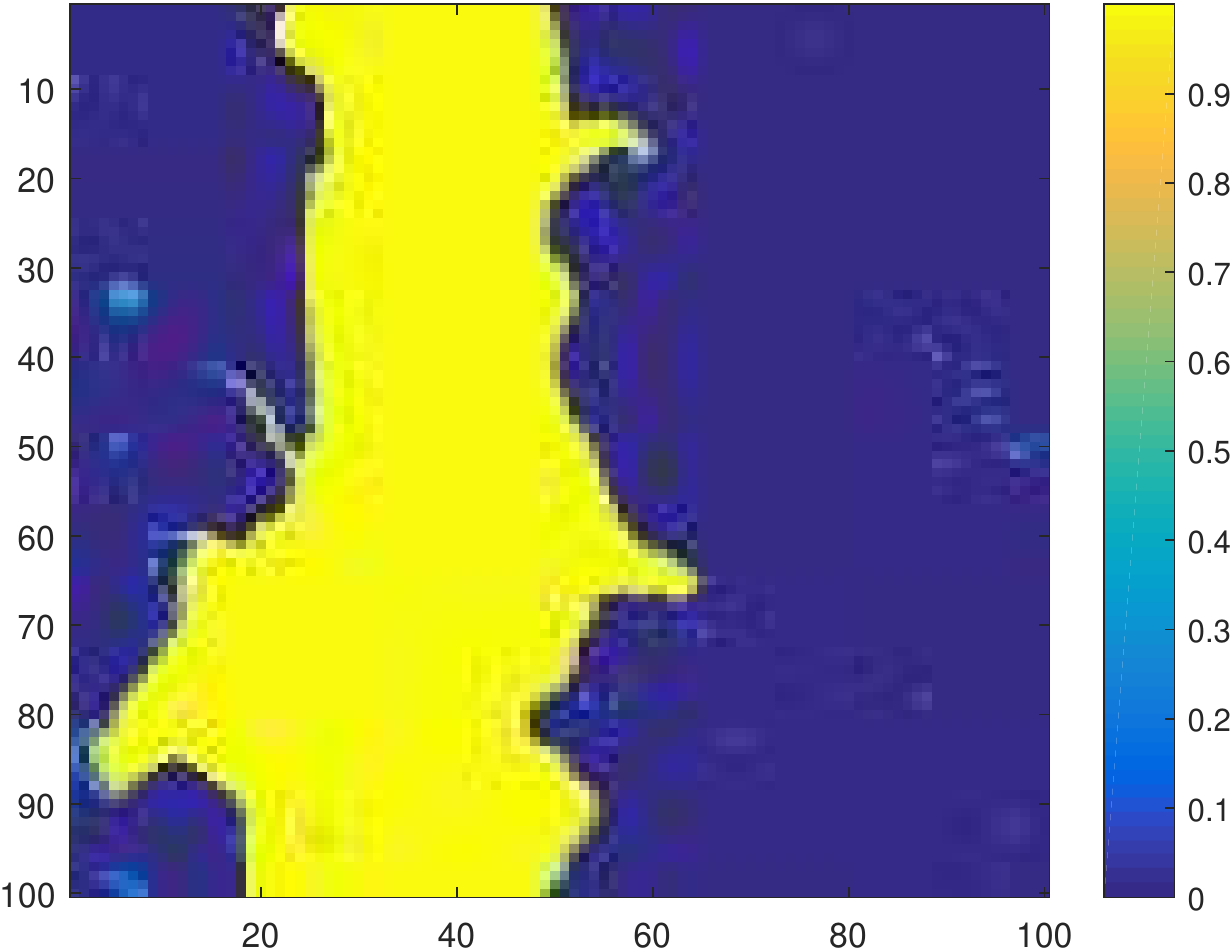}}
{\includegraphics[width=2.45cm]{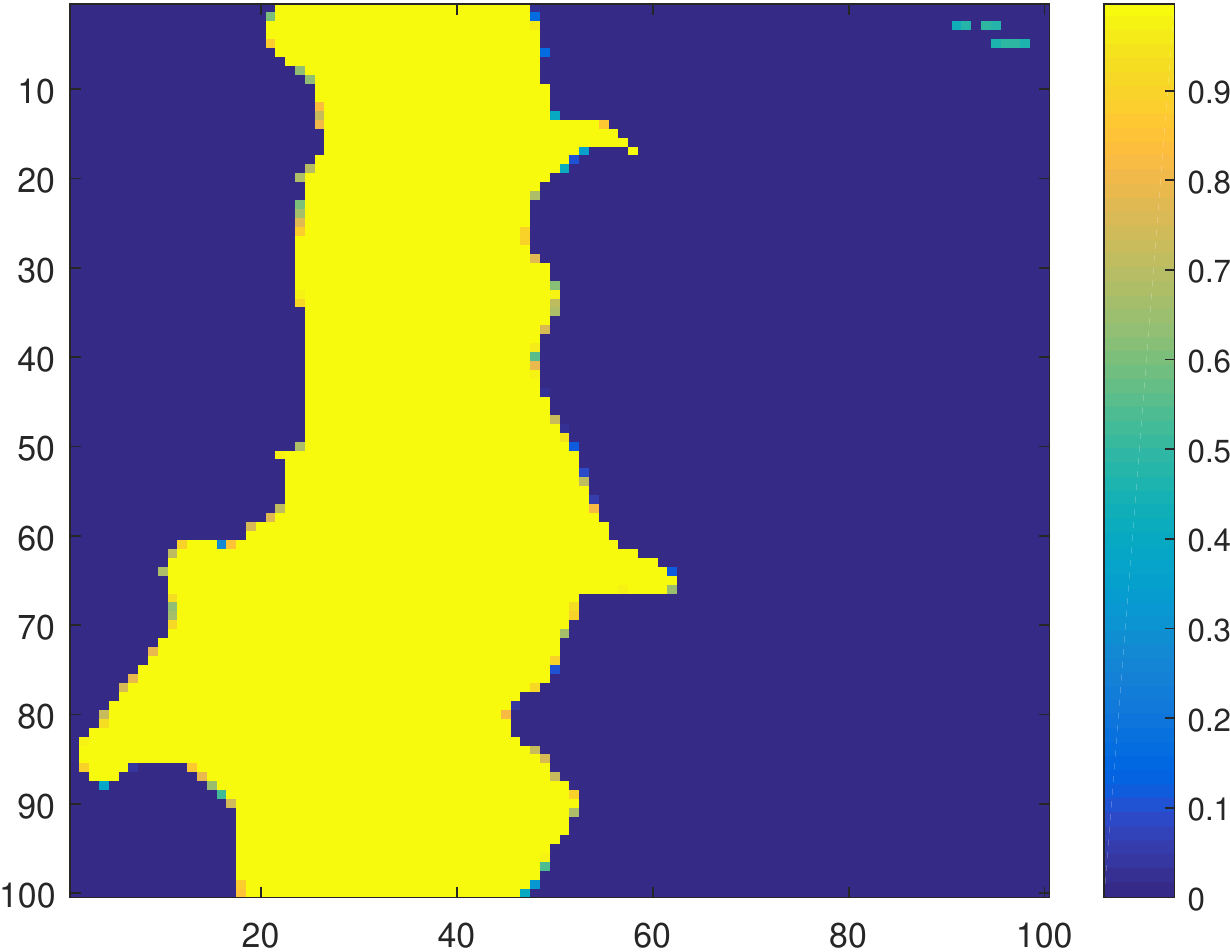}}
{\includegraphics[width=2.45cm]{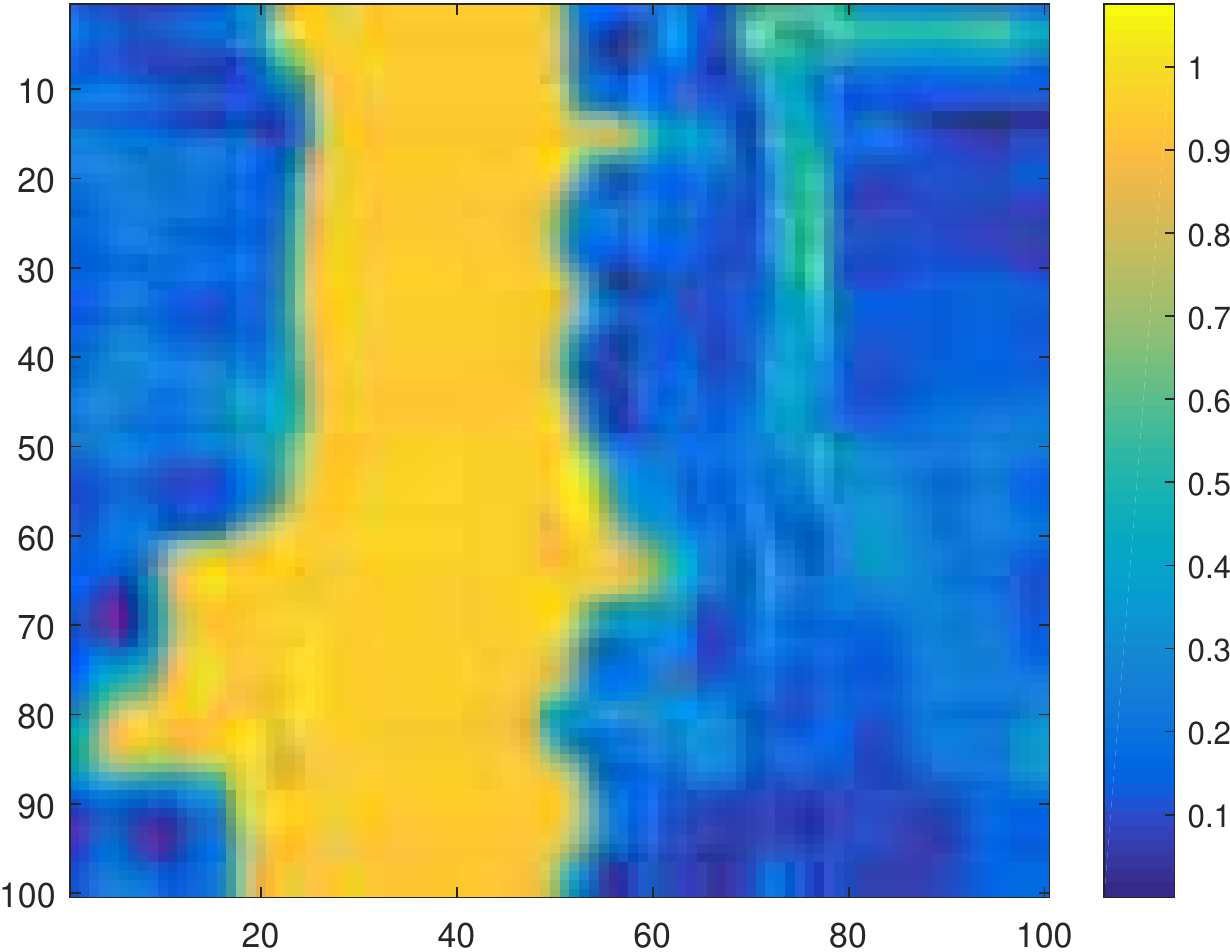}}
{\includegraphics[width=2.45cm]{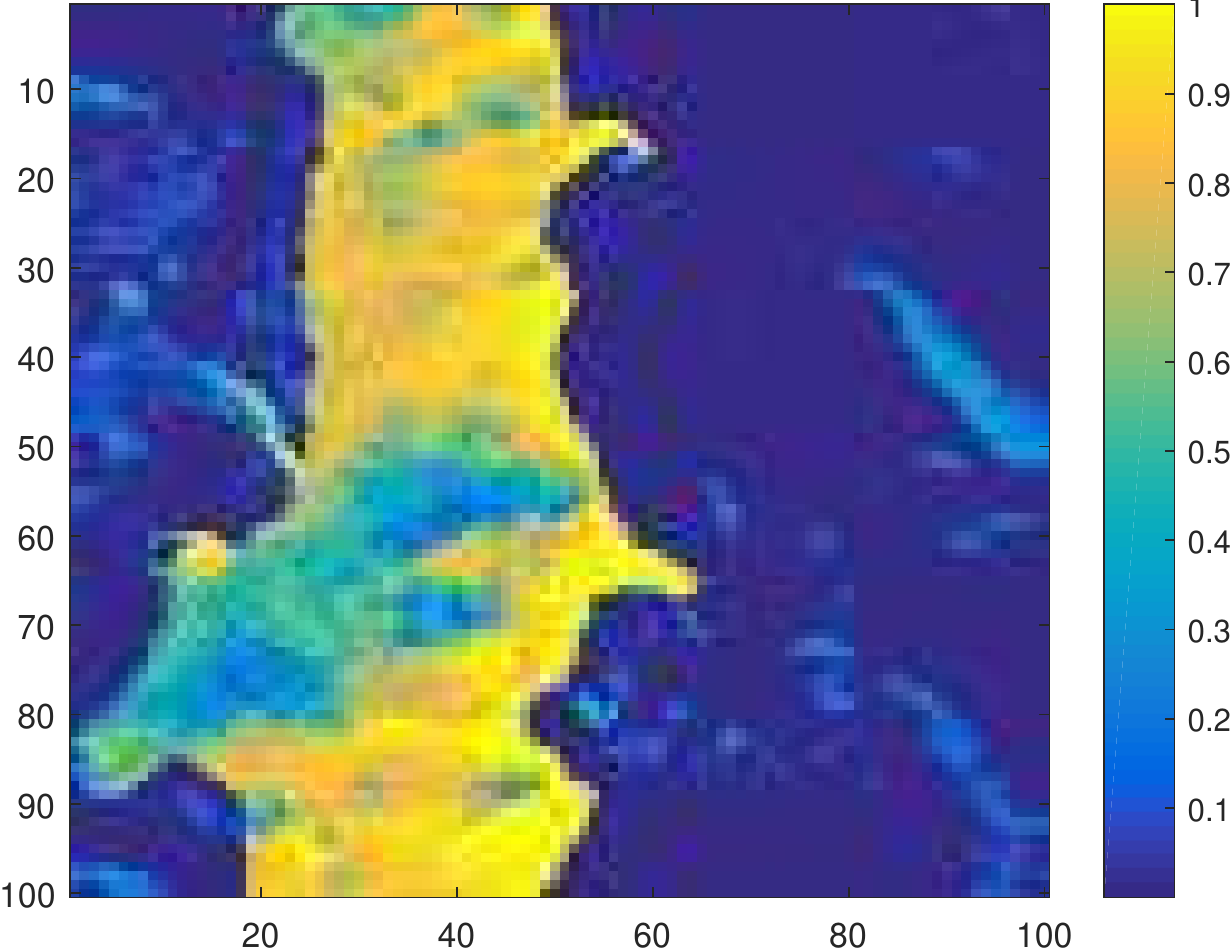}}
{\includegraphics[width=2.45cm]{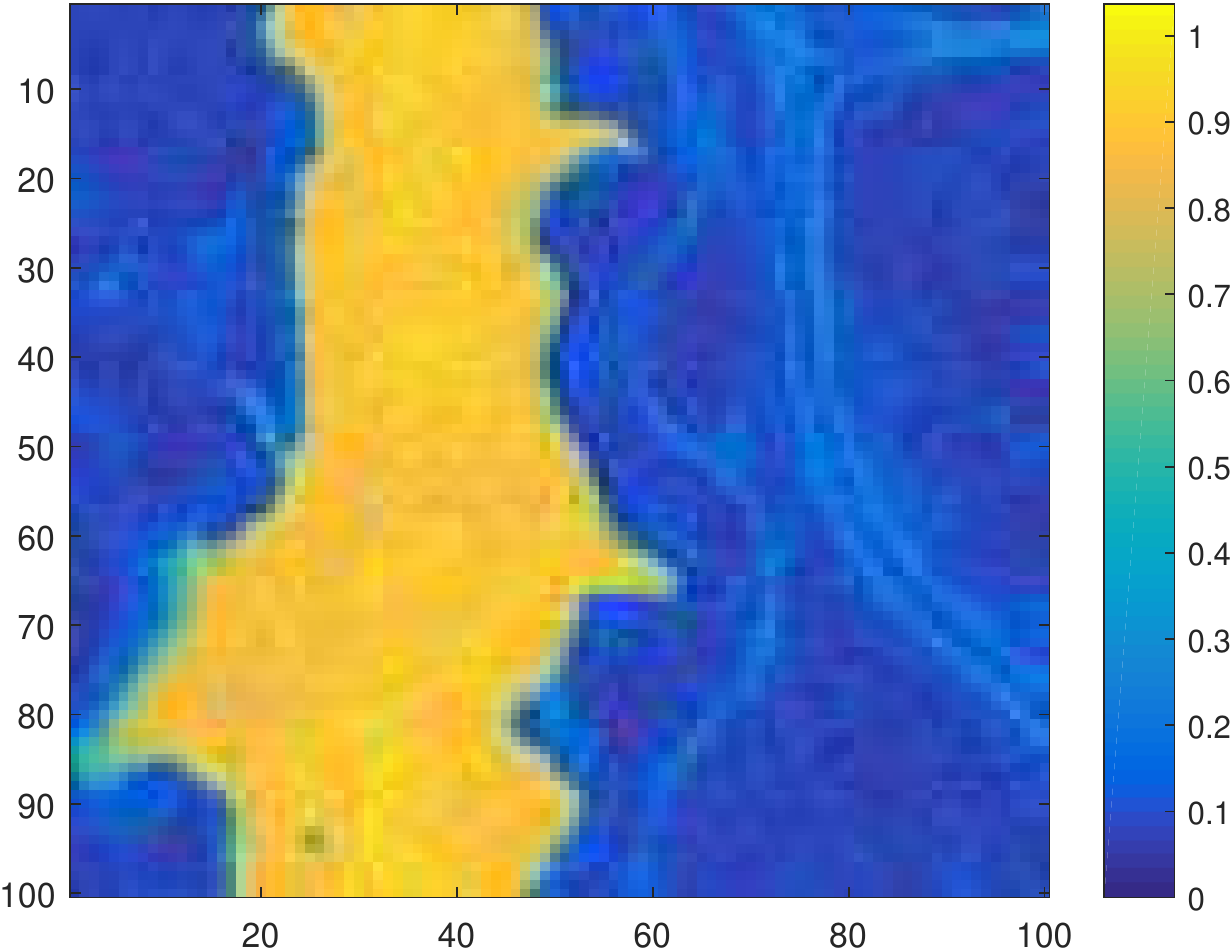}}}
\\
\mbox{
{\includegraphics[width=2.45cm]{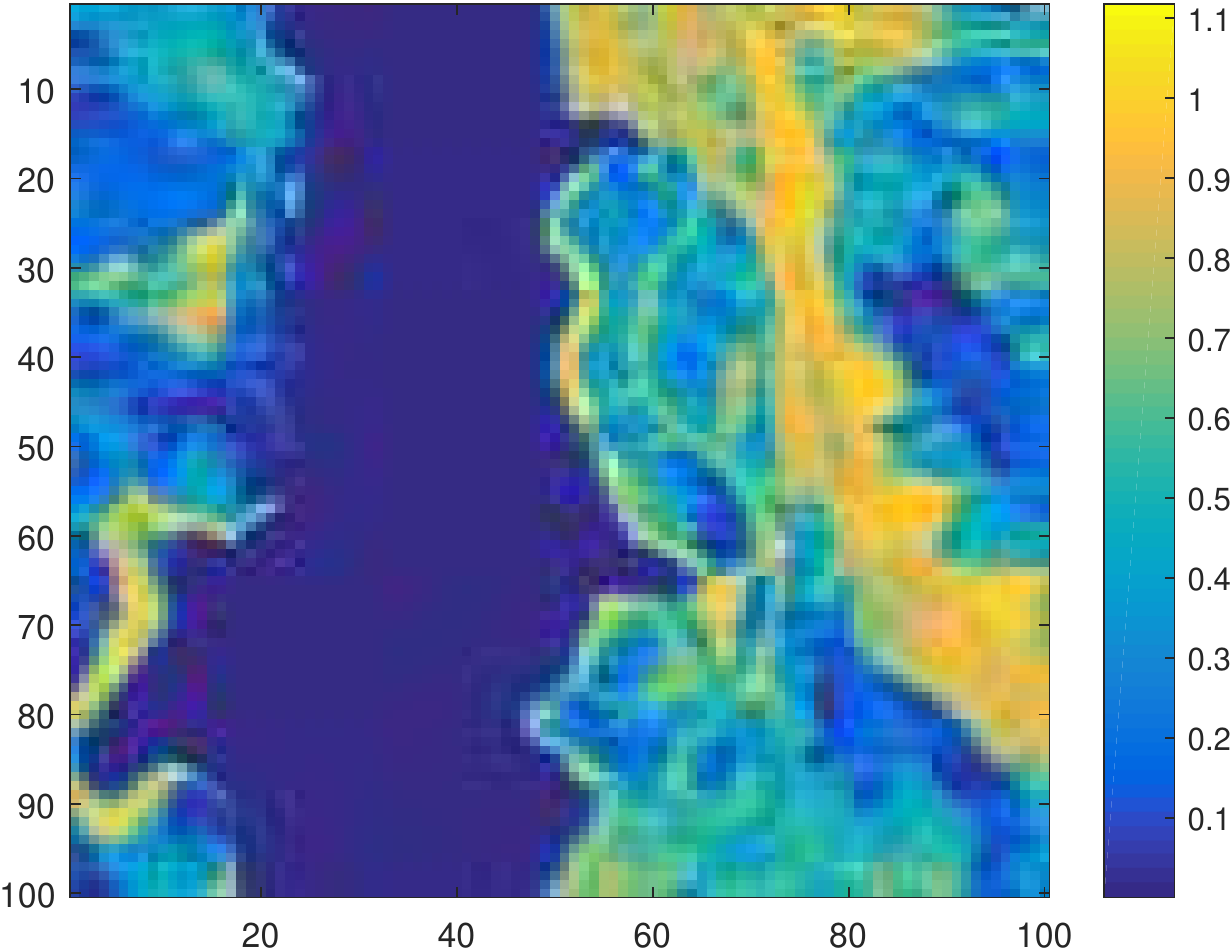}}
{\includegraphics[width=2.45cm]{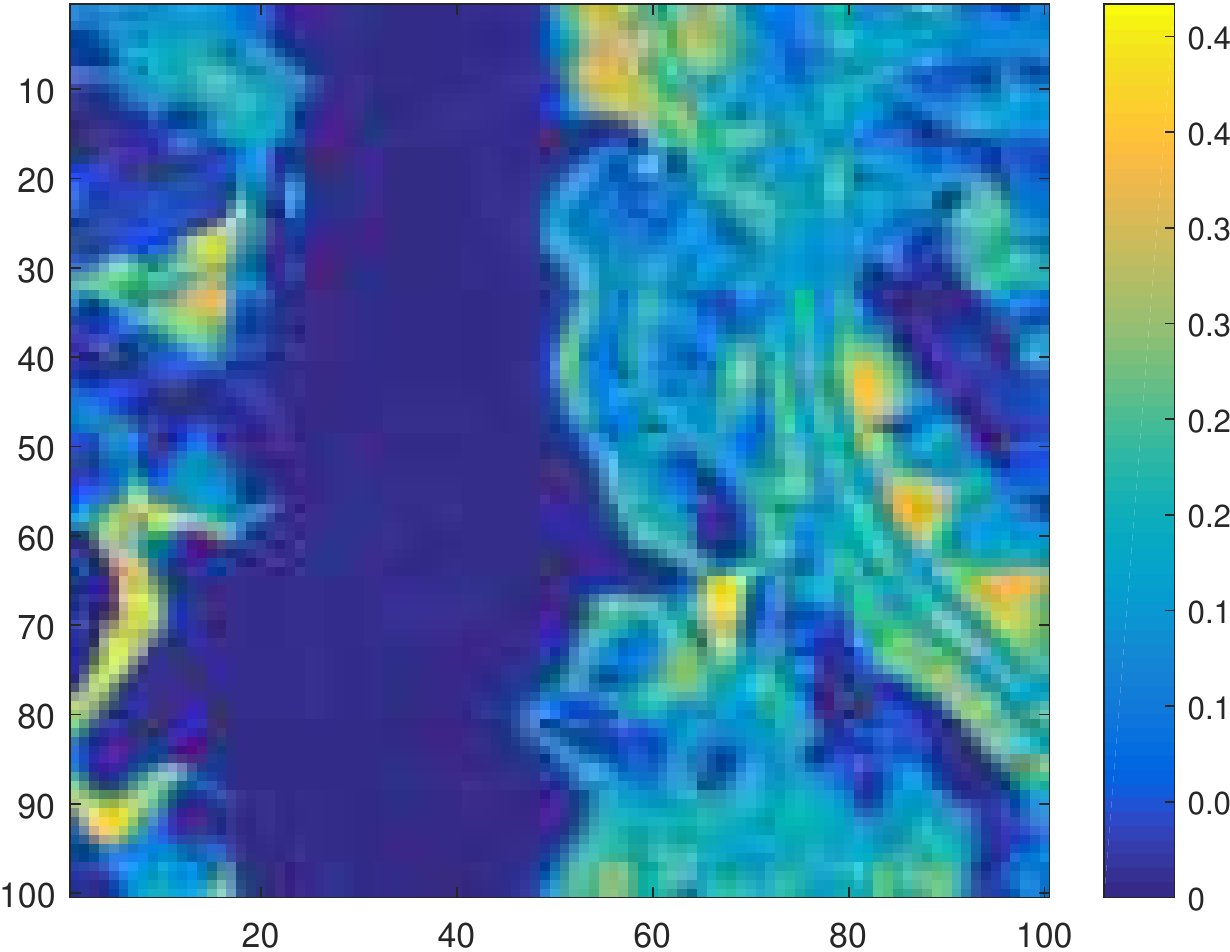}}
{\includegraphics[width=2.45cm]{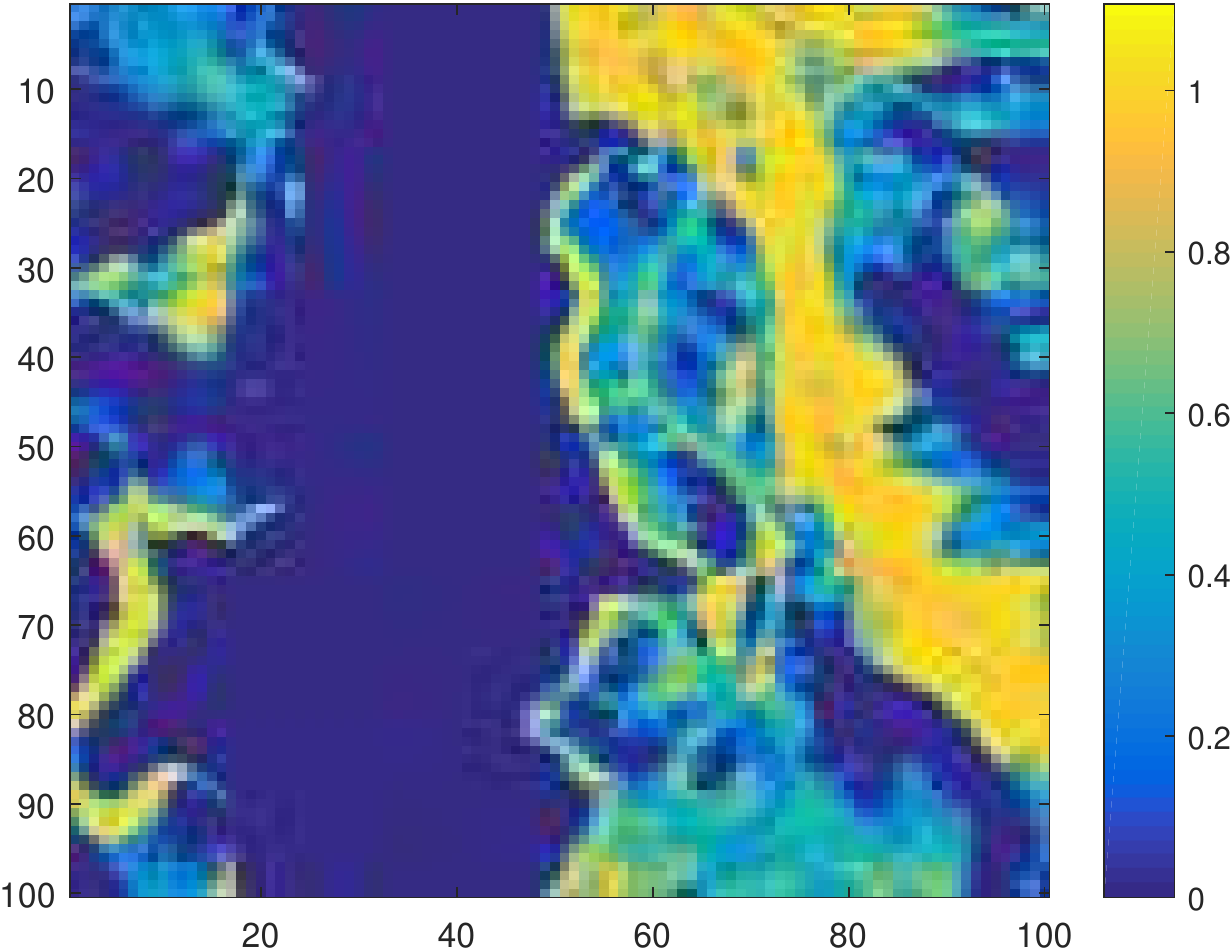}}
{\includegraphics[width=2.45cm]{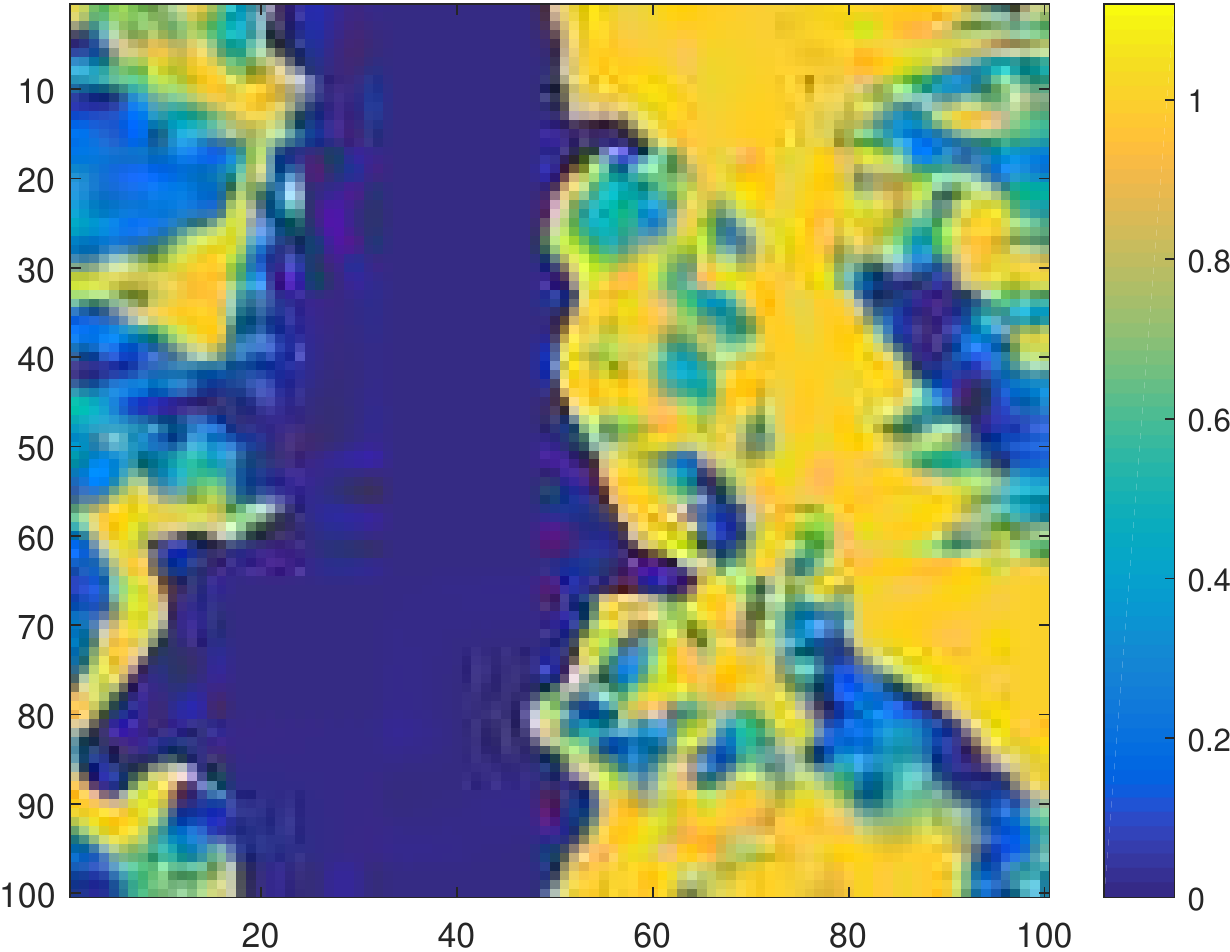}}
{\includegraphics[width=2.45cm]{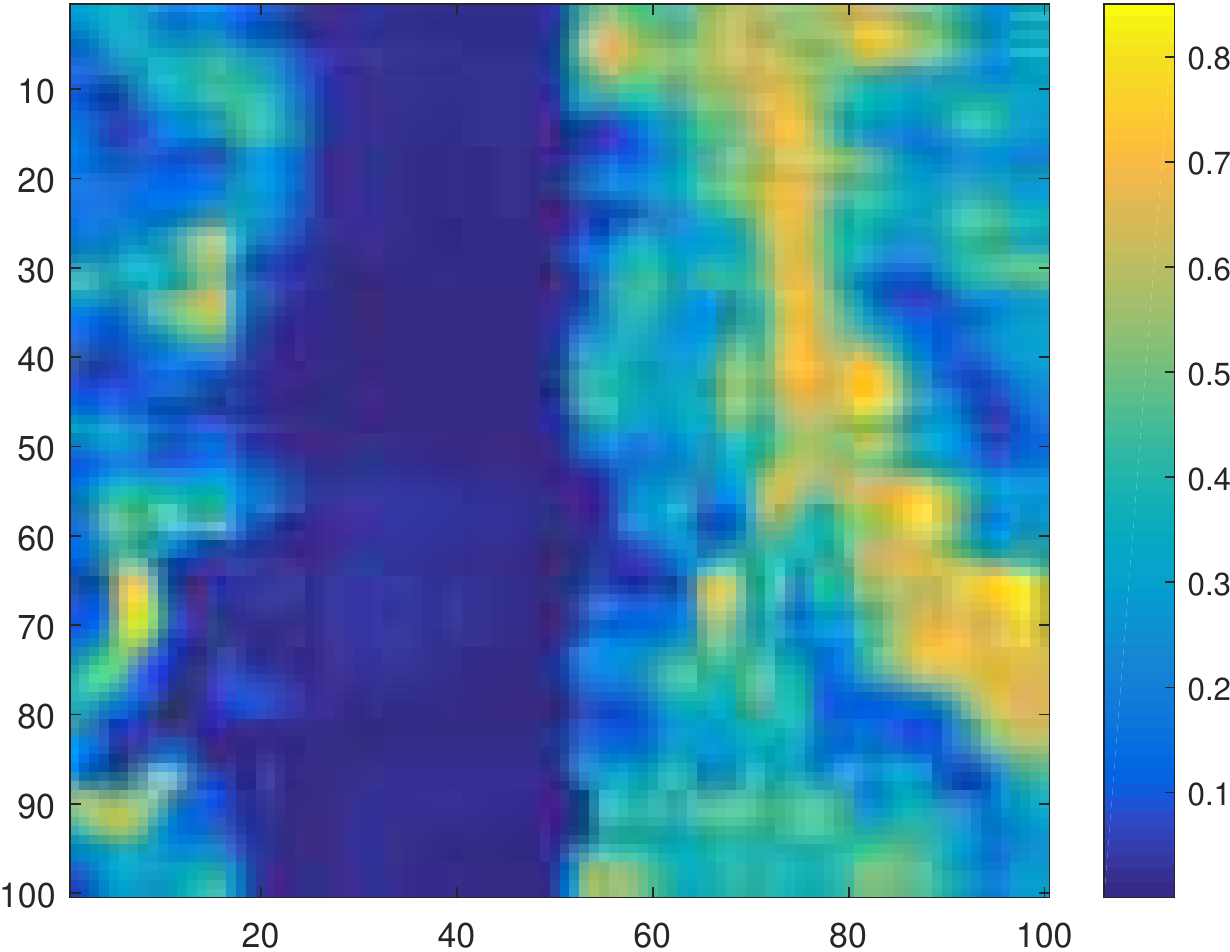}}
{\includegraphics[width=2.45cm]{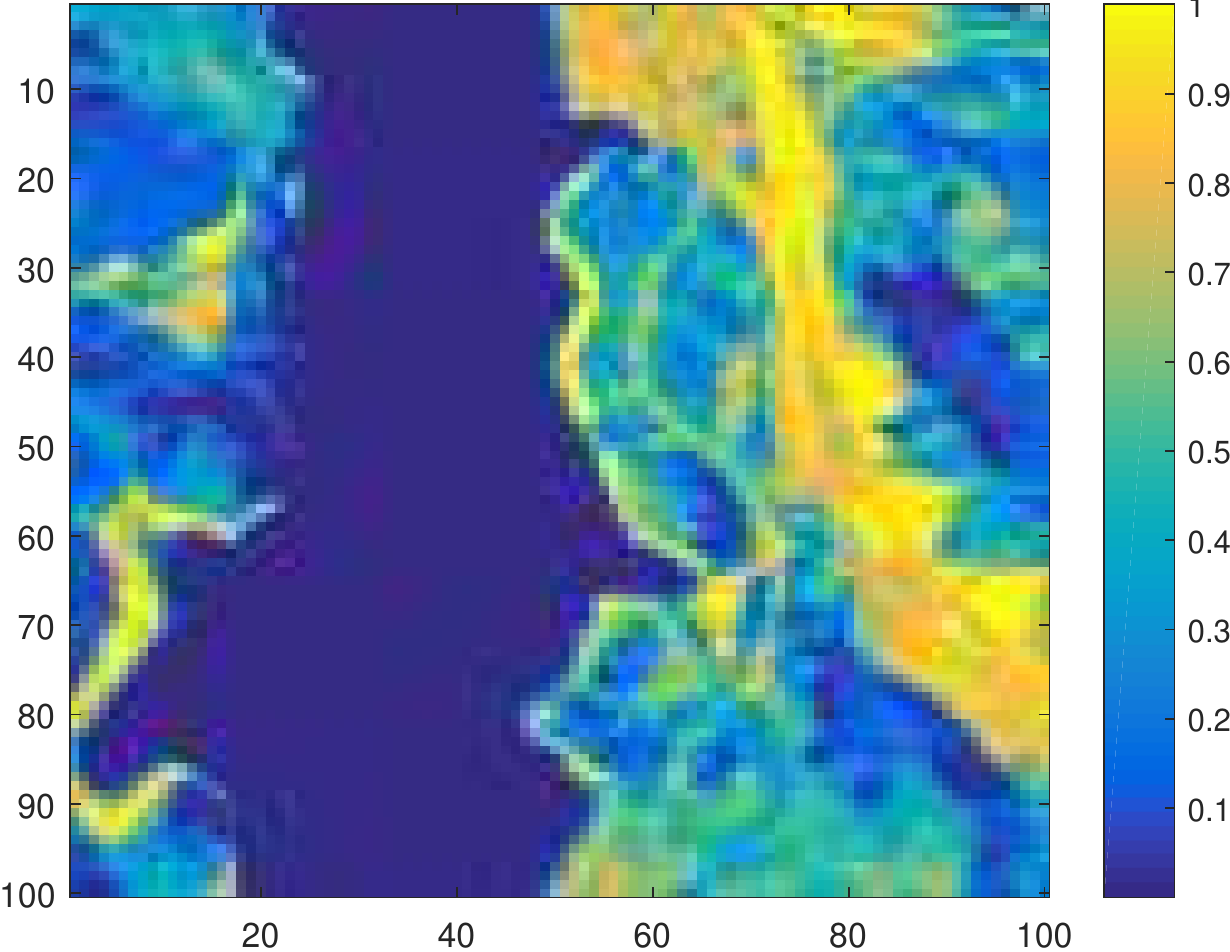}}
{\includegraphics[width=2.45cm]{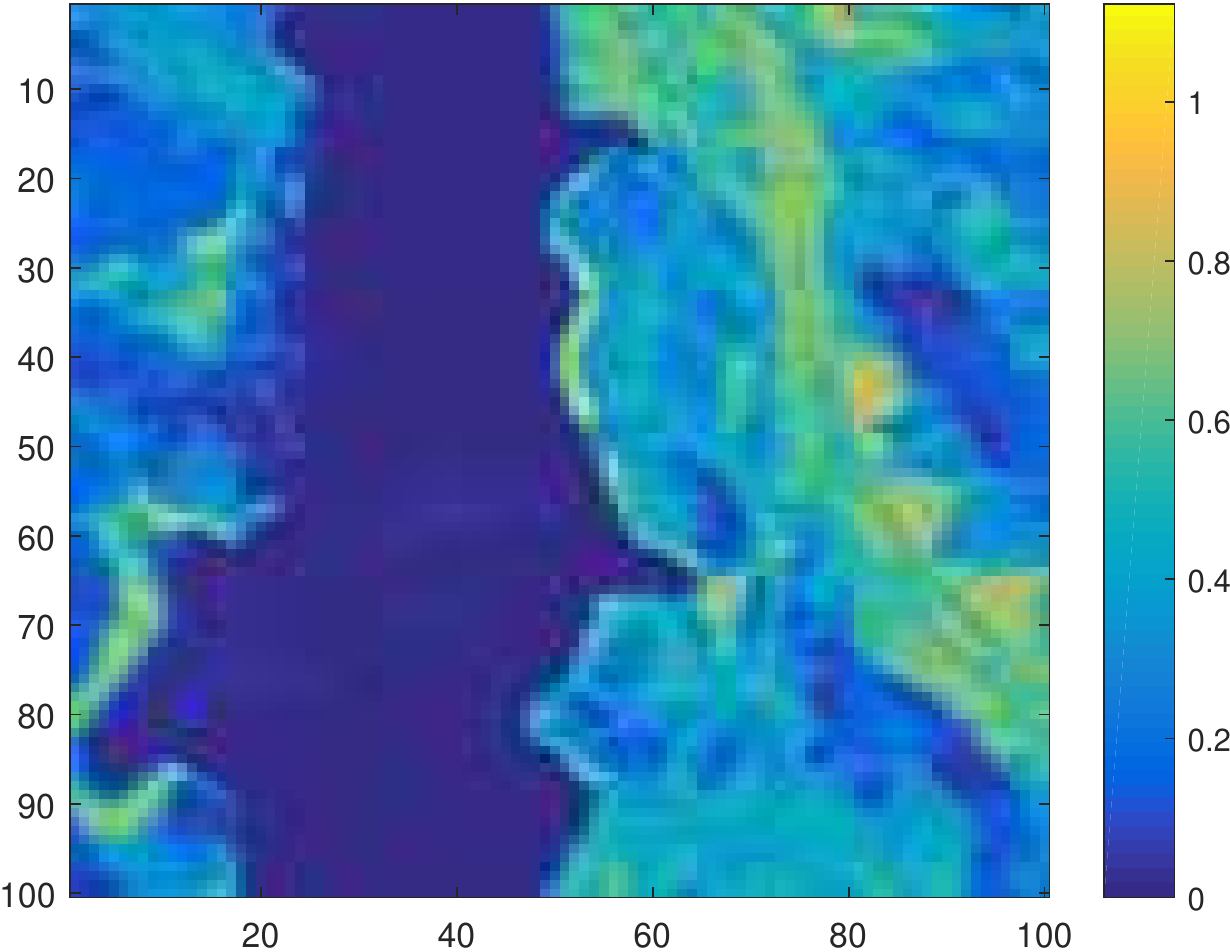}}}
\\
\mbox{
\subfigure[]{\includegraphics[width=2.45cm]{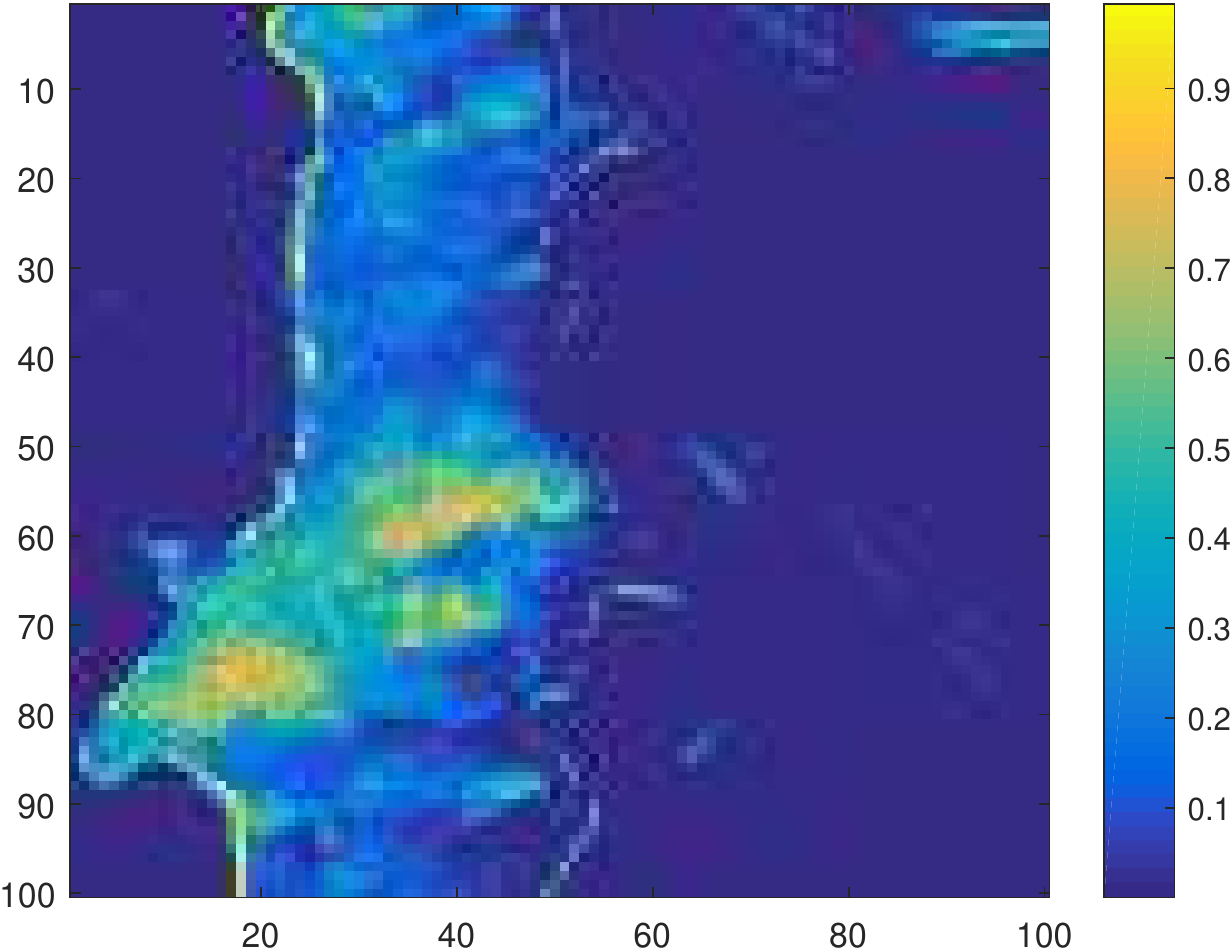}}
\subfigure[]{\includegraphics[width=2.45cm]{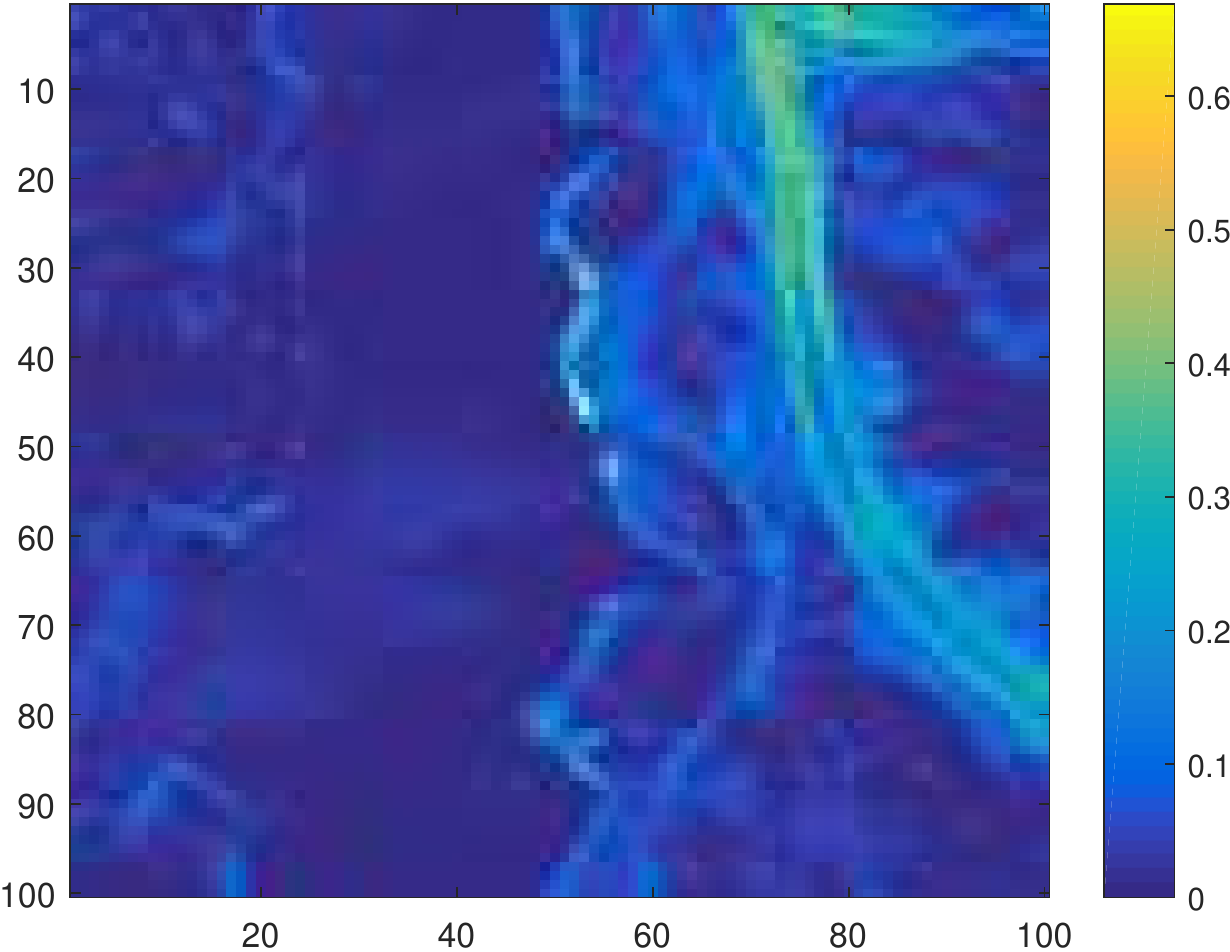}}
\subfigure[]{\includegraphics[width=2.45cm]{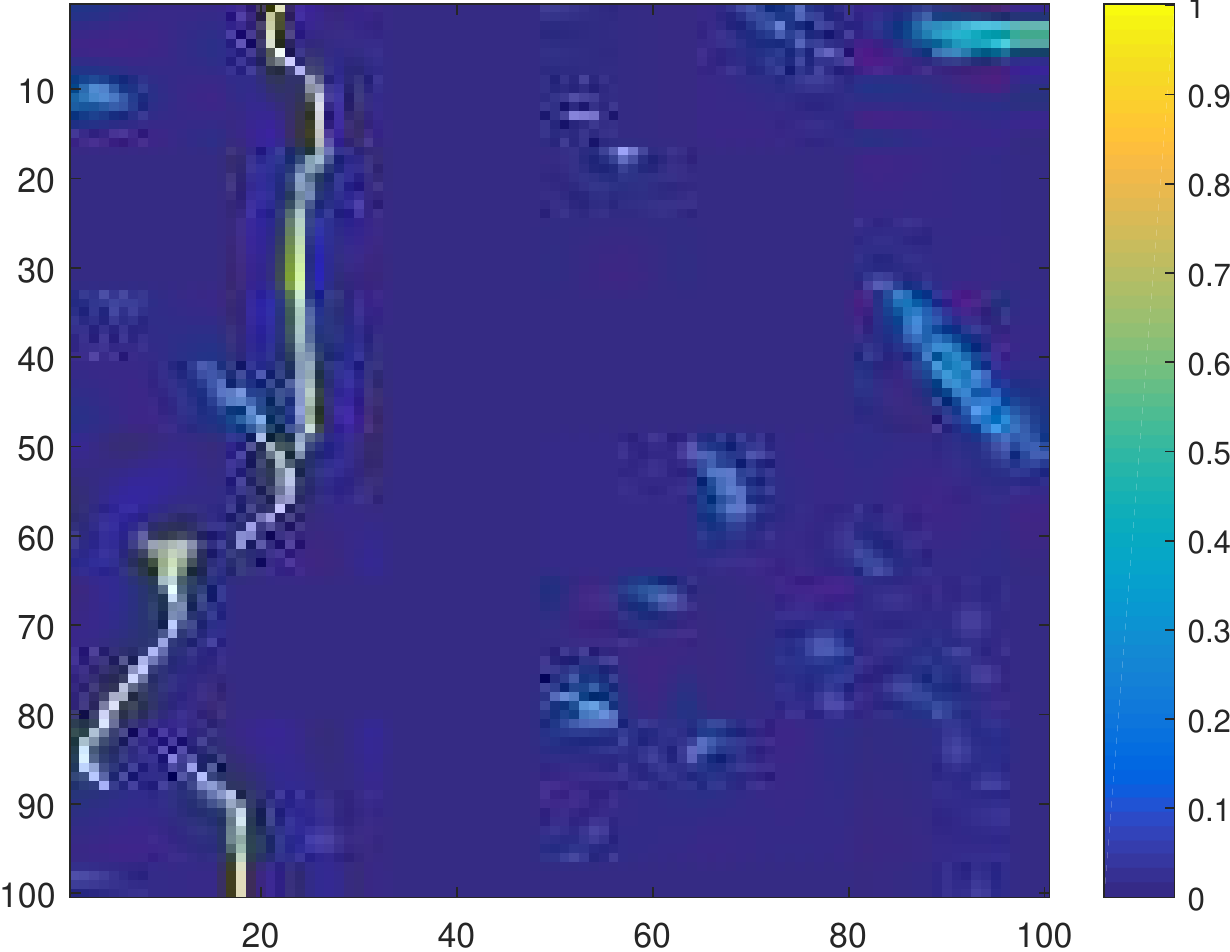}}
\subfigure[]{\includegraphics[width=2.45cm]{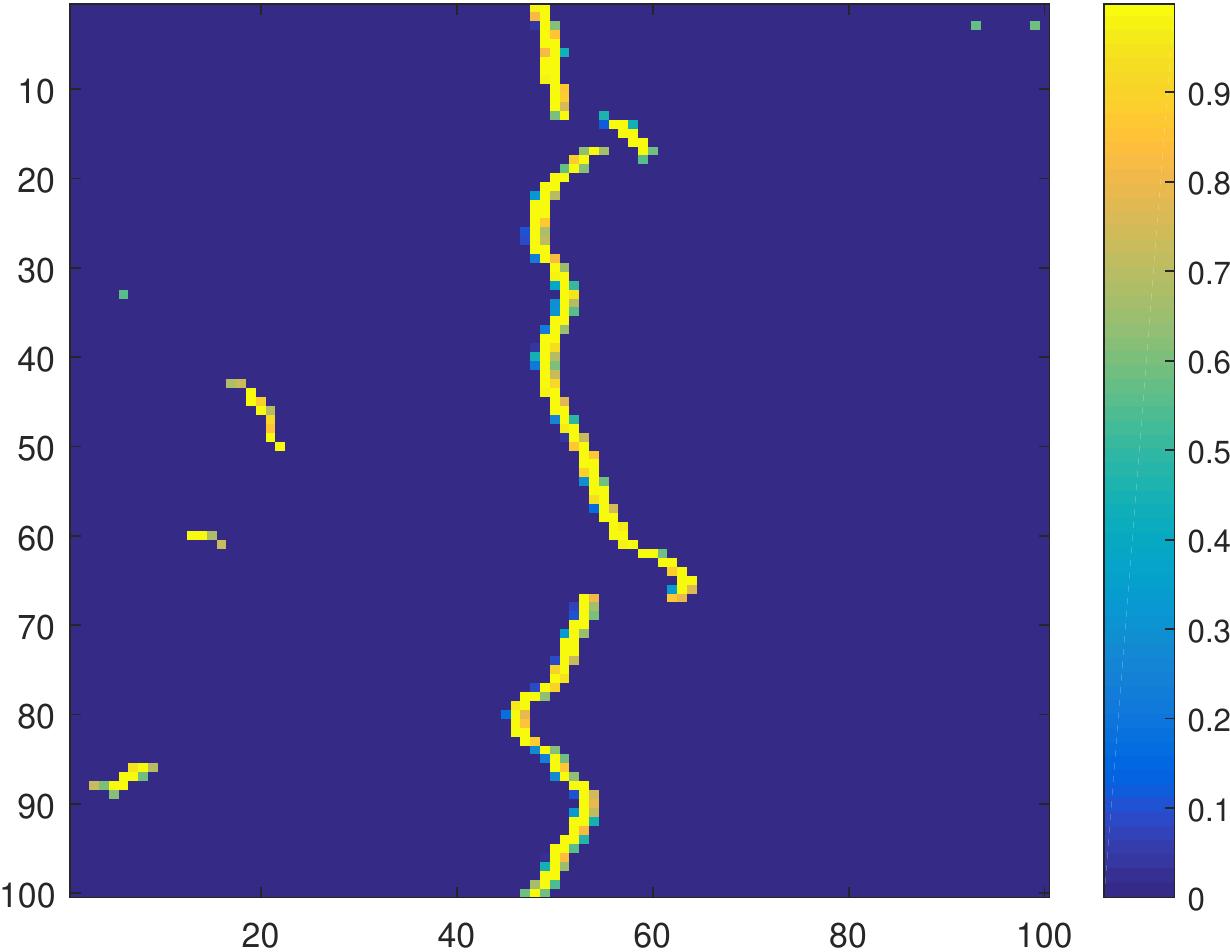}}
\subfigure[]{\includegraphics[width=2.45cm]{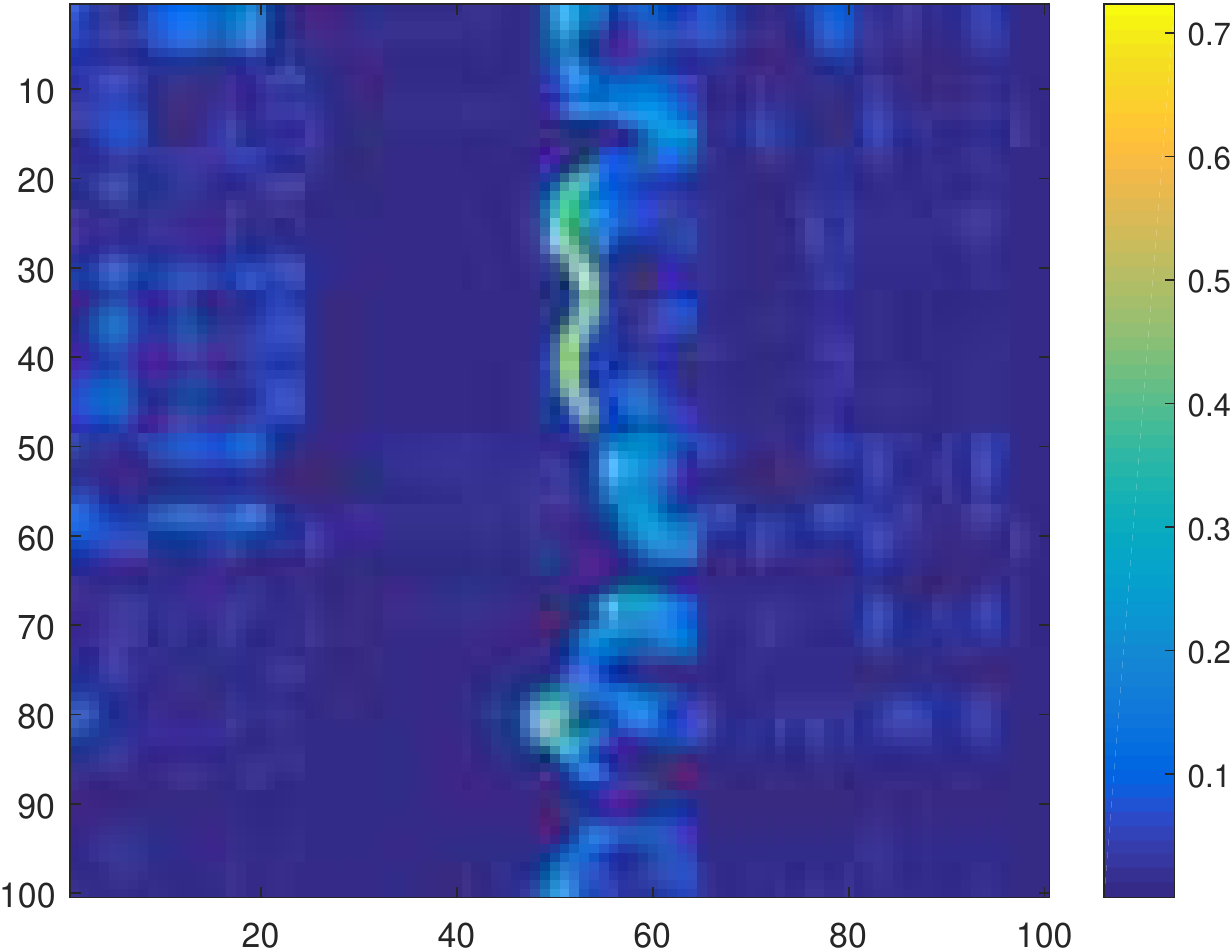}}
\subfigure[]{\includegraphics[width=2.45cm]{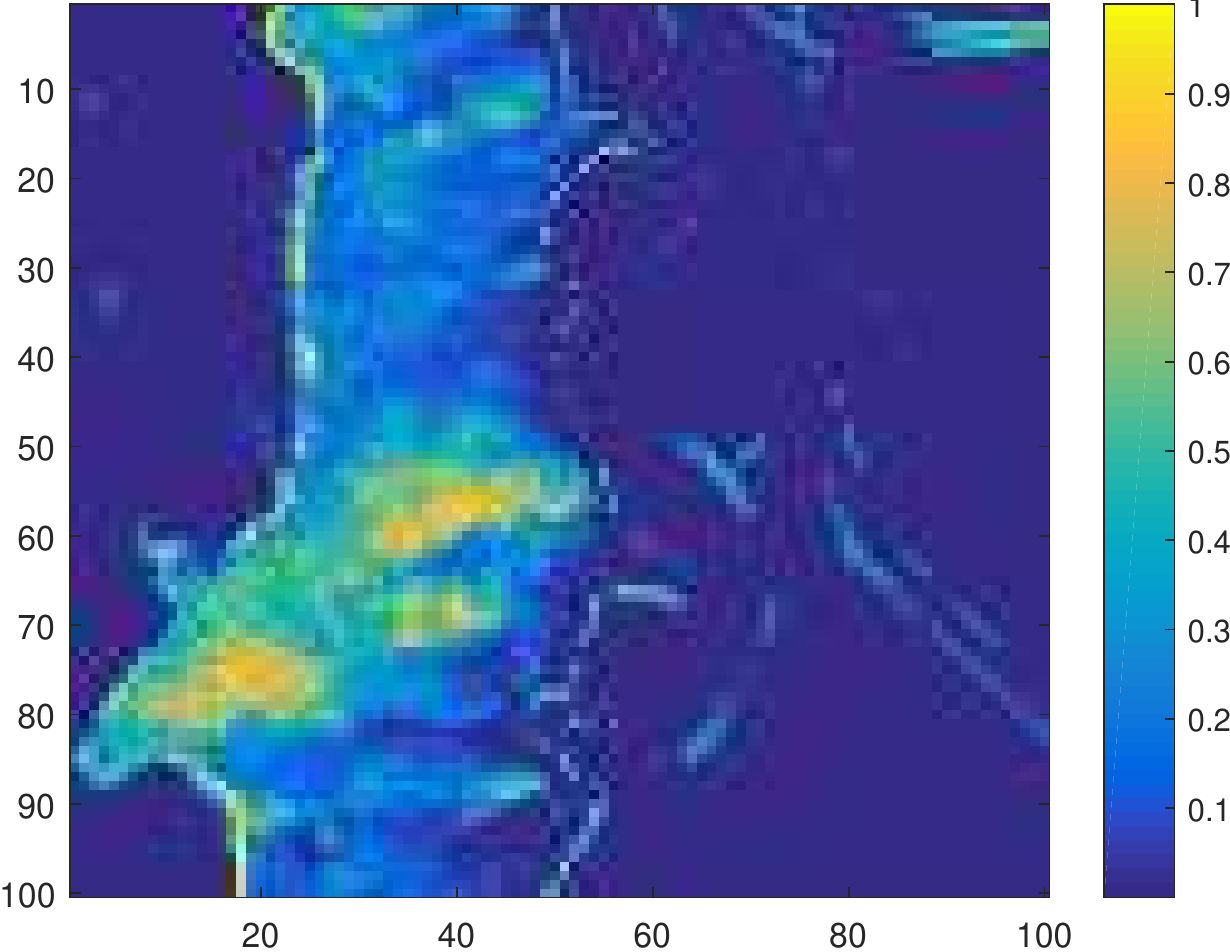}}
\subfigure[]{\includegraphics[width=2.45cm]{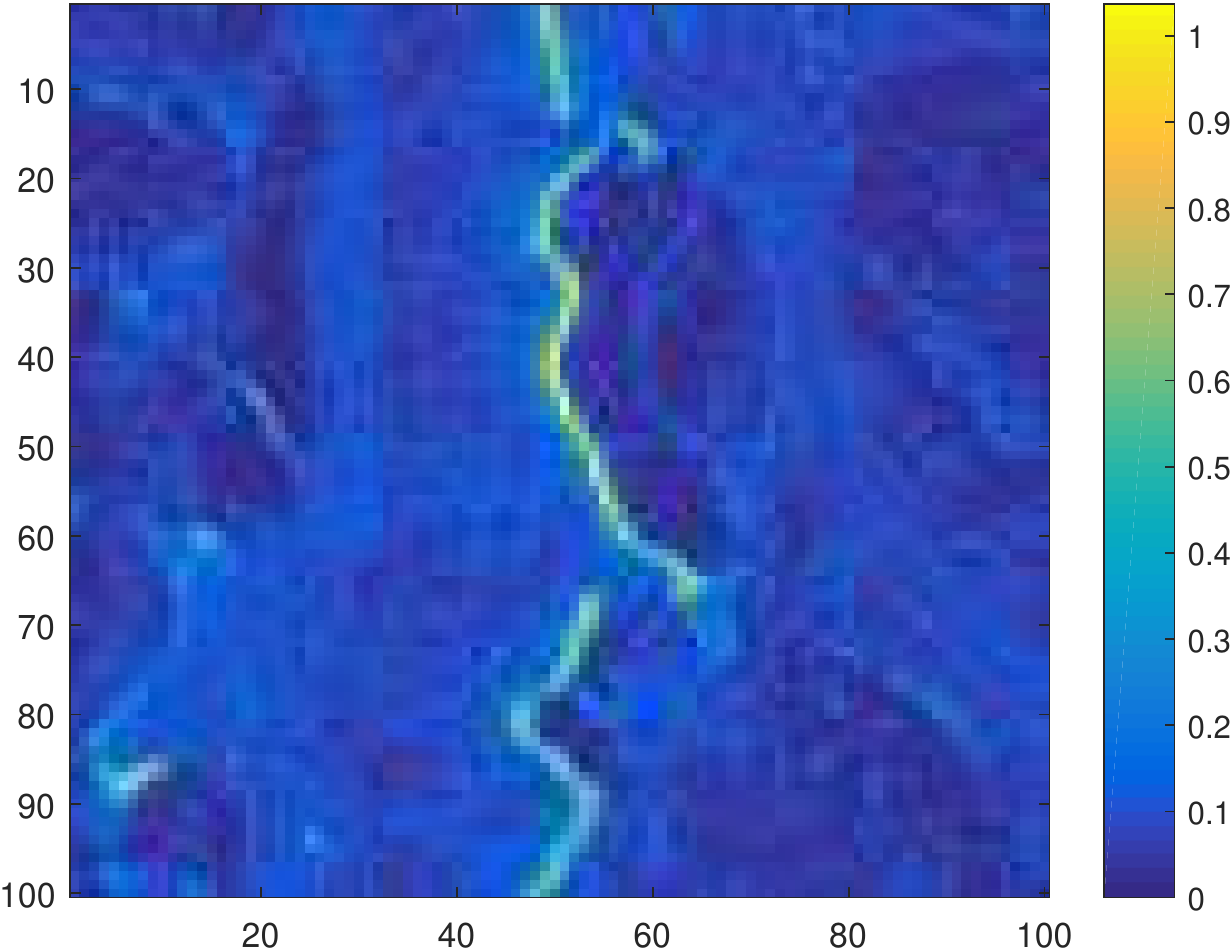}}}
\caption{Fractional abundance maps estimated by different methods of three endmembers on the Jasper Ridge data set. From top to bottom: Tree.  Water. Soil. Road. From left to right: (a) $L_{1/2}$-NMF. (b) SGSNMF. (c) TV-RSNMF. (d) $L_{1/2}$-RNMF. (e) MV-NTF-TV. (f) MLNMF. (g) SSRDMF.}
\label{fig:10}
\end{figure*}

The experimental results are summarized in Tables \ref{tab:2}-\ref{tab:4} and plotted in Figs. \ref{fig:5}-\ref{fig:10}. Tables \ref{tab:2}, \ref{tab:3}, and \ref{tab:4} list the SAD values between each reference spectrum and the endmembers extracted by each method on Cuprite, Samson, and Jasper Ridge data sets respectively. Figs. \ref{fig:5}, \ref{fig:7}, and \ref{fig:9} present the correlation of the endmember signatures obtained by these seven methods and the reference signatures. Figs. \ref{fig:6}, \ref{fig:8} and \ref{fig:10} show the visual comparison of abundance maps estimated by all algorithms.

From Tables \ref{tab:2}-\ref{tab:4}, all methods achieve satisfactory SAD mean values for most materials. It demonstrates that the introductions of sparsity regularizer, spatial information, robust constraint, and multilayer/deep architectures in hyperspectral unmixing are compelling. Besides, compared with $L_{1/2}$-NMF, methods SGSNMF, TV-RSNMF, and MV-NTF-TV achieve lower mean SAD scores, revealing the significance of preserving the spatial information. $L_{1/2}$-RNMF aims to model the sparse noise explicitly, thereby obtaining desirable unmixing results. The performance of MLNMF and SSRDMF verifies that multilayer/deep architectures offer significant advantage. In addition, the results obtained by SSRDMF are better than those provided by MLNMF. This indicates that the combination of multilayer nonlinear network and self-supervised constraint can play a significant role in the task of improving unmixing performance. In particular, from the standard deviation perspective, SGSNMF often provides the best results since the endmember matrix is initialized based on the results of the segmentation and the region-based VCA.

For illustrative purposes, Figs. \ref{fig:5}, \ref{fig:7}, and \ref{fig:9} plot the reference signatures (green solid line) along with the endmembers identified by different methods (red dash line) on the Cuprite, Samson, and  Jasper Ridge data sets, respectively. It can be seen that the endmember signatures are nearly all around correlated in spectral terms with respect to the reference partners. Meanwhile, Figs. \ref{fig:6}, \ref{fig:8}, and \ref{fig:10} show the abundance maps estimated by different methods on the Cuprite, Samson, and  Jasper Ridge data sets, respectively. Due to the ground-truth abundance maps are unavailable, we compare the obtained abundance maps with the real scene image, \emph{i.e.,} Fig. \ref{fig:4}. In the abundance maps, the higher value means the proportion of the material is larger. From Fig. \ref{fig:6}, it is obvious that similar abundance maps can be achieved in most cases for each material. From Fig. \ref{fig:8}, we can observe all estimation results have a good correlation with the geological maps of the Samson data set. For the Jasper Ridge data set, as shown in Fig. \ref{fig:9} and Fig. \ref{fig:10}, it is difficult to obtain desirable estimations in terms of Road for all methods, which may be because there are few pixels for Road so that VCA fails to extract this signature when initializing.

These seven methods have their own benefits. Figs. \ref{fig:6}(a), \ref{fig:8}(a), and \ref{fig:10}(a) show that $L_{1/2}$-NMF can obtain a sparse abundance map because $L_{1/2}$-norm regularizer is an optimal choice for the hyperspectral unmixing. Figs. \ref{fig:6}(b), \ref{fig:8}(b), and \ref{fig:10}(b) keep the spatial group structure and the sparsity within a local spatial group through utilizing the spatial group sparsity regularizer. The results in Figs. \ref{fig:6}(c), \ref{fig:8}(c), and \ref{fig:10}(c) show that it is effective to utilize the piecewise smoothness. The estimated abundances in Figs. \ref{fig:6}(d), \ref{fig:8}(d), and \ref{fig:10}(d) are robust to noise, which is mainly because $L_{1/2}$-RNMF describes the sparse noise explicitly. Figs. \ref{fig:6}(e), \ref{fig:8}(e), and \ref{fig:10}(e) preserve the local spatial structure and the global spectral-spatial information. The results listed in Figs. \ref{fig:6}(f-g), \ref{fig:8}(f-g), and \ref{fig:10}(f-g) display that the abundance maps estimated by MLNMF and SSRDMF are always in accordance with the ground truth, demonstrating the effectiveness of the multilayer/deep architectures. From the results of experiments, we find that sparse regularizer, spectral-spatial information, multilayer/deep architectures are all beneficial to the hyperspectral unmixing.

Last, we investigate the average running time of ten times on Jasper Ridge data set for $L_{1/2}$-NMF, SGSNMF, TV-RSNMF, $L_{1/2}$-RNMF, MV-NTF-TV, MLNMF, and SSRDMF, which are $4.99$, $60.81$, $25.19$, $43.31$, $107.50$, $19.38$, and $136.54$ s, respectively. Apparently, the $L_{1/2}$-NMF performs the fastest estimation since it is a single-layer factorization with efficiency sparse regularizer. SGSNMF and TV-RSNMF require more time due to the learning of spatial group structure and piecewise smooth structure, respectively. To handle sparse noise, $L_{1/2}$-RNMF models the sparse noise explicitly, increasing the running time. MLNMF is also fast since it only decomposes the observation matrix iteratively layer by layer. MV-NTF-TV and SSRDMF need much more time mainly owing to the complex factorization.

\section{Discussion And Future Directions}

NMF plays an increasingly significant role in the field of hyperspectral unmixing. In particular, the constrained NMF has the capacity of providing more accurate endmembers and abundances by integrating the spectral constraints and the spatial constraints. The structured NMF enables flexibility to account for more structures and details such as the difference of pixels, bands, and elements. By extending the decomposition form, the generalized NMF exhibits great potential in acquiring more essential characteristics, \emph{e.g.}, nonlinearities, $3$D structure information, and hidden information.

Nevertheless, there are still several drawbacks. For instance, the constrained NMF generally requires extensive parameter tunning to achieve satisfactory results. The generalized NMF often suffers from time-consuming. Secondly, the NMF-based methods rely on proper guidance or initialization to generate meaningful endmembers. Besides, it is difficult to thoroughly capture the plentiful information of HSIs. To this end, it is quite challenging to obtain excellent unmixing performance. In the future, how to design NMF methods for unmixing deserves further research. We provide some considerations as follows.

\begin{itemize}
  \item One important research direction is to make use of the spatial and spectral information simultaneously to guarantee a more reliable unmixing performance. The spectral-spatial joint has shown considerable potentials for hyperspectral unmixing.
  \item Many NMF approaches have been reported to exploit the spectral characteristics, such as the corresponding simplex volume, the endmember distance, and the signature smoothness. However, more efforts are required to describe the endmember variability under the NMF model, \emph{e.g.}, constructing a $4$D endmember tensor in \cite{TImbiriba2020}. Combining insight from spectral variability with a mathematical treatment would be valuable to improve the performance significantly.
  \item Most of the current NMF algorithms rely strongly on LMM to obtain unmixing results. In real scenarios, multi-path scattering is common due to complex landforms, resulting in nonlinear spectral mixture effects. As shown in \cite{IMeganem2014, OEches2014, NYokoya2014, XZhang2018}, a nonlinear mixture model is closely related to the LMM, indicating NMF methods that aim to solving nonlinear unmixing problems deserve to be further investigated.
  \item To achieve more reliable performance in practical scenarios, there is growing attention on improving the robustness of the methods. Although many robust NMF methods have been proposed, they mainly focus on the robustness to noise. In addition to relieve the effect of different types of noise, it is also important to investigate the robustness to the selections of the initialization methods and the tunable parameters in various application scenarios.
  \item The computational complexity is also an aspect that brings difficulties to apply most existing methods (\emph{e.g.}, graph regularized algorithms, NTF, and multilayer/deep approaches). Meanwhile, HSIs are very large in general. Considering that many applications need real- or near real-time processing, it is crucial to develop fast alternatives to reduce the computation time.
\end{itemize}

\ifCLASSOPTIONcaptionsoff
  \newpage
\fi

\bibliography{bibfile-unmixing}

\bibliographystyle{IEEEtran}

\end{document}